\documentclass[fleqn,a4paper,oneside,openany]{book}
\usepackage[british]{babel}
\usepackage{mathtools}
\usepackage{booktabs}
\usepackage{multirow}
\usepackage{subcaption}
\usepackage{amsmath}

\usepackage{amssymb}
\usepackage{graphicx}
\usepackage{marginnote}

\usepackage{xtab}
\usepackage{pdflscape}
\usepackage{placeins}
\usepackage{longtable}
\usepackage{tabu}
\usepackage{pifont}
\usepackage{csquotes}

\usepackage[numbers, sort&compress, super]{natbib}
\newcommand\id[1]{{\hfill\normalsize{\texttt{#1}}}}
\newcommand\textid[1]{{\normalsize{\texttt{#1}}}}

\usepackage{caption}
\DeclareCaptionLabelSeparator{upright}{ | }
\usepackage{geometry}
\usepackage{setspace}
\usepackage{hyperref}
\hypersetup{hidelinks}

\usepackage{titlesec}
\titlespacing\subsubsection{0pt}{8pt plus 2pt minus 2pt}{0pt plus 2pt minus 2pt}

\DeclarePairedDelimiter{\ceil}{\lceil}{\rceil}
\DeclarePairedDelimiter{\floor}{\lfloor}{\rfloor}
\DeclarePairedDelimiter{\norm}{\lVert}{\rVert}
\DeclarePairedDelimiter{\iverson}{\lbrack}{\rbrack}
\DeclarePairedDelimiter{\abs}{\lvert}{\rvert}
\newcommand{\cmark}{\ding{52}}%
\newcommand{\xmark}{\ding{53}}%

\newcommand{\layoutnewpage}{\newpage}
\renewcommand{\layoutnewpage}[1][]{}
\newcommand{\layoutvspace}{\vspace{1cm}}
\renewcommand{\layoutvspace}[1][]{}
\newcommand{\layoutclearpage}{\clearpage}
\renewcommand{\layoutclearpage}[1][]{}
\newcommand{\layoutsetvspace}[1]{\vspace{#1}}
\renewcommand{\layoutsetvspace}[1][]{}

\begin{document}

% \frontmatter
% \title{Image biomarker standardisation initiative\\
% \large Reference manual\\
% \vspace{10mm}version 1.0 (December 2019)}
% \author{}
% \date{}
% \maketitle

\frontmatter
\title{Image biomarker standardisation initiative\\
\large Reference manual}
\author{}
\date{}
\maketitle

\newpage
\chapter*{The image biomarker standardisation initiative}
The image biomarker standardisation initiative (IBSI) is an independent international collaboration which works towards standardising the extraction of image biomarkers from acquired imaging for the purpose of high-throughput quantitative image analysis (radiomics). Lack of reproducibility and validation of radiomic studies is considered to be a major challenge for the field. Part of this challenge lies in the scantiness of consensus-based guidelines and definitions for the process of translating acquired imaging into high-throughput image biomarkers. The IBSI therefore seeks to provide standardised image biomarker nomenclature and definitions, a standardised general image processing workflow, tools for verifying radiomics software implementations and reporting guidelines for radiomic studies.

\subsubsection*{Permanent identifiers}
The IBSI uses permanent identifiers for image biomarker definitions and important related concepts such as image processing. These consist of four-character codes and may be used for reference. Please do not use page numbers or section numbers as references, as these are subject to change.

\subsubsection*{Copyright}
This work is a copy-edited version of the final (v10) pre-print version of the IBSI reference manual, which was licensed under the Creative Commons Attribution 4.0 International License (CC-BY). The original work may be cited as:
Zwanenburg A, Leger S, Valli\`{e}res M, L\"{o}ck S. Image biomarker standardisation initiative. arXiv preprint arXiv:1612.07003.

This document is licensed under the Creative Commons Attribution 4.0 International License. To view a copy of this license, visit http://creativecommons.org/licenses/by/4.0/ or send a letter to Creative Commons, PO Box 1866, Mountain View, CA 94042, USA.

The digital phantom (see section \ref{sec_digital_phantom}) is licensed under the Creative Commons Attribution 4.0 International License. To view a copy of this license, visit http://creativecommons.org/licenses/by/4.0/ or send a letter to Creative Commons, PO Box 1866, Mountain View, CA 94042, USA.

The radiomics phantom (see section \ref{sec_patient_data}), which is based on a human lung cancer computed tomography image and published by cancerdata.org (\href{http://dx.doi.org/10.17195/candat.2016.08.1}{DOI:10.17195/candat.2016.08.1}), is licensed under the Creative Commons Attribution-NonCommercial 3.0 Unported Licence. To view a copy of this license, visit https://creativecommons.org/licenses/by-nc/3.0/ or send a letter to Creative Commons, PO Box 1866, Mount View, CA 94042, USA. This license pertains to both the original \texttt{DICOM} set, as well as the same data in \texttt{NIfTI} format released by the IBSI.

Copyright information regarding the reference data sets may be found on GitHub: https://github.com/theibsi/data\_sets

\subsubsection{Citation information}
To cite the document or the digital phantom, please use the following citation:
\begin{enumerate}
\item Zwanenburg A, Leger S, Valli\`{e}res M, L\"{o}ck S. Image biomarker standardisation initiative. arXiv preprint arXiv:1612.07003.
\end{enumerate}

Additionally, when using the radiomics phantom originally published on cancerdata.org, please include the following citation:
\begin{enumerate}
\item Lambin P, Leijenaar RT, Deist TM, Peerlings J, de Jong EE, van Timmeren J, Sanduleanu S, Larue RT, Even AJ, Jochems A, van Wijk Y. Radiomics: the bridge between medical imaging and personalized medicine. Nature Reviews Clinical Oncology. 2017 Dec;14(12):749.
\item Lambin P. Radiomics Digital Phantom, CancerData (2016), DOI:10.17195/candat.2016.08.1
\end{enumerate}

\subsubsection*{Contact}
\noindent Dr. Alex Zwanenburg\\
\textit{alexander.zwanenburg@nct-dresden.de}

\noindent Dr. Martin Valli\`{e}res\\
\textit{mart.vallieres@gmail.com}

\subsubsection*{IBSI contributors}
\begin{longtabu} to 0.99\linewidth {@{}lp{10cm}@{}}
\toprule
\endhead
\multicolumn{2}{r}{\textit{continued on next page}}
\endfoot
\bottomrule
\caption{Alphabetical list of IBSI contributors.\label{participantList}}
\endlastfoot
Mahmoud A. Abdalah & Image Response Assessment Team Core Facility, Moffitt Cancer Center, Tampa (FL), USA\\
Hugo J.W.L. Aerts & Dana-Farber Cancer Institute, Brigham and Women's Hospital, and Harvard Medical School, Harvard University, Boston (MA), USA\\
Vincent Andrearczyk & Institute of Information Systems, University of Applied Sciences Western Switzerland (HES-SO), Sierre, Switzerland\\
Aditya Apte & Department of Medical Physics, Memorial Sloan Kettering Cancer Center, New York (NY), USA\\
Saeed Ashrafinia & Department of Electrical and Computer Engineering, Johns Hopkins University, Baltimore (MD), USA \textit{and} Department of Radiology and Radiological Science, Johns Hopkins University, Baltimore (MD), USA\\
Spyridon Bakas & Center for Biomedical image Computing and Analytics (CBICA), University of Pennsylvania, Philadelphia (PA), USA \textit{and} Department of Radiology, Perelman School of Medicine, University of Pennsylvania, Philadelphia (PA), USA \textit{and} Department of Pathology and Laboratory Medicine, Perelman School of Medicine, University of Pennsylvania, Philadelphia (PA), USA\\
Roelof J. Beukinga & Department of Nuclear Medicine and Molecular Imaging, University of Groningen, University Medical Center Groningen (UMCG), Groningen, The Netherlands\\
Ronald Boellaard & Department of Nuclear Medicine and Molecular Imaging, University of Groningen, University Medical Center Groningen (UMCG), Groningen, The Netherlands \textit{and} Radiology and Nuclear Medicine, VU University Medical Centre (VUMC), Amsterdam, The Netherlands\\
Marta Bogowicz & Department of Radiation Oncology, University Hospital Zurich, University of Zurich, Zurich, Switzerland\\
Luca Boldrini & Fondazione Policlinico Universitario ""A. Gemelli"" IRCCS, Roma, Italia\\
Ir\`{e}ne Buvat & Imagerie Mol\'{e}culaire In Vivo, CEA, Inserm, Univ Paris Sud, CNRS, Université Paris Saclay, Orsay, France\\
Gary J.R. Cook & Cancer Imaging Dept, School of Biomedical Engineering and Imaging Sciences, King’s College London, London, United Kingdom\\
Christos Davatzikos & Center for Biomedical image Computing and Analytics (CBICA), University of Pennsylvania, Philadelphia (PA), USA \textit{and} Department of Radiology, Perelman School of Medicine, University of Pennsylvania, Philadelphia (PA), USA\\
Adrien Depeursinge & Institute of Information Systems, University of Applied Sciences Western Switzerland (HES-SO), Sierre, Switzerland \textit{and} Department of Nuclear Medicine and Molecular Imaging, Lausanne University Hospital, Lausanne, Switzerland\\
Marie-Charlotte Desseroit & Laboratory of Medical Information Processing (LaTIM) - team ACTION (image-guided therapeutic action in oncology), INSERM, UMR 1101, IBSAM, UBO, UBL, Brest, France\\
Nicola Dinapoli & Fondazione Policlinico Universitario ""A. Gemelli"" IRCCS, Roma, Italia\\
Cuong Viet Dinh & Department of Radiation Oncology, the Netherlands Cancer Institute (NKI), Amsterdam, The Netherlands\\
Sebastian Echegaray & Department of Radiology, Stanford University School of Medicine, Stanford (CA), USA\\
Issam El Naqa & Department of Radiation Oncology, Physics Division, University of Michigan, Ann Arbor (MI), USA \textit{and} Medical Physics Unit, McGill University, Montr\'{e}al, Québec, Canada\\
Andriy Y. Fedorov & Surgical Planning Laboratory, Brigham and Women's Hospital and Harvard Medical School, Harvard University, Boston (MA), USA\\
Roberto Gatta & Fondazione Policlinico Universitario ""A. Gemelli"" IRCCS, Rome, Italy\\
Robert J. Gillies & Department of Cancer Imaging and Metabolism, Moffitt Cancer Center, Tampa (FL), USA\\
Vicky Goh & Cancer Imaging Dept, School of Biomedical Engineering and Imaging Sciences, King’s College London, London, United Kingdom\\
Matthias Guckenberger & Department of Radiation Oncology, University Hospital Zurich, University of Zurich, Zurich, Switzerland\\
Michael G\"{o}tz & Department of Medical Image Computing, German Cancer Research Center (DKFZ), Heidelberg, Germany\\
Sung Min Ha & Center for Biomedical image Computing and Analytics (CBICA), University of Pennsylvania, Philadelphia (PA), USA \textit{and} Department of Radiology, Perelman School of Medicine, University of Pennsylvania, Philadelphia (PA), USA\\
Mathieu Hatt & Laboratory of Medical Information Processing (LaTIM) - team ACTION (image-guided therapeutic action in oncology), INSERM, UMR 1101, IBSAM, UBO, UBL, Brest, France\\
Fabian Isensee & Department of Medical Image Computing, German Cancer Research Center (DKFZ), Heidelberg, Germany\\
Philippe Lambin & The D-Lab, Department of Precision Medicine, GROW-School for Oncology and Developmental Biology, Maastricht University Medical Centre+, Maastricht, The Netherlands\\
Baptiste Laurent & Laboratory of Medical Information Processing (LaTIM) - team ACTION (image-guided therapeutic action in oncology), INSERM, UMR 1101, IBSAM, UBO, UBL, Brest, France\\
Stefan Leger & National Center for Tumor Diseases (NCT), Partner Site Dresden, Germany: German Cancer Research Center (DKFZ), Heidelberg, Germany; Faculty of Medicine and University Hospital Carl Gustav Carus, Technische Universit\"{a}t Dresden, Dresden, Germany, and; Helmholtz Association / Helmholtz-Zentrum Dresden - Rossendorf (HZDR), Dresden, Germany \textit{and} OncoRay -- National Center for Radiation Research in Oncology, Faculty of Medicine and University Hospital Carl Gustav Carus, Technische Universität Dresden, Helmholtz-Zentrum Dresden - Rossendorf, Dresden, Germany \textit{and} German Cancer Consortium (DKTK), Partner Site Dresden, and German Cancer Research Center (DKFZ), Heidelberg, Germany\\
Ralph T.H. Leijenaar & The D-Lab, Department of Precision Medicine, GROW-School for Oncology and Developmental Biology, Maastricht University Medical Centre+, Maastricht, The Netherlands\\
Jacopo Lenkowicz & Fondazione Policlinico Universitario ""A. Gemelli"" IRCCS, Rome, Italy\\
Fiona Lippert & Section for Biomedical Physics, Department of Radiation Oncology, University of T\"{u}bingen, Germany\\
Are Losneg\r{a}rd & Department of Clinical Medicine, University of Bergen, Bergen, Norway\\
Steffen L\"{o}ck & OncoRay -- National Center for Radiation Research in Oncology, Faculty of Medicine and University Hospital Carl Gustav Carus, Technische Universit\"{a}t Dresden, Helmholtz-Zentrum Dresden - Rossendorf, Dresden, Germany \textit{and} German Cancer Consortium (DKTK), Partner Site Dresden, and German Cancer Research Center (DKFZ), Heidelberg, Germany \textit{and} Department of Radiotherapy and Radiation Oncology, Faculty of Medicine and University Hospital Carl Gustav Carus, Technische Universität Dresden, Dresden, Germany\\
Klaus H. Maier-Hein & Department of Medical Image Computing, German Cancer Research Center (DKFZ), Heidelberg, Germany\\
Sarah A. Mattonen & Department of Medical Biophysics, The University of Western Ontario, London (ON), Canada\\
Olivier Morin & Department of Radiation Oncology, University of California, San Francisco (CA), USA\\
Henning M\"{u}ller & Institute of Information Systems, University of Applied Sciences Western Switzerland (HES-SO), Sierre, Switzerland \textit{and} University of Geneva, Geneva, Switzerland\\
Sandy Napel & Department of Radiology, Stanford University School of Medicine, Stanford (CA), USA \textit{and} Department of Electrical Engineering, Stanford University, Stanford (CA), USA \textit{and} Department of Medicine (Biomedical Informatics Research), Stanford University School of Medicine, Stanford (CA), USA\\
Christophe Nioche & Imagerie Mol\'{e}culaire In Vivo, CEA, Inserm, Univ Paris Sud, CNRS, Université Paris Saclay, Orsay, France\\
Fanny Orlhac & Imagerie Mol\'{e}culaire In Vivo, CEA, Inserm, Univ Paris Sud, CNRS, Université Paris Saclay, Orsay, France\\
Sarthak Pati & Center for Biomedical image Computing and Analytics (CBICA), University of Pennsylvania, Philadelphia (PA), USA \textit{and} Department of Radiology, Perelman School of Medicine, University of Pennsylvania, Philadelphia (PA), USA\\
Elisabeth A.G. Pfaehler & Department of Nuclear Medicine and Molecular Imaging, University of Groningen, University Medical Center Groningen (UMCG), Groningen, The Netherlands\\
Arman Rahmim & Departments of Radiology and Physics, University of British Columbia, Vancouver (BC), Canada \textit{and} Department of Radiology and Radiological Science, Johns Hopkins University, Baltimore (MD), USA\\
Arvind U.K. Rao & Department of Computational Medicine and Bioinformatics, University of Michigan, Ann Arbor (MI), USA \textit{and} Department of Radiation Oncology, Physics Division, University of Michigan, Ann Arbor (MI), USA\\
Christian Richter & OncoRay -- National Center for Radiation Research in Oncology, Faculty of Medicine and University Hospital Carl Gustav Carus, Technische Universit\"{a}t Dresden, Helmholtz-Zentrum Dresden - Rossendorf, Dresden, Germany \textit{and} German Cancer Consortium (DKTK), Partner Site Dresden, and German Cancer Research Center (DKFZ), Heidelberg, Germany \textit{and} Department of Radiotherapy and Radiation Oncology, Faculty of Medicine and University Hospital Carl Gustav Carus, Technische Universität Dresden, Dresden, Germany \textit{and} Helmholtz-Zentrum Dresden - Rossendorf, Institute of Radiooncology – OncoRay, Dresden, Germany\\
Jonas Scherer & Department of Medical Image Computing, German Cancer Research Center (DKFZ), Heidelberg, Germany\\
Muhammad Musib Siddique & Cancer Imaging Dept, School of Biomedical Engineering and Imaging Sciences, King’s College London, London, United Kingdom\\
Nanna M. Sijtsema & Department of Radiation Oncology, University of Groningen, University Medical Center Groningen (UMCG), Groningen, The Netherlands\\
Jairo Socarras Fernandez & Section for Biomedical Physics, Department of Radiation Oncology, University of T\"{u}bingen, Germany\\
Emiliano Spezi & School of Engineering, Cardiff University, Cardiff, United Kingdom \textit{and} Department of Medical Physics, Velindre Cancer Centre, Cardiff, United Kingdom\\
Roel J.H.M Steenbakkers & Department of Radiation Oncology, University of Groningen, University Medical Center Groningen (UMCG), Groningen, The Netherlands\\
Stephanie Tanadini-Lang & Department of Radiation Oncology, University Hospital Zurich, University of Zurich, Zurich, Switzerland\\
Daniela Thorwarth & Section for Biomedical Physics, Department of Radiation Oncology, University of T\"{u}bingen, Germany\\
Esther G.C. Troost & OncoRay -- National Center for Radiation Research in Oncology, Faculty of Medicine and University Hospital Carl Gustav Carus, Technische Universit\"{a}t Dresden, Helmholtz-Zentrum Dresden - Rossendorf, Dresden, Germany \textit{and} National Center for Tumor Diseases (NCT), Partner Site Dresden, Germany: German Cancer Research Center (DKFZ), Heidelberg, Germany; Faculty of Medicine and University Hospital Carl Gustav Carus, Technische Universität Dresden, Dresden, Germany, and; Helmholtz Association / Helmholtz-Zentrum Dresden - Rossendorf (HZDR), Dresden, Germany \textit{and} German Cancer Consortium (DKTK), Partner Site Dresden, and German Cancer Research Center (DKFZ), Heidelberg, Germany \textit{and} Department of Radiotherapy and Radiation Oncology, Faculty of Medicine and University Hospital Carl Gustav Carus, Technische Universität Dresden, Dresden, Germany \textit{and} Helmholtz-Zentrum Dresden - Rossendorf, Institute of Radiooncology – OncoRay, Dresden, Germany\\
Taman Upadhaya & Department of Nuclear Medicine, CHU Mil\'{e}trie, Poitiers, France \textit{and} Laboratory of Medical Information Processing (LaTIM) - team ACTION (image-guided therapeutic action in oncology), INSERM, UMR 1101, IBSAM, UBO, UBL, Brest, France\\
Vincenzo Valentini & Fondazione Policlinico Universitario ""A. Gemelli"" IRCCS, Rome, Italy\\
Martin Valli\`{e}res & Medical Physics Unit, McGill University, Montr\'{e}al, Québec, Canada\\
Lisanne V. van Dijk & Department of Radiation Oncology, University of Groningen, University Medical Center Groningen (UMCG), Groningen, The Netherlands\\
Joost van Griethuysen & Department of Radiology, the Netherlands Cancer Institute (NKI), Amsterdam, The Netherlands \textit{and} GROW-School for Oncology and Developmental Biology, Maastricht University Medical Center, Maastricht, The Netherlands \textit{and} Department of Radiation Oncology, Dana-Farber Cancer Institute, Brigham and Women’s Hospital, Harvard Medical School, Boston (MA), USA\\
Floris H.P. van Velden & Department of Radiology, Leiden University Medical Center (LUMC), Leiden, The Netherlands\\
Philip Whybra & School of Engineering, Cardiff University, Cardiff, United Kingdom\\
Alex Zwanenburg & OncoRay -- National Center for Radiation Research in Oncology, Faculty of Medicine and University Hospital Carl Gustav Carus, Technische Universit\"{a}t Dresden, Helmholtz-Zentrum Dresden - Rossendorf, Dresden, Germany \textit{and} National Center for Tumor Diseases (NCT), Partner Site Dresden, Germany: German Cancer Research Center (DKFZ), Heidelberg, Germany; Faculty of Medicine and University Hospital Carl Gustav Carus, Technische Universität Dresden, Dresden, Germany, and; Helmholtz Association / Helmholtz-Zentrum Dresden - Rossendorf (HZDR), Dresden, Germany \textit{and} German Cancer Consortium (DKTK), Partner Site Dresden, and German Cancer Research Center (DKFZ), Heidelberg, Germany\\
\end{longtabu}

\newpage
\setcounter{tocdepth}{1}
\tableofcontents

\mainmatter

\newpage
\chapter{Introduction}
A biomarker is "\textit{a characteristic that is objectively measured and evaluated as an indicator of normal biological processes, pathogenic processes, or pharmacologic responses to a therapeutic intervention}"\citep{Atkinson2001}. Biomarkers may be measured from a wide variety of sources, such as tissue samples, cell plating, and imaging. The latter are often referred to as imaging biomarkers \citep{OConnor2016}. Imaging biomarkers consist of both qualitative biomarkers, which require expert interpretation, and quantitative biomarkers which are based on mathematical definitions. Calculation of quantitative imaging biomarkers can be automated, which enables high-throughput analyses. We refer to such (high-throughput) quantitative biomarkers as image biomarkers to differentiate them from qualitative imaging biomarkers. Image biomarkers characterise the contents of (regions of) an image, such as \textit{volume} or \textit{mean intensity}. Because of the historically close relationship with the computer vision field, image biomarkers are also referred to as image features. The term \textit{features}, instead of biomarkers, will be used throughout the remainder of the reference manual, as the contents are generally applicable and not limited to life sciences and medicine only.

This work focuses specifically on the (high-throughput) extraction of image biomarkers from acquired, reconstructed and stored  imaging. High-throughput quantitative image analysis (radiomics) has shown considerable growth in e.g. cancer research \citep{Lambin2017}, but the scarceness of consensus guidelines and definitions has led to it being described as a "wild frontier" \citep{Caicedo2017}. This reference manual therefore presents an effort to chart a course through part of this frontier by presenting consensus-based recommendations, guidelines, definitions and reference values for image biomarkers and  defining a general radiomics image processing scheme. We hope use of this manual will improve reproducibility of radiomic studies.

We opted for a specific focus on the computation of image biomarkers from acquired imaging. Thus, validation of imaging biomarkers, either viewed in a broader framework such as the one presented by \citet{OConnor2016}, or within smaller-scope settings such as those presented by \citet{Caicedo2017} and by \citet{Lambin2017}, falls beyond the scope of this work. Notably, the issue of harmonising and standardising (medical) image acquisition and reconstruction is being addressed in a more comprehensive manner by groups such as the Quantitative Imaging Biomarker Alliance \citep{Sullivan2015,Mulshine2015}, the Quantitative Imaging Network \citep{Clarke2014, nordstrom2016quantitative}, and task groups and committees of the American Association of Physicists in Medicine, the European Association for Nuclear Medicine \citep{Boellaard2015}, the European Society of Radiology (ESR) \citep{EuropeanSocietyofRadiologyESR2013}, and the European Organisation for Research and Treatment of Cancer (EORTC) \citep{Waterton2012,OConnor2016}, among others. Where overlap does exists, the reference manual refers to existing recommendations and guidelines.

This reference manual is divided into several chapters that describe processing of acquired and reconstructed (medical) imaging for high-throughput computation of image biomarkers (\textbf{Chapter \ref{chap_img_proc}}); that define a diverse set of image biomarkers (\textbf{Chapter \ref{chap_image_features}}); that describe guidelines for reporting on radiomic studies and provide nomenclature for image biomarkers (\textbf{Chapter \ref{chap_report_guidelines}}); and that describe the data sets and image processing configurations used to find reference values for image biomarkers (\textbf{Chapter \ref{chap_benchmark sets}}).

\newpage
\chapter{Image processing}\label{chap_img_proc}
Image processing is the sequence of operations required to derive image biomarkers (features) from acquired images. In the context of this work an image is defined as a three-dimensional (3D) stack of two-dimensional (2D) digital image slices. Image slices are stacked along the $z$-axis. This stack is furthermore assumed to possess the same coordinate system, i.e. image slices are not rotated or translated (in the $xy$-plane) with regard to each other. Moreover, digital images typically possess a finite resolution. Intensities in an image are thus located at regular intervals, or spacing. In 2D such regular positions are called \textit{pixels}, whereas in 3D the term \textit{voxels} is used. Pixels and voxels are thus represented as the intersections on a regularly spaced grid. Alternatively, pixels and voxels may be represented as rectangles and rectangular cuboids. The centers of the pixels and voxels then coincide with the intersections of the regularly spaced grid. Both representations are used in the document.

Pixels and voxels contain an intensity value for each channel of the image. The number of channels depends on the imaging modality. Most medical imaging generates single-channel images, whereas the number of channels in microscopy may be greater, e.g. due to different stainings. In such multi-channel cases, features may be extracted for each separate channel, a subset of channels, or alternatively, channels may be combined and converted to a single-channel representation. In the remainder of the document we consider an image as if it only possesses a single channel.

The intensity of a pixel or voxel is also called a \textit{grey level} or \textit{grey tone}, particularly in single-channel images. Though practically there is no difference, the terms \textit{grey level} or \textit{grey tone} are more commonly used to refer to discrete intensities, including discretised intensities.

\begin{figure}[bp]
	\centering
	\includegraphics[scale=0.9]{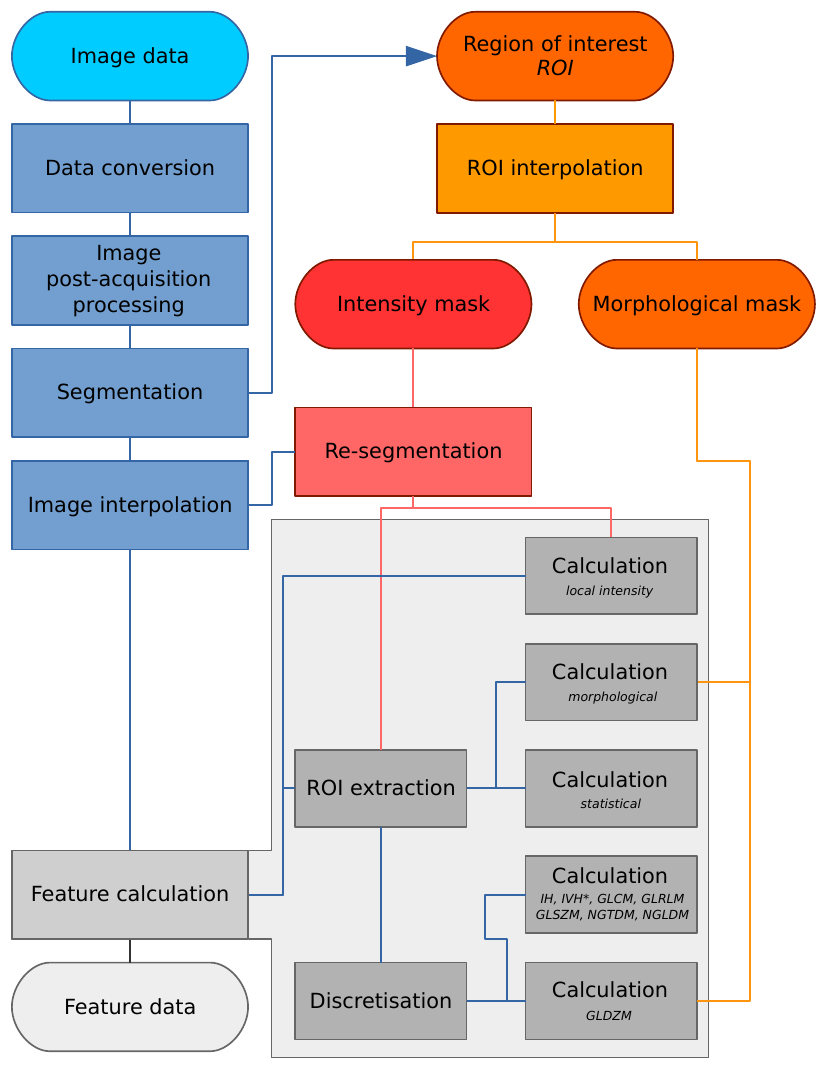}
	\caption{Image processing scheme for image feature calculation. Depending on the specific imaging modality and purpose, some steps may be omitted. The region of interest (ROI) is explicitly split into two masks, namely an intensity and morphological mask, after interpolation to the same grid as the interpolated image. Feature calculation is expanded to show the different feature families with specific pre-processing. IH: intensity histogram; IVH: intensity-volume histogram; GLCM: grey level cooccurrence matrix; GLRLM: grey level run length matrix; GLSZM: grey level size zone matrix; NGTDM: neighbourhood grey tone difference matrix; NGLDM: Neighbouring grey level dependence matrix; GLDZM: grey level distance zone matrix; *Discretisation of IVH differs from IH and texture features, see section \ref{sect_ivh}.}
	\label{figImageProc}
\end{figure}

Image processing may be conducted using a wide variety of schemes. We therefore designed a general image processing scheme for image feature calculation based on schemes used within scientific literature \citep{Hatt2016}. The image processing scheme is shown in figure \ref{figImageProc}. The processing steps referenced in the figure are described in detail within this chapter.

\section[Data conversion]{Data conversion\id{23XZ}}\label{ref_data_conversion}
Some imaging modalities require conversion of raw image data into a more meaningful presentation, e.g. standardised uptake values (SUV)\citep{Boellaard2015}. This is performed during the data conversion step. Assessment of data conversion methods falls outside the scope of the current work.

\section[Post-acquisition processing]{Image post-acquisition processing\id{PCDE}}\label{ref_image_postprocessing}
Images are post-processed to enhance image quality. For instance, magnetic resonance imaging (MRI) contains both Gaussian and Rician noise \citep{Gudbjartsson1995} and may benefit from denoising. As another example, intensities measured using MR may be non-uniform across an image and could require correction \citep{Sled1998,Vovk2007,Balafar2010}. FDG-PET-based may furthermore be corrected for partial volume effects \citep{Soret2007,Boussion2009} and noise \citep{LePogam2013,ElNaqa2017}. In CT imaging, metal objects, e.g. pacemakers and tooth implants, introduce artifacts and may require artifinterpact suppression \citep{Gjesteby2016}. Microscopy images generally benefit from field-of-view illumination correction as illumination is usually inhomogeneous due to the light-source or the optical path \citep{Caicedo2017,Smith2015}.

Evaluation and standardisation of various image post-acquisition processing methods falls outside the scope of the current work. Note that vendors may provide or implement software to perform noise reduction and other post-processing during image reconstruction. In such cases, additional post-acquisition processing may not be required.

\section[Segmentation]{Segmentation\id{OQYT}}\label{ref_segmentation}
High-throughput image analysis, within the feature-based paradigm, relies on the definition of regions of interest (ROI). ROIs are used to define the region in which features are calculated. What constitutes an ROI depends on the imaging and the study objective. For example, in 3D microscopy of cell plates, cells are natural ROIs. In medical imaging of cancer patients, the tumour volume is a common ROI. ROIs can be defined manually by experts or (semi-)automatically using algorithms.

From a process point-of-view, segmentation leads to the creation of an ROI mask $\mathbf{R}$, for which every voxel $j \in \mathbf{R}$ ($R_j$) is defined as:
\begin{displaymath}
R_j =\begin{cases}
1\qquad j \text{ in ROI}\\
0\qquad \text{otherwise}\\
\end{cases}
\end{displaymath}

ROIs are typically stored with the accompanying image. Some image formats directly store ROI masks as voxels (e.g. \texttt{NIfTI}, \texttt{NRRD} and \texttt{DICOM Segmentation}), and generating the ROI mask is conducted by loading the corresponding image. In other cases the ROI is saved as a set of $(x,y,z)$ points that define closed loops of (planar) polygons, for example within \texttt{DICOM RTSTRUCT} or \texttt{DICOM SR} files. In such cases, we should determine which voxel centers lie within the space enclosed by the contour polygon in each slice to generate the ROI mask.

\begin{figure}[thb]
	\centering
	\includegraphics[scale=0.75]{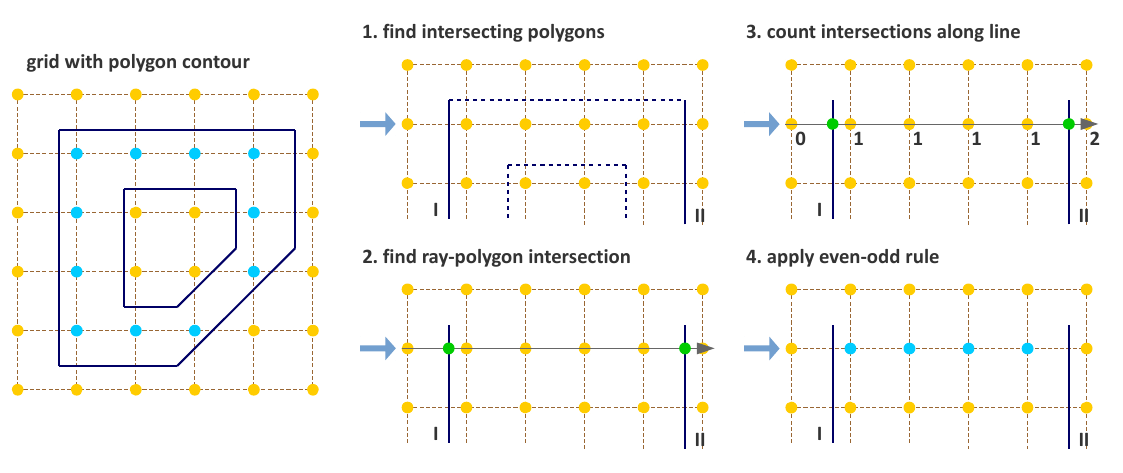}
	\caption{Simple algorithm to determine which pixels are inside a 2D polygon. The suggested implementation consists of four steps: (1) Omit edges that will not intersect with the current row of voxel centers. (2) Calculate intersection points of edges I and II with the ray for the current row. (3) Determine the number of intersections crossed from ray origin to the row voxel centers. (4) Apply \textit{even-odd} rule to determine whether voxel centers are inside the polygon.}
	\label{figImagePointGrid}
\end{figure}

A common method to determine whether a point in an image slice lies inside a 2D polygon is the \textit{crossing number} algorithm, for which several implementations exist \citep{Schirra2008}. The main concept behind this algorithm is that for any point inside the polygon, any line originating outside the polygon will cross the polygon an uneven number of times. A simple example is shown in figure \ref{figImagePointGrid}. The implementation in the example makes use of the fact that the ROI mask is a regular grid to scan entire rows at a time. The example implementation consists of the following steps:
\begin{enumerate}
\item (optional) A ray is cast horizontally from outside the polygon for each of the $n$ image rows. As we iterate over the rows, it is computationally beneficial to exclude polygon edges that will not be crossed by the ray for the current row $j$. If the current row has $y$-coordinate $y_j$, and edge $k$ has two vertices with $y$-coordinates $y_{k1}$ and $y_{k2}$, the ray will not cross the edge if both vertices lie either above or below $y_j$, i.e. $y_j < y_{k1}, y_{k2}$ or $y_j > y_{k1}, y_{k2}$. For each row $j$, find those polygon edges whose $y$-component of the vertices do not both lie on the same side of the row coordinate $y_j$. This step is used to limit calculation of intersection points to only those that cross a ray cast from outside the polygon -- e.g. ray with origin $(-1, y_j)$ and direction $(1,0)$. This an optional step.
\item Determine intersection points $x_i$ of the (remaining) polygon edges with the ray.
\item Iterate over intersection points and add $1$ to the count of each pixel center with $x \geq x_i$.
\item Apply the \textit{even-odd} rule. Pixels with an odd count are inside the polygon, whereas pixels with an even count are outside.
\end{enumerate}

Note that the example represents a relatively naive implementation that will not consistently assign voxel centers positioned on the polygon itself to the interior.

\FloatBarrier
\section[Interpolation]{Interpolation\id{VTM2}}\label{ref_interpolation}
Texture feature sets require interpolation to isotropic voxel spacing to be rotationally invariant, and to allow comparison between image data from different samples, cohorts or batches. Voxel interpolation affects image feature values as many image features are sensitive to changes in voxel size \citep{Yan2015,Bailly2016,Altazi2017,Shafiq-ul-Hassan2017,Shiri2017}. Maintaining consistent isotropic voxel spacing across different measurements and devices is therefore important for reproducibility. At the moment there are no clear indications whether upsampling or downsampling schemes are preferable. Consider, for example, an image stack of slices with $1.0 \times 1.0 \times 3.0~\text{mm}^3$ voxel spacing. Upsampling to $1.0 \times 1.0 \times 1.0~\text{mm}^3$ requires inference and introduces artificial information, while conversely downsampling to the largest dimension ($3.0 \times 3.0 \times 3.0~\text{mm}^3$) incurs information loss. Multiple-scaling strategies potentially offer a good trade-off \citep{Vallieres2017}. Note that downsampling may introduce image aliasing artifacts. Downsampling may therefore require anti-aliasing filters prior to filtering \citep{Mackin2017,Zwanenburg2018}.

While in general 3D interpolation algorithms are used to interpolate 3D images, 2D interpolation within the image slice plane may be recommended in some situations. In 2D interpolation voxels are not interpolated between slices. This may be beneficial if, for example, the spacing between slices is large compared to the desired voxel size, and/or compared to the in-plane spacing. Applying 3D interpolation would either require inferencing a large number of voxels between slices (upsampling), or the loss of a large fraction of in-plane information (downsampling). The disadvantage of 2D interpolation is that voxel spacing is no longer isotropic, and as a consequence texture features can only be calculated in-plane.

\subsubsection*{Interpolation algorithms}
Interpolation algorithms translate image intensities from the original image grid to an interpolation grid. In such grids, voxels are spatially represented by their center. Several algorithms are commonly used for interpolation, such as \textit{nearest neighbour}, \textit{trilinear}, \textit{tricubic convolution} and \textit{tricubic spline interpolation}. In short, \textit{nearest neighbour interpolation} assigns the intensity of the most nearby voxel in the original grid to each voxel in the interpolation grid. \textit{Trilinear interpolation} uses the intensities of the eight most nearby voxels in the original grid to calculate a new interpolated intensity using linear interpolation. \textit{Tricubic convolution} and \textit{tricubic spline interpolation} draw upon a larger neighbourhood to evaluate a smooth, continuous third-order polynomial at the voxel centers in the interpolation grid. The difference between \textit{tricubic convolution} and \textit{tricubic spline interpolation} lies in the implementation. Whereas \textit{tricubic spline interpolation} evaluates the smooth and continuous third-order polynomial at every voxel center, \textit{tricubic convolution} approximates the solution using a convolution filter. Though \textit{tricubic convolution} is faster, with modern hardware and common image sizes, the difference in execution speed is practically meaningless. Both interpolation algorithms produce similar results, and both are often referred to as \textit{tricubic interpolation}. 

While no consensus exists concerning the optimal choice of interpolation algorithm, \textit{trilinear interpolation} is usually seen as a conservative choice. It does not lead to the blockiness produced by \textit{nearest neighbour interpolation} that introduces bias in local textures \citep{Hatt2016}. Nor does it lead to out-of-range intensities which may occur due to overshoot with \textit{tricubic} and higher order interpolations. The latter problem can occur in acute intensity transitions, where the local neighbourhood itself is not sufficiently smooth to evaluate the polynomial within the allowed range. \textit{Tricubic} methods, however, may retain tissue contrast differences better. Particularly when upsampling, \textit{trilinear} interpolation may act as a low-pass filter which suppresses higher spatial frequencies and cause artefacts in high-pass spatial filters. Interpolation algorithms and their advantages and disadvantages are treated in more detail elsewhere, e.g. \citet{thevenaz2000image}.

In a phantom study, \citet{Larue2017} compared \textit{nearest neighbour}, \textit{trilinear} and \textit{tricubic} interpolation and indicated that feature reproducibility is dependent on the selected interpolation algorithm, i.e. some features were more reproducible using one particular algorithm. 

\subsubsection*{Rounding image intensities after interpolation \id{68QD}}
Image intensities may require rounding after interpolation, or the application of cut-off values. For example, in CT images intensities represent Hounsfield Units, and these do not take non-integer values. Following voxel interpolation, interpolated CT intensities are thus rounded to the nearest integer.

\subsubsection*{Partial volume effects in the ROI mask\id{E8H9}}
If the image on which the ROI mask was defined, is interpolated after the ROI was segmented, the ROI mask $\mathbf{R}$ should likewise be interpolated to the same dimensions. Interpolation of the ROI mask is best conducted using either the \textit{nearest neighbour} or \textit{trilinear interpolation} methods, as these are guaranteed to produce meaningful masks. \textit{Trilinear interpolation} of the ROI mask leads to partial volume effects, with some voxels containing fractions of the original voxels. Since a ROI mask is a binary mask, such fractions need to be binarised by setting a partial volume threshold $\delta$:
\begin{displaymath}
R_j=\begin{cases}
1\qquad R_{interp,j} \geq \delta\\
0\qquad R_{interp,j} < \delta
\end{cases}
\end{displaymath}
A common choice for the partial volume threshold is $\delta=0.5$. For \textit{nearest neighbour interpolation} the ROI mask does not contain partial volume fractions, and may be used directly.

Interpolation results depend on the floating point representation used for the image and ROI masks. Floating point representations should at least be full precision (\texttt{32-bit}) to avoid rounding errors.

\subsubsection*{Interpolation grid\id{UMPJ}}
Interpolated voxel centers lie on the intersections of a regularly spaced grid. Grid intersections are represented by two coordinate systems. The first coordinate system is the grid coordinate system, with origin at $(0.0, 0.0, 0.0)$ and distance between directly neighbouring voxel centers (spacing) of $1.0$. The grid coordinate system is the coordinate system typically used by computers, and consequentially, by interpolation algorithms. The second coordinate system is the world coordinate system, which is typically found in (medical) imaging and provides an image scale. As the desired isotropic spacing is commonly defined in world coordinate dimensions, conversions between world coordinates and grid coordinates are necessary, and are treated in more detail after assessing grid alignment methods.

\begin{figure}[htb]
	\centering
	\includegraphics{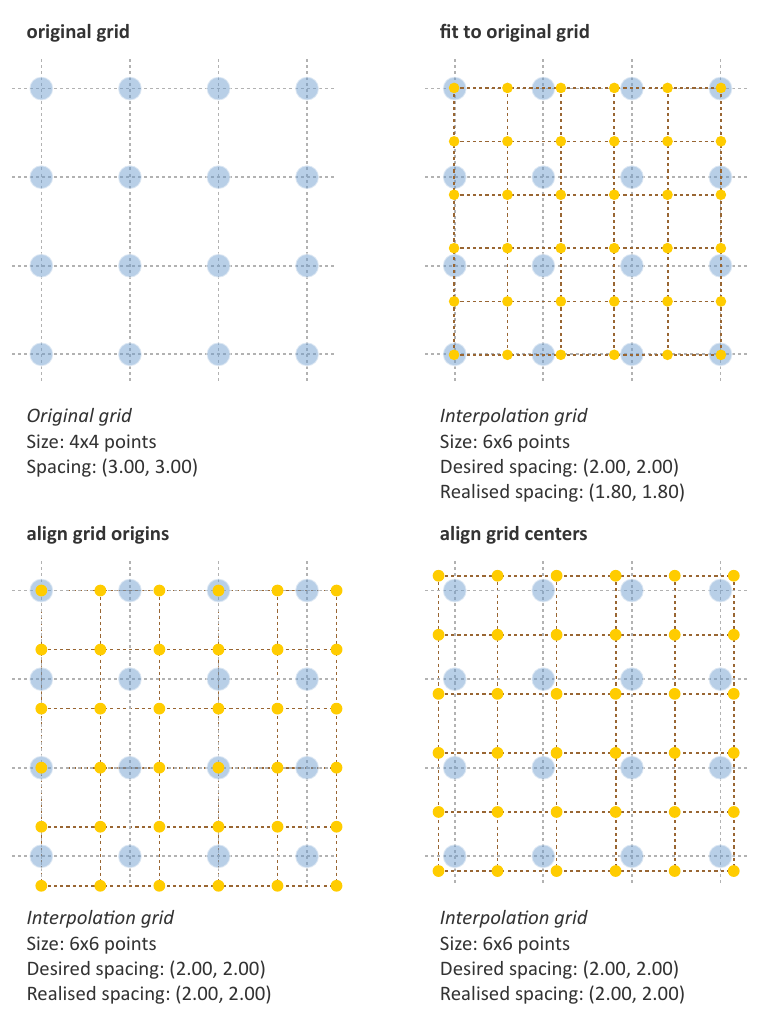}
	\caption{Different interpolation mesh grids based on an original $4\times 4$ grid with $(3.00, 3.00)$~mm spacing. The desired interpolation spacing is $(2.00, 2.00)$~mm. \textit{Fit to original grid} creates an interpolation mesh grid that overlaps with the corners of the original grid. \textit{Align grid origins} creates an interpolation mesh grid that is positioned at the origin of the original grid. \textit{Align grid centers} creates an interpolation grid that is centered on the center of original and interpolation grids.}
	\label{figMeshGrids}
\end{figure}

Grid alignment affects feature values and is non-trivial. Three common grid alignments may be identified, and are shown in figure \ref{figMeshGrids}:
\begin{enumerate}
\item \textbf{Fit to original grid} (\textid{58MB}). In this case the interpolation grid is deformed so that the voxel centers at the grid intersections overlap with the original grid vertices. For an original $4\times4$ voxel grid with spacing $(3.00, 3.00)$ mm and a desired interpolation spacing of $(2.00, 2.00)$~mm we first calculate the extent of the original voxel grid in world coordinates leading to an extent of $((4-1)\,3.00, ((4-1)\,3.00) = (9.00, 9.00)$~mm. In this case the interpolated grid will not exactly fit the original grid. Therefore we try to find the closest fitting grid, which leads to a $6\times 6$ grid by rounding up $(9.00/2.00, 9.00/2.00)$. The resulting grid has a grid spacing of $(1.80, 1.80)$~mm in world coordinates, which differs from the desired grid spacing of $(2.00, 2.00)$~mm.
\item \textbf{Align grid origins} (\textid{SBKJ}). A simple approach which conserves the desired grid spacing is the alignment of the origins of the interpolation and original grids. Keeping with the same example, the interpolation grid is $(6 \times 6)$. The resulting voxel grid has a grid spacing of $(2.00, 2.00)$~mm in world coordinates. By definition both grids are aligned at the origin, $(0.00, 0.00)$.
\item \textbf{Align grid centers} (\textid{3WE3}). The position of the origin may depend on image meta-data defining image orientation. Not all software implementations may process this meta-data the same way. An implementation-independent solution is to align both grids on the grid center. Again, keeping with the same example, the interpolation grid is $(6 \times 6)$. Thus, the resulting voxel grid has a grid spacing of $(2.00, 2.00)$~mm in world coordinates.
\end{enumerate}
\textit{Align grid centers} is recommended as it is implementation-independent and achieves the desired voxel spacing. Technical details of implementing \textit{align grid centers} are described below.

\subsubsection*{Interpolation grid dimensions\id{026Q}}
The dimensions of the interpolation grid are determined as follows. Let $n_a$ be the number of points along one axis of the original grid and $s_{a,w}$ their spacing in world coordinates. Then, let $s_{b,w}$ be the desired spacing after interpolation. The axial dimension of the interpolated mesh grid is then: 
\begin{displaymath}
n_b = \ceil*{\frac{n_a s_a}{s_b}}
\end{displaymath}
Rounding towards infinity guarantees that the interpolation grid exists even when the original grid contains few voxels. However, it also means that the interpolation mesh grid is partially located outside of the original grid. Extrapolation is thus required. Padding the original grid with the intensities at the boundary is recommended. Some implementations of interpolation algorithms may perform this padding internally.

\subsubsection*{Interpolation grid position\id{QCY4}}
For the \textit{align grid centers} method, the positions of the interpolation grid points are determined as follows. As before, let $n_a$ and $n_b$ be the dimensions of one axis in the original and interpolation grid, respectively. Moreover, let $s_{a,w}$ be the original spacing and $s_{b,w}$ the desired spacing for the same axis in world coordinates. Then, with $x_{a,w}$ the origin of the original grid in world coordinates, the origin of the interpolation grid is located at:
\begin{displaymath}
x_{b,w} = x_{a,w} + \frac{s_a (n_a - 1) - s_b (n_b - 1)}{2}
\end{displaymath}
In the grid coordinate system, the original grid origin is located at $x_{a,g} = 0$. The origin of the interpolation grid is then located at:
\begin{displaymath}
x_{b,g} = \frac{1}{2}\left(n_a - 1 - \frac{s_{b,w}}{s_{a,w}} \left(n_b -1\right) \right)
\end{displaymath}
Here the fraction $s_{b,w}/s_{a,w}= s_{b,g}$ is the desired spacing in grid coordinates. Thus, the interpolation grid points along the considered axis are located at grid coordinates:
\begin{displaymath}
x_{b,g},\,x_{b,g} + s_{b,g},\,x_{b,g} + 2s_{b,g},\,\ldots,\,x_{b,g} + (n_b-1)s_{b,g}
\end{displaymath}
Naturally, the above description applies to each grid axis.

\FloatBarrier

\begin{figure}[htb]
	\centering
	\includegraphics[scale=0.9]{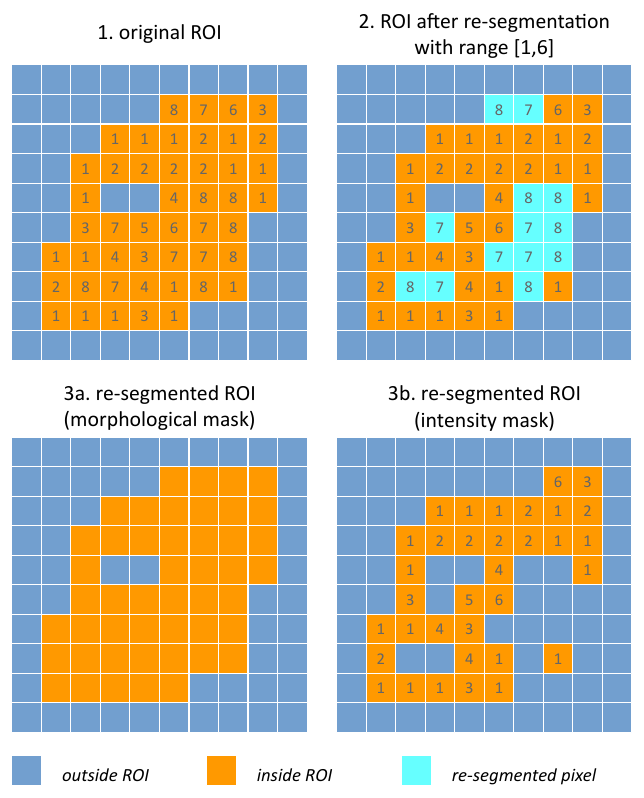}
	\caption{Example showing how intensity and morphological masks may differ due to re-segmentation. (1) The original region of interest (ROI) is shown with pixel intensities. (2) Subsequently, the ROI is re-segmented to only contain values in the range $[1,6]$. Pixels outside this range are marked for removal from the intensity mask. (3a) Resulting morphological mask, which is identical to the original ROI. (3b) Re-segmented intensity mask. Note that due to re-segmentation, intensity and morphological masks are different.}
	\label{figReSegmentationExample}
\end{figure}

\begin{figure}[htb]
    \centering
    \includegraphics{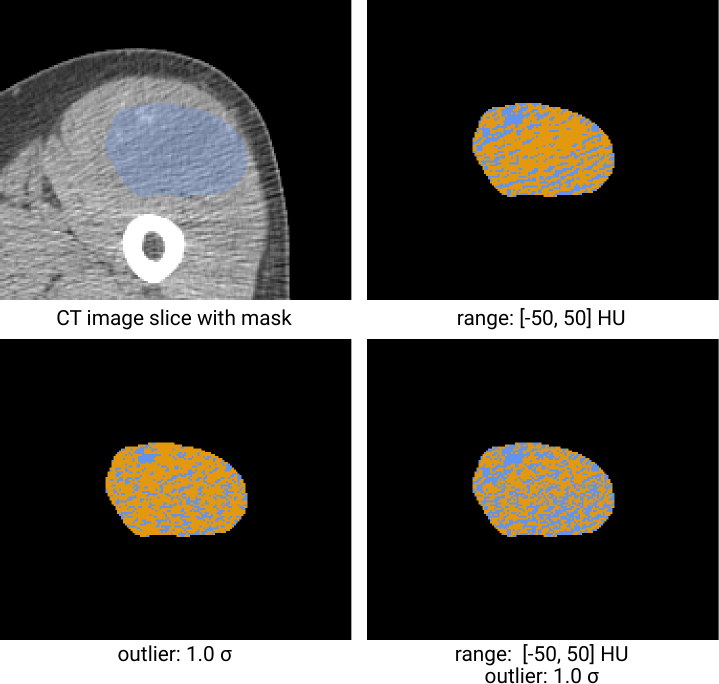}
    \caption{Re-segmentation example based on a CT-image. The masked region (blue) is re-segmented to create an intensity mask (orange). Examples using three different re-segmentation parameter sets are shown. The bottom right combines the range and outlier re-segmentation, and the resulting mask is the intersection of the masks in the other two examples. Image data from Valli\`{e}res et al. \cite{Vallieres2015,Vallieres2015-hv,Clark2013}.}
    \label{fig:resegmentation_real_example}
\end{figure}
\section[Re-segmentation]{Re-segmentation\id{IF9H}}\label{ref_resegmentation}
Re-segmentation entails updating the ROI mask $\mathbf{R}$ based on corresponding voxel intensities $\mathbf{X}_{gl}$. Re-segmentation may be performed to exclude voxels from a previously segmented ROI, and is performed after interpolation. An example use would be the exclusion of air or bone voxels from an ROI defined on CT imaging. Two common re-segmentation methods are described in this section. Combining multiple re-segmentation methods is possible. In this case, the intersection of the intensity ranges defined by the re-segmentation methods is used.

\subsubsection*{Intensity and morphological masks of an ROI\id{ECJF}}
Conventionally, an ROI consists of a single mask. However, re-segmentation may lead to exclusion of internal voxels, or divide the ROI into sub-volumes. To avoid undue complexity by again updating the re-segmented ROI for a more plausible morphology, we define two separate ROI masks.

The morphological mask (\textid{G5KJ}) is not re-segmented and maintains the original morphology as defined by an expert and/or (semi-)automatic segmentation algorithms.

The intensity mask (\textid{SEFI}) can be re-segmented and will contain only the selected voxels. For many feature families, only this is important. However, for morphological and grey level distance zone matrix (GLDZM) feature families, both intensity and morphological masks are used. A two-dimensional schematic example is shown in figure \ref{figReSegmentationExample}, and a real example is shown in figure \ref{fig:resegmentation_real_example}.

\subsubsection*{Range re-segmentation\id{USB3}}
Re-segmentation may be performed to remove voxels from the intensity mask that fall outside of a specified range. An example is the exclusion of voxels with Hounsfield Units indicating air and bone tissue in the tumour ROI within CT images, or low activity areas in PET images. Such ranges of intensities of included voxels are usually presented as a closed interval $\left[ a,b\right]$ or half-open interval $\left[a,\infty\right)$, respectively. For arbitrary intensity units (found in e.g. raw MRI data, uncalibrated microscopy images, and many spatial filters), no re-segmentation range can be provided.

When a re-segmentation range is defined by the user, it needs to be propagated and used for the calculation of features that require a specified intensity range (e.g. intensity-volume histogram features) and/or that employs \textit{fixed bin size} discretisation. Recommendations for the possible combinations of different imaging intensity definitions, re-segmentation ranges and discretisation algorithms are provided in Table \ref{table_discretisation}.

\subsubsection*{Intensity outlier filtering\id{7ACA}}
ROI voxels with outlier intensities may be removed from the intensity mask. One method for defining outliers was suggested by \citet{Vallieres2015} after \citet{Collewet2004}. The mean $\mu$ and standard deviation $\sigma$ of grey levels of voxels assigned to the ROI are calculated. Voxels outside the range $\left[\mu - 3\sigma, \mu + 3\sigma\right]$ are subsequently excluded from the intensity mask.

\FloatBarrier

\section[ROI extraction]{ROI extraction\id{1OBP}}

\begin{figure}[tb]
    \centering
    \includegraphics{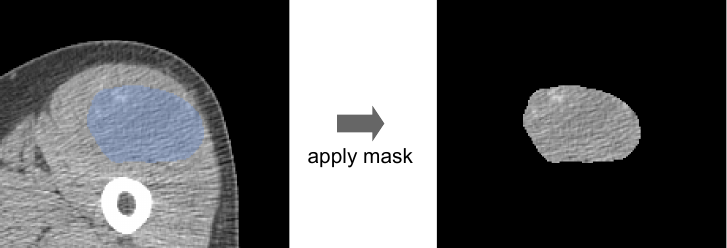}
    \caption{Masking of an image by the ROI mask during \textit{ROI extraction}. Intensities outside the ROI are excluded. Image data from Valli\`{e}res et al. \cite{Vallieres2015,Vallieres2015-hv,Clark2013}.}
    \label{fig:roi_extraction}
\end{figure}

Many feature families require that the ROI is isolated from the surrounding voxels. The ROI intensity mask is used to extract the image volume to be studied. Excluded voxels are commonly replaced by a placeholder value, often \textit{NaN}. This placeholder value may then used to exclude these voxels from calculations. Voxels included in the ROI mask retain their original intensity. An example is shown in figure \ref{fig:roi_extraction}.

\section[Intensity discretisation]{Intensity discretisation\id{4R0B}}\label{discretisation}
\begin{figure}[thb]
	\centering
	\includegraphics[width=\linewidth]{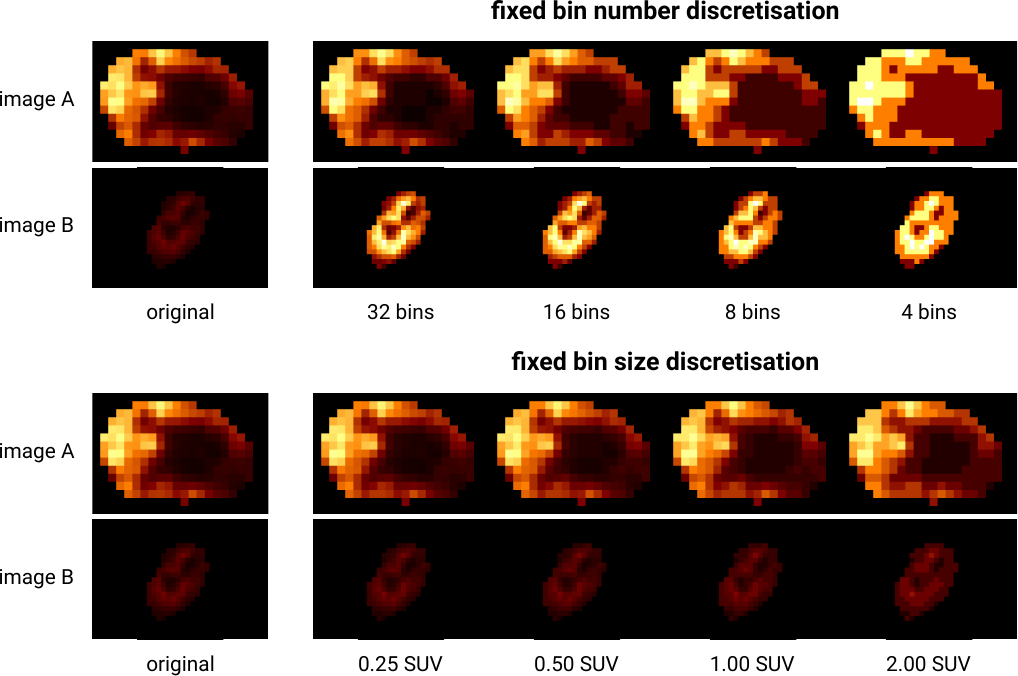}
	\caption{Discretisation of two different 18F-FDG-PET images with SUV\textsubscript{max} of $27.8$ (A) and $6.6$ (B). \textit{Fixed bin number} discretisation adjust the contrast between the two images, with the number of bins determining the coarseness of the discretised image. \textit{Fixed bin size} discretisation leaves the contrast differences between image A and B intact. Increasing the bin size increases the coarseness of the discretised image. Image data from Valli\`{e}res et al. \cite{Vallieres2015,Vallieres2015-hv,Clark2013}.}
	\label{figImageDiscretisation}
\end{figure}

Discretisation or quantisation of image intensities inside the ROI is often required to make calculation of texture features tractable \citep{Yip2016}, and possesses noise-suppressing properties as well. An example of discretisation is shown in figure \ref{figImageDiscretisation}.

Two approaches to discretisation are commonly used. One involves the discretisation to a fixed number of bins, and the other discretisation with a fixed bin width. As we will observe, there is no inherent preference for one or the other method. However, both methods have particular characteristics (as described below) that may make them better suited for specific purposes. Note that the lowest bin always has value $1$, and not $0$. This ensures consistency for calculations of texture features, where for some features grey level $0$ is not allowed .

\subsubsection*{Fixed bin number\id{K15C}}\label{par_discr_FBN}
In the \textit{fixed bin number} method, intensities $X_{gl}$ are discretised to a fixed number of $N_g$ bins. It is defined as follows:
\begin{displaymath}
X_{d,k} = \begin{cases}
\floor*{N_g \frac{X_{gl,k}-X_{gl,min}}{X_{gl,max}-X_{gl,min}}} + 1 & X_{gl,k}<X_{gl,max}\\
N_g & X_{gl,k}=X_{gl,max}
\end{cases}
\end{displaymath}
In short, the intensity $X_{gl,k}$ of voxel $k$ is corrected by the lowest occurring intensity $X_{gl,min}$ in the ROI, divided by the bin width $\left(X_{gl,max}-X_{gl,min}\right)/N_g$, and subsequently rounded down to the nearest integer (floor function).

The \textit{fixed bin number} method breaks the relationship between image intensity and physiological meaning (if any). However, it introduces a normalising effect which may be beneficial when intensity units are arbitrary (e.g. raw MRI data and many spatial filters), and where contrast is considered important. Furthermore, as values of many features depend on the number of grey levels found within a given ROI, the use of a \textit{fixed bin number} discretisation algorithm allows for a direct comparison of feature values across multiple analysed ROIs (e.g. across different samples).

\subsubsection*{Fixed bin size\id{Q3RU}}\label{par_discr_FBS}
\textit{Fixed bin size} discretisation is conceptually simple. A new bin is assigned for every intensity interval with width $w_b$; i.e. $w_b$ is the bin width, starting at a minimum $X_{gl,min}$. The minimum intensity may be a user-set value as defined by the lower bound of the re-segmentation range, or data-driven as defined by the minimum intensity in the ROI $X_{gl,min}=\text{min} \left( X_{gl} \right)$. In all cases, the method used and/or set minimum value must be clearly reported. However, to maintain consistency between samples, we strongly recommend to always set the same minimum value for all samples as defined by the lower bound of the re-segmentation range (e.g. HU of -500 for CT, SUV of 0 for PET, etc.). In the case that no re-segmentation range may be defined due to arbitrary intensity units (e.g. raw MRI data and many spatial filters), the use of the \textit{fixed bin size} discretisation algorithm is not recommended.

The \textit{fixed bin size} method has the advantage of maintaining a direct relationship with the original intensity scale, which could be useful for functional imaging modalities such as PET.

Discretised intensities are computed as follows: 
\begin{displaymath}
X_{d,k}=\floor*{\frac{X_{gl,k}-X_{gl,min}}{w_b}} + 1
\end{displaymath}
In short, the minimum intensity $X_{gl,min}$ is subtracted from intensity $X_{gl,k}$ in voxel $k$, and then divided by the bin width $w_b$. The resulting value is subsequently rounded down to the nearest integer (floor function), and $1$ is added to arrive at the discretised intensity.

\subsubsection*{Other methods}
Many other methods and variations for discretisation exist, but are not described in detail here. \citet{Vallieres2015} described the use of \textit{intensity histogram equalisation} and \textit{Lloyd-Max} algorithms for discretisation. \textit{Intensity histogram equalisation} involves redistributing intensities so that the resulting bins contain a similar number of voxels, i.e. contrast is increased by flattening the histogram as much as possible \citep{Hall1971}. Histogram equalisation of the ROI imaging intensities can be performed before any other discretisation algorithm (e.g. FBN, FSB, etc.), and it also requires the definition of a given number of bins in the histogram to be equalised. The \textit{Lloyd-Max} algorithm is an iterative clustering method that seeks to minimise mean squared discretisation errors \citep{Max1960,Lloyd1982}.

\subsubsection*{Recommendations}
The discretisation method that leads to optimal feature inter- and intra-sample reproducibility is modality-dependent. Usage recommendations for the possible combinations of different imaging intensity definitions, re-segmentation ranges and discretisation algorithms are provided in Table \ref{table_discretisation}. Overall, the discretisation choice has a substantial impact on intensity distributions, feature values and reproducibility \citep{Hatt2015,Leijenaar2015a,vanVelden2016,Desseroit2017,Hatt2016,Shafiq-ul-Hassan2017,Altazi2017}.

% The effect of the number of bins for \textit{fixed bin number} discretisation was studied by \citet{Hatt2015}, in a large methodological study with 555 pretreatment FDG-PET images covering a range of different tumours. They found that \textit{fixed bin number} discretisation using 64 bins provides the best compromise between differentiation and resolution.

% \citet{Leijenaar2015a} compared the effect of \textit{fixed bin size} and \textit{fixed bin number} discretisation methods on texture features from FDG-PET images recorded in a cohort of 35 non-small cell lung cancer patients. They concluded that \textit{fixed bin size} may be more appropriate for inter- and intra-sample comparison of texture feature values in a clinical setting.

% In another methodological study \citet{vanVelden2016} also assessed the effect of \textit{fixed bin number} versus \textit{fixed bin size} methods in FDG-PET. They concluded that texture features from FDG-PET images had better repeatability and lower sensitivity to delineation changes using the \textit{fixed bin size} discretisation method.

% It should also be noted that several studies have used \textit{fixed bin size} for CT images, e.g. \citep{Aerts2014,VanDijk2016}. Both studies used a bin size of 25 HU. However, the authors of both studies did not report on the minimum grey level used in the discretisation process, which essentially precludes the reproducibility of their findings.

% TABLE
\begin{table}
\centering
\begin{tabu} to 0.8\textwidth {@{}X[2l]X[2c]X[1c]X[1c]@{}}

\toprule
\textbf{Imaging intensity units$^{(1)}$} & \textbf{Re-segmentation range} & \textbf{FBN$^{(2)}$} &\textbf{FBS$^{(3)}$}\\
\midrule
\multirow{3}{*}{calibrated} & $[a,b]$ & \cmark & \cmark \\
						 & $[a,\infty)$ & \cmark & \cmark \\
				   & none & \cmark & \xmark \\
                   & & & \\
arbitrary & none & \cmark & \xmark \\
\bottomrule
\end{tabu}
\caption{Recommendations for the possible combinations of different imaging intensity definitions, re-segmentation ranges and discretisation algorithms. Checkmarks (\cmark) represent recommended combinations of re-segmentation range and discretisation algorithm, whereas crossmarks (\xmark) represent non-recommended combinations. \\
\textsuperscript{(1)} PET and CT are examples of imaging modalities with \textit{calibrated} intensity units (e.g. SUV and HU, respectively), and raw MRI data of arbitrary intensity units.  \\
\textsuperscript{(2)} \textit{Fixed bin number}  (FBN) discretisation uses the actual range of intensities in the analysed ROI (re-segmented or not), and not the re-segmentation range itself (when defined).\\
\textsuperscript{(3)} \textit{Fixed bin size} (FBS) discretisation uses the lower bound of the re-segmentation range as the minimum set value. When the re-segmentation range is not or cannot be defined (e.g. arbitrary intensity units), the use of the FBS algorithm is not recommended.  
}
\label{table_discretisation}
\end{table}

%\subsection{Spatial filtering}
%Image filters can be used to highlight different image aspects, for example edges and blobs. Commonly used filters are Laplacian of Gaussian, wavelet, Laws' and Gabor filters. Applying a filter to an image volume creates a new image stack. New feature values can be calculated from the transformed image. A description of various spatial filters, their merits and implementations are beyond the scope of this work, but may require future standardisation.
%
%It should be noted that spatial filtering may require the subsequent use of grey level discretisation by normalising methods, such as \textit{fixed bin number}, and that \textit{fixed bin size} methods are likely not appropriate.

\section{Feature calculation}
Feature calculation is the final processing step where feature descriptors are used to quantify characteristics of the ROI. After calculation such features may be used as image biomarkers by relating them to physiological and medical outcomes of interest. Feature calculation is handled in full details in the next chapter.

Let us recall that the image processing steps leading to image biomarker calculations can be performed in many different ways, notably in terms of spatial filtering, segmentation, interpolation and discretisation parameters. Furthermore, it is plausible that different texture features will better quantify the characteristics of the ROI when computed using different image processing parameters. For example, a lower number of grey levels in the discretisation process (e.g. 8 or 16) may allow to better characterize the sub-regions of the ROI using \textit{grey level size zone matrix} (\hyperref[sect_glszm]{GLSZM}) features, whereas \textit{grey level co-occurence matrix} (\hyperref[sect_glcm]{GLCM}) features may be better modeled with a higher number of grey levels (e.g. 32 or 64). Overall, these possible differences opens the door to the optimization of image processing parameters for each different feature in terms of a specific objective. For the specific case of the optimization of image interpolation and discretisation prior to texture analysis, Valli\`eres \textit{et al.} \citep{Vallieres2015} have named this process \textit{texture optimization}. The authors notably suggested that the \textit{texture optimization} process could have significant influence of the prognostic capability of subsequent features. In another study\citep{Vallieres2017}, the authors constructed predictive models using textures calculated from all possible combinations of PET and CT images interpolated at four isotropic resolutions and discretised with two different algorithms and four numbers of grey levels.

\newpage
\chapter{Image features}\label{chap_image_features}
In this chapter we will describe a set of quantitative image features together with the reference values established by the IBSI. This feature set builds upon the feature sets proposed by \citet{Aerts2014} and \citet{Hatt2016}, which are themselves largely derived from earlier works. References to earlier work are provided whenever they could be identified.

Reference values were derived for each feature. A table of reference values contains the values that could be reliably reproduced, within a tolerance margin, for the reference data sets (see Chapter \ref{chap_benchmark sets}). Consensus on the validity of each reference value is also noted. Consensus can have four levels, depending on the number of teams that were able to produce the same value during the standardization process: weak ($<3$ matches), moderate ($3$ to $5$ matches), strong ($6$ to $9$ matches), and very strong ($\geq 10$ matches). If consensus on a reference value was weak or if it could not be reproduced by an absolute majority of teams, it was not considered standardized. Such features do currently not have reference values, and should not be used.

The set of features can be divided into a number of families, of which intensity-based statistical, intensity histogram-based, intensity-volume histogram-based, morphological features, local intensity, and texture matrix-based features are treated here. All texture matrices are rotationally and translationally invariant. Illumination invariance of texture matrices may be achieved by particular image post-acquisition schemes, e.g. \textit{histogram matching}. None of the texture matrices are scale invariant, a property which can be useful in many (biomedical) applications. What the presented texture matrices lack, however, is directionality in combination with rotation invariance. These may be achieved by local binary patterns and steerable filters, which however fall beyond the scope of the current work. For these and other texture features, see \citet{Depeursinge2014}.

Features are calculated on the base image, as well as images transformed using wavelet or Gabor filters). To calculate features, it is assumed that an image segmentation mask exists, which identifies the voxels located within a region of interest (ROI). The ROI itself consists of two masks, an intensity mask and a morphological mask. These masks may be identical, but not necessarily so, as described in Section \ref{ref_resegmentation}.

Several feature families require additional image processing steps before feature calculation. Notably intensity histogram and texture feature families require prior discretisation of intensities into grey level bins. Other feature families do not require discretisation before calculations. For more details on image processing, see figure \ref{figImageProc} in the previous chapter.

Below is an overview table that summarises image processing requirements for the different feature families.
\newpage
\begin{center}
\begin{longtabu}{@{}X[5,l,p] X[c,p] X[c,p] X[c,p] X[c,p]@{}}
\toprule
& & \multicolumn{2}{c}{\textbf{ROI mask}} & \\
\textbf{Feature family} & \textbf{count} & \textbf{morph.} & \textbf{int.} & \textbf{discr.}\\
\midrule
\endhead

\multicolumn{5}{r}{\textit{continued on next page}}
\endfoot

\bottomrule
\caption{Feature families and required image processing. For each feature family, the number of features in the document, the required input of a morphological (morph.) and/or intensity (int.) ROI mask, as well as the requirement of image discretisation (discr.) is provided.\\
\textsuperscript{a} The entire image volume should be available when computing local intensity features.\\
\textsuperscript{b} Image discretisation for the intensity-volume histogram is performed with finer discretisation than required for e.g. textural features.}
\endlastfoot

morphology & 29 & \cmark & \cmark & \xmark\\
local intensity & 2 & \xmark & \cmark \textsuperscript{a} & \xmark\\
intensity-based statistics & 18 & \xmark & \cmark & \xmark\\
intensity histogram & 23 & \xmark & \cmark & \cmark \\
intensity-volume histogram & 5 & \xmark & \cmark & \cmark \textsuperscript{b} \\
grey level co-occurrence matrix & 25 & \xmark & \cmark & \cmark\\
grey level run length matrix & 16 & \xmark & \cmark & \cmark\\
grey level size zone matrix & 16 & \xmark & \cmark & \cmark\\
grey level distance zone matrix & 16 & \cmark & \cmark & \cmark\\
neighbourhood grey tone difference matrix & 5 & \xmark & \cmark & \cmark\\
neighbouring grey level dependence matrix & 17 & \xmark & \cmark & \cmark\\
\end{longtabu}
\end{center}

Though image processing parameters affect feature values, three other concepts influence feature values for many features: distance, feature aggregation and distance weighting. These are described below.

\subsubsection*{Grid distances\id{MPUJ}}
\begin{figure}
\centering
   \begin{minipage}[b]{120pt}
     \centering
     \includegraphics[trim = 0 0 0 0, clip, scale=0.57]{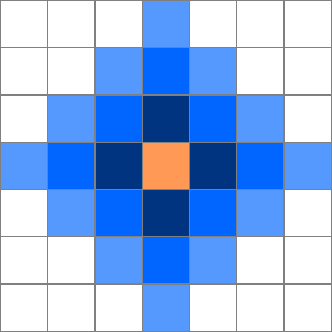}
     \subcaption{Manhattan norm}\label{fig:manhattan_distance}
     \hspace{100pt}
   \end{minipage}
   \begin{minipage}[b]{120pt}
     \centering
     \includegraphics[trim = 0 0 0 0, clip, scale=0.57]{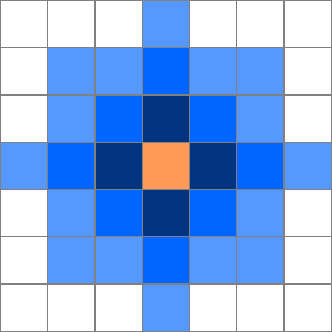}
     \subcaption{Euclidean norm}\label{fig:euclidean_distance}
     \hspace{100pt}
   \end{minipage}
   \begin{minipage}[b]{120pt}
     \centering
     \includegraphics[trim = 0 0 0 0, clip, scale=0.57]{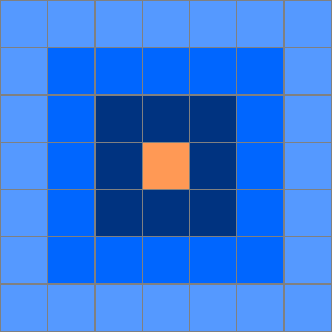}
     \subcaption{Chebyshev norm}\label{fig:chebyshev_distance}
     \hspace{100pt}
   \end{minipage}
  \caption{Grid neighbourhoods for distances up to $3$ according to Manhattan, Euclidean and Chebyshev norms. The orange pixel is considered the center pixel. Dark blue pixels have distance $\delta=1$, blue pixels $\delta\leq2$ and light blue pixels $\delta\leq3$ for the corresponding norm.
  }
  \label{fig:distance_norms}
\end{figure}

Grid distance is an important concept that is used by several feature families, particularly texture features. Grid distances can be measured in several ways. Let $\mathbf{m}=\left(m_x,m_y,m_z\right)$ be the vector from a center voxel at $\mathbf{k}=\left(k_x,k_y,k_z\right)$ to a neighbour voxel at $\mathbf{k}+\mathbf{m}$. The following norms (distances) are used:
\begin{itemize}
\item $\ell_1$ norm or \textit{Manhattan} norm (\textid{LIFZ}):
\begin{displaymath}
\norm{\mathbf{m}}_1 = |m_x| + |m_y| + |m_z|
\end{displaymath}
\item $\ell_2$ norm or \textit{Euclidean} norm (\textid{G9EV}):
\begin{displaymath}
\norm{\mathbf{m}}_2 = \sqrt{m_x^2 + m_y^2 + m_z^2}
\end{displaymath}
\item $\ell_{\infty}$ norm or \textit{Chebyshev} norm (\textid{PVMT}):
\begin{displaymath}
\norm{\mathbf{m}}_{\infty} = \text{max}(|m_x|,|m_y|,|m_z|)
\end{displaymath}
\end{itemize}
An example of how the above norms differ in practice is shown in figure \ref{fig:distance_norms}.

\subsubsection*{Feature aggregation\id{5QB6}}
Features from some families may be calculated from, e.g. slices. As a consequence, multip le values for the same feature may be computed. These different values should be combined into a single value for many common purposes. This process is referred to as feature aggregation. Feature aggregation methods depend on the family, and are detailed in the family description.

\subsubsection*{Distance weighting\id{6CK8}}
Distance weighting is not a default operation for any of the texture families, but is implemented in software such as PyRadiomics \cite{VanGriethuysen2017}. It may for example be used to put more emphasis on local intensities.

% Feature definitions
\newpage
\section[Morphological features]{Morphological features\id{HCUG}} \label{sec_morph_feat}
Morphological features describe geometric aspects of a region of interest (ROI), such as area and volume. Morphological features are based on ROI voxel representations of the volume. Three voxel representations of the volume are conceivable:
\begin{enumerate}
\item The volume is represented by a collection of voxels with each voxel taking up a certain volume (\textid{LQD8}).
\item The volume is represented by a voxel point set $\mathbf{X}_{c}$ that consists of coordinates of the voxel centers (\textid{4KW8}).
\item The volume is represented by a surface mesh (\textid{WRJH}).
\end{enumerate}
We use the second representation when the inner structure of the volume is important, and the third representation when only the outer surface structure is important. The first representation is not used outside volume approximations because it does not handle partial volume effects at the ROI edge well, and also to avoid inconsistencies in feature values introduced by mixing representations in small voxel volumes.

\subsubsection*{Mesh-based representation\id{WRJH}}
A mesh-based representation of the outer surface allows consistent evaluation of the surface volume and area independent of size. Voxel-based representations lead to partial volume effects and over-estimation of the surface area. The surface of the ROI volume is translated into a triangle mesh using a meshing algorithm. While multiple meshing algorithms exist, we suggest the use of the \textit{Marching Cubes} algorithm \citep{Lorensen1987,Lewiner2003} because of its widespread availability in different programming languages and reasonable approximation of the surface area and volume \citep{Stelldinger2007}. In practice, mesh-based feature values depend upon the meshing algorithm and small differences may occur between implementations \citep{Limkin2019-jt}.

\begin{figure}[th]
\centering
\includegraphics[scale=1.0]{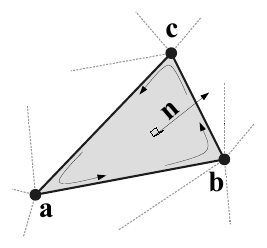}
\caption{Meshing algorithms draw faces and vertices to cover the ROI. One face, spanned by vertices $\mathbf{a}$, $\mathbf{b}$ and $\mathbf{c}$, is highlighted. Moreover, the vertices define the three edges $\mathbf{ab}=\mathbf{b}-\mathbf{a}$, $\mathbf{bc}=\mathbf{c}-\mathbf{b}$ and $\mathbf{ca}=\mathbf{a}-\mathbf{c}$. The face normal $\mathbf{n}$ is determined using the right-hand rule, and calculated as $\mathbf{n}=\left(\mathbf{ab} \times \mathbf{bc}\right) / \| \mathbf{ab} \times \mathbf{bc}\|$, i.e. the outer product of edge $\mathbf{ab}$ with edge $\mathbf{bc}$, normalised by its length.}
\label{figMorphMesh}
\end{figure}

Meshing algorithms use the ROI voxel point set $\mathbf{X}_{c}$ to create a closed mesh. Dependent on the algorithm, a parameter is required to specify where the mesh should be drawn. A default level of 0.5 times the voxel spacing is used for marching cube algorithms. Other algorithms require a so-called \textit{isovalue}, for which a value of 0.5 can be used since the ROI mask consists of $0$ and $1$ values, and we want to roughly draw the mesh half-way between voxel centers. Depending on implementation, algorithms may also require padding of the ROI mask with non-ROI ($0$) voxels to correctly estimate the mesh in places where ROI voxels would otherwise be located at the edge of the mask.

The closed mesh drawn by the meshing algorithm consists of $N_{fc}$ triangle faces spanned by $N_{vx}$ vertex points. An example triangle face is drawn in Figure \ref{figMorphMesh}. The set of vertex points is then $\mathbf{X}_{vx}$.

The calculation of the mesh volume requires that all faces have the same orientation of the face normal. Consistent orientation can be checked by the fact that in a regular, closed mesh, all edges are shared between exactly two faces. Given the edge spanned by vertices $\mathbf{a}$ and $\mathbf{b}$, the edge must be $\mathbf{ab}=\mathbf{b}-\mathbf{a}$ for one face and $\mathbf{ba}=\mathbf{a}-\mathbf{b}$ for the adjacent face. This ensures consistent application of the right-hand rule, and thus consistent orientation of the face normals. Algorithm implementations may return consistently orientated faces by default.

\subsubsection*{ROI morphological and intensity masks}
The ROI consists of a morphological and an intensity mask. The morphological mask is used to calculate many of the morphological features and to generate the voxel point set $\mathbf{X}_{c}$. Any holes within the morphological mask are understood to be the result of segmentation decisions, and thus to be intentional. The intensity mask is used to generate the voxel intensity set $\mathbf{X}_{gl}$ with corresponding point set $\mathbf{X}_{c,gl}$.

\subsubsection*{Aggregating features}
By definition, morphological features are calculated in 3D (\textid{DHQ4}), and not per slice.

\subsubsection*{Units of measurement}
By definition, morphological features are computed using the unit of length as defined in the DICOM standard, i.e. millimeter for most medical imaging modalities\footnote{DICOM PS3.3 2019a - Information Object Definitions, Section 10.7.1.3}.

If the unit of length is not defined by a standard, but is explicitly defined as meta data, this definition should be used. In this case, care should be taken that this definition is consistent across all data in the cohort.

If a feature value should be expressed as a different unit of length, e.g. cm instead of mm, such conversions should take place after computing the value using the standard units.

% VOLUME
\subsection[Mesh-based volume]{Volume (mesh)\id{RNU0}} \label{feat_morph_volume}
The mesh-based \textit{volume} $V$ is calculated from the ROI mesh as follows \citep{Zhang2001}. A tetrahedron is formed by each face $k$ and the origin. By placing the origin vertex of each tetrahedron at $(0,0,0)$, the signed volume of the tetrahedron is: 
\begin{displaymath}
V_k = \frac{\mathbf{a}\cdot\left(\mathbf{b}\times\mathbf{c}\right)}{6}
\end{displaymath}
Here $\mathbf{a}$, $\mathbf{b}$ and $\mathbf{c}$ are the vertex points of face $k$. Depending on the orientation of the normal, the signed volume may be positive or negative. Hence, the orientation of face normals should be consistent, e.g. all normals must be either pointing outward or inward. The \textit{volume} $V$ is then calculated by summing over the face volumes, and taking the absolute value:
\begin{displaymath}
F_{\mathit{morph.vol}} = V = \left|\sum_{k=1}^{N_{fc}}V_k\right|
\end{displaymath}

In positron emission tomography, the \textit{volume} of the ROI commonly receives a name related to the radioactive tracer, e.g. \textit{metabolically active tumour volume} (MATV) for \textsuperscript{18}F-FDG.

\vspace{2mm}
\begin{table}[ht]
\centering
\small
% latex table generated in R 3.6.0 by xtable 1.8-4 package
% Mon Dec 02 12:24:59 2019
\begin{tabular}{cccc}
  \toprule
{\textbf{data}} & {\textbf{value}} & {\textbf{tol.}} & {\textbf{consensus}} \\ 
  \midrule
dig. phantom &  556 &    4 & very strong \\ 
  config. A & $3.58 \times 10^{5}$ & $5 \times 10^{3}$ & very strong \\ 
  config. B & $3.58 \times 10^{5}$ & $5 \times 10^{3}$ & strong \\ 
  config. C & $3.67 \times 10^{5}$ & $6 \times 10^{3}$ & strong \\ 
  config. D & $3.67 \times 10^{5}$ & $6 \times 10^{3}$ & strong \\ 
  config. E & $3.67 \times 10^{5}$ & $6 \times 10^{3}$ & strong \\ 
   \bottomrule
\end{tabular}

\caption{Reference values for the \textit{volume (mesh)} feature.}
\end{table}
\FloatBarrier

% APPROXIMATE VOLUME
\subsection[Voxel-counting volume]{Volume (voxel counting)\id{YEKZ}} \label{feat_morph_approx_volume}
In clinical practice, volumes are commonly determined by counting voxels. For volumes consisting of a large number of voxels (1000s), the differences between \textit{voxel counting} and \textit{mesh-based} approaches are usually negligible. However for volumes with a low number of voxels (10s to 100s), \textit{voxel counting} will overestimate volume compared to the \textit{mesh-based} approach. It is therefore only used as a reference feature, and not in the calculation of other morphological features.

\noindent\textit{Voxel counting volume} is defined as:
\begin{displaymath}
F_{\mathit{morph.approx.vol}} = \sum_{k=1}^{N_v} V_k
\end{displaymath}
Here $N_v$ is the number of voxels in the morphological mask of the ROI, and $V_k$ the volume of voxel $k$.

\vspace{2mm}
\begin{table}[ht]
\centering
\small
% latex table generated in R 3.6.0 by xtable 1.8-4 package
% Mon Dec 02 12:24:59 2019
\begin{tabular}{cccc}
  \toprule
{\textbf{data}} & {\textbf{value}} & {\textbf{tol.}} & {\textbf{consensus}} \\ 
  \midrule
dig. phantom &  592 &    4 & very strong \\ 
  config. A & $3.59 \times 10^{5}$ & $5 \times 10^{3}$ & strong \\ 
  config. B & $3.58 \times 10^{5}$ & $5 \times 10^{3}$ & strong \\ 
  config. C & $3.68 \times 10^{5}$ & $6 \times 10^{3}$ & strong \\ 
  config. D & $3.68 \times 10^{5}$ & $6 \times 10^{3}$ & strong \\ 
  config. E & $3.68 \times 10^{5}$ & $6 \times 10^{3}$ & strong \\ 
   \bottomrule
\end{tabular}

\caption{Reference values for the \textit{volume (voxel counting)} feature.}
\end{table}
\FloatBarrier

% SURFACE AREA
\subsection[Mesh-based surface area]{Surface area (mesh)\id{C0JK}} \label{feat_morph_area}
The \textit{surface area} $A$ is also calculated from the ROI mesh by summing over the triangular face surface areas \citep{Aerts2014}. By definition, the area of face $k$ is:
\begin{displaymath}
A_k = \frac{|\mathbf{ab} \times \mathbf{ac}|}{2}
\end{displaymath}
As in Figure \ref{figMorphMesh}, edge $\mathbf{ab}=\mathbf{b}-\mathbf{a}$ is the vector from vertex $\mathbf{a}$ to vertex $\mathbf{b}$, and edge $\mathbf{ac}=\mathbf{c}-\mathbf{a}$ the vector from vertex $\mathbf{a}$ to vertex $\mathbf{c}$. The total \textit{surface area} $A$ is then:
\begin{displaymath}
F_{\mathit{morph.area}} = A = \sum_{k=1}^{N_{fc}} A_k
\end{displaymath}

\vspace{2mm}
\begin{table}[ht]
\centering
\small
% latex table generated in R 3.6.0 by xtable 1.8-4 package
% Mon Dec 02 12:24:58 2019
\begin{tabular}{cccc}
  \toprule
{\textbf{data}} & {\textbf{value}} & {\textbf{tol.}} & {\textbf{consensus}} \\ 
  \midrule
dig. phantom &  388 &    3 & very strong \\ 
  config. A & $3.57 \times 10^{4}$ &  300 & strong \\ 
  config. B & $3.37 \times 10^{4}$ &  300 & strong \\ 
  config. C & $3.43 \times 10^{4}$ &  400 & strong \\ 
  config. D & $3.43 \times 10^{4}$ &  400 & strong \\ 
  config. E & $3.43 \times 10^{4}$ &  400 & strong \\ 
   \bottomrule
\end{tabular}

\caption{Reference values for the \textit{surface area (mesh)} feature.}
\end{table}
\FloatBarrier

% SURFACE TO VOLUME RATIO
\subsection[Surface to volume ratio]{Surface to volume ratio\id{2PR5}}\label{feat_morph_surface_volume_ratio}
The \textit{surface to volume ratio} is given as \citep{Aerts2014}:
\begin{displaymath}
F_{\mathit{morph.av}} = \frac{A}{V}
\end{displaymath}

Note that this feature is not dimensionless.

\vspace{2mm}
\begin{table}[ht]
\centering
\small
% latex table generated in R 3.6.0 by xtable 1.8-4 package
% Mon Dec 02 12:24:58 2019
\begin{tabular}{cccc}
  \toprule
{\textbf{data}} & {\textbf{value}} & {\textbf{tol.}} & {\textbf{consensus}} \\ 
  \midrule
dig. phantom & 0.698 & 0.004 & very strong \\ 
  config. A & 0.0996 & 0.0005 & strong \\ 
  config. B & 0.0944 & 0.0005 & strong \\ 
  config. C & 0.0934 & 0.0007 & strong \\ 
  config. D & 0.0934 & 0.0007 & strong \\ 
  config. E & 0.0934 & 0.0007 & strong \\ 
   \bottomrule
\end{tabular}

\caption{Reference values for the \textit{surface to volume ratio} feature.}
\end{table}
\FloatBarrier

% COMPACTNESS 1
\subsection[Compactness 1]{Compactness 1\id{SKGS}} \label{feat_morph_comp_1}
Several features (\textit{compactness 1} and \textit{2}, \textit{spherical disproportion}, \textit{sphericity} and \textit{asphericity}) quantify the deviation of the ROI volume from a representative spheroid. All these definitions can be derived from one another. As a results these features are are highly correlated and may thus be redundant. \textit{Compactness 1} \citep{Aerts2014} is a measure for how compact, or sphere-like the volume is. It is defined as:
\begin{displaymath}
F_{\mathit{morph.comp.1}} = \frac{V}{\pi^{1/2} A^{3/2}}
\end{displaymath}
\textit{Compactness 1} is sometimes\citep{Aerts2014} defined using $A^{2/3}$ instead of $A^{3/2}$, but this does not lead to a dimensionless quantity.

\vspace{2mm}
\begin{table}[ht]
\centering
\small
% latex table generated in R 3.6.0 by xtable 1.8-4 package
% Mon Dec 02 12:24:58 2019
\begin{tabular}{cccc}
  \toprule
{\textbf{data}} & {\textbf{value}} & {\textbf{tol.}} & {\textbf{consensus}} \\ 
  \midrule
dig. phantom & 0.0411 & 0.0003 & strong \\ 
  config. A & 0.03 & 0.0001 & strong \\ 
  config. B & 0.0326 & 0.0001 & strong \\ 
  config. C & \textemdash & \textemdash & moderate \\ 
  config. D & 0.0326 & 0.0002 & strong \\ 
  config. E & 0.0326 & 0.0002 & strong \\ 
   \bottomrule
\end{tabular}

\caption{Reference values for the \textit{compactness 1} feature. An unset value (\textemdash) indicates the lack of a reference value.}
\end{table}
\FloatBarrier

% COMPACTNESS 2
\subsection[Compactness 2]{Compactness 2\id{BQWJ}} \label{feat_morph_comp_2}
Like \textit{Compactness 1}, \textit{Compactness 2} \citep{Aerts2014} quantifies how sphere-like the volume is:
\begin{displaymath}
F_{\mathit{morph.comp.2}} = 36\pi\frac{V^2}{A^3}
\end{displaymath}
By definition $F_{\mathit{morph.comp.1}} = 1/6\pi \left(F_{\mathit{morph.comp.2}}\right)^{1/2}$.

\vspace{2mm}
\begin{table}[ht]
\centering
\small
% latex table generated in R 3.6.0 by xtable 1.8-4 package
% Mon Dec 02 12:24:58 2019
\begin{tabular}{cccc}
  \toprule
{\textbf{data}} & {\textbf{value}} & {\textbf{tol.}} & {\textbf{consensus}} \\ 
  \midrule
dig. phantom & 0.599 & 0.004 & strong \\ 
  config. A & 0.319 & 0.001 & strong \\ 
  config. B & 0.377 & 0.001 & strong \\ 
  config. C & 0.378 & 0.004 & strong \\ 
  config. D & 0.378 & 0.004 & strong \\ 
  config. E & 0.378 & 0.004 & strong \\ 
   \bottomrule
\end{tabular}

\caption{Reference values for the \textit{compactness 2} feature.}
\end{table}
\FloatBarrier

% SPHERICAL DISPROPORTION
\subsection[Spherical disproportion]{Spherical disproportion \id{KRCK}} \label{feat_morph_sph_dispr}
\textit{Spherical disproportion} \citep{Aerts2014} likewise describes how sphere-like the volume is:
\begin{displaymath}
F_{\mathit{morph.sph.dispr}} = \frac{A}{4\pi R^2} = \frac{A}{\left(36\pi V^2\right)^{1/3}}
\end{displaymath}
By definition $F_{\mathit{morph.sph.dispr}} = \left(F_{\mathit{morph.comp.2}}\right)^{-1/3}$.

\vspace{2mm}
\begin{table}[ht]
\centering
\small
% latex table generated in R 3.6.0 by xtable 1.8-4 package
% Mon Dec 02 12:24:59 2019
\begin{tabular}{cccc}
  \toprule
{\textbf{data}} & {\textbf{value}} & {\textbf{tol.}} & {\textbf{consensus}} \\ 
  \midrule
dig. phantom & 1.19 & 0.01 & strong \\ 
  config. A & 1.46 & 0.01 & strong \\ 
  config. B & 1.38 & 0.01 & strong \\ 
  config. C & 1.38 & 0.01 & strong \\ 
  config. D & 1.38 & 0.01 & strong \\ 
  config. E & 1.38 & 0.01 & strong \\ 
   \bottomrule
\end{tabular}

\caption{Reference values for the \textit{spherical disproportion} feature.}
\end{table}
\FloatBarrier

% SPHERICITY
\subsection[Sphericity]{Sphericity \id{QCFX}} \label{feat_morph_sphericity}
\textit{Sphericity} \citep{Aerts2014} is a further measure to describe how sphere-like the volume is:
\begin{displaymath}
F_{\mathit{morph.sphericity}} = \frac{\left(36\pi V^2\right)^{1/3}}{A}
\end{displaymath}
By definition $F_{\mathit{morph.sphericity}} = \left(F_{\mathit{morph.comp.2}}\right)^{1/3}$.

\vspace{2mm}
\begin{table}[ht]
\centering
\small
% latex table generated in R 3.6.0 by xtable 1.8-4 package
% Mon Dec 02 12:24:59 2019
\begin{tabular}{cccc}
  \toprule
{\textbf{data}} & {\textbf{value}} & {\textbf{tol.}} & {\textbf{consensus}} \\ 
  \midrule
dig. phantom & 0.843 & 0.005 & very strong \\ 
  config. A & 0.683 & 0.001 & strong \\ 
  config. B & 0.722 & 0.001 & strong \\ 
  config. C & 0.723 & 0.003 & strong \\ 
  config. D & 0.723 & 0.003 & strong \\ 
  config. E & 0.723 & 0.003 & strong \\ 
   \bottomrule
\end{tabular}

\caption{Reference values for the \textit{sphericity} feature.}
\end{table}
\FloatBarrier

% ASPHERICITY
\subsection[Asphericity]{Asphericity \id{25C7}} \label{feat_morph_asphericity}
\textit{Asphericity} \citep{Apostolova2014} also describes how much the ROI deviates from a perfect sphere, with perfectly spherical volumes having an asphericity of 0. Asphericity is defined as:
\begin{displaymath}
F_{\mathit{morph.asphericity}}=\left(\frac{1}{36\pi}\frac{A^3}{V^2}\right)^{1/3}-1
\end{displaymath}
By definition $F_{\mathit{morph.asphericity}} = \left(F_{\mathit{morph.comp.2}}\right)^{-1/3}-1$

\vspace{2mm}
\begin{table}[ht]
\centering
\small
% latex table generated in R 3.6.0 by xtable 1.8-4 package
% Mon Dec 02 12:24:58 2019
\begin{tabular}{cccc}
  \toprule
{\textbf{data}} & {\textbf{value}} & {\textbf{tol.}} & {\textbf{consensus}} \\ 
  \midrule
dig. phantom & 0.186 & 0.001 & strong \\ 
  config. A & 0.463 & 0.002 & strong \\ 
  config. B & 0.385 & 0.001 & moderate \\ 
  config. C & 0.383 & 0.004 & strong \\ 
  config. D & 0.383 & 0.004 & strong \\ 
  config. E & 0.383 & 0.004 & strong \\ 
   \bottomrule
\end{tabular}

\caption{Reference values for the \textit{asphericity} feature.}
\end{table}
\FloatBarrier

% CENTRE OF MASS SHIFT
\subsection[Centre of mass shift]{Centre of mass shift \id{KLMA}} \label{feat_morph_centre_of_mass_shift}
The distance between the ROI volume centroid and the intensity-weighted ROI volume is an abstraction of the spatial distribution of low/high intensity regions within the ROI. Let $N_{v,m}$ be the number of voxels in the morphological mask. The ROI volume centre of mass is calculated from the morphological voxel point set $\mathbf{X}_{c}$ as follows: 
\begin{displaymath}
\overrightarrow{CoM}_{geom} = \frac{1}{N_{v,m}}\sum_{k=1}^{N_{v,m}} \vec{X}_{c,k}
\end{displaymath}
The intensity-weighted ROI volume is based on the intensity mask. The position of each voxel centre in the intensity mask voxel set $\mathbf{X}_{c,gl}$ is weighted by its corresponding intensity $\mathbf{X}_{gl}$. Therefore, with $N_{v,gl}$ the number of voxels in the intensity mask:
\begin{displaymath}
\overrightarrow{CoM}_{gl} =\frac{\sum_{k=1}^{N_{v,gl}} X_{gl,k}\vec{X}_{c,gl,k}}{\sum_{k=1}^{N_{v,gl}} X_{gl,k}}
\end{displaymath}
The distance between the two centres of mass is then:
\begin{displaymath}
F_{\mathit{morph.com}} = ||\overrightarrow{CoM}_{geom}-\overrightarrow{CoM}_{gl}||_2
\end{displaymath}

\vspace{2mm}
\begin{table}[ht]
\centering
\small
% latex table generated in R 3.6.0 by xtable 1.8-4 package
% Mon Dec 02 12:24:58 2019
\begin{tabular}{cccc}
  \toprule
{\textbf{data}} & {\textbf{value}} & {\textbf{tol.}} & {\textbf{consensus}} \\ 
  \midrule
dig. phantom & 0.672 & 0.004 & very strong \\ 
  config. A & 52.9 & 28.7 & strong \\ 
  config. B & 63.1 & 29.6 & strong \\ 
  config. C & 45.6 &  2.8 & strong \\ 
  config. D & 64.9 &  2.8 & strong \\ 
  config. E & 68.5 &  2.1 & moderate \\ 
   \bottomrule
\end{tabular}

\caption{Reference values for the \textit{centre of mass shift} feature.}
\end{table}
\FloatBarrier

% MAXIMUM 3D DIAMETER
\subsection[Maximum 3D diameter]{Maximum 3D diameter \id{L0JK}} \label{feat_morph_max_3d_diam}
The \textit{maximum 3D diameter} \citep{Aerts2014} is the distance between the two most distant vertices in the ROI mesh vertex set $\mathbf{X}_{vx}$:
\begin{displaymath}
F_{\mathit{morph.diam}} = \text{max}\left( ||\vec{X}_{vx,k_{1}}-\vec{X}_{vx,k_{2}}||_2\right),\qquad k_{1}=1,\ldots,N\qquad k_{2}=1,\ldots,N
\end{displaymath}
A practical way of determining the \textit{maximum 3D diameter} is to first construct the convex hull of the ROI mesh. The convex hull vertex set $\mathbf{X}_{vx,convex}$ is guaranteed to contain the two most distant vertices of $\mathbf{X}_{vx}$. This significantly reduces the computational cost of calculating distances between all vertices. Despite the remaining $O(n^2)$ cost of calculating distances between different vertices, $\mathbf{X}_{vx,convex}$ is usually considerably smaller in size than $\mathbf{X}_{vx}$. Moreover, the convex hull is later used for the calculation of other morphological features (\ref{feat_morph_vol_dens_conv_hull}-\ref{feat_morph_area_dens_conv_hull}).

\vspace{2mm}
\begin{table}[ht]
\centering
\small
% latex table generated in R 3.6.0 by xtable 1.8-4 package
% Mon Dec 02 12:24:58 2019
\begin{tabular}{cccc}
  \toprule
{\textbf{data}} & {\textbf{value}} & {\textbf{tol.}} & {\textbf{consensus}} \\ 
  \midrule
dig. phantom & 13.1 &  0.1 & strong \\ 
  config. A &  125 &    1 & strong \\ 
  config. B &  125 &    1 & strong \\ 
  config. C &  125 &    1 & strong \\ 
  config. D &  125 &    1 & strong \\ 
  config. E &  125 &    1 & strong \\ 
   \bottomrule
\end{tabular}

\caption{Reference values for the \textit{maximum 3D diameter} feature.}
\end{table}
\FloatBarrier

% MAJOR AXIS LENGTH
\subsection[Major axis length]{Major axis length \id{TDIC}}\label{feat_morph_pca_major}
Principal component analysis (PCA) can be used to determine the main orientation of the ROI \citep{Solomon2011}. On a three dimensional object, PCA yields three orthogonal eigenvectors $\left\lbrace e_1,e_2,e_3\right\rbrace$ and three eigenvalues $\left( \lambda_1, \lambda_2, \lambda_3\right)$.  These eigenvalues and eigenvectors geometrically describe a triaxial ellipsoid. The three eigenvectors determine the orientation of the ellipsoid, whereas the eigenvalues provide a measure of how far the ellipsoid extends along each eigenvector. Several features make use of principal component analysis, namely \textit{major}, \textit{minor} and \textit{least axis length}, \textit{elongation}, \textit{flatness}, and \textit{approximate enclosing ellipsoid volume} and {area density}.

The eigenvalues can be ordered so that $\lambda_{\mathit{major}} \geq \lambda_{\mathit{minor}}\geq \lambda_{\mathit{least}}$ correspond to the major, minor and least axes of the ellipsoid respectively. The semi-axes lengths $a$, $b$ and $c$ for the major, minor and least axes are then $2\sqrt{\lambda_{\mathit{major}}}$, $2\sqrt{\lambda_{\mathit{minor}}}$ and $2\sqrt{\lambda_{\mathit{least}}}$ respectively. The \textit{major axis length} is twice the semi-axis length $a$, determined using the largest eigenvalue obtained by PCA on the point set of voxel centers $\mathbf{X}_{c}$ \citep{Heiberger2015}:
\begin{displaymath}
F_{\mathit{morph.pca.major}} = 2a = 4\sqrt{\lambda_{\mathit{major}}}
\end{displaymath}

\vspace{2mm}
\begin{table}[ht]
\centering
\small
% latex table generated in R 3.6.0 by xtable 1.8-4 package
% Mon Dec 02 12:24:59 2019
\begin{tabular}{cccc}
  \toprule
{\textbf{data}} & {\textbf{value}} & {\textbf{tol.}} & {\textbf{consensus}} \\ 
  \midrule
dig. phantom & 11.4 &  0.1 & very strong \\ 
  config. A & 92.7 &  0.4 & very strong \\ 
  config. B & 92.6 &  0.4 & strong \\ 
  config. C & 93.3 &  0.5 & strong \\ 
  config. D & 93.3 &  0.5 & strong \\ 
  config. E & 93.3 &  0.5 & strong \\ 
   \bottomrule
\end{tabular}

\caption{Reference values for the \textit{major axis length} feature.}
\end{table}
\FloatBarrier

% MINOR AXIS LENGTH
\subsection[Minor axis length]{Minor axis length \id{P9VJ}}\label{feat_morph_pca_minor}
The \textit{minor axis length} of the ROI provides a measure of how far the volume extends along the second largest axis. The \textit{minor axis length} is twice the semi-axis length $b$, determined using the second largest eigenvalue obtained by PCA, as described in Section \ref{feat_morph_pca_major}:
\begin{displaymath}
F_{\mathit{morph.pca.minor}}= 2b =4\sqrt{\lambda_{\mathit{minor}}}
\end{displaymath}

\vspace{2mm}
\begin{table}[ht]
\centering
\small
% latex table generated in R 3.6.0 by xtable 1.8-4 package
% Mon Dec 02 12:24:59 2019
\begin{tabular}{cccc}
  \toprule
{\textbf{data}} & {\textbf{value}} & {\textbf{tol.}} & {\textbf{consensus}} \\ 
  \midrule
dig. phantom & 9.31 & 0.06 & very strong \\ 
  config. A & 81.5 &  0.4 & very strong \\ 
  config. B & 81.3 &  0.4 & strong \\ 
  config. C &   82 &  0.5 & strong \\ 
  config. D &   82 &  0.5 & strong \\ 
  config. E &   82 &  0.5 & strong \\ 
   \bottomrule
\end{tabular}

\caption{Reference values for the \textit{minor axis length} feature.}
\end{table}
\FloatBarrier

% LEAST AXIS LENGTH
\subsection[Least axis length]{Least axis length \id{7J51}}\label{feat_morph_pca_least}
The least axis is the axis along which the object is least extended. The \textit{least axis length} is twice the semi-axis length $c$, determined using the smallest eigenvalue obtained by PCA, as described in Section \ref{feat_morph_pca_major}:
\begin{displaymath}
F_{\mathit{morph.pca.least}}= 2c =4\sqrt{\lambda_{\mathit{least}}}
\end{displaymath}

\vspace{2mm}
\begin{table}[ht]
\centering
\small
% latex table generated in R 3.6.0 by xtable 1.8-4 package
% Mon Dec 02 12:24:59 2019
\begin{tabular}{cccc}
  \toprule
{\textbf{data}} & {\textbf{value}} & {\textbf{tol.}} & {\textbf{consensus}} \\ 
  \midrule
dig. phantom & 8.54 & 0.05 & very strong \\ 
  config. A & 70.1 &  0.3 & strong \\ 
  config. B & 70.2 &  0.3 & strong \\ 
  config. C & 70.9 &  0.4 & strong \\ 
  config. D & 70.9 &  0.4 & strong \\ 
  config. E & 70.9 &  0.4 & strong \\ 
   \bottomrule
\end{tabular}

\caption{Reference values for the \textit{least axis length} feature.}
\end{table}
\FloatBarrier

% ELONGATION
\subsection[Elongation]{Elongation \id{Q3CK}}\label{feat_morph_pca_elongation}
The ratio of the major and minor principal axis lengths could be viewed as the extent to which a volume is longer than it is wide, i.e. is eccentric. For computational reasons, we express \textit{elongation} as an inverse ratio. 1 is thus completely non-elongated, e.g. a sphere, and smaller values express greater elongation of the ROI volume.
\begin{displaymath}
F_{\mathit{morph.pca.elongation}} =\sqrt{\frac{\lambda_{minor}}{\lambda_{major}}}
\end{displaymath}

\vspace{2mm}
\begin{table}[ht]
\centering
\small
% latex table generated in R 3.6.0 by xtable 1.8-4 package
% Mon Dec 02 12:24:59 2019
\begin{tabular}{cccc}
  \toprule
{\textbf{data}} & {\textbf{value}} & {\textbf{tol.}} & {\textbf{consensus}} \\ 
  \midrule
dig. phantom & 0.816 & 0.005 & very strong \\ 
  config. A & 0.879 & 0.001 & strong \\ 
  config. B & 0.878 & 0.001 & strong \\ 
  config. C & 0.879 & 0.001 & strong \\ 
  config. D & 0.879 & 0.001 & strong \\ 
  config. E & 0.879 & 0.001 & strong \\ 
   \bottomrule
\end{tabular}

\caption{Reference values for the \textit{elongation} feature.}
\end{table}
\FloatBarrier

% FLATNESS
\subsection[Flatness]{Flatness\id{N17B}}\label{feat_morph_pca_flatness}
The ratio of the major and least axis lengths could be viewed as the extent to which a volume is flat relative to its length. For computational reasons, we express \textit{flatness} as an inverse ratio. 1 is thus completely non-flat, e.g. a sphere, and smaller values express objects which are increasingly flatter.
\begin{displaymath}
F_{\mathit{morph.pca.flatness}} = \sqrt{\frac{\lambda_{least}}{\lambda_{major}}}
\end{displaymath}

\vspace{2mm}
\begin{table}[ht]
\centering
\small
% latex table generated in R 3.6.0 by xtable 1.8-4 package
% Mon Dec 02 12:24:59 2019
\begin{tabular}{cccc}
  \toprule
{\textbf{data}} & {\textbf{value}} & {\textbf{tol.}} & {\textbf{consensus}} \\ 
  \midrule
dig. phantom & 0.749 & 0.005 & very strong \\ 
  config. A & 0.756 & 0.001 & strong \\ 
  config. B & 0.758 & 0.001 & strong \\ 
  config. C & 0.76 & 0.001 & strong \\ 
  config. D & 0.76 & 0.001 & strong \\ 
  config. E & 0.76 & 0.001 & strong \\ 
   \bottomrule
\end{tabular}

\caption{Reference values for the \textit{flatness} feature.}
\end{table}
\FloatBarrier

% AXIS-ALIGNED BOUNDING BOX VOLUME DENSITY
\subsection[Volume density (AABB)]{Volume density (axis-aligned bounding box)\id{PBX1}} \label{feat_morph_vol_dens_aabb}
Volume density is the fraction of the ROI volume and a comparison volume. Here the comparison volume is that of the axis-aligned bounding box (AABB) of the ROI mesh vertex set $\mathbf{X}_{vx}$ or the ROI mesh convex hull vertex set $\mathbf{X}_{vx,convex}$. Both vertex sets generate an identical bounding box, which is the smallest box enclosing the vertex set, and aligned with the axes of the reference frame.
\begin{displaymath}
F_{\mathit{morph.v.dens.aabb}} = \frac{V}{V_{\mathit{aabb}}}
\end{displaymath}

This feature is also called \textit{extent} \citep{ElNaqa2009, Solomon2011}.

\vspace{2mm}
\begin{table}[ht]
\centering
\small
% latex table generated in R 3.6.0 by xtable 1.8-4 package
% Mon Dec 02 12:24:59 2019
\begin{tabular}{cccc}
  \toprule
{\textbf{data}} & {\textbf{value}} & {\textbf{tol.}} & {\textbf{consensus}} \\ 
  \midrule
dig. phantom & 0.869 & 0.005 & strong \\ 
  config. A & 0.486 & 0.003 & strong \\ 
  config. B & 0.477 & 0.003 & strong \\ 
  config. C & 0.478 & 0.003 & strong \\ 
  config. D & 0.478 & 0.003 & strong \\ 
  config. E & 0.478 & 0.003 & strong \\ 
   \bottomrule
\end{tabular}

\caption{Reference values for the \textit{volume density (AABB)} feature.}
\end{table}
\FloatBarrier

% AXIS-ALIGNED BOUNDING BOX AREA DENSITY
\subsection[Area density (AABB)]{Area density (axis-aligned bounding box)\id{R59B}} \label{feat_morph_area_dens_aab}
Conceptually similar to the \textit{volume density (AABB)} feature, \textit{area density} considers the ratio of the ROI surface area and the surface area $A_{aabb}$ of the axis-aligned bounding box enclosing the ROI mesh vertex set $\mathbf{X}_{vx}$ \citep{VanDijk2016}. The bounding box is identical to the one used for computing the \textit{volume density (AABB)} feature. Thus:
\begin{displaymath}
F_{\mathit{morph.a.dens.aabb}} = \frac{A}{A_{aabb}}
\end{displaymath}

\vspace{2mm}
\begin{table}[ht]
\centering
\small
% latex table generated in R 3.6.0 by xtable 1.8-4 package
% Mon Dec 02 12:24:58 2019
\begin{tabular}{cccc}
  \toprule
{\textbf{data}} & {\textbf{value}} & {\textbf{tol.}} & {\textbf{consensus}} \\ 
  \midrule
dig. phantom & 0.866 & 0.005 & strong \\ 
  config. A & 0.725 & 0.003 & strong \\ 
  config. B & 0.678 & 0.003 & strong \\ 
  config. C & 0.678 & 0.003 & strong \\ 
  config. D & 0.678 & 0.003 & strong \\ 
  config. E & 0.678 & 0.003 & strong \\ 
   \bottomrule
\end{tabular}

\caption{Reference values for the \textit{area density (AABB)} feature.}
\end{table}
\FloatBarrier

% ORIENTED MINIMUM BOUNDING BOX VOLUME DENSITY
\subsection[Volume density (OMBB)]{Volume density (oriented minimum bounding box)\id{ZH1A}} \label{feat_morph_vol_dens_ombb}

\noindent\textbf{Note:} This feature currently has no reference values and should not be used.

The volume of an axis-aligned bounding box is generally not the smallest obtainable volume enclosing the ROI. By orienting the box along a different set of axes, a smaller enclosing volume may be attainable. The oriented minimum bounding box (OMBB) of the ROI mesh vertex set $\mathbf{X}_{vx}$ or $\mathbf{X}_{vx,convex}$ encloses the vertex set and has the smallest possible volume. A 3D rotating callipers technique was devised by \citet{ORourke1985} to derive the oriented minimum bounding box. Due to computational complexity of this technique, the oriented minimum bounding box is commonly approximated at lower complexity, see e.g. \citet{Barequet2001} and \citet{Chan2001}. Thus:
\begin{displaymath}
F_{\mathit{morph.v.dens.ombb}} = \frac{V}{V_{ombb}}
\end{displaymath}
Here $V_{ombb}$ is the volume of the oriented minimum bounding box.

%\input{reference_values/morph_vol_dens_ombb.txt}

% ORIENTED MINIMUM BOUNDING BOX AREA DENSITY
\subsection[Area density (OMBB)]{Area density (oriented minimum bounding box)\id{IQYR}} \label{feat_morph_area_dens_ombb}

\noindent\textbf{Note:} This feature currently has no reference values and should not be used.

\noindent The \textit{area density (OMBB)} is estimated as:
\begin{displaymath}
F_{\mathit{morph.a.dens.ombb}} = \frac{A}{A_{ombb}}
\end{displaymath}
Here $A_{ombb}$ is the surface area of the same bounding box as calculated for the \textit{volume density (OMBB)} feature.

%\input{reference_values/morph_area_dens_ombb.txt}

% APPROXIMATE ENCLOSING ELLIPSOID VOLUME DENSITY
\subsection[Volume density (AEE)]{Volume density (approximate enclosing ellipsoid)\id{6BDE}} \label{feat_morph_vol_dens_aee}
The eigenvectors and eigenvalues from PCA of the ROI voxel center point set $\mathbf{X}_{c}$ can be used to describe an ellipsoid approximating the point cloud \citep{Mazurowski2016}, i.e. the approximate enclosing ellipsoid (AEE). The volume of this ellipsoid is $V_{\mathit{aee}}=4 \pi\,a\,b\,c /3$, with $a$, $b$, and $c$ being the lengths of the ellipsoid's semi-principal axes, see Section \ref{feat_morph_pca_major}. The \textit{volume density (AEE)} is then:
\begin{displaymath}
F_{\mathit{morph.v.dens.aee}} = \frac{3V}{4\pi abc}
\end{displaymath}

\vspace{2mm}
\begin{table}[ht]
\centering
\small
% latex table generated in R 3.6.0 by xtable 1.8-4 package
% Mon Dec 02 12:24:59 2019
\begin{tabular}{cccc}
  \toprule
{\textbf{data}} & {\textbf{value}} & {\textbf{tol.}} & {\textbf{consensus}} \\ 
  \midrule
dig. phantom & 1.17 & 0.01 & moderate \\ 
  config. A & 1.29 & 0.01 & strong \\ 
  config. B & 1.29 & 0.01 & strong \\ 
  config. C & 1.29 & 0.01 & moderate \\ 
  config. D & 1.29 & 0.01 & moderate \\ 
  config. E & 1.29 & 0.01 & strong \\ 
   \bottomrule
\end{tabular}

\caption{Reference values for the \textit{volume density (AEE)} feature.}
\end{table}
\FloatBarrier

% APPROXIMATE ENCLOSING ELLIPSOID AREA DENSITY
\subsection[Area density (AEE)]{Area density (approximate enclosing ellipsoid)\id{RDD2}}\label{feat_morph_area_dens_aee}
The surface area of an ellipsoid can generally not be evaluated in an elementary form. However, it is possible to approximate the surface using an infinite series. We use the same semi-principal axes as for the \textit{volume density (AEE)} feature and define:
\begin{displaymath}
A_{\mathit{aee}}\left(a,b,c\right)=4\pi\,a\,b\sum_{\nu=0}^{\infty}\frac{\left(\alpha\,\beta\right)^{\nu}}{1-4\nu^2}P_{\nu}\left(\frac{\alpha^2+\beta^2}{2\alpha\beta}\right)
\end{displaymath}
Here $\alpha=\sqrt{1-b^2/a^2}$ and $\beta=\sqrt{1-c^2/a^2}$ are eccentricities of the ellipsoid and $P_{\nu}$ is the Legendre polynomial function for degree $\nu$. The Legendre polynomial series, though infinite, converges, and approximation may be stopped early when the incremental gains in precision become limited. By default, we stop the series after $\nu=20$.

The \textit{area density (AEE)} is then approximated as:
\begin{displaymath}
F_{\mathit{morph.a.dens.aee}} = \frac{A}{A_{\mathit{aee}}}
\end{displaymath}

\vspace{2mm}
\begin{table}[ht]
\centering
\small
% latex table generated in R 3.6.0 by xtable 1.8-4 package
% Mon Dec 02 12:24:58 2019
\begin{tabular}{cccc}
  \toprule
{\textbf{data}} & {\textbf{value}} & {\textbf{tol.}} & {\textbf{consensus}} \\ 
  \midrule
dig. phantom & 1.36 & 0.01 & moderate \\ 
  config. A & 1.71 & 0.01 & moderate \\ 
  config. B & 1.62 & 0.01 & moderate \\ 
  config. C & 1.62 & 0.01 & moderate \\ 
  config. D & 1.62 & 0.01 & moderate \\ 
  config. E & 1.62 & 0.01 & strong \\ 
   \bottomrule
\end{tabular}

\caption{Reference values for the \textit{area density (AEE)} feature.}
\end{table}
\FloatBarrier

% MINIMUM VOLUME ENCLOSING ELLIPSOID VOLUME DENSITY
\subsection[Volume density (MVEE)]{Volume density (minimum volume enclosing ellipsoid)\id{SWZ1}} \label{feat_morph_vol_dens_mvee}

\noindent\textbf{Note:} This feature currently has no reference values and should not be used.

The minimum volume enclosing ellipsoid (MVEE), unlike the approximate enclosing ellipsoid, is the smallest ellipsoid that encloses the ROI. Direct computation of the MVEE is usually unfeasible, and is therefore approximated. Various approximation algorithms have been described, e.g. \citep{Todd2007,Ahipasaoglu2015}, which are usually elaborations on Khachiyan's barycentric coordinate descent method \citep{Khachiyan1996}.

The MVEE encloses the ROI mesh vertex set $\mathbf{X}_{vx}$, and by definition $\mathbf{X}_{vx,convex}$ as well. Use of the convex mesh set $\mathbf{X}_{vx,convex}$ is recommended due to its sparsity compared to the full vertex set. The volume of the MVEE is defined by its semi-axes lengths $V_{\mathit{mvee}}=4 \pi\,a\,b\,c /3$. Then:
\begin{displaymath}
F_{\mathit{morph.v.dens.mvee}} = \frac{V}{V_{\mathit{mvee}}}
\end{displaymath}

For Khachiyan's barycentric coordinate descent-based methods we use a default tolerance $\tau=0.001$ as stopping criterion.

%\input{reference_values/morph_vol_dens_mvee.txt}

% MINIMUM VOLUME ENCLOSING ELLIPSOID AREA DENSITY
\subsection[Area density (MVEE)]{Area density (minimum volume enclosing ellipsoid)\id{BRI8}} \label{feat_morph_area_dens_mvee}

\noindent\textbf{Note:} This feature currently has no reference values and should not be used.

The surface area of an ellipsoid does not have a general elementary form, but should be approximated as noted in Section \ref{feat_morph_area_dens_aee}. Let the approximated surface area of the MVEE be $A_{\mathit{mvee}}$. Then:
\begin{displaymath}
F_{\mathit{morph.a.dens.mvee}} = \frac{A}{A_{\mathit{mvee}}}
\end{displaymath}

% \input{reference_values/morph_area_dens_mvee.txt}

% CONVEX HULL VOLUME DENSITY
\subsection[Volume density (convex hull)]{Volume density (convex hull)\id{R3ER}}\label{feat_morph_vol_dens_conv_hull}
The convex hull encloses ROI mesh vertex set $\mathbf{X}_{vx}$ and consists of the vertex set $\mathbf{X}_{vx,convex}$ and corresponding faces, see section \ref{feat_morph_max_3d_diam}. The volume of the ROI mesh convex hull set $V_{convex}$ is computed in the same way as that of the \textit{volume (mesh)} feature (\ref{feat_morph_volume}). The \textit{volume density} can then be calculated as follows:
\begin{displaymath}
F_{\mathit{morph.v.dens.conv.hull}} = \frac{V}{V_{convex}}
\end{displaymath}
This feature is also called \textit{solidity} \citep{ElNaqa2009, Solomon2011}.

\vspace{2mm}
\begin{table}[ht]
\centering
\small
% latex table generated in R 3.6.0 by xtable 1.8-4 package
% Mon Dec 02 12:24:59 2019
\begin{tabular}{cccc}
  \toprule
{\textbf{data}} & {\textbf{value}} & {\textbf{tol.}} & {\textbf{consensus}} \\ 
  \midrule
dig. phantom & 0.961 & 0.006 & strong \\ 
  config. A & 0.827 & 0.001 & moderate \\ 
  config. B & 0.829 & 0.001 & moderate \\ 
  config. C & 0.834 & 0.002 & moderate \\ 
  config. D & 0.834 & 0.002 & moderate \\ 
  config. E & 0.834 & 0.002 & moderate \\ 
   \bottomrule
\end{tabular}

\caption{Reference values for the \textit{volume density (convex hull)} feature.}
\end{table}
\FloatBarrier

% CONVEX HULL AREA DENSITY
\subsection[Area density (convex hull)]{Area density (convex hull)\id{7T7F}} \label{feat_morph_area_dens_conv_hull}
The area of the convex hull $A_{convex}$ is the sum of the areas of the faces of the convex hull, and is computed in the same way as the \textit{surface area (mesh)} feature (section \ref{feat_morph_area}). The convex hull is identical to the one used in the \textit{volume density (convex hull)} feature. Then:
\begin{displaymath}
F_{\mathit{morph.a.dens.conv.hull}} = \frac{A}{A_{convex}}
\end{displaymath}

\vspace{2mm}
\begin{table}[ht]
\centering
\small
% latex table generated in R 3.6.0 by xtable 1.8-4 package
% Mon Dec 02 12:24:58 2019
\begin{tabular}{cccc}
  \toprule
{\textbf{data}} & {\textbf{value}} & {\textbf{tol.}} & {\textbf{consensus}} \\ 
  \midrule
dig. phantom & 1.03 & 0.01 & strong \\ 
  config. A & 1.18 & 0.01 & moderate \\ 
  config. B & 1.12 & 0.01 & moderate \\ 
  config. C & 1.13 & 0.01 & moderate \\ 
  config. D & 1.13 & 0.01 & moderate \\ 
  config. E & 1.13 & 0.01 & moderate \\ 
   \bottomrule
\end{tabular}

\caption{Reference values for the \textit{area density (convex hull)} feature.}
\end{table}
\FloatBarrier

% INTEGRATED INTENSITY
\subsection[Integrated intensity]{Integrated intensity\id{99N0}} \label{feat_morph_integrated_intensity}
\textit{Integrated intensity} is the average intensity in the ROI, multiplied by the volume. In the context of \textsuperscript{18}F-FDG-PET, this feature is often called \textit{total lesion glycolysis} \citep{Vaidya2012}. Thus:
\begin{displaymath}
F_{\mathit{morph.integ.int}}=V\;\frac{1}{N_{v,gl}}\sum_{k=1}^{N_{v,gl}} X_{gl,k}
\end{displaymath}
$N_{v,gl}$ is the number of voxels in the ROI intensity mask.

\vspace{2mm}
\begin{table}[ht]
\centering
\small
% latex table generated in R 3.6.0 by xtable 1.8-4 package
% Mon Dec 02 12:24:58 2019
\begin{tabular}{cccc}
  \toprule
{\textbf{data}} & {\textbf{value}} & {\textbf{tol.}} & {\textbf{consensus}} \\ 
  \midrule
dig. phantom & $1.2 \times 10^{3}$ &   10 & moderate \\ 
  config. A & $4.81 \times 10^{6}$ & $3.2 \times 10^{5}$ & strong \\ 
  config. B & $4.12 \times 10^{6}$ & $3.2 \times 10^{5}$ & strong \\ 
  config. C & $-1.8 \times 10^{7}$ & $1.4 \times 10^{6}$ & strong \\ 
  config. D & $-8.64 \times 10^{6}$ & $1.56 \times 10^{6}$ & strong \\ 
  config. E & $-8.31 \times 10^{6}$ & $1.6 \times 10^{6}$ & strong \\ 
   \bottomrule
\end{tabular}

\caption{Reference values for the \textit{integrated intensity} feature.}
\end{table}
\FloatBarrier

% MORAN'S I INDEX
\subsection[Moran's I index]{Moran's I index\id{N365}} \label{feat_morph_moran_i}
Moran's \textit{I} index is an indicator of spatial autocorrelation \citep{Moran1950,Dale2002}. It is defined as:
\begin{displaymath}
F_{\mathit{morph.moran.i}} = \frac{N_{v,gl}}{\sum_{k_{1}=1}^{N_{v,gl}} \sum_{k_{2}=1}^{N_{v,gl}}w_{k_{1}k_{2}}} \frac{\sum_{k_{1}=1}^{N_{v,gl}}\sum_{k_{2}=1}^{N_{v,gl}} w_{k_{1}k_{2}}\left(X_{gl,k_{1}}-\mu \right) \left( X_{gl,k_{2}}-\mu \right)} {\sum_{k=1}^{N_{v,gl}} \left(X_{gl,k}-\mu \right)^2},\qquad k_{1}\neq k_{2}
\end{displaymath}
As before $N_{v,gl}$ is the number of voxels in the ROI intensity mask, $\mu$ is the mean of $\mathbf{X}_{gl}$ and $w_{k_{1}k_{2}}$ is a weight factor, equal to the inverse Euclidean distance between voxels $k_{1}$ and $k_{2}$ of the point set $\mathbf{X}_{c,gl}$ of the ROI intensity mask \citep{DaSilva2008}. Values of Moran's \textit{I} close to 1.0, 0.0 and -1.0 indicate high spatial autocorrelation, no spatial autocorrelation and high spatial anti-autocorrelation, respectively.

Note that for an ROI containing many voxels, calculating Moran's \textit{I} index may be computationally expensive due to $O(n^2)$ behaviour. Approximation by repeated subsampling of the ROI may be required to make the calculation tractable, at the cost of accuracy.

\vspace{2mm}
\begin{table}[ht]
\centering
\small
% latex table generated in R 3.6.0 by xtable 1.8-4 package
% Mon Dec 02 12:24:59 2019
\begin{tabular}{cccc}
  \toprule
{\textbf{data}} & {\textbf{value}} & {\textbf{tol.}} & {\textbf{consensus}} \\ 
  \midrule
dig. phantom & 0.0397 & 0.0003 & strong \\ 
  config. A & 0.0322 & 0.0002 & moderate \\ 
  config. B & 0.0329 & 0.0001 & moderate \\ 
  config. C & 0.0824 & 0.0003 & moderate \\ 
  config. D & 0.0622 & 0.0013 & moderate \\ 
  config. E & 0.0596 & 0.0014 & moderate \\ 
   \bottomrule
\end{tabular}

\caption{Reference values for the \textit{Moran's I index} feature.}
\end{table}
\FloatBarrier

% GEARY'S C MEASURE
\subsection[Geary's C measure]{Geary's C measure\id{NPT7}} \label{feat_morph_geary_c}
Geary's \textit{C} measure assesses spatial autocorrelation, similar to Moran's \textit{I} index \citep{Geary1954,Dale2002}. In contrast to Moran's \textit{I} index, Geary's \textit{C} measure directly assesses intensity differences between voxels and is more sensitive to local spatial autocorrelation. This measure is defined as:
\begin{displaymath}
F_{\mathit{morph.geary.c}} = \frac{N_{v,gl}-1}{2\sum_{k_{1}=1}^{N_{v,gl}} \sum_{k_{2}=1}^{N_{v,gl}}w_{k_{1}k_{2}}} \frac{\sum_{k_{1}=1}^{N_{v,gl}}\sum_{k_{2}=1}^{N_{v,gl}} w_{k_{1}k_{2}}\left(X_{gl,k_{1}}-X_{gl,k_{2}} \right)^2} {\sum_{k=1}^{N_{v,gl}} \left(X_{gl,k}-\mu \right)^2},\qquad k_{1}\neq k_{2}
\end{displaymath}
As with Moran's \textit{I}, $N_{v,gl}$ is the number of voxels in the ROI intensity mask, $\mu$ is the mean of $\mathbf{X}_{gl}$ and $w_{k_{1}k_{2}}$ is a weight factor, equal to the inverse Euclidean distance between voxels $k_{1}$ and $k_{2}$ of the ROI voxel point set $\mathbf{X}_{c,gl}$ \citep{DaSilva2008}.

Just as Moran's \textit{I}, Geary's \textit{C} measure exhibits $O(n^2)$ behaviour and an approximation scheme may be required to make calculation feasible for large ROIs.

\vspace{2mm}
\begin{table}[ht]
\centering
\small
% latex table generated in R 3.6.0 by xtable 1.8-4 package
% Mon Dec 02 12:24:58 2019
\begin{tabular}{cccc}
  \toprule
{\textbf{data}} & {\textbf{value}} & {\textbf{tol.}} & {\textbf{consensus}} \\ 
  \midrule
dig. phantom & 0.974 & 0.006 & strong \\ 
  config. A & 0.863 & 0.001 & moderate \\ 
  config. B & 0.862 & 0.001 & moderate \\ 
  config. C & 0.846 & 0.001 & moderate \\ 
  config. D & 0.851 & 0.001 & moderate \\ 
  config. E & 0.853 & 0.001 & moderate \\ 
   \bottomrule
\end{tabular}

\caption{Reference values for the \textit{Geary's C measure} feature.}
\end{table}
\FloatBarrier

\clearpage
\section[Local intensity features]{Local intensity features\id{9ST6}}
Voxel intensities within a defined neighbourhood around a center voxel are used to compute local intensity features. Unlike many other feature sets, local features do not draw solely on intensities within the ROI. While only voxels within the ROI intensity map can be used as a center voxel, the local neighbourhood draws upon all voxels regardless of being in an ROI.

\subsubsection*{Aggregating features}
\noindent By definition, local intensity features are calculated in 3D (\textid{DHQ4}), and not per slice.

% LOCAL INTENSITY PEAK
\subsection[Local intensity peak]{Local intensity peak \id{VJGA}} \label{feat_loc_int_local_peak}
The \textit{local intensity peak} was originally devised for reducing variance in determining standardised uptake values \citep{Wahl2009}. It is defined as the mean intensity in a 1 cm\textsuperscript{3} spherical volume (in world coordinates), which is centered on the voxel with the maximum intensity level in the ROI intensity mask \citep{Frings2014}.

To calculate $F_{\mathit{loc.peak.local}}$, we first select all the voxels with centers within a radius $r=\left(\frac{3}{4 \pi}\right)^{1/3} \approx 0.62$ cm of the center of the maximum intensity voxel. Subsequently, the mean intensity of the selected voxels, including the center voxel, are calculated.

In case the maximum intensity is found in multiple voxels within the ROI, \textit{local intensity peak} is calculated for each of these voxels, and the highest \textit{local intensity peak} is chosen.

\vspace{2mm}
\begin{table}[ht]
\centering
\small
% latex table generated in R 3.6.0 by xtable 1.8-4 package
% Mon Dec 02 12:24:58 2019
\begin{tabular}{cccc}
  \toprule
{\textbf{data}} & {\textbf{value}} & {\textbf{tol.}} & {\textbf{consensus}} \\ 
  \midrule
dig. phantom &  2.6 & \textemdash & strong \\ 
  config. A & $-$277 &   10 & moderate \\ 
  config. B &  178 &   10 & moderate \\ 
  config. C &  169 &   10 & moderate \\ 
  config. D &  201 &   10 & strong \\ 
  config. E &  181 &   13 & moderate \\ 
   \bottomrule
\end{tabular}

\caption{Reference values for the \textit{local intensity peak} feature.}
\end{table}
\FloatBarrier

% GLOBAL INTENSITY PEAK
\subsection[Global intensity peak]{Global intensity peak \id{0F91}} \label{feat_loc_int_global_peak}
The \textit{global intensity peak} feature $F_{\mathit{loc.peak.global}}$ is similar to the \textit{local intensity peak} \citep{Frings2014}. However, instead of calculating the mean intensity for the voxel(s) with the maximum intensity, the mean intensity is calculated within a 1 cm\textsuperscript{3} neighbourhood for every voxel in the ROI intensity mask. The highest intensity peak value is then selected.

Calculation of the \textit{global intensity peak} feature may be accelerated by construction and application of an appropriate spatial spherical mean convolution filter, due to the convolution theorem. In this case one would first construct an empty 3D filter that will fit a 1 cm\textsuperscript{3} sphere. Within this context, the filter voxels may be represented by a point set, akin to $\mathbf{X}_{c}$ in section \ref{sec_morph_feat}. Euclidean distances in world spacing between the central voxel of the filter and every remaining voxel are computed. If this distance lies within radius $r=\left(\frac{3}{4 \pi}\right)^{1/3} \approx 0.62$ the corresponding voxel receives a label $1$, and $0$ otherwise. Subsequent summation of the voxel labels yields $N_s$, the number of voxels within the 1 cm\textsuperscript{3} sphere. The filter then becomes a spherical mean filter by dividing the labels by $N_s$.

\vspace{2mm}
\begin{table}[ht]
\centering
\small
% latex table generated in R 3.6.0 by xtable 1.8-4 package
% Mon Dec 02 12:24:58 2019
\begin{tabular}{cccc}
  \toprule
{\textbf{data}} & {\textbf{value}} & {\textbf{tol.}} & {\textbf{consensus}} \\ 
  \midrule
dig. phantom &  3.1 & \textemdash & strong \\ 
  config. A &  189 &    5 & moderate \\ 
  config. B &  178 &    5 & moderate \\ 
  config. C &  180 &    5 & moderate \\ 
  config. D &  201 &    5 & moderate \\ 
  config. E &  181 &    5 & moderate \\ 
   \bottomrule
\end{tabular}

\caption{Reference values for the \textit{global intensity peak} feature.}
\end{table}
\FloatBarrier

\clearpage
\section[Intensity-based statistical features]{Intensity-based statistical features\id{UHIW}}
The intensity-based statistical features describe how intensities within the region of interest (ROI) are distributed. The features in this set do not require discretisation, and may be used to describe a continuous intensity distribution. Intensity-based statistical features are not meaningful if the intensity scale is arbitrary.

The set of intensities of the $N_v$ voxels included in the ROI intensity mask is denoted as $\mathbf{X}_{gl}=\left\lbrace X_{gl,1},X_{gl,2},\ldots,X_{gl,N_v}\right\rbrace$.

\subsubsection*{Aggregating features}
We recommend calculating intensity-based statistical features using the 3D volume (\textid{DHQ4}). An approach that computes intensity-based statistical features per slice and subsequently averages them (\textid{3IDG}) is not recommended.

% MEAN
\subsection[Mean intensity]{Mean intensity\id{Q4LE}} \label{feat_stat_mean}
The \textit{mean intensity} of $\mathbf{X}_{gl}$ is calculated as:
\begin{displaymath}
F_{\mathit{stat.mean}} = \frac{1}{N_v}\sum_{k=1}^{N_v} X_{gl,k}
\end{displaymath}

\vspace{2mm}
\begin{table}[ht]
\centering
\small
% latex table generated in R 3.6.0 by xtable 1.8-4 package
% Mon Dec 02 12:24:59 2019
\begin{tabular}{cccc}
  \toprule
{\textbf{data}} & {\textbf{value}} & {\textbf{tol.}} & {\textbf{consensus}} \\ 
  \midrule
dig. phantom & 2.15 & \textemdash & very strong \\ 
  config. A & 13.4 &  1.1 & very strong \\ 
  config. B & 11.5 &  1.1 & strong \\ 
  config. C &  $-$49 &  2.9 & very strong \\ 
  config. D & $-$23.5 &  3.9 & strong \\ 
  config. E & $-$22.6 &  4.1 & strong \\ 
   \bottomrule
\end{tabular}

\caption{Reference values for the \textit{mean} feature.}
\end{table}
\FloatBarrier

% VARIANCE
\subsection[Intensity variance]{Intensity variance\id{ECT3}} \label{feat_stat_variance}
The \textit{intensity variance} of $\mathbf{X}_{gl}$ is defined as:
\begin{displaymath}
F_{\mathit{stat.var}} = \frac{1}{N_v}\sum_{k=1}^{N_v} \left( X_{gl,k}-\mu \right)^2
\end{displaymath}

Note that we do not apply a bias correction when computing the variance.

\vspace{2mm}
\begin{table}[ht]
\centering
\small
% latex table generated in R 3.6.0 by xtable 1.8-4 package
% Mon Dec 02 12:25:00 2019
\begin{tabular}{cccc}
  \toprule
{\textbf{data}} & {\textbf{value}} & {\textbf{tol.}} & {\textbf{consensus}} \\ 
  \midrule
dig. phantom & 3.05 & \textemdash & very strong \\ 
  config. A & $1.42 \times 10^{4}$ &  400 & very strong \\ 
  config. B & $1.44 \times 10^{4}$ &  400 & strong \\ 
  config. C & $5.06 \times 10^{4}$ & $1.4 \times 10^{3}$ & very strong \\ 
  config. D & $3.28 \times 10^{4}$ & $2.1 \times 10^{3}$ & strong \\ 
  config. E & $3.51 \times 10^{4}$ & $2.2 \times 10^{3}$ & strong \\ 
   \bottomrule
\end{tabular}

\caption{Reference values for the \textit{variance} feature.}
\end{table}
\FloatBarrier

% SKEWNESS
\subsection[Intensity skewness]{Intensity skewness\id{KE2A}} \label{feat_stat_skewness}
The \textit{skewness} of the intensity distribution of $\mathbf{X}_{gl}$ is defined as:
\begin{displaymath}
F_{\mathit{stat.skew}} = \frac{\frac{1}{N_v}\sum_{k=1}^{N_v} \left( X_{gl,k}-\mu \right) ^3}{\left(\frac{1}{N_v}\sum_{k=1}^{N_v} \left( X_{gl,k}-\mu \right)^2\right)^{3/2}}
\end{displaymath}
Here $\mu=F_{\mathit{stat.mean}}$. If the \textit{intensity variance} $F_{\mathit{stat.var}} = 0$, $F_{\mathit{stat.skew}}=0$.

\vspace{2mm}
\begin{table}[ht]
\centering
\small
% latex table generated in R 3.6.0 by xtable 1.8-4 package
% Mon Dec 02 12:25:00 2019
\begin{tabular}{cccc}
  \toprule
{\textbf{data}} & {\textbf{value}} & {\textbf{tol.}} & {\textbf{consensus}} \\ 
  \midrule
dig. phantom & 1.08 & \textemdash & very strong \\ 
  config. A & $-$2.47 & 0.05 & very strong \\ 
  config. B & $-$2.49 & 0.05 & strong \\ 
  config. C & $-$2.14 & 0.05 & very strong \\ 
  config. D & $-$2.28 & 0.06 & strong \\ 
  config. E & $-$2.3 & 0.07 & strong \\ 
   \bottomrule
\end{tabular}

\caption{Reference values for the \textit{skewness} feature.}
\end{table}
\FloatBarrier

% KURTOSIS
\subsection[Excess intensity kurtosis]{(Excess) intensity kurtosis\id{IPH6}} \label{feat_stat_kurtosis}
\textit{Kurtosis}, or technically excess kurtosis, is a measure of peakedness in the intensity distribution $\mathbf{X}_{gl}$:
\begin{displaymath}
F_{\mathit{stat.kurt}} = \frac{\frac{1}{N_v}\sum_{k=1}^{N_v} \left( X_{gl,k}-\mu \right) ^4}{\left(\frac{1}{N_v}\sum_{k=1}^{N_v} \left( X_{gl,k}-\mu \right)^2\right)^{2}} -3
\end{displaymath}
Here $\mu=F_{\mathit{stat.mean}}$. Note that kurtosis is corrected by a Fisher correction of -3 to center it on 0 for normal distributions. If the \textit{intensity variance} $F_{\mathit{stat.var}} = 0$, $F_{\mathit{stat.kurt}}=0$.

\vspace{2mm}
\begin{table}[ht]
\centering
\small
% latex table generated in R 3.6.0 by xtable 1.8-4 package
% Mon Dec 02 12:24:59 2019
\begin{tabular}{cccc}
  \toprule
{\textbf{data}} & {\textbf{value}} & {\textbf{tol.}} & {\textbf{consensus}} \\ 
  \midrule
dig. phantom & $-$0.355 & \textemdash & very strong \\ 
  config. A & 5.96 & 0.24 & very strong \\ 
  config. B & 5.93 & 0.24 & strong \\ 
  config. C & 3.53 & 0.23 & very strong \\ 
  config. D & 4.35 & 0.32 & strong \\ 
  config. E & 4.44 & 0.33 & strong \\ 
   \bottomrule
\end{tabular}

\caption{Reference values for the \textit{(excess) kurtosis} feature.}
\end{table}
\FloatBarrier

% MEDIAN
\subsection[Median intensity]{Median intensity\id{Y12H}} \label{feat_stat_median}
The \textit{median intensity} $F_{\mathit{stat.median}}$ is the sample median of $\mathbf{X}_{gl}$.

\vspace{2mm}
\begin{table}[ht]
\centering
\small
% latex table generated in R 3.6.0 by xtable 1.8-4 package
% Mon Dec 02 12:24:59 2019
\begin{tabular}{cccc}
  \toprule
{\textbf{data}} & {\textbf{value}} & {\textbf{tol.}} & {\textbf{consensus}} \\ 
  \midrule
dig. phantom &    1 & \textemdash & very strong \\ 
  config. A &   46 &  0.3 & very strong \\ 
  config. B &   45 &  0.3 & strong \\ 
  config. C &   40 &  0.4 & very strong \\ 
  config. D &   42 &  0.4 & strong \\ 
  config. E &   43 &  0.5 & strong \\ 
   \bottomrule
\end{tabular}

\caption{Reference values for the \textit{median} feature.}
\end{table}
\FloatBarrier

% MINIMUM
\subsection[Minimum intensity]{Minimum intensity\id{1GSF}} \label{feat_stat_minimum}
The \textit{minimum intensity} is equal to the lowest intensity present in $\mathbf{X}_{gl}$, i.e:
\begin{displaymath}
F_{\mathit{stat.min}} = \text{min}(\mathbf{X}_{gl})
\end{displaymath}

\vspace{2mm}
\begin{table}[ht]
\centering
\small
% latex table generated in R 3.6.0 by xtable 1.8-4 package
% Mon Dec 02 12:24:59 2019
\begin{tabular}{cccc}
  \toprule
{\textbf{data}} & {\textbf{value}} & {\textbf{tol.}} & {\textbf{consensus}} \\ 
  \midrule
dig. phantom &    1 & \textemdash & very strong \\ 
  config. A & $-$500 & \textemdash & very strong \\ 
  config. B & $-$500 & \textemdash & strong \\ 
  config. C & $-$939 &    4 & very strong \\ 
  config. D & $-$724 &   12 & strong \\ 
  config. E & $-$743 &   13 & strong \\ 
   \bottomrule
\end{tabular}

\caption{Reference values for the \textit{minimum} feature.}
\end{table}
\FloatBarrier

% 10TH PERCENTILE
\subsection[10\textsuperscript{th} intensity percentile]{10\textsuperscript{th} intensity percentile\id{QG58}} \label{feat_stat_p10}
$P_{10}$ is the 10\textsuperscript{th} percentile of $\mathbf{X}_{gl}$. $P_{10}$ is a more robust alternative to the \textit{minimum intensity}.

\vspace{2mm}
\begin{table}[ht]
\centering
\small
% latex table generated in R 3.6.0 by xtable 1.8-4 package
% Mon Dec 02 12:24:59 2019
\begin{tabular}{cccc}
  \toprule
{\textbf{data}} & {\textbf{value}} & {\textbf{tol.}} & {\textbf{consensus}} \\ 
  \midrule
dig. phantom &    1 & \textemdash & very strong \\ 
  config. A & $-$129 &    8 & very strong \\ 
  config. B & $-$136 &    8 & strong \\ 
  config. C & $-$424 &   14 & very strong \\ 
  config. D & $-$304 &   20 & strong \\ 
  config. E & $-$310 &   21 & strong \\ 
   \bottomrule
\end{tabular}

\caption{Reference values for the \textit{10th percentile} feature.}
\end{table}
\FloatBarrier

% 90TH PERCENTILE
\subsection[90\textsuperscript{th} intensity percentile]{90\textsuperscript{th} intensity percentile\id{8DWT}} \label{feat_stat_p90}
$P_{90}$ is the 90\textsuperscript{th} percentile of $\mathbf{X}_{gl}$. $P_{90}$ is a more robust alternative to the \textit{maximum intensity}.

\vspace{2mm}
\begin{table}[ht]
\centering
\small
% latex table generated in R 3.6.0 by xtable 1.8-4 package
% Mon Dec 02 12:24:59 2019
\begin{tabular}{cccc}
  \toprule
{\textbf{data}} & {\textbf{value}} & {\textbf{tol.}} & {\textbf{consensus}} \\ 
  \midrule
dig. phantom &    4 & \textemdash & very strong \\ 
  config. A &   95 & \textemdash & strong \\ 
  config. B &   91 & \textemdash & strong \\ 
  config. C &   86 &  0.1 & strong \\ 
  config. D &   86 &  0.1 & strong \\ 
  config. E &   93 &  0.2 & strong \\ 
   \bottomrule
\end{tabular}

\caption{Reference values for the \textit{90th percentile} feature.}
\end{table}
\FloatBarrier

Note that the \textit{90\textsuperscript{th} intensity percentile} obtained for the digital phantom may differ from the above reference value depending on the software implementation used to compute it. For example, some implementations were found to produce a value of 4.2 instead of 4.

% MAXIMUM
\subsection[Maximum intensity]{Maximum intensity\id{84IY}} \label{feat_stat_maximum}
The \textit{maximum intensity} is equal to the highest intensity present in $\mathbf{X}_{gl}$, i.e:
\begin{displaymath}
F_{\mathit{stat.max}} = \text{max}(\mathbf{X}_{gl})
\end{displaymath}

\vspace{2mm}
\begin{table}[ht]
\centering
\small
% latex table generated in R 3.6.0 by xtable 1.8-4 package
% Mon Dec 02 12:24:59 2019
\begin{tabular}{cccc}
  \toprule
{\textbf{data}} & {\textbf{value}} & {\textbf{tol.}} & {\textbf{consensus}} \\ 
  \midrule
dig. phantom &    6 & \textemdash & very strong \\ 
  config. A &  377 &    9 & very strong \\ 
  config. B &  391 &    9 & strong \\ 
  config. C &  393 &   10 & very strong \\ 
  config. D &  521 &   22 & strong \\ 
  config. E &  345 &    9 & strong \\ 
   \bottomrule
\end{tabular}

\caption{Reference values for the \textit{maximum} feature.}
\end{table}
\FloatBarrier

% INTERQUARTILE RANGE
\subsection[Intensity interquartile range]{Intensity interquartile range\id{SALO}} \label{feat_stat_iqr}
The \textit{interquartile range} (IQR) of $\mathbf{X}_{gl}$ is defined as:
\begin{displaymath}
F_{\mathit{stat.iqr}} = P_{75}-P_{25}
\end{displaymath}
$P_{25}$ and $P_{75}$ are the 25\textsuperscript{th} and 75\textsuperscript{th} percentiles of $\mathbf{X}_{gl}$, respectively.

\vspace{2mm}
\begin{table}[ht]
\centering
\small
% latex table generated in R 3.6.0 by xtable 1.8-4 package
% Mon Dec 02 12:24:59 2019
\begin{tabular}{cccc}
  \toprule
{\textbf{data}} & {\textbf{value}} & {\textbf{tol.}} & {\textbf{consensus}} \\ 
  \midrule
dig. phantom &    3 & \textemdash & very strong \\ 
  config. A &   56 &  0.5 & very strong \\ 
  config. B &   52 &  0.5 & strong \\ 
  config. C &   67 &  4.9 & very strong \\ 
  config. D &   57 &  4.1 & strong \\ 
  config. E &   62 &  3.5 & strong \\ 
   \bottomrule
\end{tabular}

\caption{Reference values for the \textit{interquartile range} feature.}
\end{table}
\FloatBarrier

% INTENSITY RANGE
\subsection[Intensity range]{Intensity range\id{2OJQ}} \label{feat_stat_range}
The \textit{intensity range} is defined as:
\begin{displaymath}
F_{\mathit{stat.range}} = \text{max}(\mathbf{X}_{gl}) - \text{min}(\mathbf{X}_{gl})
\end{displaymath}

\vspace{2mm}
\begin{table}[ht]
\centering
\small
% latex table generated in R 3.6.0 by xtable 1.8-4 package
% Mon Dec 02 12:24:59 2019
\begin{tabular}{cccc}
  \toprule
{\textbf{data}} & {\textbf{value}} & {\textbf{tol.}} & {\textbf{consensus}} \\ 
  \midrule
dig. phantom &    5 & \textemdash & very strong \\ 
  config. A &  877 &    9 & very strong \\ 
  config. B &  891 &    9 & strong \\ 
  config. C & $1.33 \times 10^{3}$ &   20 & very strong \\ 
  config. D & $1.24 \times 10^{3}$ &   40 & strong \\ 
  config. E & $1.09 \times 10^{3}$ &   30 & strong \\ 
   \bottomrule
\end{tabular}

\caption{Reference values for the \textit{range} feature.}
\end{table}
\FloatBarrier

% MEAN ABSOLUTE DEVIATION
\subsection[Intensity-based mean absolute deviation]{Intensity-based mean absolute deviation\id{4FUA}} \label{feat_stat_mean_absolute_dev}
\textit{Mean absolute deviation} is a measure of dispersion from the mean of $\mathbf{X}_{gl}$:
\begin{displaymath}
F_{\mathit{stat.mad}} = \frac{1}{N_v}\sum_{k=1}^{N_v} \left|X_{gl,k}-\mu\right|
\end{displaymath}
Here $\mu=F_{\mathit{stat.mean}}$.

\vspace{2mm}
\begin{table}[ht]
\centering
\small
% latex table generated in R 3.6.0 by xtable 1.8-4 package
% Mon Dec 02 12:24:59 2019
\begin{tabular}{cccc}
  \toprule
{\textbf{data}} & {\textbf{value}} & {\textbf{tol.}} & {\textbf{consensus}} \\ 
  \midrule
dig. phantom & 1.55 & \textemdash & very strong \\ 
  config. A & 73.6 &  1.4 & very strong \\ 
  config. B & 74.4 &  1.4 & strong \\ 
  config. C &  158 &    4 & very strong \\ 
  config. D &  123 &    6 & strong \\ 
  config. E &  125 &    6 & strong \\ 
   \bottomrule
\end{tabular}

\caption{Reference values for the \textit{mean absolute deviation} feature.}
\end{table}
\FloatBarrier

% ROBUST MEAN ABSOLUTE DEVIATION
\subsection[Intensity-based robust mean absolute deviation]{Intensity-based robust mean absolute deviation\id{1128}} \label{feat_stat_robust_mean_absolute_dev}
The \textit{intensity-based mean absolute deviation} feature may be influenced by outliers. To increase robustness, the set of intensities can be restricted to those which lie closer to the center of the distribution. Let
\begin{displaymath}
\mathbf{X}_{gl,10-90}= \left\lbrace x \in \mathbf{X}_{gl} | P_{10}\left(\mathbf{X}_{gl}\right)\leq x \leq P_{90}\left(\mathbf{X}_{gl}\right)\right\rbrace
\end{displaymath}
Then $\mathbf{X}_{gl,10-90}$ is the set of $N_{v,10-90}\leq N_v$ voxels in $\mathbf{X}_{gl}$ whose intensities fall in the interval bounded by the 10\textsuperscript{th} and 90\textsuperscript{th} percentiles of $\mathbf{X}_{gl}$. The robust mean absolute deviation is then:
\begin{displaymath}
F_{\mathit{stat.rmad}} = \frac{1}{N_{v,10-90}}\sum_{k=1}^{N_{v,10-90}} \left|X_{gl,10-90,k}-\overline{X}_{gl,10-90}\right|
\end{displaymath}
$\overline{X}_{gl,10-90}$ denotes the sample mean of $\mathbf{X_{gl,10-90}}$.

\vspace{2mm}
\begin{table}[ht]
\centering
\small
% latex table generated in R 3.6.0 by xtable 1.8-4 package
% Mon Dec 02 12:24:59 2019
\begin{tabular}{cccc}
  \toprule
{\textbf{data}} & {\textbf{value}} & {\textbf{tol.}} & {\textbf{consensus}} \\ 
  \midrule
dig. phantom & 1.11 & \textemdash & very strong \\ 
  config. A & 27.7 &  0.8 & very strong \\ 
  config. B & 27.3 &  0.8 & strong \\ 
  config. C & 66.8 &  3.5 & very strong \\ 
  config. D & 46.8 &  3.6 & strong \\ 
  config. E & 46.5 &  3.7 & strong \\ 
   \bottomrule
\end{tabular}

\caption{Reference values for the \textit{robust mean absolute deviation} feature.}
\end{table}
\FloatBarrier

% MEDIAN ABSOLUTE DEVIATION
\subsection[Intensity-based median absolute deviation]{Intensity-based median absolute deviation\id{N72L}} \label{feat_stat_median_absolute_dev}
\textit{Median absolute deviation} is similar in concept to the \textit{intensity-based mean absolute deviation}, but measures dispersion from the median intensity instead of the mean intensity. Thus:
\begin{displaymath}
F_{\mathit{stat.medad}} = \frac{1}{N_v}\sum_{k=1}^{N_v} \left| X_{gl,k}-M\right|
\end{displaymath}
Here, median $M = F_{\mathit{stat.median}}$.

\vspace{2mm}
\begin{table}[ht]
\centering
\small
% latex table generated in R 3.6.0 by xtable 1.8-4 package
% Mon Dec 02 12:24:59 2019
\begin{tabular}{cccc}
  \toprule
{\textbf{data}} & {\textbf{value}} & {\textbf{tol.}} & {\textbf{consensus}} \\ 
  \midrule
dig. phantom & 1.15 & \textemdash & very strong \\ 
  config. A & 64.3 &    1 & strong \\ 
  config. B & 63.8 &    1 & strong \\ 
  config. C &  119 &    4 & strong \\ 
  config. D & 94.7 &  3.8 & strong \\ 
  config. E & 97.9 &  3.9 & strong \\ 
   \bottomrule
\end{tabular}

\caption{Reference values for the \textit{median absolute deviation} feature.}
\end{table}
\FloatBarrier

% COEFFICIENT OF VARIATION
\subsection[Intensity-based coefficient of variation]{Intensity-based coefficient of variation\id{7TET}} \label{feat_stat_coef_of_variation}
The \textit{coefficient of variation} measures the dispersion of $\mathbf{X}_{gl}$. It is defined as:
\begin{displaymath}
F_{\mathit{stat.cov}}=\frac{\sigma}{\mu}
\end{displaymath}
Here $\sigma={F_{\mathit{stat.var}}}^{1/2}$ and $\mu=F_{\mathit{stat.mean}}$ are the standard deviation and mean of the intensity distribution, respectively.

\vspace{2mm}
\begin{table}[ht]
\centering
\small
% latex table generated in R 3.6.0 by xtable 1.8-4 package
% Mon Dec 02 12:24:59 2019
\begin{tabular}{cccc}
  \toprule
{\textbf{data}} & {\textbf{value}} & {\textbf{tol.}} & {\textbf{consensus}} \\ 
  \midrule
dig. phantom & 0.812 & \textemdash & very strong \\ 
  config. A &  8.9 & 4.98 & strong \\ 
  config. B & 10.4 &  5.2 & strong \\ 
  config. C & $-$4.59 & 0.29 & strong \\ 
  config. D & $-$7.7 & 1.01 & strong \\ 
  config. E & $-$8.28 & 0.95 & strong \\ 
   \bottomrule
\end{tabular}

\caption{Reference values for the \textit{coefficient of variation} feature.}
\end{table}
\FloatBarrier

% QUARTILE COEFFICIENT OF DISPERSION
\subsection[Intensity-based quartile coefficient of dispersion]{Intensity-based quartile coefficient of dispersion\id{9S40}} \label{feat_stat_quartile_coef_dispersion}
The \textit{quartile coefficient of dispersion} is a more robust alternative to the \textit{intensity-based coefficient of variance}. It is defined as:
\begin{displaymath}
F_{\mathit{stat.qcod}} = \frac{P_{75}-P_{25}}{P_{75}+P_{25}}
\end{displaymath}
$P_{25}$ and $P_{75}$ are the 25\textsuperscript{th} and 75\textsuperscript{th} percentile of $\mathbf{X}_{gl}$, respectively.

\vspace{2mm}
\begin{table}[ht]
\centering
\small
% latex table generated in R 3.6.0 by xtable 1.8-4 package
% Mon Dec 02 12:24:59 2019
\begin{tabular}{cccc}
  \toprule
{\textbf{data}} & {\textbf{value}} & {\textbf{tol.}} & {\textbf{consensus}} \\ 
  \midrule
dig. phantom &  0.6 & \textemdash & very strong \\ 
  config. A & 0.636 & 0.008 & strong \\ 
  config. B & 0.591 & 0.008 & strong \\ 
  config. C & 1.03 &  0.4 & strong \\ 
  config. D & 0.74 & 0.011 & strong \\ 
  config. E & 0.795 & 0.337 & strong \\ 
   \bottomrule
\end{tabular}

\caption{Reference values for the \textit{quartile coefficient of dispersion} feature.}
\end{table}
\FloatBarrier

% ENERGY
\subsection[Intensity-based energy]{Intensity-based energy\id{N8CA}} \label{feat_stat_energy}
The \textit{energy} \citep{Aerts2014} of $\mathbf{X}_{gl}$ is defined as:
\begin{displaymath}
F_{\mathit{stat.energy}} = \sum_{k=1}^{N_v} X_{gl,k}^2
\end{displaymath}

\vspace{2mm}
\begin{table}[ht]
\centering
\small
% latex table generated in R 3.6.0 by xtable 1.8-4 package
% Mon Dec 02 12:24:59 2019
\begin{tabular}{cccc}
  \toprule
{\textbf{data}} & {\textbf{value}} & {\textbf{tol.}} & {\textbf{consensus}} \\ 
  \midrule
dig. phantom &  567 & \textemdash & very strong \\ 
  config. A & $1.65 \times 10^{9}$ & $2 \times 10^{7}$ & very strong \\ 
  config. B & $3.98 \times 10^{8}$ & $1.1 \times 10^{7}$ & strong \\ 
  config. C & $2.44 \times 10^{9}$ & $1.2 \times 10^{8}$ & strong \\ 
  config. D & $1.48 \times 10^{9}$ & $1.4 \times 10^{8}$ & strong \\ 
  config. E & $1.58 \times 10^{9}$ & $1.4 \times 10^{8}$ & strong \\ 
   \bottomrule
\end{tabular}

\caption{Reference values for the \textit{energy} feature.}
\end{table}
\FloatBarrier

% ROOT MEAN SQUARE
\subsection[Root mean square intensity]{Root mean square intensity\id{5ZWQ}} \label{feat_stat_root_mean_square}
The \textit{root mean square intensity} feature \citep{Aerts2014}, which is also called the \textit{quadratic mean}, of $\mathbf{X}_{gl}$ is defined as:
\begin{displaymath}
F_{\mathit{stat.rms}} = \sqrt{\frac{\sum_{k=1}^{N_v} X_{gl,k}^2}{N_v}}
\end{displaymath}

\vspace{2mm}
\begin{table}[ht]
\centering
\small
% latex table generated in R 3.6.0 by xtable 1.8-4 package
% Mon Dec 02 12:24:59 2019
\begin{tabular}{cccc}
  \toprule
{\textbf{data}} & {\textbf{value}} & {\textbf{tol.}} & {\textbf{consensus}} \\ 
  \midrule
dig. phantom & 2.77 & \textemdash & very strong \\ 
  config. A &  120 &    2 & very strong \\ 
  config. B &  121 &    2 & strong \\ 
  config. C &  230 &    4 & strong \\ 
  config. D &  183 &    7 & strong \\ 
  config. E &  189 &    7 & strong \\ 
   \bottomrule
\end{tabular}

\caption{Reference values for the \textit{root mean square} feature.}
\end{table}
\FloatBarrier

\clearpage
\section[Intensity histogram features]{Intensity histogram features\id{ZVCW}}
An intensity histogram is generated by discretising the original intensity distribution $\mathbf{X}_{gl}$ into intensity bins. Approaches to discretisation are described in Section \ref{discretisation}.

Let $\mathbf{X}_{d}=\left\lbrace X_{d,1},X_{d,2},\ldots,X_{d,N_v}\right\rbrace$ be the set of $N_g$ discretised intensities of the $N_v$ voxels in the ROI intensity mask. Let $\mathbf{H}=\left\lbrace n_1, n_2,\ldots, n_{N_g}\right\rbrace$ be the histogram with frequency count $n_i$ of each discretised intensity $i$ in $\mathbf{X}_{d}$. The occurrence probability $p_i$ for each discretised intensity $i$ is then approximated as $p_i=n_i/N_v$.

\subsubsection*{Aggregating features}
We recommend calculating intensity histogram features using the 3D volume (\textid{DHQ4}). An approach that computes features per slice and subsequently averages (\textid{3IDG}) is not recommended.

% MEAN
\subsection[Mean discretised intensity]{Mean discretised intensity\id{X6K6}}\label{feat_int_hist_mean}
The \textit{mean} \citep{Aerts2014} of $\mathbf{X}_{d}$ is calculated as:
\begin{displaymath}
F_{\mathit{ih.mean}} = \frac{1}{N_v}\sum_{k=1}^{N_v} X_{d,k}
\end{displaymath}
An equivalent definition is:
\begin{displaymath}
F_{\mathit{ih.mean}} = \sum_{i=1}^{N_g}i\,p_i
\end{displaymath}

\vspace{2mm}
\begin{table}[ht]
\centering
\small
% latex table generated in R 3.6.0 by xtable 1.8-4 package
% Mon Dec 02 12:24:57 2019
\begin{tabular}{cccc}
  \toprule
{\textbf{data}} & {\textbf{value}} & {\textbf{tol.}} & {\textbf{consensus}} \\ 
  \midrule
dig. phantom & 2.15 & \textemdash & very strong \\ 
  config. A & 21.1 &  0.1 & strong \\ 
  config. B & 18.9 &  0.3 & strong \\ 
  config. C & 38.6 &  0.2 & strong \\ 
  config. D & 18.5 &  0.5 & strong \\ 
  config. E & 21.7 &  0.3 & strong \\ 
   \bottomrule
\end{tabular}

\caption{Reference values for the \textit{mean} feature.}
\end{table}
\FloatBarrier

% VARIANCE
\subsection[Discretised intensity variance]{Discretised intensity variance\id{CH89}}\label{feat_int_hist_variance}
The \textit{variance} \citep{Aerts2014} of $\mathbf{X}_{d}$ is defined as:
\begin{displaymath}
F_{\mathit{ih.var}} = \frac{1}{N_v}\sum_{k=1}^{N_v} \left( X_{d,k}-\mu \right)^2
\end{displaymath}
Here $\mu=F_{\mathit{ih.mean}}$. This definition is equivalent to:
\begin{displaymath}
F_{\mathit{ih.var}} = \sum_{i=1}^{N_g}\left(i-\mu\right)^2 p_i
\end{displaymath}

Note that no bias-correction is applied when computing the variance.

\vspace{2mm}
\begin{table}[ht]
\centering
\small
% latex table generated in R 3.6.0 by xtable 1.8-4 package
% Mon Dec 02 12:24:58 2019
\begin{tabular}{cccc}
  \toprule
{\textbf{data}} & {\textbf{value}} & {\textbf{tol.}} & {\textbf{consensus}} \\ 
  \midrule
dig. phantom & 3.05 & \textemdash & strong \\ 
  config. A & 22.8 &  0.6 & strong \\ 
  config. B & 18.7 &  0.2 & strong \\ 
  config. C & 81.1 &  2.1 & strong \\ 
  config. D & 21.7 &  0.4 & strong \\ 
  config. E & 30.4 &  0.8 & strong \\ 
   \bottomrule
\end{tabular}

\caption{Reference values for the \textit{variance} feature.}
\end{table}
\FloatBarrier

% SKEWNESS
\subsection[Discretised intensity skewness]{Discretised intensity skewness\id{88K1}}\label{feat_int_hist_skewness}
The \textit{skewness} \citep{Aerts2014} of $\mathbf{X}_{d}$ is defined as:
\begin{displaymath}
F_{\mathit{ih.skew}} = \frac{\frac{1}{N_v}\sum_{k=1}^{N_v} \left( X_{d,k}-\mu \right) ^3}{\left(\frac{1}{N_v}\sum_{k=1}^{N_v} \left( X_{d,k}-\mu \right)^2\right)^{3/2}}
\end{displaymath}
Here $\mu=F_{\mathit{ih.mean}}$. This definition is equivalent to:
\begin{displaymath}
F_{\mathit{ih.skew}} = \frac{\sum_{i=1}^{N_g}\left(i-\mu\right)^3 p_i}{\left(\sum_{i=1}^{N_g}\left(i-\mu\right)^2 p_i\right)^{3/2}} 
\end{displaymath}
If the \textit{discretised intensity variance} $F_{\mathit{ih.var}} = 0$, $F_{\mathit{ih.skew}}=0$.

\vspace{2mm}
\begin{table}[ht]
\centering
\small
% latex table generated in R 3.6.0 by xtable 1.8-4 package
% Mon Dec 02 12:24:58 2019
\begin{tabular}{cccc}
  \toprule
{\textbf{data}} & {\textbf{value}} & {\textbf{tol.}} & {\textbf{consensus}} \\ 
  \midrule
dig. phantom & 1.08 & \textemdash & very strong \\ 
  config. A & $-$2.46 & 0.05 & strong \\ 
  config. B & $-$2.47 & 0.05 & strong \\ 
  config. C & $-$2.14 & 0.05 & strong \\ 
  config. D & $-$2.27 & 0.06 & strong \\ 
  config. E & $-$2.29 & 0.07 & strong \\ 
   \bottomrule
\end{tabular}

\caption{Reference values for the \textit{skewness} feature.}
\end{table}
\FloatBarrier

% KURTOSIS
\subsection[Excess discretised intensity kurtosis]{(Excess) discretised intensity kurtosis\id{C3I7}}\label{feat_int_hist_kurtosis}
\textit{Kurtosis} \citep{Aerts2014}, or technically excess kurtosis,  measures the peakedness of the $\mathbf{X}_{d}$ distribution:
\begin{displaymath}
F_{\mathit{ih.kurt}} = \frac{\frac{1}{N_v}\sum_{k=1}^{N_v} \left( X_{d,k}-\mu \right) ^4}{\left(\frac{1}{N_v}\sum_{k=1}^{N_v} \left( X_{d,k}-\mu \right)^2\right)^{2}} -3
\end{displaymath}
Here $\mu=F_{\mathit{ih.mean}}$. An alternative, but equivalent, definition is:
\begin{displaymath}
F_{\mathit{ih.kurt}} = \frac{\sum_{i=1}^{N_g}\left(i-\mu\right)^4 p_i}{\left(\sum_{i=1}^{N_g}\left(i-\mu\right)^2 p_i\right)^{2}} -3
\end{displaymath}
Note that kurtosis is corrected by a Fisher correction of -3 to center kurtosis on 0 for normal distributions. If the \textit{discretised intensity variance} $F_{\mathit{ih.var}} = 0$, $F_{\mathit{ih.kurt}}=0$.

\vspace{2mm}
\begin{table}[ht]
\centering
\small
% latex table generated in R 3.6.0 by xtable 1.8-4 package
% Mon Dec 02 12:24:57 2019
\begin{tabular}{cccc}
  \toprule
{\textbf{data}} & {\textbf{value}} & {\textbf{tol.}} & {\textbf{consensus}} \\ 
  \midrule
dig. phantom & $-$0.355 & \textemdash & very strong \\ 
  config. A &  5.9 & 0.24 & strong \\ 
  config. B & 5.84 & 0.24 & strong \\ 
  config. C & 3.52 & 0.23 & strong \\ 
  config. D & 4.31 & 0.32 & strong \\ 
  config. E &  4.4 & 0.33 & strong \\ 
   \bottomrule
\end{tabular}

\caption{Reference values for the \textit{(excess) kurtosis} feature.}
\end{table}
\FloatBarrier

% MEDIAN
\subsection[Median discretised intensity]{Median discretised intensity\id{WIFQ}}\label{feat_int_hist_median}
The \textit{median} $F_{\mathit{ih.median}}$ is the sample median of $\mathbf{X}_{d}$ \citep{Aerts2014}.

\vspace{2mm}
\begin{table}[ht]
\centering
\small
% latex table generated in R 3.6.0 by xtable 1.8-4 package
% Mon Dec 02 12:24:57 2019
\begin{tabular}{cccc}
  \toprule
{\textbf{data}} & {\textbf{value}} & {\textbf{tol.}} & {\textbf{consensus}} \\ 
  \midrule
dig. phantom &    1 & \textemdash & very strong \\ 
  config. A &   22 & \textemdash & strong \\ 
  config. B &   20 &  0.3 & strong \\ 
  config. C &   42 & \textemdash & strong \\ 
  config. D &   20 &  0.5 & strong \\ 
  config. E &   24 &  0.2 & strong \\ 
   \bottomrule
\end{tabular}

\caption{Reference values for the \textit{median} feature.}
\end{table}
\FloatBarrier

% MINIMUM
\subsection[Minimum discretised intensity]{Minimum discretised intensity\id{1PR8}}\label{feat_int_hist_minimum}
The \textit{minimum discretised intensity} \citep{Aerts2014} is equal to the lowest discretised intensity present in $\mathbf{X}_{d}$, i.e.:
\begin{displaymath}
F_{\mathit{ih.min}} = \text{min}(\mathbf{X}_{d})
\end{displaymath}

For \textit{fixed bin number} discretisation $F_{\mathit{ih.min}}=1$ by definition, but $F_{\mathit{ih.min}}>1$ is possible for \textit{fixed bin size} discretisation.

\vspace{2mm}
\begin{table}[ht]
\centering
\small
% latex table generated in R 3.6.0 by xtable 1.8-4 package
% Mon Dec 02 12:24:57 2019
\begin{tabular}{cccc}
  \toprule
{\textbf{data}} & {\textbf{value}} & {\textbf{tol.}} & {\textbf{consensus}} \\ 
  \midrule
dig. phantom &    1 & \textemdash & very strong \\ 
  config. A &    1 & \textemdash & strong \\ 
  config. B &    1 & \textemdash & strong \\ 
  config. C &    3 & 0.16 & strong \\ 
  config. D &    1 & \textemdash & strong \\ 
  config. E &    1 & \textemdash & strong \\ 
   \bottomrule
\end{tabular}

\caption{Reference values for the \textit{minimum} feature.}
\end{table}
\FloatBarrier

% 10TH PERCENTILE
\subsection[10\textsuperscript{th} discretised intensity percentile]{10\textsuperscript{th} discretised intensity  percentile\id{GPMT}}\label{feat_int_hist_p10}
$P_{10}$ is the 10\textsuperscript{th} percentile of $\mathbf{X}_{d}$.

\vspace{2mm}
\begin{table}[ht]
\centering
\small
% latex table generated in R 3.6.0 by xtable 1.8-4 package
% Mon Dec 02 12:24:58 2019
\begin{tabular}{cccc}
  \toprule
{\textbf{data}} & {\textbf{value}} & {\textbf{tol.}} & {\textbf{consensus}} \\ 
  \midrule
dig. phantom &    1 & \textemdash & very strong \\ 
  config. A &   15 &  0.4 & strong \\ 
  config. B &   14 &  0.5 & strong \\ 
  config. C &   24 &  0.7 & strong \\ 
  config. D &   11 &  0.7 & strong \\ 
  config. E &   13 &  0.7 & strong \\ 
   \bottomrule
\end{tabular}

\caption{Reference values for the \textit{10th percentile} feature.}
\end{table}
\FloatBarrier

% 90TH PERCENTILE
\subsection[90\textsuperscript{th} discretised intensity  percentile]{90\textsuperscript{th} discretised intensity  percentile\id{OZ0C}}\label{feat_int_hist_p90}
$P_{90}$ is the 90\textsuperscript{th} percentile of $\mathbf{X}_{d}$ and is defined as $F_{\mathit{ih.P90}}$.

\vspace{2mm}
\begin{table}[ht]
\centering
\small
% latex table generated in R 3.6.0 by xtable 1.8-4 package
% Mon Dec 02 12:24:58 2019
\begin{tabular}{cccc}
  \toprule
{\textbf{data}} & {\textbf{value}} & {\textbf{tol.}} & {\textbf{consensus}} \\ 
  \midrule
dig. phantom &    4 & \textemdash & strong \\ 
  config. A &   24 & \textemdash & strong \\ 
  config. B &   22 &  0.3 & strong \\ 
  config. C &   44 & \textemdash & strong \\ 
  config. D &   21 &  0.5 & strong \\ 
  config. E &   25 &  0.2 & strong \\ 
   \bottomrule
\end{tabular}

\caption{Reference values for the \textit{90th percentile} feature.}
\end{table}
\FloatBarrier

Note that the \textit{90\textsuperscript{th} discretised intensity percentile} obtained for the digital phantom may differ from the above reference value depending on the software implementation used to compute it. For example, some implementations were found to produce a value of 4.2 instead of 4 for this feature.

% MAXIMUM
\subsection[Maximum discretised intensity]{Maximum discretised intensity\id{3NCY}}\label{feat_int_hist_maximum}
The \textit{maximum discretised intensity} \citep{Aerts2014} is equal to the highest discretised intensity present in $\mathbf{X}_{d}$, i.e.:
\begin{displaymath}
F_{\mathit{ih.max}} = \text{max}(\mathbf{X}_{d})
\end{displaymath}

By definition, $F_{\mathit{ih.max}}=N_g$.

\vspace{2mm}
\begin{table}[ht]
\centering
\small
% latex table generated in R 3.6.0 by xtable 1.8-4 package
% Mon Dec 02 12:24:57 2019
\begin{tabular}{cccc}
  \toprule
{\textbf{data}} & {\textbf{value}} & {\textbf{tol.}} & {\textbf{consensus}} \\ 
  \midrule
dig. phantom &    6 & \textemdash & very strong \\ 
  config. A &   36 &  0.4 & strong \\ 
  config. B &   32 & \textemdash & strong \\ 
  config. C &   56 &  0.5 & strong \\ 
  config. D &   32 & \textemdash & strong \\ 
  config. E &   32 & \textemdash & strong \\ 
   \bottomrule
\end{tabular}

\caption{Reference values for the \textit{maximum} feature.}
\end{table}
\FloatBarrier

% MODE
\subsection[Intensity histogram mode]{Intensity histogram mode\id{AMMC}}\label{feat_int_hist_mode}
The \textit{mode} of $\mathbf{X}_{d}$ $F_{\mathit{ih.mode}}$ is the most common discretised intensity present, i.e. the value $i$ for with the highest count $n_i$. The mode may not be uniquely defined. When the highest count is found in multiple bins, the value $i$ of the bin closest to the \textit{mean discretised intensity} is chosen as \textit{intensity histogram mode}. In pathological cases with two such bins equidistant to the mean, the bin to the left of the mean is selected.

\vspace{2mm}
\begin{table}[ht]
\centering
\small
% latex table generated in R 3.6.0 by xtable 1.8-4 package
% Mon Dec 02 12:24:57 2019
\begin{tabular}{cccc}
  \toprule
{\textbf{data}} & {\textbf{value}} & {\textbf{tol.}} & {\textbf{consensus}} \\ 
  \midrule
dig. phantom &    1 & \textemdash & very strong \\ 
  config. A &   23 & \textemdash & strong \\ 
  config. B &   20 &  0.3 & strong \\ 
  config. C &   43 &  0.1 & strong \\ 
  config. D &   20 &  0.4 & strong \\ 
  config. E &   24 &  0.1 & strong \\ 
   \bottomrule
\end{tabular}

\caption{Reference values for the \textit{mode} feature.}
\end{table}
\FloatBarrier

% INTERQUARTILE RANGE
\subsection[Discretised intensity interquartile range]{Discretised intensity interquartile range\id{WR0O}}\label{feat_int_hist_iqr}
The \textit{interquartile range} (IQR) of $\mathbf{X}_{d}$ is defined as:
\begin{displaymath}
F_{\mathit{ih.iqr}} = P_{75}-P_{25}
\end{displaymath}
$P_{25}$ and $P_{75}$ are the 25\textsuperscript{th} and 75\textsuperscript{th} percentile of $\mathbf{X}_{d}$, respectively.

\vspace{2mm}
\begin{table}[ht]
\centering
\small
% latex table generated in R 3.6.0 by xtable 1.8-4 package
% Mon Dec 02 12:24:57 2019
\begin{tabular}{cccc}
  \toprule
{\textbf{data}} & {\textbf{value}} & {\textbf{tol.}} & {\textbf{consensus}} \\ 
  \midrule
dig. phantom &    3 & \textemdash & very strong \\ 
  config. A &    2 & \textemdash & strong \\ 
  config. B &    2 & \textemdash & strong \\ 
  config. C &    3 & 0.21 & strong \\ 
  config. D &    2 & 0.06 & strong \\ 
  config. E &    1 & 0.06 & strong \\ 
   \bottomrule
\end{tabular}

\caption{Reference values for the \textit{interquartile range} feature.}
\end{table}
\FloatBarrier

% INTENSITY RANGE
\subsection[Discretised intensity range]{Discretised intensity range\id{5Z3W}}\label{feat_int_hist_range}
The \textit{discretised intensity range} \citep{Aerts2014} is defined as:
\begin{displaymath}
F_{\mathit{ih.range}} = \text{max}(\mathbf{X}_{d}) - \text{min}(\mathbf{X}_{d})
\end{displaymath}
For \textit{fixed bin number} discretisation, the \textit{discretised intensity range} equals $N_g$ by definition.

\vspace{2mm}
\begin{table}[ht]
\centering
\small
% latex table generated in R 3.6.0 by xtable 1.8-4 package
% Mon Dec 02 12:24:58 2019
\begin{tabular}{cccc}
  \toprule
{\textbf{data}} & {\textbf{value}} & {\textbf{tol.}} & {\textbf{consensus}} \\ 
  \midrule
dig. phantom &    5 & \textemdash & very strong \\ 
  config. A &   35 &  0.4 & strong \\ 
  config. B &   31 & \textemdash & strong \\ 
  config. C &   53 &  0.6 & strong \\ 
  config. D &   31 & \textemdash & strong \\ 
  config. E &   31 & \textemdash & strong \\ 
   \bottomrule
\end{tabular}

\caption{Reference values for the \textit{range} feature.}
\end{table}
\FloatBarrier

% MEAN ABSOLUTE DEVIATION
\subsection[Intensity histogram mean absolute deviation]{Intensity histogram mean absolute deviation\id{D2ZX}}\label{feat_int_hist_mean_absolute_dev}
The \textit{mean absolute deviation} \citep{Aerts2014} is a measure of dispersion from the mean of $\mathbf{X}_{d}$:
\begin{displaymath}
F_{\mathit{ih.mad}} = \frac{1}{N_v}\sum_{i=1}^{N_v} \left|X_{d,i}-\mu\right|
\end{displaymath}
Here $\mu=F_{\mathit{ih.mean}}$.

\vspace{2mm}
\begin{table}[ht]
\centering
\small
% latex table generated in R 3.6.0 by xtable 1.8-4 package
% Mon Dec 02 12:24:57 2019
\begin{tabular}{cccc}
  \toprule
{\textbf{data}} & {\textbf{value}} & {\textbf{tol.}} & {\textbf{consensus}} \\ 
  \midrule
dig. phantom & 1.55 & \textemdash & very strong \\ 
  config. A & 2.94 & 0.06 & strong \\ 
  config. B & 2.67 & 0.03 & strong \\ 
  config. C & 6.32 & 0.15 & strong \\ 
  config. D & 3.15 & 0.05 & strong \\ 
  config. E & 3.69 &  0.1 & strong \\ 
   \bottomrule
\end{tabular}

\caption{Reference values for the \textit{mean absolute deviation} feature.}
\end{table}
\FloatBarrier

% ROBUST MEAN ABSOLUTE DEVIATION
\subsection[Intensity histogram robust mean absolute deviation]{Intensity histogram robust mean absolute deviation\id{WRZB}}\label{feat_int_hist_robust_mean_absolute_dev}
\textit{Intensity histogram mean absolute deviation} may be affected by outliers. To increase robustness, the set of discretised intensities under consideration can be restricted to those which are closer to the center of the distribution. Let
\begin{displaymath}
\mathbf{X}_{d,10-90}= \left\lbrace x \in \mathbf{X}_{d} | P_{10}\left(\mathbf{X}_{d}\right)\leq x \leq P_{90}\left(\mathbf{X}_{d}\right)\right\rbrace
\end{displaymath}
In short, $\mathbf{X}_{d,10-90}$ is the set of $N_{v,10-90}\leq N_v$ voxels in $\mathbf{X}_{d}$ whose discretised intensities fall in the interval bounded by the 10\textsuperscript{th} and 90\textsuperscript{th} percentiles of $\mathbf{X}_{d}$. The robust mean absolute deviation is then:
\begin{displaymath}
F_{\mathit{ih.rmad}} = \frac{1}{N_{v,10-90}}\sum_{k=1}^{N_{v,10-90}} \left|X_{d,10-90,k}-\overline{X}_{d,10-90}\right|
\end{displaymath}
$\overline{X}_{d,10-90}$ denotes the sample mean of $\mathbf{X}_{d,10-90}$.

\vspace{2mm}
\begin{table}[ht]
\centering
\small
% latex table generated in R 3.6.0 by xtable 1.8-4 package
% Mon Dec 02 12:24:58 2019
\begin{tabular}{cccc}
  \toprule
{\textbf{data}} & {\textbf{value}} & {\textbf{tol.}} & {\textbf{consensus}} \\ 
  \midrule
dig. phantom & 1.11 & \textemdash & very strong \\ 
  config. A & 1.18 & 0.04 & strong \\ 
  config. B & 1.03 & 0.03 & moderate \\ 
  config. C & 2.59 & 0.14 & strong \\ 
  config. D & 1.33 & 0.06 & strong \\ 
  config. E & 1.46 & 0.09 & moderate \\ 
   \bottomrule
\end{tabular}

\caption{Reference values for the \textit{robust mean absolute deviation} feature.}
\end{table}
\FloatBarrier

% MEDIAN ABSOLUTE DEVIATION
\subsection[Intensity histogram median absolute deviation]{Intensity histogram median absolute deviation\id{4RNL}}\label{feat_int_hist_median_absolute_dev}
\textit{Histogram median absolute deviation} is conceptually similar to \textit{histogram mean absolute deviation}, but measures dispersion from the median instead of mean. Thus:
\begin{displaymath}
F_{\mathit{ih.medad}} = \frac{1}{N_v}\sum_{k=1}^{N_v} \left| X_{d,k}-M\right|
\end{displaymath}
Here, median $M = F_{\mathit{ih.median}}$.

\vspace{2mm}
\begin{table}[ht]
\centering
\small
% latex table generated in R 3.6.0 by xtable 1.8-4 package
% Mon Dec 02 12:24:57 2019
\begin{tabular}{cccc}
  \toprule
{\textbf{data}} & {\textbf{value}} & {\textbf{tol.}} & {\textbf{consensus}} \\ 
  \midrule
dig. phantom & 1.15 & \textemdash & very strong \\ 
  config. A & 2.58 & 0.05 & strong \\ 
  config. B & 2.28 & 0.02 & strong \\ 
  config. C & 4.75 & 0.12 & strong \\ 
  config. D & 2.41 & 0.04 & strong \\ 
  config. E & 2.89 & 0.07 & strong \\ 
   \bottomrule
\end{tabular}

\caption{Reference values for the \textit{median absolute deviation} feature.}
\end{table}
\FloatBarrier

% COEFFICIENT OF VARIATION
\subsection[Intensity histogram coefficient of variation]{Intensity histogram coefficient of variation\id{CWYJ}}\label{feat_int_hist_coef_of_variation}
The \textit{coefficient of variation} measures the dispersion of the discretised intensity distribution. It is defined as:
\begin{displaymath}
F_{\mathit{ih.cov}}=\frac{\sigma}{\mu}
\end{displaymath}
Here $\sigma={F_{\mathit{ih.var}}}^{1/2}$ and $\mu=F_{\mathit{ih.mean}}$ are the standard deviation and mean of the discretised intensity distribution, respectively.

\vspace{2mm}
\begin{table}[ht]
\centering
\small
% latex table generated in R 3.6.0 by xtable 1.8-4 package
% Mon Dec 02 12:24:57 2019
\begin{tabular}{cccc}
  \toprule
{\textbf{data}} & {\textbf{value}} & {\textbf{tol.}} & {\textbf{consensus}} \\ 
  \midrule
dig. phantom & 0.812 & \textemdash & very strong \\ 
  config. A & 0.227 & 0.004 & strong \\ 
  config. B & 0.229 & 0.004 & strong \\ 
  config. C & 0.234 & 0.005 & strong \\ 
  config. D & 0.252 & 0.006 & strong \\ 
  config. E & 0.254 & 0.006 & strong \\ 
   \bottomrule
\end{tabular}

\caption{Reference values for the \textit{coefficient of variation} feature.}
\end{table}
\FloatBarrier

% QUARTILE COEFFICIENT OF DISPERSION
\subsection[Intensity histogram quartile coefficient of dispersion]{Intensity histogram quartile coefficient of dispersion\id{SLWD}}\label{feat_int_quartile_coef_dispersion}
The \textit{quartile coefficient of dispersion} is a more robust alternative to the \textit{intensity histogram coefficient of variance}. It is defined as:
\begin{displaymath}
F_{\mathit{ih.qcod}} = \frac{P_{75}-P_{25}}{P_{75}+P_{25}}
\end{displaymath}
$P_{25}$ and $P_{75}$ are the 25\textsuperscript{th} and 75\textsuperscript{th} percentile of $\mathbf{X}_{d}$, respectively.

\vspace{2mm}
\begin{table}[ht]
\centering
\small
% latex table generated in R 3.6.0 by xtable 1.8-4 package
% Mon Dec 02 12:24:58 2019
\begin{tabular}{cccc}
  \toprule
{\textbf{data}} & {\textbf{value}} & {\textbf{tol.}} & {\textbf{consensus}} \\ 
  \midrule
dig. phantom &  0.6 & \textemdash & very strong \\ 
  config. A & 0.0455 & \textemdash & strong \\ 
  config. B & 0.05 & 0.0005 & strong \\ 
  config. C & 0.0361 & 0.0027 & strong \\ 
  config. D & 0.05 & 0.0021 & strong \\ 
  config. E & 0.0213 & 0.0015 & strong \\ 
   \bottomrule
\end{tabular}

\caption{Reference values for the \textit{quartile coefficient of dispersion} feature.}
\end{table}
\FloatBarrier

% ENTROPY
\subsection[Discretised intensity entropy]{Discretised intensity entropy\id{TLU2}}\label{feat_int_hist_entropy}
\textit{Entropy} \citep{Aerts2014} is an information-theoretic concept that gives a metric for the information contained within $\mathbf{X}_{d}$. The particular metric used is Shannon entropy, which is defined as: 
\begin{displaymath}
F_{\mathit{ih.entropy}} = - \sum_{i=1}^{N_g} p_i \log_2 p_i
\end{displaymath}

Note that \textit{entropy} can only be meaningfully defined for discretised intensities as it will tend to $-\log_2 N_v$ for continuous intensity distributions.

\vspace{2mm}
\begin{table}[ht]
\centering
\small
% latex table generated in R 3.6.0 by xtable 1.8-4 package
% Mon Dec 02 12:24:57 2019
\begin{tabular}{cccc}
  \toprule
{\textbf{data}} & {\textbf{value}} & {\textbf{tol.}} & {\textbf{consensus}} \\ 
  \midrule
dig. phantom & 1.27 & \textemdash & very strong \\ 
  config. A & 3.36 & 0.03 & very strong \\ 
  config. B & 3.16 & 0.01 & strong \\ 
  config. C & 3.73 & 0.04 & strong \\ 
  config. D & 2.94 & 0.01 & strong \\ 
  config. E & 3.22 & 0.02 & strong \\ 
   \bottomrule
\end{tabular}

\caption{Reference values for the \textit{entropy} feature.}
\end{table}
\FloatBarrier

% UNIFORMITY
\subsection[Discretised intensity uniformity]{Discretised intensity uniformity\id{BJ5W}}\label{feat_int_hist_uniformity}
\textit{Uniformity} \citep{Aerts2014} of $\mathbf{X}_{d}$ is defined as: 
\begin{displaymath}
F_{\mathit{ih.uniformity}} = \sum_{i=1}^{N_g} p_i^2
\end{displaymath}

For histograms where most intensities are contained in a single bin, \textit{uniformity} approaches $1$. The lower bound is $1/N_{g}$.

Note that this feature is sometimes referred to as \textit{energy}.

\vspace{2mm}
\begin{table}[ht]
\centering
\small
% latex table generated in R 3.6.0 by xtable 1.8-4 package
% Mon Dec 02 12:24:58 2019
\begin{tabular}{cccc}
  \toprule
{\textbf{data}} & {\textbf{value}} & {\textbf{tol.}} & {\textbf{consensus}} \\ 
  \midrule
dig. phantom & 0.512 & \textemdash & very strong \\ 
  config. A & 0.15 & 0.002 & very strong \\ 
  config. B & 0.174 & 0.001 & strong \\ 
  config. C & 0.14 & 0.003 & strong \\ 
  config. D & 0.229 & 0.003 & strong \\ 
  config. E & 0.184 & 0.001 & strong \\ 
   \bottomrule
\end{tabular}

\caption{Reference values for the \textit{uniformity} feature.}
\end{table}
\FloatBarrier

% MAXIMUM HISTOGRAM GRADIENT
\subsection[Maximum histogram gradient]{Maximum histogram gradient\id{12CE}}\label{feat_int_hist_max_gradient}
The histogram gradient $\mathbf{H}'$ of intensity histogram $\mathbf{H}$ can be calculated as:
\begin{displaymath}
H'_i= \begin{cases}
n_2-n_1 & i=1\\
\left(n_{i+1}-n_{i-1}\right)/2 & 1<i<N_g\\
n_{N_g}-n_{N_g-1} & i=N_g\\
\end{cases}
\end{displaymath}
Histogram $\mathbf{H}$ should be non-sparse, i.e. bins where $n_i=0$ should not be omitted. Ostensibly, the histogram gradient can be calculated in different ways. The method above has the advantages of being easy to implement and leading to a gradient $\mathbf{H}'$ with same size as $\mathbf{H}$. This helps maintain a direct correspondence between the discretised intensities in $\mathbf{H}$ and the bins of $\mathbf{H}'$.
The \textit{maximum histogram gradient} \citep{VanDijk2016} is:
\begin{displaymath}
F_{\mathit{ih.max.grad}} = \text{max}\left(\mathbf{H}'\right)
\end{displaymath}

\vspace{2mm}
\begin{table}[ht]
\centering
\small
% latex table generated in R 3.6.0 by xtable 1.8-4 package
% Mon Dec 02 12:24:57 2019
\begin{tabular}{cccc}
  \toprule
{\textbf{data}} & {\textbf{value}} & {\textbf{tol.}} & {\textbf{consensus}} \\ 
  \midrule
dig. phantom &    8 & \textemdash & very strong \\ 
  config. A & $1.1 \times 10^{4}$ &  100 & strong \\ 
  config. B & $3.22 \times 10^{3}$ &   50 & strong \\ 
  config. C & $4.75 \times 10^{3}$ &   30 & strong \\ 
  config. D & $7.26 \times 10^{3}$ &  200 & strong \\ 
  config. E & $6.01 \times 10^{3}$ &  130 & strong \\ 
   \bottomrule
\end{tabular}

\caption{Reference values for the \textit{maximum histogram gradient} feature.}
\end{table}
\FloatBarrier

% MAXIMUM HISTOGRAM GRADIENT INTENSITY
\subsection[Maximum histogram gradient intensity]{Maximum histogram gradient intensity\id{8E6O}}\label{feat_int_hist_max_gradient_intensity}
The \textit{maximum histogram gradient intensity} \citep{VanDijk2016} $F_{\mathit{ih.max.grad.gl}}$ is the discretised intensity corresponding to the \textit{maximum histogram gradient}, i.e. the value $i$ in $\mathbf{H}$ for which $\mathbf{H}'$ is maximal.

\vspace{2mm}
\begin{table}[ht]
\centering
\small
% latex table generated in R 3.6.0 by xtable 1.8-4 package
% Mon Dec 02 12:24:57 2019
\begin{tabular}{cccc}
  \toprule
{\textbf{data}} & {\textbf{value}} & {\textbf{tol.}} & {\textbf{consensus}} \\ 
  \midrule
dig. phantom &    3 & \textemdash & strong \\ 
  config. A &   21 & \textemdash & strong \\ 
  config. B &   19 &  0.3 & strong \\ 
  config. C &   41 & \textemdash & strong \\ 
  config. D &   19 &  0.4 & strong \\ 
  config. E &   23 &  0.2 & moderate \\ 
   \bottomrule
\end{tabular}

\caption{Reference values for the \textit{maximum histogram gradient intensity} feature.}
\end{table}
\FloatBarrier

% MINIMUM HISTOGRAM GRADIENT
\subsection[Minimum histogram gradient]{Minimum histogram gradient\id{VQB3}}\label{feat_int_hist_min_gradient}
The \textit{minimum histogram gradient} \citep{VanDijk2016} is:
\begin{displaymath}
F_{\mathit{ih.min.grad}} = \text{min}\left(\mathbf{H}'\right)
\end{displaymath}

\vspace{2mm}
\begin{table}[ht]
\centering
\small
% latex table generated in R 3.6.0 by xtable 1.8-4 package
% Mon Dec 02 12:24:57 2019
\begin{tabular}{cccc}
  \toprule
{\textbf{data}} & {\textbf{value}} & {\textbf{tol.}} & {\textbf{consensus}} \\ 
  \midrule
dig. phantom &  $-$50 & \textemdash & very strong \\ 
  config. A & $-1.01 \times 10^{4}$ &  100 & strong \\ 
  config. B & $-3.02 \times 10^{3}$ &   50 & strong \\ 
  config. C & $-4.68 \times 10^{3}$ &   50 & strong \\ 
  config. D & $-6.67 \times 10^{3}$ &  230 & strong \\ 
  config. E & $-6.11 \times 10^{3}$ &  180 & strong \\ 
   \bottomrule
\end{tabular}

\caption{Reference values for the \textit{minimum histogram gradient} feature.}
\end{table}
\FloatBarrier

\layoutnewpage
% MINIMUM HISTOGRAM GRADIENT INTENSITY
\subsection[Minimum histogram gradient intensity]{Minimum histogram gradient intensity\id{RHQZ}}\label{feat_int_hist_min_gradient_intensity}
The \textit{minimum histogram gradient intensity} \citep{VanDijk2016} $F_{\mathit{ih.min.grad.gl}}$ is the discretised intensity corresponding to the \textit{minimum histogram gradient}, i.e. the value $i$ in $\mathbf{H}$ for which $\mathbf{H}'$ is minimal.

\vspace{2mm}
\begin{table}[ht]
\centering
\small
% latex table generated in R 3.6.0 by xtable 1.8-4 package
% Mon Dec 02 12:24:57 2019
\begin{tabular}{cccc}
  \toprule
{\textbf{data}} & {\textbf{value}} & {\textbf{tol.}} & {\textbf{consensus}} \\ 
  \midrule
dig. phantom &    1 & \textemdash & strong \\ 
  config. A &   24 & \textemdash & strong \\ 
  config. B &   22 &  0.3 & strong \\ 
  config. C &   44 & \textemdash & strong \\ 
  config. D &   22 &  0.4 & strong \\ 
  config. E &   25 &  0.2 & strong \\ 
   \bottomrule
\end{tabular}

\caption{Reference values for the \textit{minimum histogram gradient intensity} feature.}
\end{table}
\FloatBarrier

\clearpage
\section[Intensity-volume histogram features]{Intensity-volume histogram features\id{P88C}}\label{sect_ivh}
The (cumulative) intensity-volume histogram (IVH) of the set $\mathbf{X}_{gl}$ of voxel intensities in the ROI intensity mask describes the relationship between discretised intensity $i$ and the fraction of the volume containing at least intensity $i$, $\nu$ \citep{ElNaqa2009}. 

Depending on the imaging modality, the calculation of IVH features requires discretising $\mathbf{X}_{gl}$ to generate a new voxel set $\mathbf{X}_{d,gl}$ with discretised intensities. Moreover, the total range $\mathbf{G}$ of discretised intensities and the discretisation interval $w_d$ should be provided or determined. The total range $\mathbf{G}$ determines the range of discretised intensities to be included in the IVH, whereas the discretisation interval determines the intensity difference between adjacent discretised intensities in the IVH.

Recommendations for discretisation parameters differ depending on what type of data the image represents, and how it is represented. These recommendations are described below.

\subsubsection*{Discrete calibrated image intensities} Some imaging modalities by default generate voxels with calibrated, discrete intensities -- for example CT. In this case, the discretised ROI voxel set $\mathbf{X}_{d,gl}=\mathbf{X}_{gl}$, i.e. no discretisation required. If a re-segmentation range is provided (see Section \ref{ref_resegmentation}), the total range $\mathbf{G}$ is equal to the re-segmentation range. In the case of a half-open re-segmentation range, the upper limit of the range is $\text{max}(\mathbf{X}_{gl})$. When no range is provided, $\mathbf{G}=[\text{min}(\mathbf{X}_{gl}),\text{max}(\mathbf{X}_{gl})]$. The discretisation interval is $w_d=1$.

\subsubsection*{Continuous calibrated image intensities} 
Imaging with calibrated, continuous intensities such as PET requires discretisation to determine the IVH, while preserving the quantitative intensity information. The use of a \textit{fixed bin size} discretisation method is thus recommended, see Section \ref{discretisation}. This method requires a minimum intensity $X_{gl,min}$, a maximum intensity $X_{gl,max}$ and the bin width $w_b$. If a re-segmentation range is defined (see Section \ref{ref_resegmentation}), $X_{gl,min}$ is set to the lower bound of the re-segmentation range and $X_{gl,max}$ to the upper bound; otherwise $X_{gl,min} = \text{min}(\mathbf{X}_{gl})$ and $X_{gl,max} = \text{max}(\mathbf{X}_{gl})$ (i.e. the minimum and maximum intensities in the imaging volume prior to discretisation). The bin width $w_b$ is modality dependent, but should be small relative to the intensity range, e.g. 0.10 SUV for \textsuperscript{18}F-FDG-PET. 

Next, \textit{fixed bin size} discretisation produces the voxel set $\mathbf{X}_{d}$ of bin numbers, which needs to be converted to bin centers in order to maintain a direct relationship with the original intensities. We thus replace bin numbers $\mathbf{X}_{d}$ with the intensity corresponding to the bin center:
\begin{displaymath}
\mathbf{X}_{d,gl} = X_{gl,min} + \left(\mathbf{X}_{d}-0.5\right)w_b
\end{displaymath}
The total range is then $\mathbf{G}=[X_{gl,min}+0.5w_b, X_{gl,max}-0.5w_b]$. In this case, the discretisation interval matches the bin width, i.e. $w_d=w_b$.

\subsubsection*{Arbitrary intensity units} Some imaging modalities, such as many MRI sequences, produce arbitrary intensities. In such cases, a \textit{fixed bin number} discretisation method with $N_g=1000$ bins is recommended, see Section \ref{discretisation}. The discretisation bin width is $w_b=\left(X_{gl,max}-X_{gl,min}\right)/N_g$, with $X_{gl,max}=\text{max}\left(\mathbf{X}_{gl}\right)$ and $X_{gl,min}=\text{min}\left(\mathbf{X}_{gl}\right)$, as re-segmentation ranges generally cannot be provided for non-calibrated intensities. The \textit{fixed bin number} discretisation produces the voxel set $\mathbf{X}_{d} \in \{1,2,\ldots,N_g\}$. Because of the lack of calibration, $\mathbf{X}_{d,gl}=\mathbf{X}_{d}$, and consequentially the discretisation interval is $w_d=1$ and the total range is $\mathbf{G}=[1,N_g]$

\subsubsection*{Calculating the IV histogram} We use $\mathbf{X}_{d,gl}$ to calculate fractional volumes and fractional intensities. 

As voxels for the same image stack generally all have the same dimensions, we may define fractional volume $\nu$ for discretised intensity $i$:
\begin{displaymath}
\nu_i = 1 - \frac{1}{N_v}\sum_{k=1}^{N_v}\left[X_{d,gl,k}< i\right]
\end{displaymath}
Here $\left[\ldots\right]$ is an Iverson bracket, yielding $1$ if the condition is true and $0$ otherwise. In essence, we count the voxels containing a discretised intensity smaller than $i$, divide by the total number of voxels, and then subtract this volume fraction to find $\nu_i$.

The intensity fraction $\gamma$ for discretised intensity $i$ in the range $\mathbf{G}$ is calculated as:
\begin{displaymath}
\gamma_i=\frac{i-\text{min}\left(\mathbf{G}\right)} {\text{max}\left(\mathbf{G}\right) - \text{min}\left(\mathbf{G}\right)}
\end{displaymath}
Note that intensity fractions are also calculated for discretised intensities that are absent in $\mathbf{X}_{d,gl}$. For example intensities 2 and 5 are absent in the digital phantom (see Chapter \ref{chap_benchmark sets}), but are still evaluated to determine both the fractional volume and the intensity fraction. An example IVH for the digital phantom is shown in Table \ref{TableAUC-CVH}.

\begin{table}[t]
\centering
\begin{tabular}{@{}ccc@{}}
\toprule
$i$ & $\gamma$ & $\nu$\\
\midrule
1 & 0.0 & 1.000\\
2 & 0.2 & 0.324\\
3 & 0.4 & 0.324\\
4 & 0.6 & 0.311\\
5 & 0.8 & 0.095\\
6 & 1.0 & 0.095\\
\bottomrule
\end{tabular}
\caption{Example intensity-volume histogram evaluated at discrete intensities $i$ of the digital phantom. The total range $\mathbf{G}=[1,6]$, with discretisation interval $w=1$. Thus $\gamma$ is the intensity fraction and $\nu$ is the corresponding volume fraction that contains intensity $i$ or greater.}
\label{TableAUC-CVH}
\end{table}

\subsubsection*{Aggregating features}
We recommend calculating intensity-volume histogram features using the 3D volume (\textid{DHQ4}). Computing features per slice and subsequently averaging (\textid{3IDG}) is not recommended.

% VOLUME AT INTENSITY FRACTION
\subsection[Volume at intensity fraction]{Volume at intensity fraction\id{BC2M}}\label{feat_int_vol_hist_volume_at_intensity}
The \textit{volume at intensity fraction} $V_x$ is the largest volume fraction $\nu$ that has an intensity fraction $\gamma$ of at least $x\%$. This differs from conceptually similar dose-volume histograms used in radiotherapy planning, where $V_{10}$ would indicate the volume fraction receiving at least 10 Gy planned dose. \citet{ElNaqa2009} defined both $V_{10}$ and $V_{90}$ as features. In the context of this work, these two features are defined as $F_{\mathit{ivh.V10}}$ and $F_{\mathit{ivh.V90}}$, respectively. 

\vspace{2mm}
\begin{table}[ht]
\centering
\small
% latex table generated in R 3.6.0 by xtable 1.8-4 package
% Mon Dec 02 12:24:58 2019
\begin{tabular}{cccc}
  \toprule
{\textbf{data}} & {\textbf{value}} & {\textbf{tol.}} & {\textbf{consensus}} \\ 
  \midrule
dig. phantom & 0.324 & \textemdash & very strong \\ 
  config. A & 0.978 & 0.001 & strong \\ 
  config. B & 0.977 & 0.001 & strong \\ 
  config. C & 0.998 & 0.001 & moderate \\ 
  config. D & 0.972 & 0.003 & strong \\ 
  config. E & 0.975 & 0.002 & strong \\ 
   \bottomrule
\end{tabular}

\caption{Reference values for the \textit{volume fraction at 10\% intensity} feature.}
\end{table}
\FloatBarrier
\vspace{2mm}
\begin{table}[ht]
\centering
\small
% latex table generated in R 3.6.0 by xtable 1.8-4 package
% Mon Dec 02 12:24:58 2019
\begin{tabular}{cccc}
  \toprule
{\textbf{data}} & {\textbf{value}} & {\textbf{tol.}} & {\textbf{consensus}} \\ 
  \midrule
dig. phantom & 0.0946 & \textemdash & very strong \\ 
  config. A & $6.98 \times 10^{-5}$ & $1.03 \times 10^{-5}$ & strong \\ 
  config. B & $7.31 \times 10^{-5}$ & $1.03 \times 10^{-5}$ & strong \\ 
  config. C & 0.000152 & $2 \times 10^{-5}$ & strong \\ 
  config. D & $9 \times 10^{-5}$ & 0.000415 & strong \\ 
  config. E & 0.000157 & 0.000248 & strong \\ 
   \bottomrule
\end{tabular}

\caption{Reference values for the \textit{volume fraction at 90\% intensity} feature.}
\end{table}
\FloatBarrier

% INTENSITY AT VOLUME FRACTION
\subsection[Intensity at volume fraction]{Intensity at volume fraction\id{GBPN}}\label{feat_int_vol_hist_intensity_at_volume}
The \textit{intensity at volume fraction} $I_x$ is the minimum discretised intensity $i$ present in at most $x\%$ of the volume. \citet{ElNaqa2009} defined both $I_{10}$ and $I_{90}$ as features. In the context of this work, these two features are defined as $F_{\mathit{ivh.I10}}$ and $F_{\mathit{ivh.I90}}$, respectively.

\vspace{2mm}
\begin{table}[ht]
\centering
\small
% latex table generated in R 3.6.0 by xtable 1.8-4 package
% Mon Dec 02 12:24:58 2019
\begin{tabular}{cccc}
  \toprule
{\textbf{data}} & {\textbf{value}} & {\textbf{tol.}} & {\textbf{consensus}} \\ 
  \midrule
dig. phantom &    5 & \textemdash & very strong \\ 
  config. A &   96 & \textemdash & strong \\ 
  config. B &   92 & \textemdash & strong \\ 
  config. C & 88.8 &  0.2 & moderate \\ 
  config. D &   87 &  0.1 & strong \\ 
  config. E &  770 &    5 & moderate \\ 
   \bottomrule
\end{tabular}

\caption{Reference values for the \textit{intensity at 10\% volume} feature.}
\end{table}
\FloatBarrier
\vspace{2mm}
\begin{table}[ht]
\centering
\small
% latex table generated in R 3.6.0 by xtable 1.8-4 package
% Mon Dec 02 12:24:58 2019
\begin{tabular}{cccc}
  \toprule
{\textbf{data}} & {\textbf{value}} & {\textbf{tol.}} & {\textbf{consensus}} \\ 
  \midrule
dig. phantom &    2 & \textemdash & very strong \\ 
  config. A & $-$128 &    8 & strong \\ 
  config. B & $-$135 &    8 & strong \\ 
  config. C & $-$421 &   14 & strong \\ 
  config. D & $-$303 &   20 & strong \\ 
  config. E &  399 &   17 & moderate \\ 
   \bottomrule
\end{tabular}

\caption{Reference values for the \textit{intensity at 90\% volume} feature.}
\end{table}
\FloatBarrier

% VOLUME DIFFERENCE BETWEEN INTENSITY FRACTIONS
\subsection[Volume fraction difference between intensity fractions]{Volume fraction difference between intensity fractions\id{DDTU}}\label{feat_int_vol_hist_volume_at_intensity_difference}
This feature is the difference between the volume fractions at two different intensity fractions, e.g. $V_{10}-V_{90}$ \citep{ElNaqa2009}. In the context of this work, this feature is defined as $F_{\mathit{ivh.V10minusV90}}$.

\vspace{2mm}
\begin{table}[ht]
\centering
\small
% latex table generated in R 3.6.0 by xtable 1.8-4 package
% Mon Dec 02 12:24:58 2019
\begin{tabular}{cccc}
  \toprule
{\textbf{data}} & {\textbf{value}} & {\textbf{tol.}} & {\textbf{consensus}} \\ 
  \midrule
dig. phantom & 0.23 & \textemdash & very strong \\ 
  config. A & 0.978 & 0.001 & strong \\ 
  config. B & 0.977 & 0.001 & strong \\ 
  config. C & 0.997 & 0.001 & strong \\ 
  config. D & 0.971 & 0.001 & strong \\ 
  config. E & 0.974 & 0.001 & strong \\ 
   \bottomrule
\end{tabular}

\caption{Reference values for the \textit{volume fraction difference between 10\% and 90\% intensity} feature.}
\end{table}
\FloatBarrier

% INTENSITY AT VOLUME FRACTION DIFFERENCE
\subsection[Intensity fraction difference between volume fractions]{Intensity fraction difference between volume fractions\id{CNV2}}\label{feat_int_vol_hist_intensity_at_volume_difference}
This feature is the difference between discretised intensities at two different fractional volumes, e.g. $I_{10} - I_{90}$ \citep{ElNaqa2009}. In the context of this work, this feature is defined as $F_{\mathit{ivh.I10minusI90}}$.

\vspace{2mm}
\begin{table}[ht]
\centering
\small
% latex table generated in R 3.6.0 by xtable 1.8-4 package
% Mon Dec 02 12:24:58 2019
\begin{tabular}{cccc}
  \toprule
{\textbf{data}} & {\textbf{value}} & {\textbf{tol.}} & {\textbf{consensus}} \\ 
  \midrule
dig. phantom &    3 & \textemdash & very strong \\ 
  config. A &  224 &    8 & strong \\ 
  config. B &  227 &    8 & strong \\ 
  config. C &  510 &   14 & strong \\ 
  config. D &  390 &   20 & strong \\ 
  config. E &  371 &   13 & moderate \\ 
   \bottomrule
\end{tabular}

\caption{Reference values for the \textit{intensity difference between 10\% and 90\% volume} feature.}
\end{table}
\FloatBarrier

% AREA UNDER THE INTENSITY-VOLUME HISTOGRAM
\subsection[Area under the IVH curve]{Area under the IVH curve\id{9CMM}}\label{feat_int_vol_hist_area}

\noindent\textbf{Note:} This feature currently has no reference values and should not be used.

The \textit{area under the IVH curve} $F_{\mathit{ivh.auc}}$ was defined by \citet{VanVelden2011}. The \textit{area under the IVH curve} can be approximated by calculating the Riemann sum using the trapezoidal rule. Note that if there is only one discretised intensity in the ROI, we define the \textit{area under the IVH curve} $F_{\mathit{ivh.auc}}=0$.

\clearpage
\section[Grey level co-occurrence based features]{Grey level co-occurrence based features\id{LFYI}} \label{sect_glcm}
In image analysis, texture is one of the defining sets of features. Texture features were originally designed to assess surface texture in 2D images. Texture analysis is however not restricted to 2D slices and can be extended to 3D objects. Image intensities are generally discretised before calculation of texture features, see Section \ref{discretisation}.

The grey level co-occurrence matrix (GLCM) is a matrix that expresses how combinations of discretised intensities (grey levels) of neighbouring pixels, or voxels in a 3D volume, are distributed along one of the image directions. Generally, the neighbourhood for GLCM is a 26-connected neighbourhood in 3D and a 8-connected neighbourhood in 2D. Thus, in 3D there are 13 unique direction vectors within the neighbourhood for Chebyshev distance $\delta=1$, i.e. $(0,0,1)$, $(0,1,0)$, $(1,0,0)$, $(0,1,1)$, $(0,1,-1)$, $(1,0,1)$, $(1,0,-1)$, $(1,1,0)$, $(1,-1,0)$, $(1,1,1)$, $(1,1,-1)$, $(1,-1,1)$ and $(1,-1,-1)$, whereas in 2D the direction vectors are $(1,0,0)$, $(1,1,0)$, $(0,1,0)$ and $(-1,1,0)$.

A GLCM is calculated for each direction vector, as follows. Let $\mathbf{M}_{\mathbf{m}}$ be the $N_g \times N_g$ grey level co-occurrence matrix, with $N_g$ the number of discretised grey levels present in the ROI intensity mask, and $\mathbf{m}$ the particular direction vector. Element $(i,j)$ of the GLCM contains the frequency at which combinations of discretised grey levels $i$ and $j$ occur in neighbouring voxels along direction $\mathbf{m}_{+}=\mathbf{m}$ and along direction $\mathbf{m}_{-}= -\mathbf{m}$. Then, $\mathbf{M}_{\mathbf{m}} = \mathbf{M}_{\mathbf{m}_{+}} + \mathbf{M}_{\mathbf{m}_{-}} = \mathbf{M}_{\mathbf{m}_{+}} + \mathbf{M}_{\mathbf{m}_{+}}^T$ \citep{Haralick1973}. As a consequence the GLCM matrix $\mathbf{M}_{\mathbf{m}}$ is symmetric. An example of the calculation of a GLCM is shown in Table \ref{figGLCM1}. Corresponding grey level co-occurrence matrices for each direction are shown in Table \ref{figGLCM2}.
\begin{table}[th]
\centering
\subcaptionbox{Grey levels}{
	\begin{tabular}{@{}cccc@{}}
		\toprule
		1 & 2 & 2 & 3\\
		1 & 2 & 3 & 3\\
		4 & 2 & 4 & 1\\
		4 & 1 & 2 & 3\\
		\bottomrule
	\end{tabular}}\qquad
\subcaptionbox{$\mathbf{M}_{\mathbf{m}_{+} = \rightarrow}$}{
	\begin{tabular}{@{}ccccc@{}}
		\toprule
		&\multicolumn{4}{c}{$j$}\\
		\midrule
		\multirow{4}{*}{$i$} & 0 & 3 & 0 & 0\\
		& 0 & 1 & 3 & 1\\
		& 0 & 0 & 1 & 0\\
		& 2 & 1 & 0 & 0\\
		\bottomrule
	\end{tabular}}\qquad
\subcaptionbox{$\mathbf{M}_{\mathbf{m}_{-} = \leftarrow}$}{
	\begin{tabular}{@{}ccccc@{}}
		\toprule
		&\multicolumn{4}{c}{$j$}\\
		\midrule
		\multirow{4}{*}{$i$} & 0 & 0 & 0 & 2\\
		& 3 & 1 & 0 & 1\\
		& 0 & 3 & 1 & 0\\
		& 0 & 1 & 0 & 0\\
		\bottomrule
	\end{tabular}}
\caption{Grey levels (a) and corresponding grey level co-occurrence matrices for the $0^{\circ}$ (b) and $180^{\circ}$ directions (c). In vector notation these directions are $\mathbf{m}_{+}=(1,0)$ and $\mathbf{m}_{-}=(-1,0)$. To calculate the symmetrical co-occurrence matrix $\mathbf{M}_{\mathbf{m}}$ both matrices are summed by element.}
\label{figGLCM1}
\end{table}

\begin{table}
\centering
\subcaptionbox{$\mathbf{M}_{\mathbf{m}=\rightarrow}$}{
	\begin{tabular}{@{}ccccc@{}}
		\toprule
		&\multicolumn{4}{c}{$j$}\\
		\midrule
		\multirow{4}{*}{$i$} & 0 & 3 & 0 & 2\\
		& 3 & 2 & 3 & 2\\
		& 0 & 3 & 2 & 0\\
		& 2 & 2 & 0 & 0\\
		\bottomrule
	\end{tabular}}\qquad
\subcaptionbox{$\mathbf{M}_{\mathbf{m}=\nearrow}$}{
	\begin{tabular}{@{}ccccc@{}}
		\toprule
		&\multicolumn{4}{c}{$j$}\\
		\midrule
		\multirow{4}{*}{$i$} & 0 & 2 & 0 & 1\\
		& 2 & 2 & 1 & 2\\
		& 0 & 1 & 2 & 1\\
		& 1 & 2 & 1 & 0\\
		\bottomrule
	\end{tabular}}\vspace{1cm}\\
\subcaptionbox{$\mathbf{M}_{\mathbf{m}=\uparrow}$}{
	\begin{tabular}{@{}ccccc@{}}
		\toprule
		&\multicolumn{4}{c}{$j$}\\
		\midrule
		\multirow{4}{*}{$i$} & 2 & 1 & 2 & 1\\
		& 1 & 4 & 1 & 1\\
		& 2 & 1 & 2 & 1\\
		& 1 & 1 & 1 & 2\\
		\bottomrule
	\end{tabular}}\qquad
\subcaptionbox{$\mathbf{M}_{\mathbf{m}=\nwarrow}$}{
	\begin{tabular}{@{}ccccc@{}}
		\toprule
		&\multicolumn{4}{c}{$j$}\\
		\midrule
		\multirow{4}{*}{$i$} & 0 & 2 & 1 & 1\\
		& 2 & 2 & 2 & 1\\
		& 1 & 2 & 0 & 1\\
		& 1 & 1 & 1 & 0\\
		\bottomrule
	\end{tabular}}
\caption{Grey level co-occurrence matrices for the $0^{\circ}$ (a), $45^{\circ}$ (b), $90^{\circ}$ (c) and $135^{\circ}$ (d) directions. In vector notation these directions are $\mathbf{m}=(1,0)$, $\mathbf{m}=(1,1)$, $\mathbf{m}=(0,1)$ and $\mathbf{m}=(-1,1)$, respectively.}
\label{figGLCM2}
\end{table}

GLCM features rely on the probability distribution for the elements of the GLCM. Let us consider $\mathbf{M}_{\mathbf{m}=(1,0)}$ from the example, as shown in Table \ref{figGLCM3}. We derive a probability distribution for grey level co-occurrences, $\mathbf{P}_{\mathbf{m}}$, by normalising $\mathbf{M}_{\mathbf{m}}$ by the sum of its elements. Each element $p_{ij}$ of $\mathbf{P}_{\mathbf{m}}$ is then the joint probability of grey levels $i$ and $j$ occurring in neighbouring voxels along direction $\mathbf{m}$. Then $p_{i.} = \sum_{j=1}^{N_g} p_{ij}$ is the row marginal probability, and $p_{.j}=\sum_{i=1}^{N_g} p_{ij}$ is the column marginal probability. As $\mathbf{P}_{\mathbf{m}}$ is by definition symmetric, $p_{i.} = p_{.j}$. Furthermore, let us consider diagonal and cross-diagonal probabilities $p_{i-j}$ and $p_{i+j}$ \citep{Haralick1973,Unser1986}:
\begin{align*}
p_{i-j,k}&=\sum_{i=1}^{N_g} \sum_{j=1}^{N_g} p_{ij}\, \left[k=|i-j|\right]\qquad k=0,\ldots ,N_g-1\\
p_{i+j,k}&=\sum_{i=1}^{N_g} \sum_{j=1}^{N_g} p_{ij}\, \left[k=i+j\right]\qquad k=2,\ldots ,2N_g
\end{align*}
Here, $\left[\ldots\right]$ is an Iverson bracket, which equals $1$ when the condition within the brackets is true and $0$ otherwise. In effect we select only combinations of elements $(i,j)$ for which the condition holds.

It should be noted that while a distance $\delta=1$ is commonly used for GLCM, other distances are possible. However, this does not change the number of  For example, for $\delta=3$ (in 3D) the voxels at $(0,0,3)$, $(0,3,0)$, $(3,0,0)$, $(0,3,3)$, $(0,3,-3)$, $(3,0,3)$, $(3,0,-3)$, $(3,3,0)$, $(3,-3,0)$, $(3,3,3)$, $(3,3,-3)$, $(3,-3,3)$ and $(3,-3,-3)$ from the center voxel are considered.

\subsubsection*{Aggregating features}
\begin{figure}[ht]
\centering
\begin{subfigure}[h]{0.45\textwidth}\centering
	\includegraphics[scale=0.55]{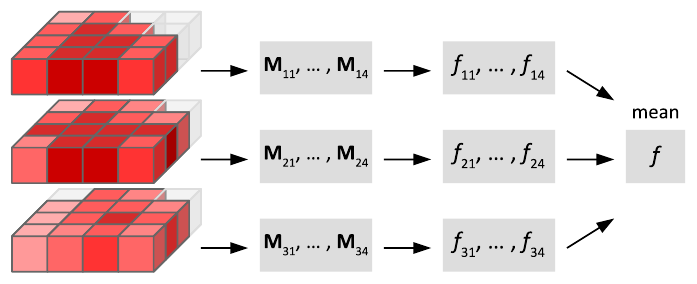}
	\caption{2D: by slice, without merging}
\end{subfigure}
\hfill
\begin{subfigure}[h]{0.45\textwidth}\centering
	\includegraphics[scale=0.55]{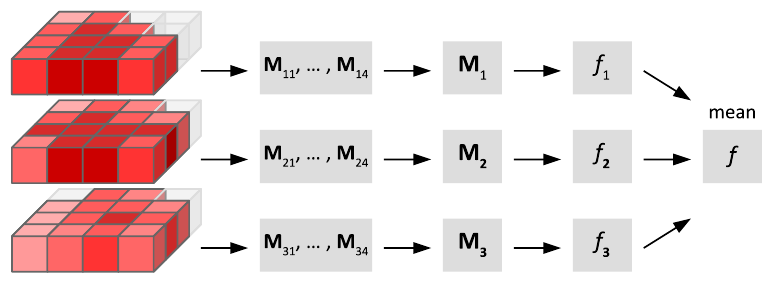}
	\caption{2D: by slice, with merging by slice}
\end{subfigure}
\vspace{1cm}\\
\begin{subfigure}[h]{0.45\textwidth}\centering
	\includegraphics[scale=0.55]{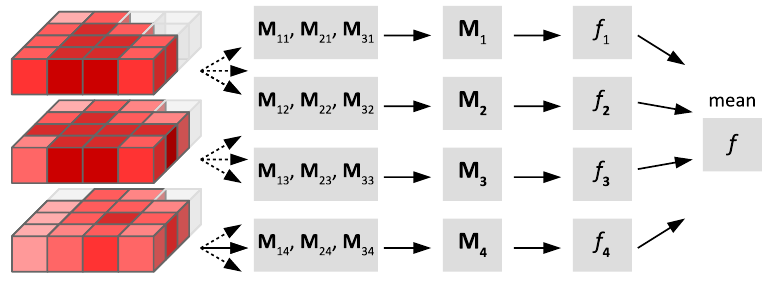}
	\caption{2.5D: by slice, with merging by direction}
\end{subfigure}
\hfill
\begin{subfigure}[h]{0.45\textwidth}\centering
	\includegraphics[scale=0.55]{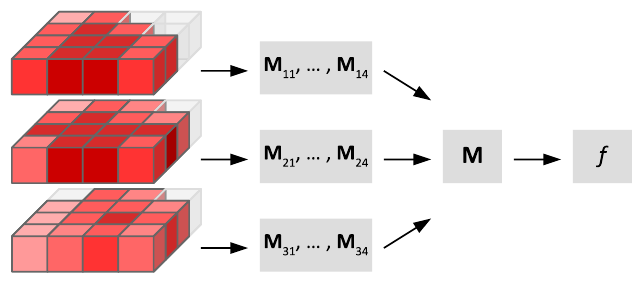}
	\caption{2.5D: by slice, with full merging}
\end{subfigure}
\vspace{1cm}\\
\begin{subfigure}[h]{0.45\textwidth}\centering
	\includegraphics[scale=0.55]{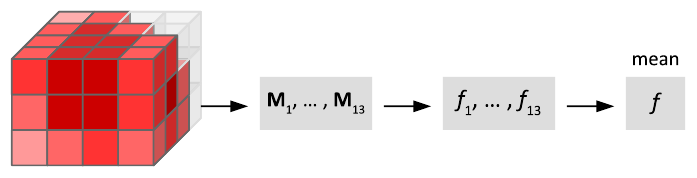}
	\caption{3D: as volume, without merging}
\end{subfigure}
\hfill
\begin{subfigure}[h]{0.45\textwidth}\centering
	\includegraphics[scale=0.55]{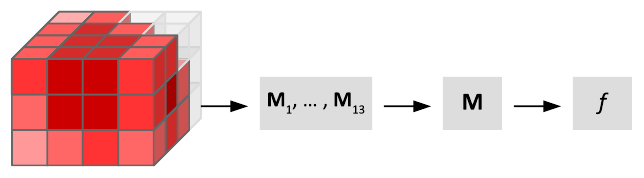}
	\caption{3D: as volume, with full merging}
\end{subfigure}
\caption{Approaches to calculating grey level co-occurrence matrix-based features. $\mathbf{M}_{\Delta k}$ are texture matrices calculated for direction $\Delta$ in slice $k$ (if applicable), and $f_{\Delta k}$ is the corresponding feature value. In (b-d) and (e) the matrices are merged prior to feature calculation.}
\label{figGLCMCalcApproaches}
\end{figure}

To improve rotational invariance, GLCM feature values are computed by aggregating information from the different underlying directional matrices \citep{Depeursinge2017a}. Five methods can be used to aggregate GLCMs and arrive at a single feature value. A schematic example is shown in Figure \ref{figGLCMCalcApproaches}. A feature may be aggregated as follows:
\begin{enumerate}
\item Features are computed from each 2D directional matrix and averaged over 2D directions and slices (\textid{BTW3}).
\item Features are computed from a single matrix after merging 2D directional matrices per slice, and then averaged over slices (\textid{SUJT}).
\item Features are computed from a single matrix after merging 2D directional matrices per direction, and then averaged over directions (\textid{JJUI}).
\item The feature is computed from a single matrix after merging all 2D directional matrices (\textid{ZW7Z}).
\item Features are computed from each 3D directional matrix and averaged over the 3D directions (\textid{ITBB}).
\item The feature is computed from a single matrix after merging all 3D directional matrices (\textid{IAZD}).
\end{enumerate}
In methods 2,3,4 and 6, matrices are merged by summing the co-occurrence counts in each matrix element $(i,j)$ over the different matrices. Probability distributions are subsequently calculated for the merged GLCM, which is then used to calculate GLCM features. Feature values may dependent strongly on the aggregation method.

\begin{table}[th]
\centering
\subcaptionbox{$\mathbf{M}_{\mathbf{m}=(1,0)}$ with margins}{
	\begin{tabular}{@{}cccccc@{}}
		\toprule
		&\multicolumn{4}{c}{$j$} & $\sum_j$\\
		\midrule
		\multirow{4}{*}{$i$} & 0 & 3 & 0 & 2 & \textbf{5}\\
		& 3 & 2 & 3 & 2 & \textbf{10}\\
		& 0 & 3 & 2 & 0 & \textbf{5}\\
		& 2 & 2 & 0 & 0 & \textbf{4}\\
		$\sum_i$ & \textbf{5} & \textbf{10} & \textbf{5} & \textbf{4} & \textbf{24}\\
		\bottomrule
	\end{tabular}}\qquad
\subcaptionbox{$\mathbf{P}_{\mathbf{m}=(1,0)}$ with margins}{
	\begin{tabular}{@{}cccccc@{}}
		\toprule
		&\multicolumn{4}{c}{$j$} & $p_{i.}$\\
		\midrule
		\multirow{4}{*}{$i$} & 0.00 & 0.13 & 0.00 & 0.08 & \textbf{0.21}\\
		& 0.13 & 0.08 & 0.13 & 0.08 & \textbf{0.42}\\
		& 0.00 & 0.13 & 0.08 & 0.00 & \textbf{0.21}\\
		& 0.08 & 0.08 & 0.00 & 0.00 & \textbf{0.17}\\
		$p_{.j}$ & \textbf{0.21} & \textbf{0.42} & \textbf{0.21} & \textbf{0.17} & \textbf{1.00}\\
		\bottomrule
	\end{tabular}}\vspace{1cm}\\
\subcaptionbox{Diagonal probability for $\mathbf{P}_{\mathbf{m}=(1,0)}$}{
	\begin{tabular}{@{}ccccc@{}}
		\toprule
		$k=|i-j|$ & 0 & 1 & 2 & 3\\
		\cmidrule(lr){2-5}
		$p_{i-j}$ & 0.17 & 0.50 & 0.17 & 0.17\\
		\bottomrule
	\end{tabular}}\vspace{1cm}\\
\subcaptionbox{Cross-diagonal probability for $\mathbf{P}_{\mathbf{m}=(1,0)}$}{
	\begin{tabular}{@{}cccccccc@{}}
		\toprule
		$k=i+j$ & 2 & 3 & 4 & 5 & 6 & 7 & 8\\
		\cmidrule(lr){2-8}
		$p_{i+j}$ & 0.00 & 0.25 & 0.08 & 0.42 & 0.25 & 0.00 & 0.00\\
		\bottomrule
	\end{tabular}}
\caption{Grey level co-occurrence matrix for the $0^{\circ}$ direction (a); its corresponding probability matrix $\mathbf{P}_{\mathbf{m}=(1,0)}$ with marginal probabilities $p_{i.}$ and $p_{.j}$(b); the diagonal probabilities $p_{i-j}$ (c); and the cross-diagonal probabilities $p_{i+j}$ (d). Discrepancies in panels b, c, and d are due to rounding errors caused by showing only two decimal places. Also, note that due to GLCM symmetry marginal probabilities $p_{i.}$ and $p_{.j}$ are the same in both row and column margins of panel b.}
\label{figGLCM3}
\end{table}

\subsubsection*{Distances and distance weighting}
The default neighbourhood includes all voxels within Chebyshev distance $1$. The corresponding direction vectors are multiplied by the desired distance $\delta$. From a technical point-of-view, direction vectors may also be determined differently, using any distance norm. In this case, direction vectors are the vectors to the voxels at $\delta$, or between $\delta$ and $\delta-1$ for the Euclidean norm. Such usage is however rare and we caution against it due to potential reproducibility issues.

GLCMs may be weighted for distance by multiplying $\mathbf{M}$ with a weighting factor $w$. By default $w=1$, but $w$ may also be an inverse distance function to weight each GLCM, e.g. $w=\norm{\mathbf{m}}^{-1}$ or $w=\exp(-\norm{\mathbf{m}}^2)$ \citep{VanGriethuysen2017}, with $\norm{\mathbf{m}}$ the length of direction vector $m$. Whether distance weighting yields different feature values depends on several factors. When aggregating the feature values, matrices have to be merged first, otherwise weighting has no effect. Also, it has no effect if the default neighbourhood is used and the Chebyshev norm is using for weighting. Nor does weighting have an effect if either Manhattan or Chebyshev norms are used both for constructing a non-default neighbourhood and for weighting. Weighting may furthermore have no effect for distance $\delta=1$, dependent on distance norms. Because of these exceptions, we recommend against using distance weighting for GLCM.

% JOINT MAXIMUM
\subsection[Joint maximum]{Joint maximum\id{GYBY}}\label{feat_cm_joint_maximum}
\textit{Joint maximum} \citep{Haralick1979} is the probability corresponding to the most common grey level co-occurrence in the GLCM:
\begin{displaymath}
F_{\mathit{cm.joint.max}}=\text{max}(p_{ij})
\end{displaymath}

\vspace{2mm}
\small
% latex table generated in R 3.6.0 by xtable 1.8-4 package
% Mon Dec 02 14:11:54 2019
\begin{longtable}{ccccc}
  \toprule
{\textbf{data}} & {\textbf{aggr. method}} & {\textbf{value}} & {\textbf{tol.}} & {\textbf{consensus}} \\ 
  \midrule
dig. phantom & 2D, averaged & 0.519 & \textemdash & very strong \\ 
  dig. phantom & 2D, slice-merged & 0.512 & \textemdash & strong \\ 
  dig. phantom & 2.5D, direction-merged & 0.489 & \textemdash & strong \\ 
  dig. phantom & 2.5D, merged & 0.492 & \textemdash & strong \\ 
  dig. phantom & 3D, averaged & 0.503 & \textemdash & very strong \\ 
  dig. phantom & 3D, merged & 0.509 & \textemdash & very strong \\ 
  config. A & 2D, averaged & 0.109 & 0.001 & strong \\ 
  config. A & 2D, slice-merged & 0.109 & 0.001 & strong \\ 
  config. A & 2.5D, direction-merged & 0.0943 & 0.0008 & strong \\ 
  config. A & 2.5D, merged & 0.0943 & 0.0008 & strong \\ 
  config. B & 2D, averaged & 0.156 & 0.002 & strong \\ 
  config. B & 2D, slice-merged & 0.156 & 0.002 & strong \\ 
  config. B & 2.5D, direction-merged & 0.126 & 0.002 & strong \\ 
  config. B & 2.5D, merged & 0.126 & 0.002 & strong \\ 
  config. C & 3D, averaged & 0.111 & 0.002 & strong \\ 
  config. C & 3D, merged & 0.111 & 0.002 & very strong \\ 
  config. D & 3D, averaged & 0.232 & 0.007 & strong \\ 
  config. D & 3D, merged & 0.232 & 0.007 & strong \\ 
  config. E & 3D, averaged & 0.153 & 0.003 & moderate \\ 
  config. E & 3D, merged & 0.153 & 0.003 & strong \\ 
   \bottomrule
\caption{Reference values for the \textit{joint maximum} feature.} 
\end{longtable}

\normalsize
\FloatBarrier

% JOINT AVERAGE
\subsection[Joint average]{Joint average\id{60VM}}\label{feat_cm_joint_average}
\textit{Joint average} \citep{Unser1986} is the grey level weighted sum of joint probabilities:
\begin{displaymath}
F_{\mathit{cm.joint.avg}}=\sum_{i=1}^{N_g} \sum_{j=1}^{N_g} i\, p_{ij}
\end{displaymath}

\input{reference_values/cm_joint_avg.txt}

% JOINT VARIANCE
\subsection[Joint variance]{Joint variance\id{UR99}}\label{feat_cm_joint_variance}
The \textit{joint variance} \citep{Unser1986}, which is also called \textit{sum of squares} \citep{Haralick1973}, is defined as:
\begin{displaymath}
F_{\mathit{cm.joint.var}}=\sum_{i=1}^{N_g} \sum_{j=1}^{N_g} \left(i-\mu\right)^2 p_{ij}
\end{displaymath}
Here $\mu$ is equal to the value of $F_{\mathit{cm.joint.avg}}$, which was defined above.

\input{reference_values/cm_joint_var.txt}

% JOINT ENTROPY
\subsection[Joint entropy]{Joint entropy\id{TU9B}}\label{feat_cm_joint_entropy}
\textit{Joint entropy} \citep{Haralick1973} is defined as:
\begin{displaymath}
F_{\mathit{cm.joint.entr}}=-\sum_{i=1}^{N_g} \sum_{j=1}^{N_g} p_{ij} \log_2 p_{ij}
\end{displaymath}

\input{reference_values/cm_joint_entr.txt}

% DIFFERENCE AVERAGE
\subsection[Difference average]{Difference average\id{TF7R}}\label{feat_cm_difference_average}
The \textit{difference average} \citep{Unser1986} for the diagonal probabilities is defined as:
\begin{displaymath}
F_{\mathit{cm.diff.avg}}=\sum_{k=0}^{N_g-1} k\, p_{i-j,k}
\end{displaymath}
By definition \textit{difference average} is equivalent to the \textit{dissimilarity} feature \citep{VanGriethuysen2017}.

\input{reference_values/cm_diff_avg.txt}

% DIFFERENCE VARIANCE
\subsection[Difference variance]{Difference variance\id{D3YU}}\label{feat_cm_difference_variance}
The \textit{difference variance} for the diagonal probabilities \citep{Haralick1973} is defined as:
\begin{displaymath}
F_{\mathit{cm.diff.var}}=\sum_{k=0}^{N_g-1} (k-\mu)^2 p_{i-j,k}
\end{displaymath}
Here $\mu$ is equal to the value of \textit{difference average}.

\input{reference_values/cm_diff_var.txt}

% DIFFERENCE ENTROPY
\subsection[Difference entropy]{Difference entropy\id{NTRS}}\label{feat_cm_difference_entropy}
The \textit{difference entropy} for the diagonal probabilities \citep{Haralick1973} is defined as:
\begin{displaymath}
F_{\mathit{cm.diff.entr}}=-\sum_{k=0}^{N_g-1} p_{i-j,k} \log_2 p_{i-j,k}
\end{displaymath}

\input{reference_values/cm_diff_entr.txt}

% SUM AVERAGE
\subsection[Sum average]{Sum average\id{ZGXS}}\label{feat_cm_sum_average}
The \textit{sum average} for the cross-diagonal probabilities \citep{Haralick1973} is defined as:
\begin{displaymath}
F_{\mathit{cm.sum.avg}}=\sum_{k=2}^{2N_g} k\, p_{i+j,k}
\end{displaymath}
By definition, $F_{\mathit{cm.sum.avg}} = 2 F_{\mathit{cm.joint.avg}}$ \citep{VanGriethuysen2017}.

\input{reference_values/cm_sum_avg.txt}

% SUM VARIANCE
\subsection[Sum variance]{Sum variance\id{OEEB}}\label{feat_cm_sum_variance}
The \textit{sum variance} for the cross-diagonal probabilities \citep{Haralick1973} is defined as:
\begin{displaymath}
F_{\mathit{cm.sum.var}}=\sum_{k=2}^{2N_g} (k-\mu)^2 p_{i+j,k}
\end{displaymath}
Here $\mu$ is equal to the value of \textit{sum average}. \textit{Sum variance} is mathematically identical to the \textit{cluster tendency} feature \citep{VanGriethuysen2017}.

\input{reference_values/cm_sum_var.txt}

% SUM ENTROPY
\subsection[Sum entropy]{Sum entropy\id{P6QZ}}\label{feat_cm_sum_ENTROPY}
The \textit{sum entropy} for the cross-diagonal probabilities \citep{Haralick1973} is defined as:
\begin{displaymath}
F_{\mathit{cm.sum.entr}}=-\sum_{k=2}^{2N_g} p_{i+j,k} \log_2 p_{i+j,k}
\end{displaymath}

\input{reference_values/cm_sum_entr.txt}

% UNIFORMITY
\subsection[Angular second moment]{Angular second moment\id{8ZQL}}\label{feat_cm_uniformity}
The \textit{angular second moment} \citep{Haralick1973}, which represents the energy of $\mathbf{P}_{\Delta}$, is defined as:
\begin{displaymath}
F_{\mathit{cm.energy}} = \sum_{i=1}^{N_g} \sum_{j=1}^{N_g} p_{ij}^2
\end{displaymath}
This feature is also called \textit{energy} \citep{Unser1986,Aerts2014} and \textit{uniformity} \citep{Clausi2002}.

\input{reference_values/cm_energy.txt}

% CONTRAST
\subsection[Contrast]{Contrast\id{ACUI}}\label{feat_cm_contrast}
\textit{Contrast} assesses grey level variations \citep{Haralick1973}. Hence elements of $\mathbf{M}_{\Delta}$ that represent large grey level differences receive greater weight. \textit{Contrast} is defined as \citep{Clausi2002}:
\begin{displaymath}
F_{\mathit{cm.contrast}}= \sum_{i=1}^{N_g} \sum_{j=1}^{N_g} \left(i-j\right)^2 p_{ij}
\end{displaymath}
Note that the original definition by \citet{Haralick1973} is seemingly more complex, but rearranging and simplifying terms leads to the above formulation of \textit{contrast}.

\input{reference_values/cm_contrast.txt}

% DISSIMILARITY
\subsection[Dissimilarity]{Dissimilarity\id{8S9J}}\label{feat_cm_dissimilarity}
\textit{Dissimilarity} \citep{Clausi2002} is conceptually similar to the \textit{contrast} feature, and is defined as:
\begin{displaymath}
F_{\mathit{cm.dissimilarity}}= \sum_{i=1}^{N_g} \sum_{j=1}^{N_g} |i-j|\, p_{ij}
\end{displaymath}
By definition \textit{dissimilarity} is equivalent to the \textit{difference average} feature \citep{VanGriethuysen2017}.

\input{reference_values/cm_dissimilarity.txt}

% INVERSE DIFFERENCE
\subsection[Inverse difference]{Inverse difference\id{IB1Z}}\label{feat_cm_inverse_difference}
\textit{Inverse difference} is a measure of homogeneity \citep{Clausi2002}. Grey level co-occurrences with a large difference in levels are weighed less, thus lowering the total feature value. The feature score is maximal if all grey levels are the same. Inverse difference is defined as:
\begin{displaymath}
F_{\mathit{cm.inv.diff}}=\sum_{i=1}^{N_g} \sum_{j=1}^{N_g} \frac{p_{ij}}{1+|i-j|}
\end{displaymath}
The equation above may also be expressed in terms of diagonal probabilities\citep{VanGriethuysen2017}:
\begin{displaymath}
F_{\mathit{cm.inv.diff}}=\sum_{k=0}^{N_g-1} \frac{p_{i-j,k}}{1+k}
\end{displaymath}

\input{reference_values/cm_inv_diff.txt}

% NORMALISED INVERSE DIFFERENCE 
\subsection[Normalised inverse difference]{Normalised inverse difference\id{NDRX}}\label{feat_cm_inverse_difference_normalised}
\citet{Clausi2002} suggested normalising \textit{inverse difference} to improve classification ability. The normalised feature is then defined as:
\begin{displaymath}
F_{\mathit{cm.inv.diff.norm}}=\sum_{i=1}^{N_g} \sum_{j=1}^{N_g} \frac{p_{ij}}{1+|i-j|/N_g}
\end{displaymath}
Note that in Clausi's definition, $|i-j|^2/N_g^2$ is used instead of $|i-j|/N_g$, which is likely an oversight, as this exactly matches the definition of the \textit{normalised inverse difference moment} feature.

\noindent The equation may also be expressed in terms of diagonal probabilities\citep{VanGriethuysen2017}:
\begin{displaymath}
F_{\mathit{cm.inv.diff.norm}}=\sum_{k=0}^{N_g-1} \frac{p_{i-j,k}}{1+k/N_g}
\end{displaymath}

\input{reference_values/cm_inv_diff_norm.txt}

\layoutvspace

% INVERSE DIFFERENCE MOMENT
\subsection[Inverse difference moment]{Inverse difference moment\id{WF0Z}}\label{feat_cm_inverse_difference_moment}
\textit{Inverse difference moment} \citep{Haralick1973} is similar in concept to the \textit{inverse difference} feature, but with lower weights for elements that are further from the diagonal:
\begin{displaymath}
F_{\mathit{cm.inv.diff.mom}}=\sum_{i=1}^{N_g} \sum_{j=1}^{N_g} \frac{p_{ij}}{1+\left(i-j\right)^2}
\end{displaymath}

\noindent The equation above may also be expressed in terms of diagonal probabilities\citep{VanGriethuysen2017}:
\begin{displaymath}
F_{\mathit{cm.inv.diff.mom}}=\sum_{k=0}^{N_g-1} \frac{p_{i-j,k}}{1+k^2}
\end{displaymath}

\noindent This feature is also called \textit{homogeneity} \citep{Unser1986}.

\input{reference_values/cm_inv_diff_mom.txt}

% NORMALISED INVERSE DIFFERENCE MOMENT
\subsection[Normalised inverse difference moment]{Normalised inverse difference moment\id{1QCO}}\label{feat_cm_inverse_difference_moment_normalised}
\citet{Clausi2002} suggested normalising \textit{inverse difference moment} to improve classification performance. This leads to the following definition:
\begin{displaymath}
F_{\mathit{cm.inv.diff.mom.norm}}=\sum_{i=1}^{N_g} \sum_{j=1}^{N_g} \frac{p_{ij}}{1+\left(i-j\right)^2/N_g^2}
\end{displaymath}

\noindent The equation above may also be expressed in terms of diagonal probabilities\citep{VanGriethuysen2017}:
\begin{displaymath}
F_{\mathit{cm.inv.diff.mom.norm}}=\sum_{k=0}^{N_g-1} \frac{p_{i-j,k}}{1+\left(k/N_g\right)^2}
\end{displaymath}

\input{reference_values/cm_inv_diff_mom_norm.txt}

% INVERSE VARIANCE
\subsection[Inverse variance]{Inverse variance\id{E8JP}}\label{feat_cm_inverse_variance}
The \textit{inverse variance} \citep{Aerts2014} feature is defined as:
\begin{displaymath}
F_{\mathit{cm.inv.var}}=2\sum_{i=1}^{N_g} \sum_{j>i}^{N_g} \frac{p_{ij}}{\left(i-j\right)^2}
\end{displaymath}

The equation above may also be expressed in terms of diagonal probabilities. Note that in this case, summation starts at $k=1$ instead of $k=0$\citep{VanGriethuysen2017}:
\begin{displaymath}
F_{\mathit{cm.inv.var}}=\sum_{k=1}^{N_g-1} \frac{p_{i-j,k}}{k^2}
\end{displaymath}

\input{reference_values/cm_inv_var.txt}

% CORRELATION
\subsection[Correlation]{Correlation\id{NI2N}}\label{feat_cm_correlation}
\textit{Correlation} \citep{Haralick1973} is defined as:
\begin{displaymath}
F_{\mathit{cm.corr}}=\frac{1}{\sigma_{i.}\,\sigma_{.j}} \left(-\mu_{i.}\,\mu_{.j} + \sum_{i=1}^{N_g} \sum_{j=1}^{N_g} i\,j\,p_{ij}\right)
\end{displaymath}
$\mu_{i.}=\sum_{i=1}^{N_g}i\,p_{i.}$ and $\sigma_{i.}=\left(\sum_{i=1}^{N_g} (i-\mu_{i.})^2 p_{i.}\right)^{1/2}$ are the mean and standard deviation of row marginal probability $p_{i.}$, respectively. Likewise, $\mu_{.j}$ and $\sigma_{.j}$ are the mean and standard deviation of the column marginal probability $p_{.j}$, respectively. The calculation of \textit{correlation} can be simplified since $\mathbf{P}_{\Delta}$ is symmetrical:
\begin{displaymath}
F_{\mathit{cm.corr}}=\frac{1}{\sigma_{i.}^2} \left(-\mu_{i.}^2 + \sum_{i=1}^{N_g} \sum_{j=1}^{N_g} i\,j\,p_{ij}\right)
\end{displaymath}
An equivalent formulation of \textit{correlation} is:
\begin{displaymath}
F_{\mathit{cm.corr}}=\frac{1}{\sigma_{i.}\,\sigma_{.j}} \sum_{i=1}^{N_g} \sum_{j=1}^{N_g} \left(i-\mu_{i.}\right) \left(j-\mu_{.j}\right)p_{ij}
\end{displaymath}
Again, simplifying due to matrix symmetry yields:
\begin{displaymath}
F_{\mathit{cm.corr}}=\frac{1}{\sigma_{i.}^2} \sum_{i=1}^{N_g} \sum_{j=1}^{N_g} \left(i-\mu_{i.}\right) \left(j-\mu_{i.}\right)p_{ij}
\end{displaymath}

\input{reference_values/cm_corr.txt}

% AUTOCORRELATION
\subsection[Autocorrelation]{Autocorrelation\id{QWB0}}\label{feat_cm_autocorrelation}
\citet{soh1999texture} defined \textit{autocorrelation} as:
\begin{displaymath}
F_{\mathit{cm.auto.corr}}=\sum_{i=1}^{N_g} \sum_{j=1}^{N_g} i\,j\,p_{ij}
\end{displaymath}

\input{reference_values/cm_auto_corr.txt}

% CLUSTER TENDENCY
\subsection[Cluster tendency]{Cluster tendency\id{DG8W}}\label{feat_cm_cluster_tendency}
\textit{Cluster tendency} \citep{Aerts2014} is defined as:
\begin{displaymath}
F_{\mathit{cm.clust.tend}}=\sum_{i=1}^{N_g} \sum_{j=1}^{N_g} \left(i+j-\mu_{i.}-\mu_{.j}\right)^2 p_{ij}
\end{displaymath}
Here $\mu_{i.}=\sum_{i=1}^{N_g} i\, p_{i.}$ and $\mu_{.j}=\sum_{j=1}^{N_g} j\, p_{.j}$. Because of the symmetric nature of $\mathbf{P}_{\Delta}$, the feature can also be formulated as:
\begin{displaymath}
F_{\mathit{cm.clust.tend}}=\sum_{i=1}^{N_g} \sum_{j=1}^{N_g} \left(i+j-2\mu_{i.}\right)^2 p_{ij}
\end{displaymath}
\textit{Cluster tendency} is mathematically equal to the \textit{sum variance} feature \citep{VanGriethuysen2017}.

\input{reference_values/cm_clust_tend.txt}

% CLUSTER SHADE
\subsection[Cluster shade]{Cluster shade\id{7NFM}}\label{feat_cm_cluster_shade}
\textit{Cluster shade} \citep{Unser1986} is defined as:
\begin{displaymath}
F_{\mathit{cm.clust.shade}}=\sum_{i=1}^{N_g} \sum_{j=1}^{N_g} \left(i+j-\mu_{i.}-\mu_{.j}\right)^3 p_{ij}
\end{displaymath}
As with \textit{cluster tendency}, $\mu_{i.}=\sum_{i=1}^{N_g} i\, p_{i.}$ and $\mu_{.j}=\sum_{j=1}^{N_g} j\, p_{.j}$. Because of the symmetric nature of $\mathbf{P}_{\Delta}$, the feature can also be formulated as:
\begin{displaymath}
F_{\mathit{cm.clust.shade}}=\sum_{i=1}^{N_g} \sum_{j=1}^{N_g} \left(i+j-2\mu_{i.}\right)^3 p_{ij}
\end{displaymath}

\input{reference_values/cm_clust_shade.txt}

\layoutvspace

% CLUSTER PROMINENCE
\subsection[Cluster prominence]{Cluster prominence\id{AE86}}\label{feat_cm_cluster_prominence}
\textit{Cluster prominence} \citep{Unser1986} is defined as:
\begin{displaymath}
F_{\mathit{cm.clust.prom}}=\sum_{i=1}^{N_g} \sum_{j=1}^{N_g} \left(i+j-\mu_{i.}-\mu_{.j}\right)^4 p_{ij}
\end{displaymath}
As before, $\mu_{i.}=\sum_{i=1}^{N_g} i\, p_{i.}$ and $\mu_{.j}=\sum_{j=1}^{N_g} j\, p_{.j}$. Because of the symmetric nature of $\mathbf{P}_{\Delta}$, the feature can also be formulated as:
\begin{displaymath}
F_{\mathit{cm.clust.prom}}=\sum_{i=1}^{N_g} \sum_{j=1}^{N_g} \left(i+j-2\mu_{i.}\right)^4 p_{ij}
\end{displaymath}

\input{reference_values/cm_clust_prom.txt}

% FIRST MEASURE OF INFORMATION CORRELATION
\subsection[Information correlation 1]{Information correlation 1\id{R8DG}}\label{feat_cm_information_corr_1}
\textit{Information theoretic correlation} is estimated using two different measures \citep{Haralick1973}. For symmetric $\mathbf{P}_{\Delta}$ the first measure is defined as:
\begin{displaymath}
F_{\mathit{cm.info.corr.1}}=\frac{\mathit{HXY}-\mathit{HXY_1}}{\mathit{HX}}
\end{displaymath}
$\mathit{HXY} = -\sum_{i=1}^{N_g} \sum_{j=1}^{N_g} p_{ij} \log_2 p_{ij}$ is the entropy for the joint probability. $\mathit{HX}=-\sum_{i=1}^{N_g} p_{i.} \log_2 p_{i.}$ is the entropy for the row marginal probability, which due to symmetry is equal to the entropy of the column marginal probability. $\mathit{HXY}_1$ is a type of entropy that is defined as:
\begin{displaymath}
\mathit{HXY}_1 = -\sum_{i=1}^{N_g} \sum_{j=1}^{N_g} p_{ij} \log_2 \left(p_{i.} p_{.j}\right)
\end{displaymath}

\input{reference_values/cm_info_corr1.txt}

% SECOND MEASURE OF INFORMATION CORRELATION
\subsection[Information correlation 2]{Information correlation 2\id{JN9H}}\label{feat_cm_information_corr_2}
The \textit{second measure of information theoretic correlation} \citep{Haralick1973} is estimated as follows for symmetric $\mathbf{P}_{\Delta}$:
\begin{displaymath}
F_{\mathit{cm.info.corr.2}}=\sqrt{1-\exp\left(-2\left(\mathit{HXY}_2-\mathit{HXY}\right)\right)}
\end{displaymath}
As earlier, $\mathit{HXY} = -\sum_{i=1}^{N_g} \sum_{j=1}^{N_g} p_{ij} \log_2 p_{ij}$. $\mathit{HXY}_2$ is a type of entropy defined as:
\begin{displaymath}
\mathit{HXY}_2=-\sum_{i=1}^{N_g} \sum_{j=1}^{N_g} p_{i.} p_{.j} \log_2 \left(p_{i.} p_{.j} \right)
\end{displaymath}

\input{reference_values/cm_info_corr2.txt}

\clearpage
\section[Grey level run length based features]{Grey level run length based features\id{TP0I}}\label{sect_glrlm}
The grey level run length matrix (GLRLM) was introduced by \citet{Galloway1975} to define various texture features. Like the grey level co-occurrence matrix, GLRLM also assesses the distribution of discretised grey levels in an image or in a stack of images. However, whereas GLCM assesses co-occurrence of grey levels within neighbouring pixels or voxels, GLRLM assesses run lengths. A run length is defined as the length of a consecutive sequence of pixels or voxels with the same grey level along direction $\mathbf{m}$, which was previously defined in Section \ref{sect_glcm}. The GLRLM then contains the occurrences of runs with length $j$ for a discretised grey level $i$.

A complete example for GLRLM construction from a 2D image is shown in Table \ref{figGLRLM1}. Let $\mathbf{M}_{\mathbf{m}}$ be the $N_g \times N_r$ grey level run length matrix, where $N_g$ is the number of discretised grey levels present in the ROI intensity mask and $N_r$ the maximal possible run length along direction $\mathbf{m}$. Matrix element $r_{ij}$ of the GLRLM is the occurrence of grev level $i$ with run length $j$. Then, let $N_v$ be the total number of voxels in the ROI intensity mask, and $N_s=\sum_{i=1}^{N_g}\sum_{j=1}^{N_r}r_{ij}$ the sum over all elements in $\mathbf{M}_{\mathbf{m}}$. Marginal sums are also defined. Let $r_{i.}$ be the marginal sum of the runs over run lengths $j$ for grey value $i$, that is $r_{i.}=\sum_{j=1}^{N_r} r_{ij}$. Similarly, the marginal sum of the runs over the grey values $i$ for run length $j$ is $r_{.j}=\sum_{i=1}^{N_g} r_{ij}$.

\subsubsection*{Aggregating features}
To improve rotational invariance, GLRLM feature values are computed by aggregating information from the different underlying directional matrices \citep{Depeursinge2017a}. Five methods can be used to aggregate GLRLMs and arrive at a single feature value. A schematic example was previously shown in Figure \ref{figGLCMCalcApproaches}. A feature may be aggregated as follows:
\begin{enumerate}
\item Features are computed from each 2D directional matrix and averaged over 2D directions and slices (\textid{BTW3}).
\item Features are computed from a single matrix after merging 2D directional matrices per slice, and then averaged over slices (\textid{SUJT}).
\item Features are computed from a single matrix after merging 2D directional matrices per direction, and then averaged over directions (\textid{JJUI}).
\item The feature is computed from a single matrix after merging all 2D directional matrices (\textid{ZW7Z}).
\item Features are computed from each 3D directional matrix and averaged over the 3D directions (\textid{ITBB}).
\item The feature is computed from a single matrix after merging all 3D directional matrices (\textid{IAZD}).
\end{enumerate}
In methods 2,3,4 and 6 matrices are merged by summing the run counts of each matrix element $(i,j)$ over the different matrices. Note that when matrices are merged, $N_v$ should likewise be summed to retain consistency. Feature values may dependent strongly on the aggregation method.

\subsubsection*{Distance weighting}
GLRLMs may be weighted for distance by multiplying the run lengths with a weighting factor $w$. By default $w=1$, but $w$ may also be an inverse distance function, e.g. $w=\norm{\mathbf{m}}^{-1}$ or $w=\exp(-\norm{\mathbf{m}}^2)$ \citep{VanGriethuysen2017}, with $\norm{\mathbf{m}}$ the length of direction vector $m$. Whether distance weighting yields different feature values depends on several factors. When aggregating the feature values, matrices have to be merged first, otherwise weighting has no effect. It also has no effect if the Chebyshev norm is used for weighting. Distance weighting is non-standard use, and we caution against it due to potential reproducibility issues.

\begin{table}[th]
\centering
\subcaptionbox{Grey levels}{
	\begin{tabular}{@{}cccc@{}}
		\toprule
		1 & 2 & 2 & 3\\
		1 & 2 & 3 & 3\\
		4 & 2 & 4 & 1\\
		4 & 1 & 2 & 3\\
		\bottomrule
	\end{tabular}}\vspace{1cm}\\
\subcaptionbox{$\mathbf{M}_{\mathbf{m}=\rightarrow}$}{
	\begin{tabular}{@{}cccccc@{}}
		\toprule
		& &\multicolumn{4}{c}{Run length $j$}\\
		& & 1 & 2 & 3 & 4\\
		\midrule
		\multirow{4}{*}{$i$} & 1 & 4 & 0 & 0 & 0\\
		& 2 & 3 & 1 & 0 & 0\\
		& 3 & 2 & 1 & 0 & 0\\
		& 4 & 3 & 0 & 0 & 0\\
		\bottomrule
	\end{tabular}}\qquad
\subcaptionbox{$\mathbf{M}_{\mathbf{m}=\nearrow}$}{
	\begin{tabular}{@{}cccccc@{}}
		\toprule
		& &\multicolumn{4}{c}{Run length $j$}\\
		& & 1 & 2 & 3 & 4\\
		\midrule
		\multirow{4}{*}{$i$} & 1 & 4 & 0 & 0 & 0\\
		& 2 & 3 & 1 & 0 & 0\\
		& 3 & 2 & 1 & 0 & 0\\
		& 4 & 3 & 0 & 0 & 0\\
		\bottomrule
	\end{tabular}}\vspace{1cm}\\
\subcaptionbox{$\mathbf{M}_{\mathbf{m}=\uparrow}$}{
	\begin{tabular}{@{}cccccc@{}}
		\toprule
		& &\multicolumn{4}{c}{Run length $j$}\\
		& & 1 & 2 & 3 & 4\\
		\midrule
		\multirow{4}{*}{$i$} & 1 & 2 & 1 & 0 & 0\\
		& 2 & 2 & 0 & 1 & 0\\
		& 3 & 2 & 1 & 0 & 0\\
		& 4 & 1 & 1 & 0 & 0\\
		\bottomrule
	\end{tabular}}\qquad
\subcaptionbox{$\mathbf{M}_{\mathbf{m}=\nwarrow}$}{
	\begin{tabular}{@{}cccccc@{}}
		\toprule
		& &\multicolumn{4}{c}{Run length $j$}\\
		& & 1 & 2 & 3 & 4\\
		\midrule
		\multirow{4}{*}{$i$} & 1 & 4 & 0 & 0 & 0\\
		& 2 & 3 & 1 & 0 & 0\\
		& 3 & 4 & 0 & 0 & 0\\
		& 4 & 3 & 0 & 0 & 0\\
		\bottomrule
	\end{tabular}}
\caption{Grey level run length matrices for the $0^{\circ}$ (a), $45^{\circ}$ (b), $90^{\circ}$ (c) and $135^{\circ}$ (d) directions. In vector notation these directions are $\mathbf{m}=(1,0)$, $\mathbf{m}=(1,1)$, $\mathbf{m}=(0,1)$ and $\mathbf{m}=(-1,1)$, respectively.}
\label{figGLRLM1}
\end{table}

% SHORT RUNS EMPHASIS
\subsection[Short runs emphasis]{Short runs emphasis\id{22OV}}\label{feat_rlm_short_runs_emphasis}
This feature emphasises short run lengths \citep{Galloway1975}. It is defined as:
\begin{displaymath}
F_{\mathit{rlm.sre}} = \frac{1}{N_s} \sum_{j=1}^{N_r} \frac{r_{.j}}{j^2}
\end{displaymath}

\input{reference_values/rlm_sre.txt}

% LONG RUNS EMPHASIS
\subsection[Long runs emphasis]{Long runs emphasis\id{W4KF}}\label{feat_rlm_long_runs_emphasis}
This feature emphasises long run lengths \citep{Galloway1975}. It is defined as:
\begin{displaymath}
F_{\mathit{rlm.lre}} = \frac{1}{N_s} \sum_{j=1}^{N_r} j^2 r_{.j}
\end{displaymath}

\input{reference_values/rlm_lre.txt}

% LOW GREY LEVEL RUN EMPHASIS
\subsection[Low grey level run emphasis]{Low grey level run emphasis\id{V3SW}}\label{feat_rlm_low_grey_level_run_emphasis}
This feature is a grey level analogue to \textit{short runs emphasis} \citep{Chu1990}. Instead of short run lengths, low grey levels are emphasised. The feature is defined as:
\begin{displaymath}
F_{\mathit{rlm.lgre}}=\frac{1}{N_s} \sum_{i=1}^{N_g} \frac{r_{i.}}{i^2}
\end{displaymath}

\input{reference_values/rlm_lgre.txt}

% HIGH GREY LEVEL RUN EMPHASIS
\subsection[High grey level run emphasis]{High grey level run emphasis\id{G3QZ}}\label{feat_rlm_high_grey_level_run_emphasis}
The \textit{high grey level run emphasis} feature is a grey level analogue to \textit{long runs emphasis} \citep{Chu1990}. The feature emphasises high grey levels, and is defined as:
\begin{displaymath}
F_{\mathit{rlm.hgre}}=\frac{1}{N_s} \sum_{i=1}^{N_g} i^2 r_{i.}
\end{displaymath}

\input{reference_values/rlm_hgre.txt}

% SHORT RUN LOW GREY LEVEL EMPHASIS
\subsection[Short run low grey level emphasis]{Short run low grey level emphasis\id{HTZT}}\label{feat_rlm_short_run_low_grey_level_emphasis}
This feature emphasises runs in the upper left quadrant of the GLRLM, where short run lengths and low grey levels are located \citep{Dasarathy1991}. It is defined as:
\begin{displaymath}
F_{\mathit{rlm.srlge}}=\frac{1}{N_s} \sum_{i=1}^{N_g} \sum_{j=1}^{N_r} \frac{r_{ij}}{i^2 j^2}
\end{displaymath}

\input{reference_values/rlm_srlge.txt}

% SHORT RUN HIGH GREY LEVEL EMPHASIS
\subsection[Short run high grey level emphasis]{Short run high grey level emphasis\id{GD3A}}\label{feat_rlm_short_run_high_grey_level_emphasis}
This feature emphasises runs in the lower left quadrant of the GLRLM, where short run lengths and high grey levels are located \citep{Dasarathy1991}. The feature is defined as:
\begin{displaymath}
F_{\mathit{rlm.srhge}}=\frac{1}{N_s} \sum_{i=1}^{N_g} \sum_{j=1}^{N_r} \frac{i^2 r_{ij}}{j^2}
\end{displaymath}

\input{reference_values/rlm_srhge.txt}

% LONG RUN LOW GREY LEVEL EMPHASIS
\subsection[Long run low grey level emphasis]{Long run low grey level emphasis\id{IVPO}}\label{feat_rlm_long_run_low_grey_level_emphasis}
This feature emphasises runs in the upper right quadrant of the GLRLM, where long run lengths and low grey levels are located \citep{Dasarathy1991}. The feature is defined as:
\begin{displaymath}
F_{\mathit{rlm.lrlge}}=\frac{1}{N_s} \sum_{i=1}^{N_g} \sum_{j=1}^{N_r} \frac{j^2 r_{ij}}{i^2}
\end{displaymath}

\input{reference_values/rlm_lrlge.txt}

% LONG RUN HIGH GREY LEVEL EMPHASIS
\subsection[Long run high grey level emphasis]{Long run high grey level emphasis\id{3KUM}}\label{feat_rlm_long_run_high_grey_level_emphasis}
This feature emphasises runs in the lower right quadrant of the GLRLM, where long run lengths and high grey levels are located \citep{Dasarathy1991}. The feature is defined as:
\begin{displaymath}
F_{\mathit{rlm.lrhge}}=\frac{1}{N_s} \sum_{i=1}^{N_g} \sum_{j=1}^{N_r} i^2 j^2 r_{ij}
\end{displaymath}

\input{reference_values/rlm_lrhge.txt}

% GREY LEVEL NON-UNIFORMITY
\subsection[Grey level non-uniformity]{Grey level non-uniformity\id{R5YN}}\label{feat_rlm_grey_level_non_uniformity}
This feature assesses the distribution of runs over the grey values \citep{Galloway1975}. The feature value is low when runs are equally distributed along grey levels. The feature is defined as:
\begin{displaymath}
F_{\mathit{rlm.glnu}}= \frac{1}{N_s} \sum_{i=1}^{N_g} r_{i.}^2
\end{displaymath}

\input{reference_values/rlm_glnu.txt}

% NORMALISED GREY LEVEL NON-UNIFORMITY
\subsection[Normalised grey level non-uniformity]{Normalised grey level non-uniformity\id{OVBL}}\label{feat_rlm_grey_level_non_uniformity_normalised}
This is a normalised version of the \textit{grey level non-uniformity} feature. It is defined as:
\begin{displaymath}
F_{\mathit{rlm.glnu.norm}}= \frac{1}{N_s^2} \sum_{i=1}^{N_g} r_{i.}^2
\end{displaymath}

\input{reference_values/rlm_glnu_norm.txt}

% RUN LENGTH NON-UNIFORMITY
\subsection[Run length non-uniformity]{Run length non-uniformity\id{W92Y}}\label{feat_rlm_run_length_non_uniformity}
This features assesses the distribution of runs over the run lengths \citep{Galloway1975}. The feature value is low when runs are equally distributed along run lengths. It is defined as:
\begin{displaymath}
F_{\mathit{rlm.rlnu}}= \frac{1}{N_s} \sum_{j=1}^{N_r} r_{.j}^2
\end{displaymath}

\input{reference_values/rlm_rlnu.txt}

% NORMALISED RUN LENGTH NON-UNIFORMITY
\subsection[Normalised run length non-uniformity]{Normalised run length non-uniformity\id{IC23}}\label{feat_rlm_run_length_non_uniformity_normalised}
This is normalised version of the \textit{run length non-uniformity} feature. It is defined as:
\begin{displaymath}
F_{\mathit{rlm.rlnu.norm}}= \frac{1}{N_s^2} \sum_{j=1}^{N_r} r_{.j}^2
\end{displaymath}

\input{reference_values/rlm_rlnu_norm.txt}

% RUN PERCENTAGE
\subsection[Run percentage]{Run percentage\id{9ZK5}}\label{feat_rlm_run_percentage}
This feature measures the fraction of the number of realised runs and the maximum number of potential runs \citep{Galloway1975}. Strongly linear or highly uniform ROI volumes produce a low \textit{run percentage}. It is defined as:
\begin{displaymath}
F_{\mathit{rlm.r.perc}}=\frac{N_s}{N_v}
\end{displaymath}
As noted before, when this feature is calculated using a merged GLRLM, $N_v$ should be the sum of the number of voxels of the underlying matrices to allow proper normalisation.

\input{reference_values/rlm_r_perc.txt}

% GREY LEVEL VARIANCE
\subsection[Grey level variance]{Grey level variance\id{8CE5}}\label{feat_rlm_grey_level_variance}
This feature estimates the variance in runs over the grey levels. Let $p_{ij} = r_{ij}/N_s$ be the joint probability estimate for finding discretised grey level $i$ with run length $j$. \textit{Grey level variance} is then defined as:
\begin{displaymath}
F_{\mathit{rlm.gl.var}}=  \sum_{i=1}^{N_g} \sum_{j=1}^{N_r} (i-\mu)^2 p_{ij}
\end{displaymath}
Here, $\mu = \sum_{i=1}^{N_g} \sum_{j=1}^{N_r} i\,p_{ij}$.

\input{reference_values/rlm_gl_var.txt}

\layoutvspace

% RUN LENGTH VARIANCE
\subsection[Run length variance]{Run length variance\id{SXLW}}\label{feat_rlm_run_length_variance}
This feature estimates the variance in runs over the run lengths. As before let $p_{ij} = r_{ij}/N_s$. The feature is defined as:
\begin{displaymath}
F_{\mathit{rlm.rl.var}}= \sum_{i=1}^{N_g} \sum_{j=1}^{N_r} (j-\mu)^2 p_{ij}
\end{displaymath}
Mean run length is defined as $\mu = \sum_{i=1}^{N_g} \sum_{j=1}^{N_r} j\,p_{ij}$.

\input{reference_values/rlm_rl_var.txt}

% RUN ENTROPY
\subsection[Run entropy]{Run entropy\id{HJ9O}}\label{feat_rlm_run_entropy}
\textit{Run entropy} was investigated by \citet{Albregtsen2000}. Again, let $p_{ij} = r_{ij}/N_s$. The entropy is then defined as:
\begin{displaymath}
F_{\mathit{rlm.rl.entr}} = - \sum_{i=1}^{N_g} \sum_{j=1}^{N_r} p_{ij} \log_2 p_{ij}
\end{displaymath}

\input{reference_values/rlm_rl_entr.txt}

\clearpage
\section[Grey level size zone based features]{Grey level size zone based features\id{9SAK}}\label{sect_glszm}
The grey level size zone matrix (GLSZM) counts the number of groups (or zones) of linked voxels \citep{Thibault2014}. Voxels are linked if the neighbouring voxel has an identical discretised grey level. Whether a voxel classifies as a neighbour depends on its connectedness. In a 3D approach to texture analysis we consider 26-connectedness, which indicates that a center voxel is linked to all of the 26 neighbouring voxels with the same grey level. In the 2 dimensional approach, 8-connectedness is used. A potential issue for the 2D approach is that voxels which may otherwise be considered to belong to the same zone by linking across slices, are now two or more separate zones within the slice plane. Whether this issue negatively affects predictive performance of GLSZM-based features or their reproducibility has not been determined.

Let $\mathbf{M}$ be the $N_g \times N_z$ grey level size zone matrix, where $N_g$ is the number of discretised grey levels present in the ROI intensity mask and $N_z$ the maximum zone size of any group of linked voxels. Element $s_{ij}$ of $\mathbf{M}$ is then the number of zones with discretised grey level $i$ and size $j$. Furthermore, let $N_v$ be the number of voxels in the intensity mask and $N_s=\sum_{i=1}^{N_g}\sum_{j=1}^{N_z}s_{ij}$ be the total number of zones. Marginal sums can likewise be defined. Let $s_{i.}=\sum_{j=1}^{N_z}s_{ij}$ be the number of zones with discretised grey level $i$, regardless of size. Likewise, let $s_{.j}=\sum_{i=1}^{N_g}s_{ij}$ be the number of zones with size $j$, regardless of grey level. A two dimensional example is shown in Table \ref{figGLSZM1}.

\subsubsection*{Aggregating features}
Three methods can be used to aggregate GLSZMs and arrive at a single feature value. A schematic example is shown in Figure \ref{figGLSZMCalcApproaches}. A feature may be aggregated as follows:
\begin{enumerate}
\item Features are computed from 2D matrices and averaged over slices (\textid{8QNN}).
\item The feature is computed from a single matrix after merging all 2D matrices (\textid{62GR}).
\item The feature is computed from a 3D matrix (\textid{KOBO}).
\end{enumerate}
Method 2 involves merging GLSZMs by summing the number of zones $s_{ij}$ over the GLSZM for the different slices. Note that when matrices are merged, $N_v$ should likewise be summed to retain consistency. Feature values may dependent strongly on the aggregation method.

\subsubsection*{Distances}
The default neighbourhood for GLSZM is constructed using Chebyshev distance $\delta=1$. Manhattan or Euclidean norms may also be used to construct a neighbourhood, and both lead to a 6-connected (3D) and 4-connected (2D) neighbourhoods. Larger distances are also technically possible, but will occasionally cause separate zones with the same intensity to be considered as belonging to the same zone. Using different neighbourhoods for determining voxel linkage is non-standard use, and we caution against it due to potential reproducibility issues.

\subsubsection*{Note on feature references}
GLSZM feature definitions are based on the definitions of GLRLM features \citep{Thibault2014}. Hence, references may be found in the section on GLRLM (\ref{sect_glrlm}).

\begin{table}[th]
\centering
\subcaptionbox{Grey levels}{
	\begin{tabular}{@{}cccc@{}}
		\toprule
		1 & 2 & 2 & 3\\
		1 & 2 & 3 & 3\\
		4 & 2 & 4 & 1\\
		4 & 1 & 2 & 3\\
		\bottomrule
	\end{tabular}}\qquad
\subcaptionbox{Grey level size zone matrix}{
	\begin{tabular}{@{}ccccccc@{}}
		\toprule
		& &\multicolumn{5}{c}{Zone size $j$}\\
		& & 1 & 2 & 3 & 4 & 5\\
		\midrule
		\multirow{4}{*}{$i$} & 1 & 2 & 1 & 0 & 0 & 0\\
		& 2 & 0 & 0 & 0 & 0 & 1\\
		& 3 & 1 & 0 & 1 & 0 & 0\\
		& 4 & 1 & 1 & 0 & 0 & 0\\
		\bottomrule
	\end{tabular}}
\caption{Original image with grey levels (a); and corresponding grey level size zone matrix (GLSZM) under 8-connectedness (b). Element $s(i,j)$ of the GLSZM indicates the number of times a zone of $j$ linked pixels and grey level $i$ occurs within the image.}
\label{figGLSZM1}
\end{table}

\begin{figure}[ht]
\centering
\begin{subfigure}[h]{0.45\textwidth}\centering
	\includegraphics[scale=0.55]{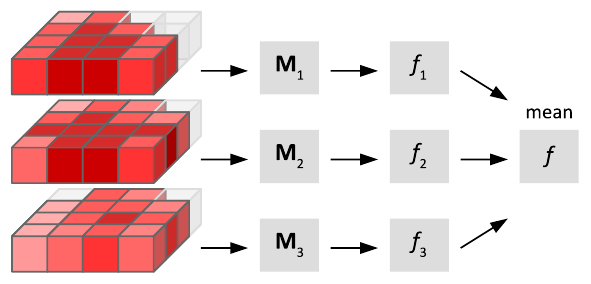}
	\caption{2D: by slice, without merging}
\end{subfigure}
\hfill
\begin{subfigure}[h]{0.45\textwidth}\centering
	\includegraphics[scale=0.55]{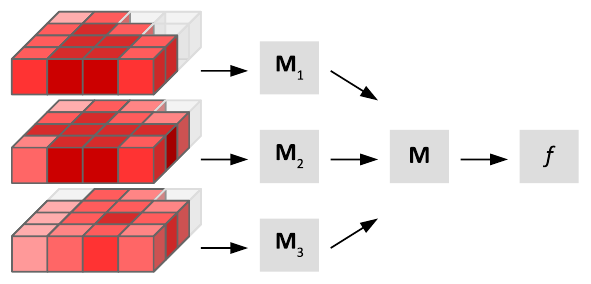}
	\caption{2.5D: by slice, with merging}
\end{subfigure}
\vspace{1cm}\\
\begin{subfigure}[h]{0.45\textwidth}\centering
	\includegraphics[scale=0.55]{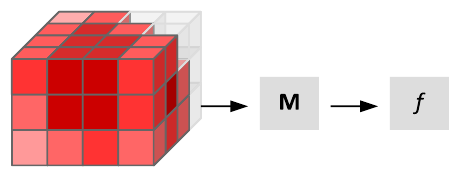}
	\caption{3D: as volume}
\end{subfigure}
\caption{Approaches to calculating grey level size zone matrix-based features. $\mathbf{M}_{k}$ are texture matrices calculated for slice $k$ (if applicable), and $f_{k}$ is the corresponding feature value. In (b) the matrices from the different slices are merged prior to feature calculation.}
\label{figGLSZMCalcApproaches}
\end{figure}

\FloatBarrier
\layoutclearpage

% SMALL ZONE EMPHASIS
\subsection[Small zone emphasis]{Small zone emphasis\id{5QRC}}\label{feat_szm_small_zone_emphasis}
This feature emphasises small zones. It is defined as:
\begin{displaymath}
F_{\mathit{szm.sze}} = \frac{1}{N_s} \sum_{j=1}^{N_z} \frac{s_{.j}}{j^2}
\end{displaymath}

\vspace{2mm}
\small
% latex table generated in R 3.6.0 by xtable 1.8-4 package
% Mon Dec 02 14:11:56 2019
\begin{longtable}{ccccc}
  \toprule
{\textbf{data}} & {\textbf{aggr. method}} & {\textbf{value}} & {\textbf{tol.}} & {\textbf{consensus}} \\ 
  \midrule
dig. phantom & 2D & 0.363 & \textemdash & strong \\ 
  dig. phantom & 2.5D & 0.368 & \textemdash & strong \\ 
  dig. phantom & 3D & 0.255 & \textemdash & very strong \\ 
  config. A & 2D & 0.688 & 0.003 & strong \\ 
  config. A & 2.5D & 0.68 & 0.003 & strong \\ 
  config. B & 2D & 0.745 & 0.003 & strong \\ 
  config. B & 2.5D & 0.741 & 0.003 & strong \\ 
  config. C & 3D & 0.695 & 0.001 & strong \\ 
  config. D & 3D & 0.637 & 0.005 & strong \\ 
  config. E & 3D & 0.676 & 0.003 & strong \\ 
   \bottomrule
\caption{Reference values for the \textit{small zone emphasis} feature.} 
\end{longtable}

\normalsize
\FloatBarrier

% LARGE ZONE EMPHASIS
\subsection[Large zone emphasis]{Large zone emphasis\id{48P8}}\label{feat_szm_large_zone_emphasis}
This feature emphasises large zones. It is defined as:
\begin{displaymath}
F_{\mathit{szm.lze}} = \frac{1}{N_s} \sum_{j=1}^{N_z} j^2 s_{.j}
\end{displaymath}

\input{reference_values/szm_lze.txt}

% LOW GREY LEVEL ZONE EMPHASIS
\subsection[Low grey level zone emphasis]{Low grey level zone emphasis\id{XMSY}}\label{feat_szm_low_grey_level_zone_emphasis}
This feature is a grey level analogue to \textit{small zone emphasis}. Instead of small zone sizes, low grey levels are emphasised. The feature is defined as:
\begin{displaymath}
F_{\mathit{szm.lgze}}=\frac{1}{N_s} \sum_{i=1}^{N_g} \frac{s_{i.}}{i^2}
\end{displaymath}

\input{reference_values/szm_lgze.txt}

% HIGH GREY LEVEL ZONE EMPHASIS
\subsection[High grey level zone emphasis]{High grey level zone emphasis\id{5GN9}}\label{feat_szm_high_grey_level_zone_emphasis}
The\textit{ high grey level zone emphasis} feature is a grey level analogue to \textit{large zone emphasis}. The feature emphasises high grey levels, and is defined as:
\begin{displaymath}
F_{\mathit{szm.hgze}}=\frac{1}{N_s} \sum_{i=1}^{N_g} i^2 s_{i.}
\end{displaymath}

\input{reference_values/szm_hgze.txt}

% SMALL ZONE LOW GREY LEVEL EMPHASIS
\subsection[Small zone low grey level emphasis]{Small zone low grey level emphasis\id{5RAI}}\label{feat_szm_small_zone_low_grey_level_emphasis}
This feature emphasises zone counts within the upper left quadrant of the GLSZM, where small zone sizes and low grey levels are located. It is defined as:
\begin{displaymath}
F_{\mathit{szm.szlge}}=\frac{1}{N_s} \sum_{i=1}^{N_g} \sum_{j=1}^{N_z} \frac{s_{ij}}{i^2 j^2}
\end{displaymath}

\input{reference_values/szm_szlge.txt}

% SMALL ZONE HIGH GREY LEVEL EMPHASIS
\subsection[Small zone high grey level emphasis]{Small zone high grey level emphasis\id{HW1V}}\label{feat_szm_small_zone_high_grey_level_emphasis}
This feature emphasises zone counts in the lower left quadrant of the GLSZM, where small zone sizes and high grey levels are located. The feature is defined as:
\begin{displaymath}
F_{\mathit{szm.szhge}}=\frac{1}{N_s} \sum_{i=1}^{N_g} \sum_{j=1}^{N_z} \frac{i^2 s_{ij}}{j^2}
\end{displaymath}

\input{reference_values/szm_szhge.txt}

% LARGE ZONE LOW GREY LEVEL EMPHASIS
\subsection[Large zone low grey level emphasis]{Large zone low grey level emphasis\id{YH51}}\label{feat_szm_large_zone_low_grey_level_emphasis}
This feature emphasises zone counts in the upper right quadrant of the GLSZM, where large zone sizes and low grey levels are located. The feature is defined as:
\begin{displaymath}
F_{\mathit{szm.lzlge}}=\frac{1}{N_s} \sum_{i=1}^{N_g} \sum_{j=1}^{N_z} \frac{j^2 s_{ij}}{i^2}
\end{displaymath}

\input{reference_values/szm_lzlge.txt}

% LARGE ZONE HIGH GREY LEVEL EMPHASIS
\subsection[Large zone high grey level emphasis]{Large zone high grey level emphasis\id{J17V}}\label{feat_szm_large_zone_high_grey_level_emphasis}
This feature emphasises zone counts in the lower right quadrant of the GLSZM, where large zone sizes and high grey levels are located. The feature is defined as:
\begin{displaymath}
F_{\mathit{szm.lzhge}}=\frac{1}{N_s} \sum_{i=1}^{N_g} \sum_{j=1}^{N_z} i^2 j^2 s_{ij}
\end{displaymath}

\input{reference_values/szm_lzhge.txt}

% GREY LEVEL NON-UNIFORMITY
\subsection[Grey level non-uniformity]{Grey level non-uniformity\id{JNSA}}\label{feat_szm_grey_level_non_uniformity}
This feature assesses the distribution of zone counts over the grey values. The feature value is low when zone counts are equally distributed along grey levels. The feature is defined as:
\begin{displaymath}
F_{\mathit{szm.glnu}}= \frac{1}{N_s} \sum_{i=1}^{N_g} s_{i.}^2
\end{displaymath}

\input{reference_values/szm_glnu.txt}

% NORMALISED GREY LEVEL NON-UNIFORMITY
\subsection[Normalised grey level non-uniformity]{Normalised grey level non-uniformity\id{Y1RO}}\label{feat_szm_grey_level_non_uniformity_normalised}
This is a normalised version of the \textit{grey level non-uniformity} feature. It is defined as:
\begin{displaymath}
F_{\mathit{szm.glnu.norm}}= \frac{1}{N_s^2} \sum_{i=1}^{N_g} s_{i.}^2
\end{displaymath}

\input{reference_values/szm_glnu_norm.txt}

% ZONE SIZE NON-UNIFORMITY
\subsection[Zone size non-uniformity]{Zone size non-uniformity\id{4JP3}}\label{feat_szm_zone_size_non_uniformity}
This features assesses the distribution of zone counts over the different zone sizes. \textit{Zone size non-uniformity} is low when zone counts are equally distributed along zone sizes. It is defined as:
\begin{displaymath}
F_{\mathit{szm.zsnu}}= \frac{1}{N_s} \sum_{j=1}^{N_z} s_{.j}^2
\end{displaymath}

\input{reference_values/szm_zsnu.txt}

% NORMALISED ZONE SIZE NON-UNIFORMITY
\subsection[Normalised zone size non-uniformity]{Normalised zone size non-uniformity\id{VB3A}}\label{feat_szm_zone_size_non_uniformity_normalised}
This is a normalised version of \textit{zone size non-uniformity}. It is defined as:
\begin{displaymath}
F_{\mathit{szm.zsnu.norm}}= \frac{1}{N_s^2} \sum_{i=1}^{N_z} s_{.j}^2
\end{displaymath}

\input{reference_values/szm_zsnu_norm.txt}

% ZONE PERCENTAGE
\subsection[Zone percentage]{Zone percentage\id{P30P}}\label{feat_szm_zone_percentage}
This feature measures the fraction of the number of realised zones and the maximum number of potential zones. Highly uniform ROIs produce a low \textit{zone percentage}. It is defined as:
\begin{displaymath}
F_{\mathit{szm.z.perc}}=\frac{N_s}{N_v}
\end{displaymath}

\input{reference_values/szm_z_perc.txt}

% GREY LEVEL VARIANCE
\subsection[Grey level variance]{Grey level variance\id{BYLV}}\label{feat_szm_grey_level_variance}
This feature estimates the variance in zone counts over the grey levels. Let $p_{ij} = s_{ij}/N_s$ be the joint probability estimate for finding zones with discretised grey level $i$ and size $j$. The feature is then defined as:
\begin{displaymath}
F_{\mathit{szm.gl.var}}=  \sum_{i=1}^{N_g} \sum_{j=1}^{N_z} (i-\mu)^2 p_{ij}
\end{displaymath}
Here, $\mu = \sum_{i=1}^{N_g} \sum_{j=1}^{N_z} i\,p_{ij}$.

\input{reference_values/szm_gl_var.txt}

% ZONE SIZE VARIANCE
\subsection[Zone size variance]{Zone size variance\id{3NSA}}\label{feat_szm_zone_size_variance}
This feature estimates the variance in zone counts over the different zone sizes. As before let $p_{ij} = s_{ij}/N_s$. The feature is defined as:
\begin{displaymath}
F_{\mathit{szm.zs.var}}= \sum_{i=1}^{N_g} \sum_{j=1}^{N_z} (j-\mu)^2 p_{ij}
\end{displaymath}
Mean zone size is defined as $\mu = \sum_{i=1}^{N_g} \sum_{j=1}^{N_z} j\,p_{ij}$.

\input{reference_values/szm_zs_var.txt}

% ZONE SIZE ENTROPY
\subsection[Zone size entropy]{Zone size entropy\id{GU8N}}\label{feat_szm_zone_size_entropy}
Let $p_{ij} = s_{ij}/N_s$. \textit{Zone size entropy} is then defined as:
\begin{displaymath}
F_{\mathit{szm.zs.entr}} = - \sum_{i=1}^{N_g} \sum_{j=1}^{N_z} p_{ij} \log_2 p_{ij}
\end{displaymath}

\input{reference_values/szm_zs_entr.txt}

\clearpage
\section[Grey level distance zone based features]{Grey level distance zone based features\id{VMDZ}} \label{sect_gldzm}
The grey level distance zone matrix (GLDZM) counts the number of groups (or zones) of linked voxels which share a specific discretised grey level value and possess the same distance to ROI edge \citep{Thibault2014}. The GLDZM thus captures the relation between location and grey level. Two maps are required to calculate the GLDZM. The first is a grey level zone map, which is identical to the one created for the grey level size zone matrix (GLSZM), see Section \ref{sect_glszm}. The second is a distance map, which will be described in detail later.

As with GSLZM, neighbouring voxels are linked if they share the same grey level value. Whether a voxel classifies as a neighbour depends on its connectedness. We consider 26-connectedness for a 3D approach and 8-connectedness in the 2D approach.

The distance to the ROI edge is defined according to 6 and 4-connectedness for 3D and 2D, respectively. Because of the connectedness definition used, the distance of a voxel to the outer border is equal to the minimum number edges of neighbouring voxels that need to be crossed to reach the ROI edge. The distance for a linked group of voxels with the same grey value is equal to the minimum distance for the respective voxels in the distance map.

Our definition deviates from the original by \citet{Thibault2014}. The original was defined in a rectangular 2D image, whereas ROIs are rarely rectangular cuboids. Approximating distance using Chamfer maps is then no longer a fast and easy solution. Determining distance iteratively in 6 or 4-connectedness is a relatively efficient solution, implemented as follows:
\begin{enumerate}
\item The ROI mask is morphologically eroded using the appropriate (6 or 4-connected) structure element.
\item All eroded ROI voxels are updated in the distance map by adding 1.
\item The above steps are performed iteratively until the ROI mask is empty.
\end{enumerate}
A second difference with the original definition is that the lowest possible distance is $1$ instead of $0$ for voxels directly on the ROI edge. This prevents division by $0$ for some features.

Let $\mathbf{M}$ be the $N_g \times N_d$ grey level size zone matrix, where $N_g$ is the number of discretised grey levels present in the ROI intensity mask and $N_d$ the largest distance of any zone. Element $d_{ij}=d(i,j)$ of $\mathbf{M}$ is then number of zones with discretised grey level $i$ and distance $j$. Furthermore, let $N_v$ be the number of voxels and $N_s=\sum_{i=1}^{N_g}\sum_{j=1}^{N_d}d_{ij}$ be the total zone count. Marginal sums can likewise be defined. Let $d_{i.}=\sum_{j=1}^{N_d}d_{ij}$ be the number of zones with discretised grey level $i$, regardless of distance. Likewise, let $d_{.j}=\sum_{i=1}^{N_g}d_{ij}$ be the number of zones with distance $j$, regardless of grey level. A two dimensional example is shown in Table \ref{figGLDZM1}.

\begin{table}[th]
\centering
\subcaptionbox{Grey levels}{
	\begin{tabular}{@{}cccc@{}}
		\toprule
		1 & 2 & 2 & 3\\
		1 & 2 & 3 & 3\\
		4 & 2 & 4 & 1\\
		4 & 1 & 2 & 3\\
		\bottomrule
	\end{tabular}}\qquad
\subcaptionbox{Distance map}{
	\begin{tabular}{@{}cccc@{}}
		\toprule
		1 & 1 & 1 & 1\\
		1 & 2 & 2 & 1\\
		1 & 2 & 2 & 1\\
		1 & 1 & 1 & 1\\
		\bottomrule
	\end{tabular}}\qquad
\subcaptionbox{Grey level distance zone matrix}{
	\begin{tabular}{@{}cccc@{}}
		\toprule
		& &\multicolumn{2}{c}{$j$}\\
		& & 1 & 2\\
		\midrule
		\multirow{4}{*}{$i$} & 1 & 3 & 0\\
		& 2 & 2 & 0\\
		& 3 & 2 & 0\\
		& 4 & 1 & 1\\
		\bottomrule
	\end{tabular}}
\caption{Original image with grey levels (a); corresponding distance map for distance to border (b); and corresponding grey level distance zone matrix (GLDZM) under 4-connectedness (c). Element $d(i,j)$ of the GLDZM indicates the number of times a zone with grey level $i$ and a minimum distance to border $j$ occurs within the image.}
\label{figGLDZM1}
\end{table}

\subsubsection*{Morphological and intensity masks.}
The GLDZM is special in that it uses both ROI masks. The distance map is determined using the morphological ROI mask, whereas the intensity mask is used for determining the zones, as with the GLSZM.

\subsubsection*{Aggregating features}
Three methods can be used to aggregate GLDZMs and arrive at a single feature value. A schematic example was previously shown in Figure \ref{figGLSZMCalcApproaches}. A feature may be aggregated as follows:
\begin{enumerate}
\item Features are computed from 2D matrices and averaged over slices (\textid{8QNN}).
\item The feature is computed from a single matrix after merging all 2D matrices (\textid{62GR}).
\item The feature is computed from a 3D matrix (\textid{KOBO}).
\end{enumerate}
Method 2 involves merging GLDZMs by summing the number of zones $d_{ij}$ over the GLDZM for the different slices. Note that when matrices are merged, $N_v$ should likewise be summed to retain consistency. Feature values may dependent strongly on the aggregation method.

\subsubsection*{Distances}
In addition to the use of different distance norms to determine voxel linkage, as described in section \ref{sect_glszm}, different distance norms may be used to determine distance of zones to the boundary. The default is to use the Manhattan norm which allows for a computationally efficient implementation, as described above. A similar implementation is possible using the Chebyshev norm, as it merely changes connectedness of the structure element. Implementations using an Euclidean distance norm are less efficient as this demands searching for the nearest non-ROI voxel for each of the $N_v$ voxels in the ROI. An added issue is that Euclidean norms may lead to a wide range of different distances $j$ that require rounding before constructing the grey level distance zone matrix $\mathbf{M}$. Using different distance norms is non-standard use, and we caution against it due to potential reproducibility issues.

\subsubsection*{Note on feature references}
GLDZM feature definitions are based on the definitions of GLRLM features \citep{Thibault2014}. Hence, references may be found in the section on GLRLM (\ref{sect_glrlm}).

\layoutnewpage

% SMALL DISTANCE EMPHASIS
\subsection{Small distance emphasis\id{0GBI}}\label{feat_dzm_small_distance_emphasis}
This feature emphasises small distances. It is defined as:
\begin{displaymath}
F_{\mathit{dzm.sde}} = \frac{1}{N_s} \sum_{j=1}^{N_d} \frac{d_{.j}}{j^2}
\end{displaymath}

\input{reference_values/dzm_sde.txt}

% LARGE DISTANCE EMPHASIS
\subsection{Large distance emphasis\id{MB4I}}\label{feat_dzm_large_distance_emphasis}
This feature emphasises large distances. It is defined as:
\begin{displaymath}
F_{\mathit{dzm.lde}} = \frac{1}{N_s} \sum_{j=1}^{N_d} j^2 d_{.j}
\end{displaymath}

\input{reference_values/dzm_lde.txt}

% LOW GREY LEVEL ZONE EMPHASIS
\subsection[Low grey level zone emphasis]{Low grey level zone emphasis\id{S1RA}}\label{feat_dzm_low_grey_level_zone_emphasis}
This feature is a grey level analogue to \textit{small distance emphasis}. Instead of small zone distances, low grey levels are emphasised. The feature is defined as:
\begin{displaymath}
F_{\mathit{dzm.lgze}}=\frac{1}{N_s} \sum_{i=1}^{N_g} \frac{d_{i.}}{i^2}
\end{displaymath}

\input{reference_values/dzm_lgze.txt}

% HIGH GREY LEVEL ZONE EMPHASIS
\subsection[High grey level zone emphasis]{High grey level zone emphasis\id{K26C}}\label{feat_dzm_high_grey_level_zone_emphasis}
The \textit{high grey level zone emphasis} feature is a grey level analogue to \textit{large distance emphasis}. The feature emphasises high grey levels, and is defined as:
\begin{displaymath}
F_{\mathit{dzm.hgze}}=\frac{1}{N_s} \sum_{i=1}^{N_g} i^2 d_{i.}
\end{displaymath}

\input{reference_values/dzm_hgze.txt}

% SMALL DISTANCE LOW GREY LEVEL EMPHASIS
\subsection[Small distance low grey level emphasis]{Small distance low grey level emphasis\id{RUVG}}\label{feat_dzm_small_distance_low_grey_level_emphasis}
This feature emphasises runs in the upper left quadrant of the GLDZM, where small zone distances and low grey levels are located. It is defined as:
\begin{displaymath}
F_{\mathit{dzm.sdlge}}=\frac{1}{N_s} \sum_{i=1}^{N_g} \sum_{j=1}^{N_d} \frac{d_{ij}}{i^2 j^2}
\end{displaymath}

\input{reference_values/dzm_sdlge.txt}

% SMALL DISTANCE HIGH GREY LEVEL EMPHASIS
\subsection[Small distance high grey level emphasis]{Small distance high grey level emphasis\id{DKNJ}}\label{feat_dzm_small_distance_high_grey_level_emphasis}
This feature emphasises runs in the lower left quadrant of the GLDZM, where small zone distances and high grey levels are located. \textit{Small distance high grey level emphasis} is defined as:
\begin{displaymath}
F_{\mathit{dzm.sdhge}}=\frac{1}{N_s} \sum_{i=1}^{N_g} \sum_{j=1}^{N_d} \frac{i^2 d_{ij}}{j^2}
\end{displaymath}

\input{reference_values/dzm_sdhge.txt}

% LARGE DISTANCE LOW GREY LEVEL EMPHASIS
\subsection[Large distance low grey level emphasis]{Large distance low grey level emphasis\id{A7WM}}\label{feat_dzm_large_distance_low_grey_level_emphasis}
This feature emphasises runs in the upper right quadrant of the GLDZM, where large zone distances and low grey levels are located. The feature is defined as:
\begin{displaymath}
F_{\mathit{dzm.ldlge}}=\frac{1}{N_s} \sum_{i=1}^{N_g} \sum_{j=1}^{N_d} \frac{j^2 d_{ij}}{i^2}
\end{displaymath}

\input{reference_values/dzm_ldlge.txt}

% LARGE DISTANCE HIGH GREY LEVEL EMPHASIS
\subsection[Large distance high grey level emphasis]{Large distance high grey level emphasis\id{KLTH}}\label{feat_dzm_large_distance_high_grey_level_emphasis}
This feature emphasises runs in the lower right quadrant of the GLDZM, where large zone distances and high grey levels are located. The \textit{large distance high grey level emphasis} feature is defined as:
\begin{displaymath}
F_{\mathit{dzm.ldhge}}=\frac{1}{N_s} \sum_{i=1}^{N_g} \sum_{j=1}^{N_d} i^2 j^2 d_{ij}
\end{displaymath}

\input{reference_values/dzm_ldhge.txt}

% GREY LEVEL NON-UNIFORMITY
\subsection[Grey level non-uniformity]{Grey level non-uniformity\id{VFT7}}\label{feat_dzm_grey_level_non_uniformity}
This feature measures the distribution of zone counts over the grey values. \textit{Grey level non-uniformity} is low when zone counts are equally distributed along grey levels. The feature is defined as:
\begin{displaymath}
F_{\mathit{dzm.glnu}}= \frac{1}{N_s} \sum_{i=1}^{N_g} d_{i.}^2
\end{displaymath}

\input{reference_values/dzm_glnu.txt}

% NORMALIZED GREY LEVEL NON-UNIFORMITY
\subsection[Normalised grey level non-uniformity]{Normalised grey level non-uniformity\id{7HP3}}\label{feat_dzm_grey_level_non_uniformity_normalised}
This is a normalised version of the \textit{grey level non-uniformity} feature. It is defined as:
\begin{displaymath}
F_{\mathit{dzm.glnu.norm}}= \frac{1}{N_s^2} \sum_{i=1}^{N_g} d_{i.}^2
\end{displaymath}

\input{reference_values/dzm_glnu_norm.txt}

% ZONE DISTANCE NON-UNIFORMITY
\subsection[Zone distance non-uniformity]{Zone distance non-uniformity\id{V294}}\label{feat_dzm_zone_distance_non_uniformity}
\textit{Zone distance non-uniformity} measures the distribution of zone counts over the different zone distances. \textit{Zone distance non-uniformity} is low when zone counts are equally distributed along zone distances. It is defined as:
\begin{displaymath}
F_{\mathit{dzm.zdnu}}= \frac{1}{N_s} \sum_{j=1}^{N_d} d_{.j}^2
\end{displaymath}

\input{reference_values/dzm_zdnu.txt}

% NORMALISED ZONE DISTANCE NON-UNIFORMITY
\subsection[Normalised zone distance non-uniformity]{Normalised zone distance non-uniformity \id{IATH}}\label{feat_dzm_zone_distance_non_uniformity_normalised}
This is a normalised version of the \textit{zone distance non-uniformity} feature. It is defined as: 
\begin{displaymath}
F_{\mathit{dzm.zdnu.norm}}= \frac{1}{N_s^2} \sum_{i=1}^{N_d} d_{.j}^2
\end{displaymath}

\input{reference_values/dzm_zdnu_norm.txt}

% ZONE PERCENTAGE
\subsection[Zone percentage]{Zone percentage\id{VIWW}}\label{feat_dzm_zone_percentage}
This feature measures the fraction of the number of realised zones and the maximum number of potential zones. Highly uniform ROIs produce a low \textit{zone percentage}. It is defined as:
\begin{displaymath}
F_{\mathit{dzm.z.perc}}=\frac{N_s}{N_v}
\end{displaymath}

\input{reference_values/dzm_z_perc.txt}

% GREY LEVEL VARIANCE
\subsection[Grey level variance]{Grey level variance\id{QK93}}\label{feat_dzm_grey_level_variance}
This feature estimates the variance in zone counts over the grey levels. Let $p_{ij} = d_{ij}/N_s$ be the joint probability estimate for finding zones with discretised grey level $i$ at distance $j$. The feature is then defined as:
\begin{displaymath}
F_{\mathit{dzm.gl.var}}=  \sum_{i=1}^{N_g} \sum_{j=1}^{N_d} (i-\mu)^2 p_{ij}
\end{displaymath}
Here, $\mu = \sum_{i=1}^{N_g} \sum_{j=1}^{N_d} i\,p_{ij}$.

\input{reference_values/dzm_gl_var.txt}

% ZONE DISTANCE VARIANCE
\subsection[Zone distance variance]{Zone distance variance\id{7WT1}}\label{feat_dzm_zone_distance_variance}
This feature estimates the variance in zone counts for the different zone distances. As before let $p_{ij} = d_{ij}/N_s$. The feature is defined as:
\begin{displaymath}
F_{\mathit{dzm.zd.var}}= \sum_{i=1}^{N_g} \sum_{j=1}^{N_d} (j-\mu)^2 p_{ij}
\end{displaymath}
Mean zone size is defined as $\mu = \sum_{i=1}^{N_g} \sum_{j=1}^{N_d} j\,p_{ij}$.

\input{reference_values/dzm_zd_var.txt}

% ZONE DISTANCE ENTROPY
\subsection[Zone distance entropy]{Zone distance entropy\id{GBDU}}\label{feat_dzm_zone_distance_entropy}
Again, let $p_{ij} = d_{ij}/N_s$. Zone distance entropy is then defined as:
\begin{displaymath}
F_{\mathit{dzm.zd.entr}} = - \sum_{i=1}^{N_g} \sum_{j=1}^{N_d} p_{ij} \log_2 p_{ij}
\end{displaymath}

\input{reference_values/dzm_zd_entr.txt}

\clearpage
\section[Neighbourhood grey tone difference based features]{Neighbourhood grey tone difference based features\id{IPET}}
\citet{Amadasun1989} introduced an alternative to the grey level co-occurrence matrix. The neighbourhood grey tone difference matrix (NGTDM) contains the sum of grey level differences of pixels/voxels with discretised grey level $i$ and the average discretised grey level of neighbouring pixels/voxels within a Chebyshev distance $\delta$. For 3D volumes, we can extend the original definition by Amadasun and King. Let $X_{d,k}$ be the discretised grey level of a voxel at position $\mathbf{k}=(k_x,k_y,k_z)$. Then the average grey level within a neighbourhood centred at $(k_x,k_y,k_z)$, but excluding $(k_x,k_y,k_z)$ itself is:
\begin{align*}
\overline{X}_k& =\frac{1}{W}\sum_{m_z{=}-\delta}^\delta \sum_{m_y{=}-\delta}^\delta \sum_{m_x{=}-\delta}^\delta X_{d}(k_x{+}m_x, k_y{+}m_y, k_z{+}m_z)\\
& \hspace{6cm} (m_x,m_y,m_z)\neq (0,0,0)
\end{align*}
$W=(2\delta+1)^3-1$ is the size of the 3D neighbourhood. For 2D $W=(2\delta+1)^2-1$, and averages are not calculated between different slices.
Neighbourhood grey tone difference $s_i$ for discretised grey level $i$ is then:
\begin{displaymath}
s_i=\sum_{k}^{N_v} \abs{i-\overline{X}_k} \, \iverson{X_d(\mathbf{k})=i \text{ and } k \text{ has a valid neighbourhood}}
\end{displaymath}
Here, $[\ldots]$ is an Iverson bracket, which is $1$ if the conditions that the grey level $X_{d,k}$ of voxel $k$ is equal to $i$ and the voxel has a valid neighbourhood are both true; it is $0$ otherwise. $N_v$ is the number of voxels in the ROI intensity mask.

A 2D example is shown in Table \ref{figNGTDM1}. A distance of $\delta=1$ is used in this example, leading to 8 neighbouring pixels. Entry $s_1=0$ because there are no valid pixels with grey level $1$. Two pixels have grey level $2$. The average value of their neighbours are $19/8$ and $21/8$. Thus $s_2=|2-19/8|+|2-21/8|=1$. Similarly $s_3=|3-19/8|=0.625$ and $s_4=|4-17/8|=1.825$.

We deviate from the original definition by \citet{Amadasun1989} as we do not demand that valid neighbourhoods are  completely inside the ROI. In an irregular ROI mask, valid neighbourhoods may simply not exist for a distance $\delta$. Instead, we consider a valid neighbourhood to exist if there is at least one neighbouring voxel included in the ROI mask. The average grey level for voxel $k$ within a valid neighbourhood is then:
\begin{displaymath}
\overline{X}_k =\frac{1}{W_k}\sum_{m_z{=}-\delta}^\delta \sum_{m_y{=}-\delta}^\delta \sum_{m_x{=}-\delta}^\delta X_{d}(\mathbf{k}+\mathbf{m}) \iverson{\mathbf{m\neq\mathbf{0}} \text{ and } \mathbf{k}+\mathbf{m} \text{ in ROI}}
\end{displaymath}
The neighbourhood size $W_k$ for this voxel is equal to the number of voxels in the neighbourhood that are part of the ROI mask:
\begin{displaymath}
W_k = \sum_{m_z{=}-\delta}^\delta \sum_{m_y{=}-\delta}^\delta \sum_{m_x{=}-\delta}^\delta \iverson{\mathbf{m\neq\mathbf{0}} \text{ and } \mathbf{k}+\mathbf{m} \text{ in ROI}}
\end{displaymath}
Under our definition, neighbourhood grey tone difference $s_i$ for discretised grey level $i$ can be directly expressed using neighbourhood size $W_k$ of voxel $k$:
\begin{displaymath}
s_i=\sum_{k}^{N_v} \abs{i-\overline{X}_k} \, \iverson{X_d(\mathbf{k})=i \text{ and } W_k\neq0}
\end{displaymath}
Consequentially, $n_i$ is the total number of voxels with grey level $i$ which have a non-zero neighbourhood size.

Many NGTDM-based features depend on the $N_g$ grey level probabilities $p_i=n_i/N_{v,c}$, where $N_g$ is the number of discretised grey levels in the ROI intensity mask and $N_{v,c}=\sum n_i$ is total number of voxels that have at least one neighbour. If all voxels have at least one neighbour $N_{v,c}=N_v$. Furthermore, let $N_{g,p} \leq N_g$ be the number of discretised grey levels with $p_i>0$. In the above example, $N_g=4$ and $N_{g,p}=3$.

\begin{table}[t]
\centering
\subcaptionbox{Grey levels}{
	\begin{tabular}{@{}cccc@{}}
		\toprule
		1 & 2 & 2 & 3\\
		\cline{2-3}
		\multicolumn{1}{@{}c|}{1} & 2 & \multicolumn{1}{c|}{3} & 3\\
		\multicolumn{1}{@{}c|}{4} & 2 & \multicolumn{1}{c|}{4} & 1\\
		\cline{2-3}
		4 & 1 & 2 & 3\\
		\bottomrule
	\end{tabular}}\qquad
\subcaptionbox{Neighbourhood grey tone difference matrix}{
	\begin{tabular}{@{}ccccc@{}}
		\toprule
		& & $n_i$ & $p_i$ & $s_i$\\
		\midrule
		\multirow{4}{*}{$i$} & 1 & 0 & 0.00 & 0.000\\
		& 2 & 2 & 0.50 & 1.000\\
		& 3 & 1 & 0.25 & 0.625\\
		& 4 & 1 & 0.25 & 1.875\\
		\bottomrule
	\end{tabular}}
\caption{Original image with grey levels (a) and corresponding neighbourhood grey tone difference matrix (NGTDM) (b). The $N_{v,c}$ pixels with valid neighbours at distance 1 are located within the rectangle in (a). The grey level voxel count $n_i$, the grey level probability $p_i = n_i/N_{v,c}$, and the neighbourhood grey level difference $s_i$ for pixels with grey level $i$ are included in the NGTDM. Note that our actual definition deviates from the original definition of \citet{Amadasun1989}, which is used here. In our definition complete neighbourhood are no longer required. In our definition the NGTDM would be calculated on the entire pixel area, and not solely on those pixels within the rectangle of panel (a).}
\label{figNGTDM1}
\end{table}

\subsubsection*{Aggregating features}
Three methods can be used to aggregate NGTDMs and arrive at a single feature value. A schematic example was previously shown in Figure \ref{figGLSZMCalcApproaches}. A feature may be aggregated as follows:
\begin{enumerate}
\item Features are computed from 2D matrices and averaged over slices (\textid{8QNN}).
\item The feature is computed from a single matrix after merging all 2D matrices (\textid{62GR}).
\item The feature is computed from a 3D matrix (\textid{KOBO}).
\end{enumerate}
Method 2 involves merging NGTDMs by summing the neighbourhood grey tone difference $s_i$ and the number of voxels with a valid neighbourhood $n_i$ and grey level $i$ for NGTDMs of the different slices. Note that when NGTDMs are merged, $N_{v,c}$ and $p_i$ should be updated based on the merged NGTDM. Feature values may dependent strongly on the aggregation method.

\subsubsection*{Distances and distance weighting}
The default neighbourhood is defined using the Chebyshev norm. Manhattan or Euclidean norms may be used as well. This requires a more general definition for the average grey level $\overline{X}_k$:
\begin{displaymath}
\overline{X}_k =\frac{1}{W_k}\sum_{m_z{=}-\delta}^\delta \sum_{m_y{=}-\delta}^\delta \sum_{m_x{=}-\delta}^\delta X_{d}(\mathbf{k}+\mathbf{m}) \iverson{\norm{\mathbf{m}}\leq\delta \text{ and } \mathbf{m\neq\mathbf{0}} \text{ and } \mathbf{k}+\mathbf{m} \text{ in ROI}}
\end{displaymath}
The neighbourhood size $W_k$ is:
\begin{displaymath}
W_k = \sum_{m_z{=}-\delta}^\delta \sum_{m_y{=}-\delta}^\delta \sum_{m_x{=}-\delta}^\delta \iverson{\norm{\mathbf{m}}\leq\delta \text{ and } \mathbf{m\neq\mathbf{0}} \text{ and } \mathbf{k}+\mathbf{m} \text{ in ROI}}
\end{displaymath}
As before, $\iverson{\ldots}$ is an Iverson bracket.

Distance weighting for NGTDM is relatively straightforward. Let $w$ be a weight dependent on $\mathbf{m}$, e.g. $w=\norm{\mathbf{m}}^{-1}$ or $w=\exp(-\norm{\mathbf{m}}^2)$. The average grey level is then:
\begin{displaymath}
\overline{X}_k =\frac{1}{W_k}\sum_{m_z{=}-\delta}^\delta \sum_{m_y{=}-\delta}^\delta \sum_{m_x{=}-\delta}^\delta w(\mathbf{m}) X_{d}(\mathbf{k}+\mathbf{m}) \iverson{\norm{\mathbf{m}}\leq\delta \text{ and } \mathbf{m\neq\mathbf{0}} \text{ and } \mathbf{k}+\mathbf{m} \text{ in ROI}}
\end{displaymath}
The neighbourhood size $W_k$ becomes a general weight:
\begin{displaymath}
W_k = \sum_{m_z{=}-\delta}^\delta \sum_{m_y{=}-\delta}^\delta \sum_{m_x{=}-\delta}^\delta w(\mathbf{m}) \iverson{\norm{\mathbf{m}}\leq\delta \text{ and } \mathbf{m\neq\mathbf{0}} \text{ and } \mathbf{k}+\mathbf{m} \text{ in ROI}}
\end{displaymath}

Employing different distance norms and distance weighting is considered non-standard use, and we caution against them due to potential reproducibility issues.

\FloatBarrier

% COARSENESS
\subsection[Coarseness]{Coarseness\id{QCDE}}\label{feat_ngtdm_coarseness}
Grey level differences in coarse textures are generally small due to large-scale patterns. Summing differences gives an indication of the level of the spatial rate of change in intensity \citep{Amadasun1989}. \textit{Coarseness} is defined as:
\begin{displaymath}
F_{\mathit{ngt.coarseness}}=\frac{1}{\sum_{i=1}^{N_g} p_i\,s_i }
\end{displaymath}
Because $\sum_{i=1}^{N_g} p_i\,s_i$ potentially evaluates to 0, the maximum \textit{coarseness} value is set to an arbitrary number of $10^6$. Amadasun and King originally circumvented this issue by adding a unspecified small number $\epsilon$ to the denominator, but an explicit, though arbitrary, maximum value should allow for more consistency.

\input{reference_values/ngt_coarseness.txt}

% newpage command to prevent a very odd page-break where the Contrast subsection extends below the actual page edge.

% CONTRAST
\subsection[Contrast]{Contrast\id{65HE}}\label{feat_ngtdm_contrast}
\textit{Contrast} depends on the dynamic range of the grey levels as well as the spatial frequency of intensity changes \citep{Amadasun1989}. Thus, \textit{contrast} is defined as:
\begin{displaymath}
F_{\mathit{ngt.contrast}}=\left(\frac{1}{N_{g,p}\left(N_{g,p}-1\right)} \sum_{i_{1}=1}^{N_g} \sum_{i_{2}=1}^{N_g} p_{i_{1}} p_{i_{2}}\,(i_{1}-i_{2})^2 \right) \left( \frac{1}{N_{v,c}}\sum_{i=1}^{N_g} s_i \right)
\end{displaymath}

Grey level probabilities $p_{i_{1}}$ and $p_{i_{2}}$ are copies of $p_i$ with different iterators, i.e. $p_{i_{1}}=p_{i_{2}}$ for $i_{1}=i_{2}$. The first term considers the grey level dynamic range, whereas the second term is a measure for intensity changes within the volume. If $N_{g,p}=1$, $F_{\mathit{ngt.contrast}}=0$.

\input{reference_values/ngt_contrast.txt}

% BUSYNESS
\subsection[Busyness]{Busyness\id{NQ30}}\label{feat_ngtdm_busyness}
Textures with large changes in grey levels between neighbouring voxels are said to be busy \citep{Amadasun1989}. \textit{Busyness} was defined as:
\begin{displaymath}
F_{\mathit{ngt.busyness}}=\frac{\sum_{i=1}^{N_g}p_i\,s_i}{\sum_{i_{1}=1}^{N_g}\sum_{i_2=1}^{N_g} i_{1} \, p_{i_{1}}- i_{2} \, p_{i_{2}}},\qquad \text{$p_{i_{1}}\neq 0$ \text{and} $p_{i_{2}}\neq 0$}
\end{displaymath}
As before, $p_{i_{1}}=p_{i_{2}}$ for $i_{1}=i_{2}$. The original definition was erroneously formulated as the denominator will always evaluate to 0. Therefore we use a slightly different definition \citep{Hatt2016}:
\begin{displaymath}
F_{\mathit{ngt.busyness}}=\frac{\sum_{i=1}^{N_g}p_i\,s_i}{\sum_{i_{1}=1}^{N_g}\sum_{i_{2}=1}^{N_g} \left| i_{1} \, p_{i_{1}}-i_{2} \, p_{i_{2}}\right|},\qquad \text{$p_{i_{1}}\neq 0$ \text{and} $p_{i_{2}}\neq 0$}
\end{displaymath}
If $N_{g,p}=1$, $F_{\mathit{ngt.busyness}}=0$.

\input{reference_values/ngt_busyness.txt}

% COMPLEXITY
\subsection[Complexity]{Complexity\id{HDEZ}}\label{feat_ngtdm_complexity}
Complex textures are non-uniform and rapid changes in grey levels are common \citep{Amadasun1989}. Texture \textit{complexity} is defined as:
\begin{displaymath}
F_{\mathit{ntg.complexity}}=\frac{1}{N_{v,c}}\sum_{i_{1}=1}^{N_g}\sum_{i_{2}=1}^{N_g} \left| i_{1} - i_{2}\right| \frac{p_{i_{1}}\, s_{i_{1}} + p_{i_{2}}\,s_{i_{2}}}{p_{i_{1}} + p_{i_{2}}}, \qquad \text{$p_{i_{1}}\neq 0$ \text{and} $p_{i_{2}}\neq 0$}
\end{displaymath}
As before, $p_{i_{1}}=p_{i_{2}}$ for $i_{1}=i_{2}$, and likewise $s_{i_{1}}=s_{i_{2}}$ for $i_{1}=i_{2}$.

\input{reference_values/ngt_complexity.txt}

\layoutnewpage

% STRENGTH
\subsection[Strength]{Strength\id{1X9X}}\label{feat_ngtdm_strength}
\citet{Amadasun1989} defined texture \textit{strength} as:
\begin{displaymath}
F_{\mathit{ngt.strength}}=\frac{\sum_{i_{1}=1}^{N_g}\sum_{i_{2}=1}^{N_g}\left( p_{i_{1}} + p_{i_{2}} \right) \left( i_{1} - i_{2}\right)^2 }{\sum_{i=1}^{N_g}s_i},\qquad \text{$p_{i_{1}}\neq 0$ \text{and} $p_{i_{2}}\neq 0$}
\end{displaymath}
As before, $p_{i_{1}}=p_{i_{2}}$ for $i_{1}=i_{2}$. If $\sum_{i=1}^{N_g}s_i=0$, $F_{\mathit{ngt.strength}}=0$.

\input{reference_values/ngt_strength.txt}

\clearpage
\section[Neighbouring grey level dependence based features]{Neighbouring grey level dependence based features\id{REK0}}
\citet{Sun1983} defined the neighbouring grey level dependence matrix (NGLDM) as an alternative to the grey level co-occurrence matrix. The NGLDM aims to capture the coarseness of the overall texture and is rotationally invariant. 

NGLDM also involves the concept of a neighbourhood around a central voxel. All voxels within Chebyshev distance $\delta$ are considered to belong to the neighbourhood of the center voxel. The discretised grey levels of the center voxel $k$ at position $\mathbf{k}$ and a neighbouring voxel $m$ at $\mathbf{k}+\mathbf{m}$ are said to be dependent if $\abs{X_d(\mathbf{k}) - X_d(\mathbf{k}+\mathbf{m}) } \leq \alpha$, with $\alpha$ being a non-negative integer coarseness parameter. The number of grey level dependent voxels $j$ within the neighbourhood is then counted as:
\begin{displaymath}
j_k = 1+\sum_{m_z{=}-\delta}^\delta \sum_{m_y{=}-\delta}^\delta \sum_{m_x{=}-\delta}^\delta \iverson{\abs{X_{d}(\mathbf{k})-X_{d}(\mathbf{k}+\mathbf{m})} \leq \alpha \text{ and } \mathbf{m}\neq\mathbf{0}}
\end{displaymath}
Here, $\iverson{\ldots}$ is an Iverson bracket, which is $1$ if the aforementioned condition is fulfilled, and $0$ otherwise. Note that the minimum dependence $j_k=1$ and not $j_k=0$. This is done because some feature definitions require a minimum dependence of 1 or are undefined otherwise. One may therefore also simplify the expression for $j_k$ by including the center voxel:
\begin{displaymath}
j_k = \sum_{m_z{=}-\delta}^\delta \sum_{m_y{=}-\delta}^\delta \sum_{m_x{=}-\delta}^\delta \iverson{\abs{X_{d}(\mathbf{k})-X_{d}(\mathbf{k}+\mathbf{m})} \leq \alpha}
\end{displaymath}

Dependence $j_k$ is iteratively determined for each voxel $k$ in the ROI intensity mask. $\mathbf{M}$ is then the $N_g \times N_n$ neighbouring grey level dependence matrix, where $N_g$ is the number of discretised grey levels present in the ROI intensity mask and $N_n=\text{max}(j_k)$ the maximum grey level dependence count found. Element $s_{ij}$ of $\mathbf{M}$ is then the number of neighbourhoods with a center voxel with discretised grey level $i$ and a neighbouring voxel dependence $j$. Furthermore, let $N_v$ be the number of voxels in the ROI intensity mask, and $N_s = \sum_{i=1}^{N_g}\sum_{j=1}^{N_n} s_{ij}$ the number of neighbourhoods. Marginal sums can likewise be defined. Let $s_{i.}=\sum_{j=1}^{N_n}$ be the number of neighbourhoods with discretised grey level $i$, and let $s_{j.}=\sum_{i=1}^{N_g}s_{ij}$ be the number of neighbourhoods with dependence $j$, regardless of grey level. A two dimensional example is shown in Table \ref{figNGLDM1}.

The definition we actually use deviates from the original by \citet{Sun1983}. Because regions of interest are rarely cuboid, omission of neighbourhoods which contain voxels outside the ROI mask may lead to inconsistent results, especially for larger distance $\delta$. Hence the neighbourhoods of all voxels in the within the ROI intensity mask are considered,
and consequently $N_v=N_s$. Neighbourhood voxels located outside the ROI do not add to dependence $j$:
\begin{displaymath}
j_k = \sum_{m_z{=}-\delta}^\delta \sum_{m_y{=}-\delta}^\delta \sum_{m_x{=}-\delta}^\delta \iverson{\abs{X_{d}(\mathbf{k})-X_{d}(\mathbf{k}+\mathbf{m})} \leq \alpha \text{ and } \mathbf{k}+\mathbf{m} \text{ in ROI}}
\end{displaymath}

Note that while $\alpha=0$ is a typical choice for the coarseness parameter, different $\alpha$ are possible. Likewise, a typical choice for neighbourhood radius $\delta$ is Chebyshev distance $\delta=1$ but larger values are possible as well.

\begin{table}[tbh]
\centering
\subcaptionbox{Grey levels}{
	\begin{tabular}{@{}cccc@{}}
		\toprule
		1 & 2 & 2 & 3\\
		\cline{2-3}
		\multicolumn{1}{@{}c|}{1} & 2 & \multicolumn{1}{c|}{3} & 3\\
		\multicolumn{1}{@{}c|}{4} & 2 & \multicolumn{1}{c|}{4} & 1\\
		\cline{2-3}
		4 & 1 & 2 & 3\\
		\bottomrule
	\end{tabular}}\qquad
\subcaptionbox{Neighbouring grey level dependence matrix}{
	\begin{tabular}{@{}cccccc@{}}
		\toprule
		& \multicolumn{5}{r}{dependence $k$}\\
		& & 0 & 1 & 2 & 3\\
		\midrule
		\multirow{4}{*}{$i$} & 1 & 0 & 0 & 0 & 0\\
		& 2 & 0 & 0 & 1 & 1\\
		& 3 & 0 & 0 & 1 & 0\\
		& 4 & 1 & 0 & 0 & 0\\
		\bottomrule
	\end{tabular}}
\caption{Original image with grey levels and pixels with a complete neighbourhood within the square (a); corresponding neighbouring grey level dependence matrix for distance $d=\sqrt{2}$ and coarseness parameter $a=0$ (b). Element $s(i,j)$ of the NGLDM indicates the number of neighbourhoods with a center pixel with grey level $i$ and neighbouring grey level dependence $k$ within the image. Note that in our definition a complete neighbourhood is no longer required. Thus every voxel is considered as a center voxel with a neighbourhood, instead of being constrained to the voxels within the square in panel (a).}
\label{figNGLDM1}
\end{table}

\subsubsection*{Aggregating features}
Three methods can be used to aggregate NGLDMs and arrive at a single feature value. A schematic example was previously shown in Figure \ref{figGLSZMCalcApproaches}. A feature may be aggregated as follows:
\begin{enumerate}
\item Features are computed from 2D matrices and averaged over slices (\textid{8QNN}).
\item The feature is computed from a single matrix after merging all 2D matrices (\textid{62GR}).
\item The feature is computed from a 3D matrix (\textid{KOBO}).
\end{enumerate}
Method 2 involves merging NGLDMs by summing the dependence count $s_{ij}$ by element over the NGLDM of the different slices. Note that when NGLDMs are merged, $N_v$ and $N_s$ should likewise be summed to retain consistency. Feature values may dependent strongly on the aggregation method.

\subsubsection*{Distances and distance weighting}
Default neighbourhoods are constructed using the Chebyshev norm, but other norms can be used as well. For this purpose it is useful to generalise the dependence count equation to:
\begin{displaymath}
j_k = \sum_{m_z{=}-\delta}^\delta \sum_{m_y{=}-\delta}^\delta \sum_{m_x{=}-\delta}^\delta \iverson{\norm{\mathbf{m}}\leq\delta \text{ and } \abs{X_{d}(\mathbf{k})-X_{d}(\mathbf{k}+\mathbf{m})} \leq \alpha \text{ and } \mathbf{k}+\mathbf{m} \text{ in ROI}}
\end{displaymath}
with $\mathbf{m}$ the vector between voxels $k$ and $m$ and $\norm{\mathbf{m}}$ its length according to the particular norm.

In addition, dependence may be weighted by distance. Let $w$ be a weight dependent on $\mathbf{m}$, e.g. $w=\norm{\mathbf{m}}^{-1}$ or $w=\exp(-\norm{\mathbf{m}}^2)$. The dependence of voxel $k$ is then:
\begin{displaymath}
j_k = \sum_{m_z{=}-\delta}^\delta \sum_{m_y{=}-\delta}^\delta \sum_{m_x{=}-\delta}^\delta w(\mathbf{m}) \iverson{\norm{\mathbf{m}}\leq\delta \text{ and } \abs{X_{d}(\mathbf{k})-X_{d}(\mathbf{k}+\mathbf{m})} \leq \alpha \text{ and } \mathbf{k}+\mathbf{m} \text{ in ROI}}
\end{displaymath}

Employing different distance norms and distance weighting is considered non-standard use, and we caution against them due to potential reproducibility issues.

\subsubsection*{Note on feature references}
The NGLDM is structured similarly to the GLRLM, GLSZM and GLDZM. NGLDM feature definitions are therefore based on the definitions of GLRLM features, and references may be found in Section \ref{sect_glrlm}, except for the features originally defined by \citet{Sun1983}.

\FloatBarrier

% LOW DEPENDENCE EMPHASIS
\subsection[Low dependence emphasis]{Low dependence emphasis\id{SODN}}\label{feat_ngldm_low_dependence_emphasis}
This feature emphasises low neighbouring grey level dependence counts. \citet{Sun1983} refer to this feature as \textit{small number emphasis}. It is defined as:
\begin{displaymath}
F_{\mathit{ngl.lde}} = \frac{1}{N_s} \sum_{j=1}^{N_n} \frac{s_{.j}}{j^2}
\end{displaymath}

\input{reference_values/ngl_lde.txt}

% HIGH DEPENDENCE EMPHASIS
\subsection[High dependence emphasis]{High dependence emphasis\id{IMOQ}}\label{feat_ngldm_high_dependence_emphasis}
This feature emphasises high neighbouring grey level dependence counts. \citet{Sun1983} refer to this feature as \textit{large number emphasis}. It is defined as:
\begin{displaymath}
F_{\mathit{ngl.hde}} = \frac{1}{N_s} \sum_{j=1}^{N_n} j^2 s_{.j}
\end{displaymath}

\input{reference_values/ngl_hde.txt}

% LOW GREY LEVEL COUNT EMPHASIS
\subsection[Low grey level count emphasis]{Low grey level count emphasis\id{TL9H}}\label{feat_ngldm_low_grey_level_count_emphasis}
This feature is a grey level analogue to \textit{low dependence emphasis}. Instead of low neighbouring grey level dependence counts, low grey levels are emphasised. The feature is defined as:
\begin{displaymath}
F_{\mathit{ngl.lgce}}=\frac{1}{N_s} \sum_{i=1}^{N_g} \frac{s_{i.}}{i^2}
\end{displaymath}

\input{reference_values/ngl_lgce.txt}

% HIGH GREY LEVEL COUNT EMPHASIS
\subsection[High grey level count emphasis]{High grey level count emphasis\id{OAE7}}\label{feat_ngldm_high_grey_level_count_emphasis}
The \textit{high grey level count emphasis} feature is a grey level analogue to \textit{high dependence emphasis}. The feature emphasises high grey levels, and is defined as:
\begin{displaymath}
F_{\mathit{ngl.hgce}}=\frac{1}{N_s} \sum_{i=1}^{N_g} i^2 s_{i.}
\end{displaymath}

\input{reference_values/ngl_hgce.txt}

% LOW DEPENDENCE LOW GREY LEVEL EMPHASIS
\subsection[Low dependence low grey level emphasis]{Low dependence low grey level emphasis\id{EQ3F}}\label{feat_ngldm_low_dependence_low_grey_level_emphasis}
This feature emphasises neighbouring grey level dependence counts in the upper left quadrant of the NGLDM, where low dependence counts and low grey levels are located. It is defined as:
\begin{displaymath}
F_{\mathit{ngl.ldlge}}=\frac{1}{N_s} \sum_{i=1}^{N_g} \sum_{j=1}^{N_n} \frac{s_{ij}}{i^2 j^2}
\end{displaymath}

\input{reference_values/ngl_ldlge.txt}

% LOW DEPENDENCE HIGH GREY LEVEL EMPHASIS
\subsection[Low dependence high grey level emphasis]{Low dependence high grey level emphasis\id{JA6D}}\label{feat_ngldm_low_dependence_high_grey_level_emphasis}
This feature emphasises neighbouring grey level dependence counts in the lower left quadrant of the NGLDM, where low dependence counts and high grey levels are located. The feature is defined as:
\begin{displaymath}
F_{\mathit{ngl.ldhge}}=\frac{1}{N_s} \sum_{i=1}^{N_g} \sum_{j=1}^{N_n} \frac{i^2 s_{ij}}{j^2}
\end{displaymath}

\layoutnewpage

\input{reference_values/ngl_ldhge.txt}

% HIGH DEPENDENCE LOW GREY LEVEL EMPHASIS
\subsection[High dependence low grey level emphasis]{High dependence low grey level emphasis\id{NBZI}}\label{feat_ngldm_high_dependence_low_grey_level_emphasis}
This feature emphasises neighbouring grey level dependence counts in the upper right quadrant of the NGLDM, where high dependence counts and low grey levels are located. The feature is defined as:
\begin{displaymath}
F_{\mathit{ngl.hdlge}}=\frac{1}{N_s} \sum_{i=1}^{N_g} \sum_{j=1}^{N_n} \frac{j^2 s_{ij}}{i^2}
\end{displaymath}

\input{reference_values/ngl_hdlge.txt}

\layoutnewpage

% HIGH DEPENDENCE HIGH GREY LEVEL EMPHASIS
\subsection[High dependence high grey level emphasis]{High dependence high grey level emphasis\id{9QMG}}\label{feat_ngldm_high_dependence_high_grey_level_emphasis}
The \textit{high dependence high grey level emphasis} feature emphasises neighbouring grey level dependence counts in the lower right quadrant of the NGLDM, where high dependence counts and high grey levels are located. The feature is defined as:
\begin{displaymath}
F_{\mathit{ngl.hdhge}}=\frac{1}{N_s} \sum_{i=1}^{N_g} \sum_{j=1}^{N_n} i^2 j^2 s_{ij}
\end{displaymath}

\input{reference_values/ngl_hdhge.txt}

% GREY LEVEL NON-UNIFORMITY
\subsection[Grey level non-uniformity]{Grey level non-uniformity\id{FP8K}}\label{feat_ngldm_grey_level_non_uniformity}
\textit{Grey level non-uniformity} assesses the distribution of neighbouring grey level dependence counts over the grey values. The feature value is low when dependence counts are equally distributed along grey levels. The feature is defined as:
\begin{displaymath}
F_{\mathit{ngl.glnu}}= \frac{1}{N_s} \sum_{i=1}^{N_g} s_{i.}^2
\end{displaymath}

\input{reference_values/ngl_glnu.txt}

% NORMALISED GREY LEVEL NON-UNIFORMITY
\subsection[Normalised grey level non-uniformity]{Normalised grey level non-uniformity\id{5SPA}}\label{feat_ngldm_grey_level_non_uniformity_normalised}
This is a normalised version of the \textit{grey level non-uniformity} feature. It is defined as: 
\begin{displaymath}
F_{\mathit{ngl.glnu.norm}}= \frac{1}{N_s^2} \sum_{i=1}^{N_g} s_{i.}^2
\end{displaymath}
The \textit{normalised grey level non-uniformity} computed from a single 3D NGLDM matrix is equivalent to the \textit{intensity histogram uniformity} feature \citep{VanGriethuysen2017}.

\input{reference_values/ngl_glnu_norm.txt}

% DEPENDENCE COUNT NON-UNIFORMITY
\subsection[Dependence count non-uniformity]{Dependence count non-uniformity\id{Z87G}}\label{feat_ngldm_dependence_count_non_uniformity}
This features assesses the distribution of neighbouring grey level dependence counts over the different dependence counts. The feature value is low when dependence counts are equally distributed. \citet{Sun1983} refer to this feature as \textit{number non-uniformity}. It is defined as:
\begin{displaymath}
F_{\mathit{ngl.dcnu}}= \frac{1}{N_s} \sum_{j=1}^{N_n} s_{.j}^2
\end{displaymath}

\layoutsetvspace{-5mm}

\input{reference_values/ngl_dcnu.txt}

% NORMALISED DEPENDENCE COUNT NON-UNIFORMITY
\subsection[Normalised dependence count non-uniformity]{Normalised dependence count non-uniformity\id{OKJI}}\label{feat_ngldm_normalised_dependence_count_non_uniformity}
This is a normalised version of the \textit{dependence count non-uniformity} feature. It is defined as:
\begin{displaymath}
F_{\mathit{ngl.dcnu.norm}}= \frac{1}{N_s^2} \sum_{i=1}^{N_n} s_{.j}^2
\end{displaymath}

\input{reference_values/ngl_dcnu_norm.txt}

% DEPENDENCE COUNT PERCENTAGE
\subsection[Dependence count percentage]{Dependence count percentage\id{6XV8}}\label{feat_ngldm_dependence_count_percentage}
This feature measures the fraction of the number of realised neighbourhoods and the maximum number of potential neighbourhoods. \textit{Dependence count percentage} may be completely omitted as it evaluates to $1$ when complete neighbourhoods are not required, as is the case under our definition. It is defined as:
\begin{displaymath}
F_{\mathit{ngl.dc.perc}}=\frac{N_s}{N_v}
\end{displaymath}

\input{reference_values/ngl_dc_perc.txt}

% GREY LEVEL VARIANCE
\subsection[Grey level variance]{Grey level variance\id{1PFV}}\label{feat_ngldm_grey_level_variance}
This feature estimates the variance in dependence counts over the grey levels. Let $p_{ij} = s_{ij}/N_s$ be the joint probability estimate for finding discretised grey level $i$ with dependence $j$. The feature is then defined as:
\begin{displaymath}
F_{\mathit{ngl.gl.var}}=  \sum_{i=1}^{N_g} \sum_{j=1}^{N_n} (i-\mu)^2 p_{ij}
\end{displaymath}
Here, $\mu = \sum_{i=1}^{N_g} \sum_{j=1}^{N_n} i\,p_{ij}$.

\input{reference_values/ngl_gl_var.txt}

\layoutsetvspace{5mm}

% DEPENDENCE COUNT VARIANCE
\subsection[Dependence count variance]{Dependence count variance\id{DNX2}}\label{feat_ngldm_dependence_count_variance}
This feature estimates the variance in dependence counts over the different possible dependence counts. As before let $p_{ij} = s_{ij}/N_s$. The feature is defined as:
\begin{displaymath}
F_{\mathit{ngl.dc.var}}= \sum_{i=1}^{N_g} \sum_{j=1}^{N_n} (j-\mu)^2 p_{ij}
\end{displaymath}
Mean dependence count is defined as $\mu = \sum_{i=1}^{N_g} \sum_{j=1}^{N_n} j\,p_{ij}$.

\input{reference_values/ngl_dc_var.txt}

% DEPENDENCE COUNT ENTROPY
\subsection[Dependence count entropy]{Dependence count entropy\id{FCBV}}\label{feat_ngldm_dependence_count_entropy}
This feature is referred to as \textit{entropy} by \citet{Sun1983}. Let $p_{ij} = s_{ij}/N_s$. \textit{Dependence count entropy} is then defined as:
\begin{displaymath}
F_{\mathit{ngl.dc.entr}} = - \sum_{i=1}^{N_g} \sum_{j=1}^{N_n} p_{ij} \log_2 p_{ij}
\end{displaymath}
This definition remedies an error in the definition of \citet{Sun1983}, where the term within the logarithm is dependence count $s_{ij}$ instead of count probability $p_{ij}$.

\input{reference_values/ngl_dc_entr.txt}

% DEPENDENCE COUNT ENERGY
\subsection[Dependence count energy]{Dependence count energy\id{CAS9}}\label{feat_ngldm_dependence_count_energy}
This feature is called \textit{second moment} by \citet{Sun1983}. Let $p_{ij} = s_{ij}/N_s$. Then \textit{dependence count energy} is defined as:
\begin{displaymath}
F_{\mathit{ngl.dc.energy}} = \sum_{i=1}^{N_g} \sum_{j=1}^{N_n} p_{ij}^2
\end{displaymath}
This definition also remedies an error in the original definition, where squared dependence count $s_{ij}^2$ is divided by $N_s$ only, thus leaving a major volume dependency. In the definition given here, $s_{ij}^2$ is normalised by $N_s^2$ through the use of count probability $p_{ij}$.

\input{reference_values/ngl_dc_energy.txt}

\newpage
\chapter{Radiomics reporting guidelines and nomenclature} \label{chap_report_guidelines}
Reliable and complete reporting is necessary to ensure reproducibility and validation of results. To help provide a complete report on image processing and image biomarker extraction, we present the guidelines below, as well as a nomenclature system to uniquely features.

\section{Reporting guidelines}\label{sec_reporting_guidelines}
These guidelines are partially based on the work of \citet{Sollini2017,Lambin2017,Sanduleanu2018-iu,Traverso2018-yr}. Additionally, guidelines are derived from the image processing and feature calculation steps described within this document. An earlier version was reported elsewhere \cite{vallieres2017responsible}.

\small
\begin{longtable}{p{3.5cm}ccp{7cm}}

\toprule
\textbf{topic} & & \textbf{item} & \textbf{description}\\
\midrule
\endhead

\bottomrule
\multicolumn{4}{r}{\textit{continued on next page}}
\endfoot

\bottomrule
\caption[Reporting guidelines]{Guidelines for reporting on radiomic studies. Not all items may be applicable.} \label{table_guidelines}
\endlastfoot

\multicolumn{4}{l}{\textbf{Patient}}\\
\midrule
Region of interest\footnote{Also referred to as volume of interest.} & &
1 & Describe the region of interest that is being imaged.
\\
Patient preparation & & 2a & Describe specific instructions given to
patients prior to image acquisition, e.g. fasting prior to imaging.
\\
& & 2b & Describe administration of drugs to the patient prior to image
acquisition, e.g. muscle relaxants. \\
& & 2c & Describe the use of specific equipment for patient comfort
during scanning, e.g. ear plugs. \\
Radioactive tracer & PET, SPECT & 3a & Describe which radioactive tracer
was administered to the patient, e.g. 18F-FDG. \\
& PET, SPECT & 3b & Describe the administration method. \\
& PET, SPECT & 3c & Describe the injected activity of the radioactive
tracer at administration. \\
& PET, SPECT & 3d & Describe the uptake time prior to image acquisition.
\\
& PET, SPECT & 3e & Describe how competing substance levels were
controlled.\footnote{An example is glucose present in the blood which
  competes with the uptake of 18F-FDG tracer in tumour tissue. To reduce
  competition with the tracer, patients are usually asked to fast for
  several hours and a blood glucose measurement may be conducted prior
  to tracer administration.} \\
Contrast agent & & 4a & Describe which contrast agent was administered
to the patient. \\
& & 4b & Describe the administration method. \\
& & 4c & Describe the injected quantity of contrast agent.
\\
& & 4d & Describe the uptake time prior to image acquisition.
\\
& & 4e & Describe how competing substance levels were controlled.
\\
Comorbidities & & 5 & Describe if the patients have comorbidities that
affect imaging.\footnote{An example of a comorbidity that may affect
  image quality in 18F-FDG PET scans are type I and type II diabetes
  melitus, as well as kidney failure.} \\
 
\multicolumn{4}{l}{\textbf{Acquisition}\footnote{Many acquisition parameters may be
  extracted from DICOM header meta-data, or calculated from them.}}\\
\midrule
Acquisition protocol & & 6 & Describe whether a standard imaging
protocol was used, and where its description may be found.
\\
Scanner type & & 7 & Describe the scanner type(s) and vendor(s) used in
the study. \\
Imaging modality & & 8 & Clearly state the imaging modality that was
used in the study, e.g. CT, MRI. \\
Static/dynamic scans & & 9a & State if the scans were static or dynamic.
\\
& Dynamic scans & 9b & Describe the acquisition time per time frame.
\\
& Dynamic scans & 9c & Describe any temporal modelling technique that
was used. \\
Scanner calibration & & 10 & Describe how and when the scanner was
calibrated. \\
Patient instructions & & 11 & Describe specific instructions given to
the patient during acquisition, e.g. breath holding. \\
Anatomical motion correction & & 12 & Describe the method used to
minimise the effect of anatomical motion. \\
Scan duration & & 13 & Describe the duration of the complete scan or the
time per bed position. \\
Tube voltage & CT & 14 & Describe the peak kilo voltage output of the
X-ray source. \\
Tube current & CT & 15 & Describe the tube current in mA.
\\
Time-of-flight & PET & 16 & State if scanner time-of-flight capabilities
are used during acquisition. \\
RF coil & MRI & 17 & Describe what kind RF coil used for acquisition,
incl. vendor. \\
Scanning sequence & MRI & 18a & Describe which scanning sequence was
acquired. \\
& MRI & 18b & Describe which sequence variant was acquired.
\\
& MRI & 18c & Describe which scan options apply to the current sequence,
e.g. flow compensation, cardiac gating. \\
Repetition time & MRI & 19 & Describe the time in ms between subsequent
pulse sequences. \\
Echo time & MRI & 20 & Describe the echo time in ms. \\
Echo train length & MRI & 21 & Describe the number of lines in k-space
that are acquired per excitation pulse. \\
Inversion time & MRI & 22 & Describe the time in ms between the middle
of the inverting RF pulse to the middle of the excitation pulse.
\\
Flip angle & MRI & 23 & Describe the flip angle produced by the RF
pulses. \\
Acquisition type & MRI & 24 & Describe the acquisition type of the MRI
scan, e.g. 3D. \\
k-space traversal & MRI & 25 & Describe the acquisition trajectory of
the k-space. \\
Number of averages/ excitations & MRI & 26 & Describe the number of
times each point in k-space is sampled. \\
Magnetic field strength & MRI & 27 & Describe the nominal strength of
the MR magnetic field. \\

\multicolumn{4}{l}{\textbf{Reconstruction}\footnote{Many reconstruction parameters may be
  extracted from DICOM header meta-data.}}\\
\midrule
In-plane resolution & & 28 & Describe the distance between pixels, or
alternatively the field of view and matrix size. \\
Image slice thickness & & 29 & Describe the slice thickness.
\\
Image slice spacing & & 30 & Describe the distance between image
slices.\footnote{Spacing between image slicing is commonly, but not
  necessarily, the same as the slice thickness,.} \\
Convolution kernel & CT & 31a & Describe the convolution kernel used to
reconstruct the image. \\
& CT & 31b & Describe settings pertaining to iterative reconstruction
algorithms. \\
Exposure & CT & 31c & Describe the exposure (in mAs) in slices
containing the region of interest. \\
Reconstruction method & PET & 32a & Describe which reconstruction method
was used, e.g. 3D OSEM. \\
& PET & 32b & Describe the number of iterations for iterative
reconstruction. \\
& PET & 32c & Describe the number of subsets for iterative
reconstruction. \\
Point spread function modelling & PET & 33 & Describe if and how
point-spread function modelling was performed. \\
Image corrections & PET & 34a & Describe if and how attenuation
correction was performed. \\
& PET & 34b & Describe if and how other forms of correction were
performed, e.g. scatter correction, randoms correction, dead time
correction etc. \\
Reconstruction method & MRI & 35a & Describe the reconstruction method
used to reconstruct the image from the k-space information.
\\
& MRI & 35b & Describe any artifact suppression methods used during
reconstruction to suppress artifacts due to undersampling of k-space.
\\
Diffusion-weighted imaging & DWI-MRI & 36 & Describe the b-values used
for diffusion-weighting. \\

\multicolumn{4}{l}{\textbf{Image registration}}\\
\midrule
Registration method & & 37 & Describe the method used to register
multi-modality imaging. \\

\multicolumn{4}{l}{\textbf{Image processing - data conversion}} \\
\midrule
SUV normalisation & PET & 38 & Describe which standardised uptake value
(SUV) normalisation method is used. \\
ADC computation & DWI-MRI & 39 & Describe how apparent diffusion
coefficient (ADC) values were calculated. \\
Other data conversions & & 40 & Describe any other conversions that are
performed to generate e.g. perfusion maps. \\

\multicolumn{4}{l}{\textbf{Image processing - post-acquisition processing}} \\
\midrule
Anti-aliasing & & 41 & Describe the method used to deal with
anti-aliasing when down-sampling during interpolation. \\
Noise suppression & & 42 & Describe methods used to suppress image
noise. \\
Post-reconstruction smoothing filter & PET & 43 & Describe the width of
the Gaussian filter (FWHM) to spatially smooth intensities.
\\
Skull stripping & MRI (brain) & 44 & Describe method used to perform
skull stripping. \\
Non-uniformity correction\footnote{Also known as bias-field correction.}
& MRI & 45 & Describe the method and settings used to perform
non-uniformity correction. \\
Intensity normalisation & & 46 & Describe the method and settings used
to normalise intensity distributions within a patient or patient cohort.
\\
Other post-acquisition processing methods & & 47 & Describe any other
methods that were used to process the image and are not mentioned
separately in this list. \\

\multicolumn{4}{l}{\textbf{Segmentation}} \\
\midrule
Segmentation method & & 48a & Describe how regions of interest were
segmented, e.g. manually. \\
& & 48b & Describe the number of experts, their expertise and consensus
strategies for manual delineation. \\
& & 48c & Describe methods and settings used for semi-automatic and
fully automatic segmentation. \\
& & 48d & Describe which image was used to define segmentation in case
of multi-modality imaging. \\
Conversion to mask & & 49 & Describe the method used to convert
polygonal or mesh-based segmentations to a voxel-based mask.
\\

\multicolumn{4}{l}{\textbf{Image processing - image interpolation}} \\
\midrule
Interpolation method & & 50a & Describe which interpolation algorithm
was used to interpolate the image. \\
& & 50b & Describe how the position of the interpolation grid was
defined, e.g. align by center. \\
& & 50c & Describe how the dimensions of the interpolation grid were
defined, e.g. rounded to nearest integer. \\
& & 50d & Describe how extrapolation beyond the original image was
handled. \\
Voxel dimensions & & 51 & Describe the size of the interpolated voxels.
\\
Intensity rounding & CT & 52 & Describe how fractional Hounsfield Units
are rounded to integer values after interpolation. \\

\multicolumn{4}{l}{\textbf{Image processing - ROI interpolation}} \\
\midrule
Interpolation method & & 53 & Describe which interpolation algorithm was
used to interpolate the region of interest mask. \\
Partially masked voxels & & 54 & Describe how partially masked voxels
after interpolation are handled. \\

\multicolumn{4}{l}{\textbf{Image processing - re-segmentation}} \\
\midrule
Re-segmentation methods & & 55 & Describe which methods and settings are
used to re-segment the ROI intensity mask. \\

\multicolumn{4}{l}{\textbf{Image processing - discretisation}} \\
\midrule
Discretisation method\footnote{Discretisation may be performed
  separately to create intensity-volume histograms. If this is indeed
  the case, this should be described as well.} & & 56a & Describe the
method used to discretise image intensities. \\
& & 56b & Describe the number of bins (FBN) or the bin size (FBS) used
for discretisation. \\
& & 56c & Describe the lowest intensity in the first bin for FBS
discretisation.\footnote{This is typically set by range re-segmentation.}
\\

\multicolumn{4}{l}{\textbf{Image processing - image transformation}} \\
\midrule
Image filter\footnote{The IBSI has not introduced image transformation
  into the standardised image processing scheme, and is in the process
  of benchmarking various common filters. This section may therefore be
  expanded in the future.} & & 57 & Describe the methods and settings
used to filter images, e.g. Laplacian-of-Gaussian. \\

\multicolumn{4}{l}{\textbf{Image biomarker computation}} \\
\midrule
Biomarker set & & 58 & Describe which set of image biomarkers is
computed and refer to their definitions or provide these.
\\
IBSI compliance & & 59 & State if the software used to extract the set
of image biomarkers is able to reproduce the IBSI feature reference values.\footnote{A
  software is compliant if and only if it is able to reproduce image biomarker reference values for the digital phantom and for one or more image processing configurations using the radiomics CT phantom. Reviewers
  may demand that you provide the IBSI compliance spreadsheet for your software.} \\
Robustness & & 60 & Describe how robustness of the image biomarkers was
assessed, e.g. test-retest analysis. \\
Software availability & & 61 & Describe which software and version was
used to compute image biomarkers. \\

\multicolumn{4}{l}{\textbf{Image biomarker computation - texture parameters}}\\
\midrule
Texture matrix aggregation & & 62 & Define how texture-matrix based
biomarkers were computed from underlying texture matrices.
\\
Distance weighting & & 63 & Define how CM, RLM, NGTDM and NGLDM weight
distances, e.g. no weighting. \\
CM symmetry & & 64 & Define whether symmetric or asymmetric
co-occurrence matrices were computed. \\
CM distance & & 65 & Define the (Chebyshev) distance at which
co-occurrence of intensities is determined, e.g. 1. \\
SZM linkage distance & & 66 & Define the distance and distance norm for
which voxels with the same intensity are considered to belong to the
same zone for the purpose of constructing an SZM, e.g. Chebyshev
distance of 1. \\
DZM linkage distance & & 67 & Define the distance and distance norm for
which voxels with the same intensity are considered to belong to the
same zone for the purpose of constructing a DZM, e.g. Chebyshev distance
of 1. \\
DZM zone distance norm & & 68 & Define the distance norm for determining
the distance of zones to the border of the ROI, e.g. Manhattan distance.
\\
NGTDM distance & & 69 & Define the neighbourhood distance and distance
norm for the NGTDM, e.g. Chebyshev distance of 1. \\
NGLDM distance & & 70 & Define the neighbourhood distance and distance
norm for the NGLDM, e.g. Chebyshev distance of 1. \\
NGLDM coarseness & & 71 & Define the coarseness parameter for the NGLDM,
e.g. 0. \\

\multicolumn{4}{l}{\textbf{Machine learning and radiomics analysis}} \\
\midrule
Diagnostic and prognostic modelling & & 72 & See the TRIPOD guidelines
for reporting on diagnostic and prognostic modelling. \\
Comparison with known factors & & 73 & Describe where performance of
radiomics models is compared with known (clinical) factors.
\\
Multicollinearity & & 74 & Describe where the multicollinearity between
image biomarkers in the signature is assessed. \\
Model availability & & 75 & Describe where radiomics models with the
necessary pre-processing information may be found. \\
Data availability & & 76 & Describe where imaging data and relevant
meta-data used in the study may be found. \\
\end{longtable}

\normalsize
\FloatBarrier

\section{Feature nomenclature}
Image features may be extracted using a variety of different settings, and may even share the same name. A feature nomenclature is thus required. Let us take the example of differentiating the following features: \textit{i}) intensity histogram-based entropy, discretised using a \textit{fixed bin size} algorithm with 25~HU bins, extracted from a CT image; and \textit{ii}) grey level run length matrix entropy, discretised using a \textit{fixed bin number} algorithm with 32 bins, extracted from a PET image. To refer to both as \textit{entropy} would be ambiguous, whereas to add a full textual description would be cumbersome. In the nomenclature proposed below, the features would be called \textit{entropy\textsubscript{IH, CT, FBS:25HU}} and \textit{entropy\textsubscript{RLM, PET, FBN:32}}, respectively.

Features are thus indicated by a feature name and a subscript. As the nomenclature is designed to both concise and complete, only details for which ambiguity may exist are to be explicitly incorporated in the subscript. The subscript of a feature name may contain the following items to address ambiguous naming:

\begin{enumerate}
\item An abbreviation of the feature family (required).
\item The aggregation method of a feature (optional).
\item A descriptor describing the modality the feature is based on, the specific channel (for microscopy images), the specific imaging data (in the case of repeat imaging or delta-features) sets, conversions (such as SUV and SUL), and/or the specific ROI. For example, one could write \textit{PET:SUV} to separate it from \textit{CT} and \textit{PET:SUL} features (optional).
\item Spatial filters and settings (optional).
\item The interpolation algorithm and uniform interpolation grid spacing (optional).
\item The re-segmentation range and outlier filtering (optional).
\item The discretisation method and relevant discretisation parameters, i.e. number of bins or bin size (optional).
\item Feature specific parameters, such as distance for some texture features (optional).
\end{enumerate}
Optional descriptors are only added to the subscript if there are multiple possibilities. For example, if only CT data is used, adding the modality to the subscript is not required. Nonetheless, such details must be reported as well (see section \ref{sec_reporting_guidelines}).

The sections below have tables with permanent IBSI identifiers for concepts that were defined within this document.

\subsection{Abbreviating feature families}
The following is a list of the feature families in this document and their suggested abbreviations:

\begin{table}[th]
\centering
\small
\begin{tabular}{lcc}
\toprule
\textbf{feature family} & \textbf{abbreviation} &\\
\midrule
morphology & MORPH & \textid{HCUG}\\
local intensity & LI & \textid{9ST6}\\
intensity-based statistics & IS, STAT  & \textid{UHIW}\\
intensity histogram & IH & \textid{ZVCW}\\
intensity-volume histogram & IVH & \textid{P88C}\\
grey level co-occurrence matrix & GLCM, CM & \textid{LFYI}\\
grey level run length matrix & GLRLM, RLM & \textid{TP0I}\\
grey level size zone matrix & GLSZM, SZM& \textid{9SAK} \\
grey level distance zone matrix & GLDZM, DZM & \textid{VMDZ}\\
neighbourhood grey tone difference matrix & NGTDM & \textid{IPET}\\
neighbouring grey level dependence matrix & NGLDM & \textid{REK0}\\
\bottomrule
\end{tabular}
\end{table}

\FloatBarrier

\subsection{Abbreviating feature aggregation}
The following is a list of feature families and the possible aggregation methods:

\begin{table}[th]
\centering
\small
\begin{tabular}{clc}
\toprule
\multicolumn{3}{l}{\textbf{morphology, LI}}\\
\midrule
-- & features are 3D by definition & \textid{DHQ4}\\
\noalign{\vskip 2mm}
\multicolumn{2}{l}{\textbf{IS, IH, IVH}}\\
\midrule
2D & averaged over slices (rare) & \textid{3IDG}\\
--, 3D & calculated over the volume (default) & \textid{DHQ4}\\
\noalign{\vskip 2mm}
\multicolumn{2}{l}{\textbf{GLCM, GLRLM}}\\
\midrule
2D:avg & averaged over slices and directions & \textid{BTW3}\\
2D:mrg, 2D:smrg & merged directions per slice and averaged & \textid{SUJT}\\
2.5D:avg, 2.5D:dmrg & merged per direction and averaged & \textid{JJUI}\\
2.5D:mrg, 2.5D:vmrg & merged over all slices& \textid{ZW7Z}\\
3D:avg & averaged over 3D directions& \textid{ITBB}\\
3D:mrg & merged 3D directions& \textid{IAZD}\\
\noalign{\vskip 2mm}
\multicolumn{2}{l}{\textbf{GLSZM, GLDZM, NGTDM, NGLDM}}\\
\midrule
2D & averaged over slices & \textid{8QNN}\\
2.5D & merged over all slices & \textid{62GR}\\
3D & calculated from single 3D matrix & \textid{KOBO}\\
\bottomrule
\end{tabular}
\end{table}

\FloatBarrier

In the list above, '--' signifies an empty entry which does not need to be added to the subscript. The following examples highlight the nomenclature used above:
\begin{itemize}
\item joint maximum\textsubscript{CM, 2D:avg}: GLCM-based \textit{joint maximum} feature, calculated by averaging the feature for every in-slice GLCM.
\item short runs emphasis\textsubscript{RLM, 3D:mrg}: RLM-based \textit{short runs emphasis} feature, calculated from an RLM that was aggregated by merging the RLM of each 3D direction.
\item mean\textsubscript{IS}: intensity statistical \textit{mean} feature, calculated over the 3D ROI volume.
\item grey level variance\textsubscript{SZM, 2D}: SZM-based \textit{grey level variance} feature, calculated by averaging the feature value from the SZM in each slice over all the slices.
\end{itemize}

\subsection{Abbreviating interpolation}
The following is a list of interpolation methods and the suggested notation. Note that \# is the interpolation spacing, including units, and \textit{dim} is 2D for interpolation with the slice plane and 3D for volumetric interpolation.

\begin{table}[ht]
\centering
\small
\begin{tabular}{lc}
\toprule
\textbf{interpolation method} & \textbf{notation}\\
\midrule
none & INT:-- \\
nearest neighbour interpolation & NNB:\textit{dim}:\# \\
linear interpolation & LIN:\textit{dim}:\# \\
cubic convolution interpolation & CCI:\textit{dim}:\# \\
cubic spline interpolation & CSI:\textit{dim}:\#, SI3:\textit{dim}:\# \\
\bottomrule
\end{tabular}
\end{table}

\FloatBarrier

The dimension attribute and interpolation spacing may be omitted if this is clear from the context. The following examples highlight the nomenclature introduced above:
\begin{itemize}
\item mean\textsubscript{IS, LIN:2D:2mm}: intensity statistical \textit{mean} feature, calculated after \textit{bilinear} interpolation with the slice planes to uniform voxel sizes of 2mm.
\item mean\textsubscript{IH, NNB:3D:1mm}: intensity histogram \textit{mean} feature, calculated after \textit{trilinear} interpolation to uniform voxel sizes of 1mm.
\item joint maximum\textsubscript{CM, 2D:mrg, CSI:2D:2mm}: GLCM-based \textit{joint maximum} feature, calculated by first merging all GLCM within a slice to single GLCM, calculating the feature and then averaging the feature values over the slices. GLCMs were determined in the image interpolated within the slice plane to 2 $\times$ 2mm voxels using \textit{cubic spline} interpolation.
\end{itemize}

\subsection{Describing re-segmentation}
Re-segmentation can be noted as follows:

\begin{table}[ht]
\centering
\small
\begin{tabular}{lcc}
\toprule
\textbf{re-segmentation method} & \textbf{notation}\\
\midrule
none & RS:-- & \\
range & RS:[\#,\#] & \textid{USB3}\\
outlier filtering & RS:\#$\sigma$ & \textid{7ACA}\\
\bottomrule
\end{tabular}
\end{table}

\FloatBarrier

In the table above \# signify numbers. A re-segmentation range can be half-open, i.e. RS:[\#,$\infty$). Re-segmentation methods may be combined, i.e. both range and outlier filtering methods may be used. This is noted as RS:[\#,\#]+\#$\sigma$ or RS:\#$\sigma$+[\#,\#]. The following are examples of the application of the above notation:
\begin{itemize}
\item mean\textsubscript{IS, CT, RS:[-200,150]}: intensity statistical \textit{mean} feature, based on an ROI in a CT image that was re-segmented within a [-200,150] HU range.
\item mean\textsubscript{IS, PET:SUV, RS:[3,$\infty$)}: intensity statistical \textit{mean} feature, based on an ROI in a PET image with SUV values, that was re-segmented to contain only SUV of 3 and above.
\item mean\textsubscript{IS, MRI:T1, RS:3$\sigma$}: intensity statistical \textit{mean} feature, based on an ROI in a T1-weighted MR image where the ROI was re-segmented by removing voxels with an intensity outside a $\mu \pm 3\sigma$ range.
\end{itemize}

\subsection{Abbreviating discretisation}
The following is a list of discretisation methods and the suggested notation. Note that \# is the value of the relevant discretisation parameter, e.g. number of bins or bin size, including units.

\begin{table}[ht]
\centering
\small
\begin{tabular}{lcc}
\toprule
\textbf{discretisation method} & \textbf{notation} &\\
\midrule
none & DIS:-- &\\
fixed bin size & FBS:\# & \textid{Q3RU}\\
fixed bin number & FBN:\# & \textid{K15C}\\
histogram equalisation & EQ:\#&\\
Lloyd-Max, minimum mean squared & LM:\#, MMS:\# &\\
\bottomrule
\end{tabular}
\end{table}

\FloatBarrier

In the table above, \# signify numbers such as the number of bins or their width. Histogram equalisation of the ROI intensities can be performed before the "none", "fixed bin size", "fixed bin number" or "Lloyd-Max, minimum mean squared" algorithms defined above, with \# specifying the number of bins in the histogram to be equalised. The following are examples of the application of the above notation:
\begin{itemize}
\item mean\textsubscript{IH,PET:SUV,RS[0,$\infty$],FBS:0.2}: intensity histogram \textit{mean} feature, based on an ROI in a SUV-PET image, with bin-width of 0.2 SUV, and binning from 0.0 SUV.
\item grey level variance\textsubscript{SZM,MR:T1,RS:3$\sigma$,FBN:64}: size zone matrix-based \textit{grey level variance} feature, based on an ROI in a T1-weighted MR image, with $3\sigma$ re-segmentation and subsequent binning into 64 bins.
\end{itemize}

\subsection{Abbreviating feature-specific parameters}
Some features and feature families require additional parameters, which may be varied. These are the following:
\begin{center}
\small
\begin{longtable}{clc}
\toprule
\endhead

\multicolumn{3}{r}{\textit{continued on next page}}
\endfoot

\bottomrule
\endlastfoot

\multicolumn{3}{l}{\textbf{grey level co-occurrence matrix}}\\
\midrule
\multicolumn{3}{l}{\textit{co-occurrence matrix symmetry}}\\
--, SYM & symmetrical co-occurrence matrices &\\
ASYM & asymmetrical co-occurrence matrices (not recommended) &\\
\noalign{\vskip 2mm}
\multicolumn{3}{l}{\textit{distance}}\\
$\delta$:\#, $\delta$-$\infty$:\# &  Chebyshev ($\ell_{\infty}$) norm with distance \# (default) & \textid{PVMT}\\
$\delta$-$2$:\# &  Euclidean ($\ell_{2}$) norm with distance \# & \textid{G9EV}\\
$\delta$-$1$:\# &  Manhattan ($\ell_{1}$) norm with distance \# & \textid{LIFZ}\\
\noalign{\vskip 2mm}
\multicolumn{3}{l}{\textit{distance weighting}}\\
--, w:1 & no weighting (default) &\\
w:f & weighting with function $f$ &\\
\noalign{\vskip 2mm}
\multicolumn{3}{l}{\textbf{grey level run length matrix}}\\
\midrule
\multicolumn{3}{l}{\textit{distance weighting}}\\
--, w:1 & no weighting (default) &\\
w:f & weighting with function $f$ &\\
\noalign{\vskip 2mm}
\multicolumn{3}{l}{\textbf{grey level size zone matrix}}\\
\midrule
\multicolumn{3}{l}{\textit{linkage distance}}\\
$\delta$:\#, $\delta$-$\infty$:\# &  Chebyshev ($\ell_{\infty}$) norm with distance (default) \# & \textid{PVMT}\\
$\delta$-$2$:\# &  Euclidean ($\ell_{2}$) norm with distance \# & \textid{G9EV}\\
$\delta$-$1$:\# &  Manhattan ($\ell_{1}$) norm with distance \# & \textid{LIFZ}\\
\noalign{\vskip 2mm}
\multicolumn{3}{l}{\textbf{grey level distance zone matrix}}\\
\midrule
\multicolumn{3}{l}{\textit{linkage distance}}\\
$\delta$:\#, $\delta$-$\infty$:\# &  Chebyshev ($\ell_{\infty}$) norm with distance (default) \# & \textid{PVMT}\\
$\delta$-$2$:\# &  Euclidean ($\ell_{2}$) norm with distance \# & \textid{G9EV}\\
$\delta$-$1$:\# &  Manhattan ($\ell_{1}$) norm with distance \# & \textid{LIFZ}\\
\noalign{\vskip 2mm}
\multicolumn{3}{l}{\textit{zone distance norm}}\\
$l$-$\infty$:\# &  Chebyshev ($\ell_{\infty}$) norm & \textid{PVMT}\\
$l$-$2$:\# &  Euclidean ($\ell_{2}$) norm & \textid{G9EV}\\
--, $l$-$1$:\# &  Manhattan ($\ell_{1}$) norm (default) & \textid{LIFZ}\\
\noalign{\vskip 2mm}
\multicolumn{3}{l}{\textbf{neighbourhood grey tone difference matrix}}\\
\midrule
\multicolumn{3}{l}{\textit{distance}}\\
$\delta$:\#, $\delta$-$\infty$:\# &  Chebyshev ($\ell_{\infty}$) norm with distance \# (default) & \textid{PVMT}\\
$\delta$-$2$:\# &  Euclidean ($\ell_{2}$) norm with distance \# & \textid{G9EV}\\
$\delta$-$1$:\# &  Manhattan ($\ell_{1}$) norm with distance \# & \textid{LIFZ}\\
\noalign{\vskip 2mm}
\multicolumn{3}{l}{\textit{distance weighting}}\\
--, w:1 & no weighting (default) &\\
w:f & weighting with function $f$ &\\
\noalign{\vskip 2mm}
\multicolumn{3}{l}{\textbf{neighbouring grey level dependence matrix}}\\
\midrule
\multicolumn{3}{l}{\textit{dependence coarseness}}\\
$\alpha$:\# & dependence coarseness parameter with value \# &\\
\noalign{\vskip 2mm}
\multicolumn{3}{l}{\textit{distance}}\\
$\delta$:\#, $\delta$-$\infty$:\# &  Chebyshev ($\ell_{\infty}$) norm with distance \# (default) & \textid{PVMT}\\
$\delta$-$2$:\# &  Euclidean ($\ell_{2}$) norm with distance \# & \textid{G9EV}\\
$\delta$-$1$:\# &  Manhattan ($\ell_{1}$) norm with distance \# & \textid{LIFZ}\\
\noalign{\vskip 2mm}
\multicolumn{3}{l}{\textit{distance weighting}}\\
--, w:1 & no weighting (default) &\\
w:f & weighting with function $f$ &\\
\end{longtable}
\end{center}

\FloatBarrier

In the above table, \# represents numbers.

\clearpage
\chapter{Reference data sets}\label{chap_benchmark sets}
Reference values for features were obtained using a digital image phantom and the CT image of a lung cancer patient, which are described below. The same data sets can be used to verify radiomics software implementations. The data sets themselves may be found here: https://github.com/theibsi/data\_sets.

\section{Digital phantom}\label{sec_digital_phantom}
\begin{figure}[p]
	\centering
	\includegraphics[scale=0.6]{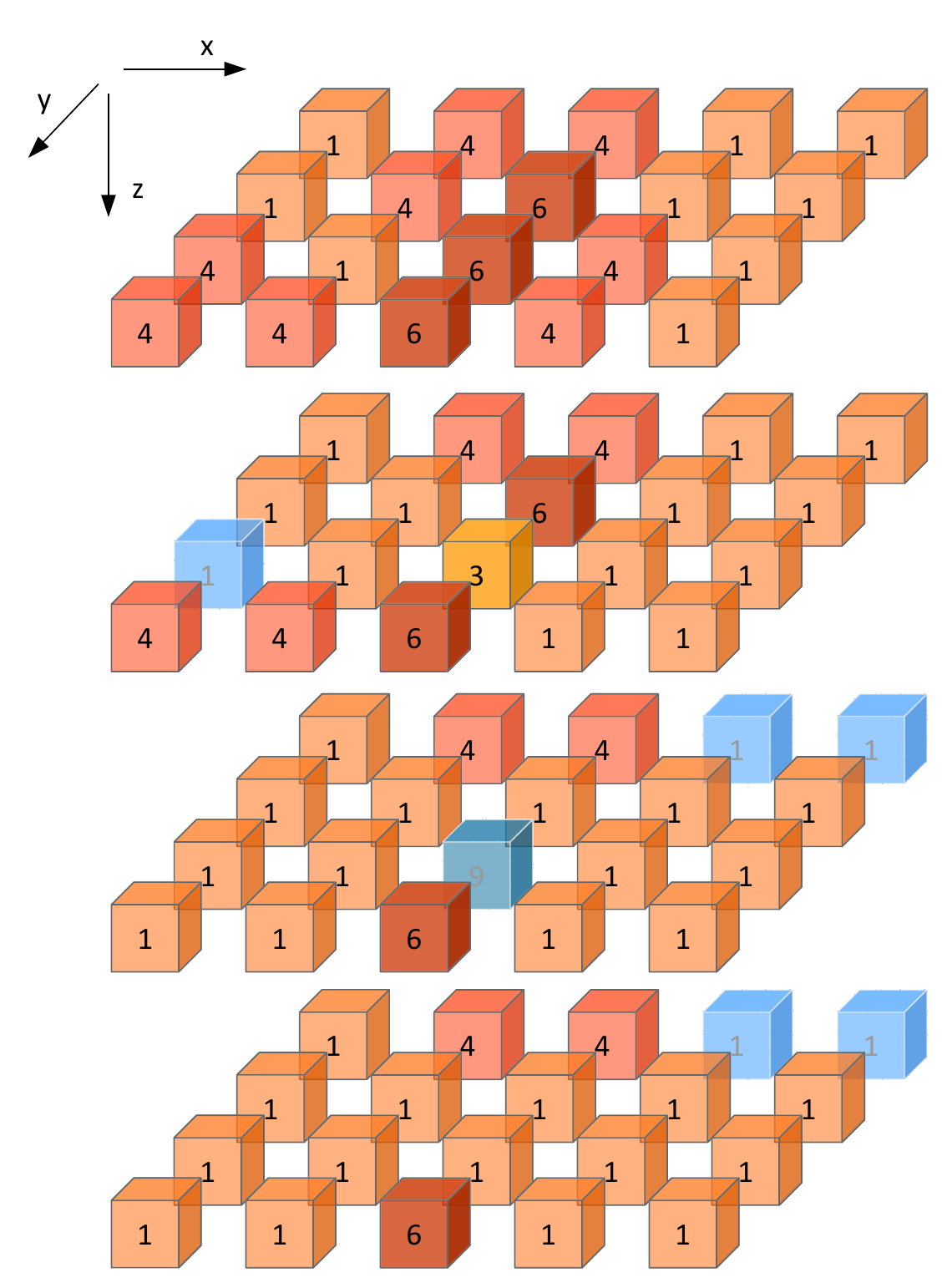}
	\caption{Exploded view of the test volume. The number in each voxel corresponds with its grey level. Blue voxels are excluded from the region of interest. The coordinate system is so that $x$ increases from left to right, $y$ increases from back to front and $z$ increases from top to bottom, as is indicated by the axis definition in the top-left.}
	\label{figTestVolume}
\end{figure}

A small digital phantom was developed to derive image features manually and compare these values with values obtained from radiomics software implementations. The phantom is shown in figure \ref{figTestVolume}. The phantom has the following characteristics:
\begin{itemize}
\item The phantom consists of $5 \times 4 \times 4$ $(x,y,z)$ voxels.
\item A slice consists of the voxels in $(x,y)$ plane for a particular slice at position $z$. Slices are therefore stacked in the $z$ direction.
\item Voxels are $2.0 \times 2.0 \times 2.0$ mm in size.
\item Not all voxels are included in the region of interest. Several excluded voxels are located on the outside of the ROI, and one internal voxel was excluded as well. Voxels excluded from the ROI are shown in blue in figure \ref{figTestVolume}.
\item Some intensities are not present in the phantom. Notably, grey levels $2$ and $5$ are absent. $1$ is the lowest grey level present in the ROI, and $6$ the highest.
\end{itemize}

\subsection{Computing image features}
The digital phantom was designed to not require image processing prior to calculating the features. Thus, feature calculation is done directly on the phantom itself. The following should be taken into account for calculating image features:
\begin{itemize}
\item Discretisation is not required. All features are to be calculated using the phantom as it is. Alternatively, one could use a \textit{fixed bin size} discretisation of 1 or \textit{fixed bin number} discretisation of 6 bins, which does not alter the contents of the phantom.
\item Grey level co-occurrence matrices are symmetrical and calculated for (Chebyshev) distance $\delta=1$.
\item Neighbouring grey level dependence and neighbourhood grey tone difference matrices are likewise calculated for (Chebyshev) distance $\delta=1$. Additionally, the neighbouring grey level dependence coarseness parameter has the value $\alpha=0$.
\item Because discretisation is lacking, most intensity-based statistical features will match their intensity histogram-based analogues in value.
\item The ROI morphological and intensity masks are identical for the digital phantom, due to lack of re-segmentation.
\end{itemize}

\section{Lung cancer CT image}\label{sec_patient_data}
A small data set of CT images from four non-small-cell lung carcinoma patients was made publicly available to serve as radiomics phantoms \href{http://dx.doi.org/10.17195/candat.2016.08.1}{(DOI:10.17195/candat.2016.08.1)}. We use the image for the first patient (\texttt{PAT1}) to obtain feature reference values for different configurations of the image processing scheme, as detailed below.

The CT image set is stored as a stack of slices in \texttt{DICOM} format. The image slices can be identified by the \texttt{DCM\_IMG} prefix. The gross tumour volume (GTV) was delineated and is used as the region of interest (ROI). Contour information is stored as an RT structure set in the DICOM file starting with \texttt{DCM\_RS}. For broader use, both the \texttt{DICOM} set and segmentation mask have been converted to the \texttt{NIfTI} format. When using the data in \texttt{NIfTI} format, both image stacks should be converted to (at least) 32-bit floating point and rounded to the nearest integer before further processing.

We defined five image processing configurations to test different image processing methods, see Table \ref{list_summary_cases}. While most settings are self-explanatory, there are several aspects that require some attention. Configurations are divided in 2D and 3D approaches. For the 2D configurations (A, B), image interpolation is conducted within the slice, and likewise texture features are extracted from the in-slice plane, and not volumetrically (3D). For the 3D configurations (C-E) interpolation is conducted in three dimensions, and features are likewise extracted volumetrically. Discretisation is moreover required for texture, intensity histogram and intensity-volume histogram features, and both \textit{fixed bin number} and \textit{fixed bin size} algorithms are tested.

\subsection{Notes on interpolation}\label{sec_benchmark_interpolation_notes}
Interpolation has a major influence on feature values. Different implementations of the same interpolation method may ostensibly provide the same functionality, but may use different interpolation grids. It is therefore recommended to read the documentation of the particular implementation to assess if the implementation allows or implements the following:
\begin{itemize}
\item The spatial origin of the original (input) grid in world coordinates matches the \texttt{DICOM} origin by definition.
\item The size of the interpolation grid is determined by rounding the fractional grid size towards infinity, i.e. a ceiling operation. This prevents the interpolation grid from disappearing for very small images, but is otherwise an arbitrary choice.
\item The centers of the interpolation and original image grids should be aligned, i.e. the interpolation grid is centered on the center of the original image grid. This prevents spacing inconsistencies in the interpolation grid and avoids potential issues with grid orientation.
\item The extent of the interpolation grid is, by definition, always equal or larger than that of the original grid. This means that intensities at the grid boundary are extrapolated. To facilitate this process, the image should be sufficiently padded with voxels that take on the nearest boundary intensity.
\item The floating point representation of the image and the ROI masks affects interpolation precision, and consequentially feature values. Image and ROI masks should at least be represented at full precision (\texttt{32-bit}) to avoid rounding errors. One example is the unintended exclusion of voxels from the interpolated ROI mask, which occurs when interpolation yields 0.4999\ldots instead of 0.5. When images and ROI masks are converted to full precision from lower precision (e.g. \texttt{16-bit}), values may require rounding if the original data were integer values, such as Hounsfield Units or the ROI mask labels.
\end{itemize}
More details are provided in Section \ref{ref_interpolation}.

\subsection{Diagnostic features}\label{sub_sect_diag_feat}
Identifying issues with an implementation of the image processing sequence may be challenging. Multiple steps follow one another and differences propagate. Hence we define a small number of diagnostic features that describe how the image and ROI masks change with each image processing step. These diagnostic features also have reference values that may be found in IBSI compliance check spreadsheet.

\paragraph{Initial image stack.}
The following features may be used to describe the initial image stack (i.e. after loading image data for processing):
\begin{itemize}
\item \textit{Image dimensions.} This describes the image dimensions in voxels along the different image axes.
\item \textit{Voxel dimensions.} This describes the voxel dimensions in mm. The dimension along the z-axis is equal to the distance between the origin voxels of two adjacent slices, and is generally equal to the slice thickness.
\item \textit{Mean intensity.} This is the average intensity within the entire image.
\item \textit{Minimum intensity.} This is the lowest intensity within the entire image.
\item \textit{Maximum intensity.} This is the highest intensity within the entire image.
\end{itemize}

\paragraph{Interpolated image stack.}
The above features may also be used to describe the image stack after image interpolation.

\paragraph{Initial region of interest.}
The following descriptors are used to describe the region of interest (ROI) directly after segmentation of the image:
\begin{itemize}
\item \textit{ROI intensity mask dimensions.} This describes the dimensions, in voxels, of the ROI intensity mask.
\item \textit{ROI intensity mask bounding box dimensions.} This describes the dimensions, in voxels, of the bounding box of the ROI intensity mask.
\item \textit{ROI morphological mask bounding box dimensions.} This describes the dimensions, in voxels, of the bounding box of the ROI morphological mask.
\item \textit{Number of voxels in the ROI intensity mask.} This describes the number of voxels included in the ROI intensity mask.
\item \textit{Number of voxels in the ROI morphological mask.} This describes the number of voxels included in the ROI intensity mask.
\item \textit{Mean ROI intensity.} This is the mean intensity of image voxels within the ROI intensity mask.
\item \textit{Minimum ROI intensity.} This is the lowest intensity of image voxels within the ROI intensity mask.
\item \textit{Maximum ROI intensity.} This is the highest intensity of image voxels within the ROI intensity mask.
\end{itemize}

\paragraph{Interpolated region of interest.}
The same features can be used to describe the ROI after interpolation of the ROI mask.

\paragraph{Re-segmented region of interest.}
Again, the same features as above can be used to describe the ROI after re-segmentation.

\subsection{Computing image features}
Unlike the digital phantom, the lung cancer CT image does require additional image processing, which is done according to the processing configurations described in Table \ref{list_summary_cases}. The following should be taken into account when calculating image features:
\begin{itemize}
\item Grey level co-occurrence matrices are symmetrical and calculated for (Chebyshev) distance $\delta=1$.
\item Neighbouring grey level dependence and neighbourhood grey tone difference matrices are likewise calculated for (Chebyshev) distance $\delta=1$. Additionally, the neighbouring grey level dependence coarseness parameter $\alpha=0$.
\item Intensity-based statistical features and their intensity histogram-based analogues will differ in value due to discretisation, in contrast to the same features for the digital phantom.
\item Due to re-segmentation, the ROI morphological and intensity masks are not identical.
\item Calculation of IVH feature: since by default CT contains calibrated and discrete intensities, no separate discretisation prior to the calculation of intensity-volume histogram features is required. This is the case for configurations A, B and D (i.e. \enquote{calibrated intensity units -- discrete case}). However, for configurations C and E, we re-discretise the ROI intensities prior to calculation of intensity-volume histogram features to allow for testing of of these methods. Configuration C simulates the \enquote{calibrated intensity units -- continuous case}, while configuration E simulates the \enquote{arbitrary intensity units} case where the re-segmentation range is not used. For details, please consult section \ref{sect_ivh}.
\end{itemize}

\begin{landscape}
\begin{table}
\centering
\small
\begin{tabular}{llllll}
%\begin{tabu} to 0.99 \linewidth {@{}X[1.5,l,p] X[1,l,p] X[1,l,p] X[1,l,p] X[1,l,p] X[1,l,p]@{}}
\toprule
\textbf{Parameter} & \textbf{Config. A} & \textbf{Config. B} & \textbf{Config. C} & \textbf{Config. D} & \textbf{Config. E}\\
\midrule
sample identifier & \texttt{PAT1} & \texttt{PAT1} & \texttt{PAT1} & \texttt{PAT1} & \texttt{PAT1}\\
ROI name & \texttt{GTV-1} & \texttt{GTV-1} & \texttt{GTV-1} & \texttt{GTV-1} & \texttt{GTV-1}\\
slice-wise or single volume (3D) & 2D & 2D & 3D & 3D & 3D\\
interpolation & no & yes & yes & yes & yes\\
\qquad resampled voxel spacing (mm)& & $2\times 2$ (axial) & $2\times 2\times 2$ & $2\times 2\times 2$ & $2\times 2\times 2$\\
\qquad interpolation method & & bilinear & trilinear & trilinear & tricubic spline\\
\qquad intensity rounding & & nearest integer & nearest integer & nearest integer & nearest integer\\
\qquad ROI interpolation method & & bilinear & trilinear & trilinear & trilinear\\
\qquad ROI partial mask volume & & $0.5$ & $0.5$ & $0.5$ & $0.5$\\
re-segmentation & & & & &\\
\qquad range (HU) & $\left[-500,400\right]$ & $\left[-500,400\right]$ & $\left[-1000,400\right]$ & no & $\left[-1000,400\right]$\\
\qquad outlier filtering & no & no & no & $3\sigma$ & $3\sigma$\\
discretisation & & & & &\\
\qquad texture and IH & FBS: 25 HU & FBN: 32 bins & FBS: 25 HU & FBN: 32 bins & FBN: 32 bins\\
\qquad IVH & no & no &  FBS: 2.5 HU & no & FBN: 1000 bins\\
texture parameters & & & & &\\
\qquad GLCM, NGTDM, NGLDM distance & 1 & 1 & 1 & 1 & 1\\
\qquad GLSZM, GLDZM linkage distance & 1 & 1 & 1 & 1 & 1\\
\qquad NGLDM coarseness & 0.0 & 0.0 & 0.0 & 0.0 & 0.0\\
\bottomrule
\end{tabular}
\caption{Different configurations for image processing. For details, refer to the corresponding sections in chapter \ref{chap_img_proc}. ROI: region of interest; HU: Hounsfield Unit; IH: intensity histogram; FBS: fixed bin size; FBN: fixed bin number; IVH: intensity-volume histogram; GLCM: grey level co-occurrence matrix; NGTDM: neighborhood grey tone difference matrix; NGLDM: neighbouring grey level dependence matrix; GLSZM: grey level size zone matrix; GLDZM: grey level distance zone matrix.}
\label{list_summary_cases}
\end{table}
\end{landscape}

\newpage
\appendix
\chapter{Digital phantom texture matrices}
This section contains the texture matrices extracted from the digital phantom for reference purposes.

\section{Grey level co-occurrence matrix (2D)}
\begin{table}[ht]
\centering
\small
\begin{subtable}[t]{3cm}\centering
\begin{tabular}[t]{@{}rrr@{}}
\toprule
   i &    j &   n \\
\midrule
 1.0 &  1.0 &  10 \\
 1.0 &  4.0 &   4 \\
 4.0 &  1.0 &   4 \\
 4.0 &  4.0 &   6 \\
 4.0 &  6.0 &   1 \\
 6.0 &  4.0 &   1 \\
 6.0 &  6.0 &   4 \\
\bottomrule
\end{tabular}
\caption{\textbf{x}: (0,1,0)\\slice: 1 of 4}
\end{subtable}
\hfill
\begin{subtable}[t]{3cm}\centering
\begin{tabular}[t]{@{}rrr@{}}
\toprule
   i &    j &   n \\
\midrule
 1.0 &  1.0 &  16 \\
 1.0 &  4.0 &   2 \\
 3.0 &  6.0 &   2 \\
 4.0 &  1.0 &   2 \\
 4.0 &  6.0 &   1 \\
 6.0 &  3.0 &   2 \\
 6.0 &  4.0 &   1 \\
\bottomrule
\end{tabular}
\caption{\textbf{x}: (0,1,0)\\slice: 2 of 4}
\end{subtable}
\hfill
\begin{subtable}[t]{3cm}\centering
\begin{tabular}[t]{@{}rrr@{}}
\toprule
   i &    j &   n \\
\midrule
 1.0 &  1.0 &  18 \\
 1.0 &  4.0 &   2 \\
 4.0 &  1.0 &   2 \\
\bottomrule
\end{tabular}
\caption{\textbf{x}: (0,1,0)\\slice: 3 of 4}
\end{subtable}
\hfill
\begin{subtable}[t]{3cm}\centering
\begin{tabular}[t]{@{}rrr@{}}
\toprule
   i &    j &   n \\
\midrule
 1.0 &  1.0 &  20 \\
 1.0 &  4.0 &   2 \\
 1.0 &  6.0 &   1 \\
 4.0 &  1.0 &   2 \\
 6.0 &  1.0 &   1 \\
\bottomrule
\end{tabular}
\caption{\textbf{x}: (0,1,0)\\slice: 4 of 4}
\end{subtable}
\end{table}
\begin{table}[h]\ContinuedFloat
\centering
\small
\begin{subtable}[t]{3cm}\centering
\begin{tabular}[t]{@{}rrr@{}}
\toprule
   i &    j &  n \\
\midrule
 1.0 &  1.0 &  2 \\
 1.0 &  4.0 &  4 \\
 1.0 &  6.0 &  3 \\
 4.0 &  1.0 &  4 \\
 4.0 &  4.0 &  4 \\
 4.0 &  6.0 &  2 \\
 6.0 &  1.0 &  3 \\
 6.0 &  4.0 &  2 \\
\bottomrule
\end{tabular}
\caption{\textbf{x}: (1,-1,0)\\slice: 1 of 4}
\end{subtable}
\hfill
\begin{subtable}[t]{3cm}\centering
\begin{tabular}[t]{@{}rrr@{}}
\toprule
   i &    j &  n \\
\midrule
 1.0 &  1.0 &  6 \\
 1.0 &  3.0 &  1 \\
 1.0 &  4.0 &  3 \\
 1.0 &  6.0 &  3 \\
 3.0 &  1.0 &  1 \\
 3.0 &  4.0 &  1 \\
 4.0 &  1.0 &  3 \\
 4.0 &  3.0 &  1 \\
 6.0 &  1.0 &  3 \\
\bottomrule
\end{tabular}
\caption{\textbf{x}: (1,-1,0)\\slice: 2 of 4}
\end{subtable}
\hfill
\begin{subtable}[t]{3cm}\centering
\begin{tabular}[t]{@{}rrr@{}}
\toprule
   i &    j &   n \\
\midrule
 1.0 &  1.0 &  10 \\
 1.0 &  4.0 &   2 \\
 1.0 &  6.0 &   1 \\
 4.0 &  1.0 &   2 \\
 6.0 &  1.0 &   1 \\
\bottomrule
\end{tabular}
\caption{\textbf{x}: (1,-1,0)\\slice: 3 of 4}
\end{subtable}
\hfill
\begin{subtable}[t]{3cm}\centering
\begin{tabular}[t]{@{}rrr@{}}
\toprule
   i &    j &   n \\
\midrule
 1.0 &  1.0 &  14 \\
 1.0 &  4.0 &   2 \\
 1.0 &  6.0 &   1 \\
 4.0 &  1.0 &   2 \\
 6.0 &  1.0 &   1 \\
\bottomrule
\end{tabular}
\caption{\textbf{x}: (1,-1,0)\\slice: 4 of 4}
\end{subtable}
\end{table}
\begin{table}[ht]\ContinuedFloat
\centering
\small
\begin{subtable}[t]{3cm}\centering
\begin{tabular}[t]{@{}rrr@{}}
\toprule
   i &    j &  n \\
\midrule
 1.0 &  1.0 &  4 \\
 1.0 &  4.0 &  6 \\
 1.0 &  6.0 &  2 \\
 4.0 &  1.0 &  6 \\
 4.0 &  4.0 &  4 \\
 4.0 &  6.0 &  4 \\
 6.0 &  1.0 &  2 \\
 6.0 &  4.0 &  4 \\
\bottomrule
\end{tabular}
\caption{\textbf{d}: (1,0,0)\\slice: 1 of 4}
\end{subtable}
\hfill
\begin{subtable}[t]{3cm}\centering
\begin{tabular}[t]{@{}rrr@{}}
\toprule
   i &    j &   n \\
\midrule
 1.0 &  1.0 &  10 \\
 1.0 &  3.0 &   2 \\
 1.0 &  4.0 &   2 \\
 1.0 &  6.0 &   3 \\
 3.0 &  1.0 &   2 \\
 4.0 &  1.0 &   2 \\
 4.0 &  4.0 &   4 \\
 4.0 &  6.0 &   1 \\
 6.0 &  1.0 &   3 \\
 6.0 &  4.0 &   1 \\
\bottomrule
\end{tabular}
\caption{\textbf{d}: (1,0,0)\\slice: 2 of 4}
\end{subtable}
\hfill
\begin{subtable}[t]{3cm}\centering
\begin{tabular}[t]{@{}rrr@{}}
\toprule
   i &    j &   n \\
\midrule
 1.0 &  1.0 &  16 \\
 1.0 &  4.0 &   1 \\
 1.0 &  6.0 &   2 \\
 4.0 &  1.0 &   1 \\
 4.0 &  4.0 &   2 \\
 6.0 &  1.0 &   2 \\
\bottomrule
\end{tabular}
\caption{\textbf{d}: (1,0,0)\\slice: 3 of 4}
\end{subtable}
\hfill
\begin{subtable}[t]{3cm}\centering
\begin{tabular}[t]{@{}rrr@{}}
\toprule
   i &    j &   n \\
\midrule
 1.0 &  1.0 &  20 \\
 1.0 &  4.0 &   1 \\
 1.0 &  6.0 &   2 \\
 4.0 &  1.0 &   1 \\
 4.0 &  4.0 &   2 \\
 6.0 &  1.0 &   2 \\
\bottomrule
\end{tabular}
\caption{\textbf{d}: (1,0,0)\\slice: 4 of 4}
\end{subtable}
\end{table}
\begin{table}[ht]\ContinuedFloat
\centering
\small
\begin{subtable}[t]{3cm}\centering
\begin{tabular}[t]{@{}rrr@{}}
\toprule
   i &    j &  n \\
\midrule
 1.0 &  1.0 &  6 \\
 1.0 &  4.0 &  3 \\
 1.0 &  6.0 &  1 \\
 4.0 &  1.0 &  3 \\
 4.0 &  4.0 &  2 \\
 4.0 &  6.0 &  4 \\
 6.0 &  1.0 &  1 \\
 6.0 &  4.0 &  4 \\
\bottomrule
\end{tabular}
\caption{\textbf{d}: (1,1,0)\\slice: 1 of 4}
\end{subtable}
\hfill
\begin{subtable}[t]{3cm}\centering
\begin{tabular}[t]{@{}rrr@{}}
\toprule
   i &    j &   n \\
\midrule
 1.0 &  1.0 &  10 \\
 1.0 &  3.0 &   2 \\
 1.0 &  4.0 &   1 \\
 1.0 &  6.0 &   2 \\
 3.0 &  1.0 &   2 \\
 4.0 &  1.0 &   1 \\
 4.0 &  6.0 &   1 \\
 6.0 &  1.0 &   2 \\
 6.0 &  4.0 &   1 \\
\bottomrule
\end{tabular}
\caption{\textbf{d}: (1,1,0)\\slice: 2 of 4}
\end{subtable}
\hfill
\begin{subtable}[t]{3cm}\centering
\begin{tabular}[t]{@{}rrr@{}}
\toprule
   i &    j &   n \\
\midrule
 1.0 &  1.0 &  12 \\
 1.0 &  4.0 &   2 \\
 1.0 &  6.0 &   1 \\
 4.0 &  1.0 &   2 \\
 6.0 &  1.0 &   1 \\
\bottomrule
\end{tabular}
\caption{\textbf{d}: (1,1,0)\\slice: 3 of 4}
\end{subtable}
\hfill
\begin{subtable}[t]{3cm}\centering
\begin{tabular}[t]{@{}rrr@{}}
\toprule
   i &    j &   n \\
\midrule
 1.0 &  1.0 &  16 \\
 1.0 &  4.0 &   2 \\
 1.0 &  6.0 &   1 \\
 4.0 &  1.0 &   2 \\
 6.0 &  1.0 &   1 \\
\bottomrule
\end{tabular}
\caption{\textbf{d}: (1,1,0)\\slice: 4 of 4}
\end{subtable}
\caption{Grey-level co-occurrence matrices extracted from the $xy$ plane (2D) of the digital phantom using Chebyshev distance 1. \textbf{x} indicates the direction in $(x,y,z)$ coordinates.}
\end{table}

\FloatBarrier

\section{Grey level co-occurrence matrix (2D, merged)}
\begin{table}[ht]
\centering
\small
\begin{subtable}[t]{3cm}\centering
\begin{tabular}[t]{@{}rrr@{}}
\toprule
   i &    j &   n \\
\midrule
 1.0 &  1.0 &  22 \\
 1.0 &  4.0 &  17 \\
 1.0 &  6.0 &   6 \\
 4.0 &  1.0 &  17 \\
 4.0 &  4.0 &  16 \\
 4.0 &  6.0 &  11 \\
 6.0 &  1.0 &   6 \\
 6.0 &  4.0 &  11 \\
 6.0 &  6.0 &   4 \\
\bottomrule
\end{tabular}
\caption{slice: 1 of 4}
\end{subtable}
\hfill
\begin{subtable}[t]{3cm}\centering
\begin{tabular}[t]{@{}rrr@{}}
\toprule
   i &    j &   n \\
\midrule
 1.0 &  1.0 &  42 \\
 1.0 &  3.0 &   5 \\
 1.0 &  4.0 &   8 \\
 1.0 &  6.0 &   8 \\
 3.0 &  1.0 &   5 \\
 3.0 &  4.0 &   1 \\
 3.0 &  6.0 &   2 \\
 4.0 &  1.0 &   8 \\
 4.0 &  3.0 &   1 \\
 4.0 &  4.0 &   4 \\
 4.0 &  6.0 &   3 \\
 6.0 &  1.0 &   8 \\
 6.0 &  3.0 &   2 \\
 6.0 &  4.0 &   3 \\
\bottomrule
\end{tabular}
\caption{slice: 2 of 4}
\end{subtable}
\hfill
\begin{subtable}[t]{3cm}\centering
\begin{tabular}[t]{@{}rrr@{}}
\toprule
   i &    j &   n \\
\midrule
 1.0 &  1.0 &  56 \\
 1.0 &  4.0 &   7 \\
 1.0 &  6.0 &   4 \\
 4.0 &  1.0 &   7 \\
 4.0 &  4.0 &   2 \\
 6.0 &  1.0 &   4 \\
\bottomrule
\end{tabular}
\caption{slice: 3 of 4}
\end{subtable}
\hfill
\begin{subtable}[t]{3cm}\centering
\begin{tabular}[t]{@{}rrr@{}}
\toprule
   i &    j &   n \\
\midrule
 1.0 &  1.0 &  70 \\
 1.0 &  4.0 &   7 \\
 1.0 &  6.0 &   5 \\
 4.0 &  1.0 &   7 \\
 4.0 &  4.0 &   2 \\
 6.0 &  1.0 &   5 \\
\bottomrule
\end{tabular}
\caption{slice: 4 of 4}
\end{subtable}
\caption{Merged grey-level co-occurrence matrices extracted from the $xy$ plane (2D) of the digital phantom using Chebyshev distance 1.}
\end{table}

\FloatBarrier

\section{Grey level co-occurrence matrix (3D)}
\begin{table}[ht]
\centering
\small
\begin{subtable}[t]{3cm}\centering
\begin{tabular}[t]{@{}rrr@{}}
\toprule
   i &    j &   n \\
\midrule
 1.0 &  1.0 &  66 \\
 1.0 &  4.0 &   5 \\
 1.0 &  6.0 &   1 \\
 3.0 &  6.0 &   1 \\
 4.0 &  1.0 &   5 \\
 4.0 &  4.0 &  16 \\
 6.0 &  1.0 &   1 \\
 6.0 &  3.0 &   1 \\
 6.0 &  6.0 &   8 \\
\bottomrule
\end{tabular}
\caption{\textbf{x}: (0,0,1)}
\end{subtable}
\hfill
\begin{subtable}[t]{3cm}\centering
\begin{tabular}[t]{@{}rrr@{}}
\toprule
   i &    j &   n \\
\midrule
 1.0 &  1.0 &  42 \\
 1.0 &  3.0 &   1 \\
 1.0 &  4.0 &   9 \\
 1.0 &  6.0 &   1 \\
 3.0 &  1.0 &   1 \\
 3.0 &  6.0 &   1 \\
 4.0 &  1.0 &   9 \\
 4.0 &  4.0 &   2 \\
 4.0 &  6.0 &   2 \\
 6.0 &  1.0 &   1 \\
 6.0 &  3.0 &   1 \\
 6.0 &  4.0 &   2 \\
 6.0 &  6.0 &   2 \\
\bottomrule
\end{tabular}
\caption{\textbf{x}: (0,1,-1)}
\end{subtable}
\hfill
\begin{subtable}[t]{3cm}\centering
\begin{tabular}[t]{@{}rrr@{}}
\toprule
   i &    j &   n \\
\midrule
 1.0 &  1.0 &  64 \\
 1.0 &  4.0 &  10 \\
 1.0 &  6.0 &   1 \\
 3.0 &  6.0 &   2 \\
 4.0 &  1.0 &  10 \\
 4.0 &  4.0 &   6 \\
 4.0 &  6.0 &   2 \\
 6.0 &  1.0 &   1 \\
 6.0 &  3.0 &   2 \\
 6.0 &  4.0 &   2 \\
 6.0 &  6.0 &   4 \\
\bottomrule
\end{tabular}
\caption{\textbf{x}: (0,1,0)}
\end{subtable}
\hfill
\begin{subtable}[t]{3cm}\centering
\begin{tabular}[t]{@{}rrr@{}}
\toprule
   i &    j &   n \\
\midrule
 1.0 &  1.0 &  52 \\
 1.0 &  4.0 &   8 \\
 3.0 &  6.0 &   2 \\
 4.0 &  1.0 &   8 \\
 4.0 &  4.0 &   2 \\
 4.0 &  6.0 &   1 \\
 6.0 &  3.0 &   2 \\
 6.0 &  4.0 &   1 \\
 6.0 &  6.0 &   2 \\
\bottomrule
\end{tabular}
\caption{\textbf{x}: (0,1,1)}
\end{subtable}
\end{table}
\begin{table}[ht]\ContinuedFloat
\centering
\small
\begin{subtable}[t]{3cm}\centering
\begin{tabular}[t]{@{}rrr@{}}
\toprule
   i &    j &   n \\
\midrule
 1.0 &  1.0 &  30 \\
 1.0 &  3.0 &   2 \\
 1.0 &  4.0 &   7 \\
 1.0 &  6.0 &   5 \\
 3.0 &  1.0 &   2 \\
 4.0 &  1.0 &   7 \\
 4.0 &  6.0 &   2 \\
 6.0 &  1.0 &   5 \\
 6.0 &  4.0 &   2 \\
\bottomrule
\end{tabular}
\caption{\textbf{x}: (1,-1,-1)}
\end{subtable}
\hfill
\begin{subtable}[t]{3cm}\centering
\begin{tabular}[t]{@{}rrr@{}}
\toprule
   i &    j &   n \\
\midrule
 1.0 &  1.0 &  32 \\
 1.0 &  3.0 &   1 \\
 1.0 &  4.0 &  11 \\
 1.0 &  6.0 &   8 \\
 3.0 &  1.0 &   1 \\
 3.0 &  4.0 &   1 \\
 4.0 &  1.0 &  11 \\
 4.0 &  3.0 &   1 \\
 4.0 &  4.0 &   4 \\
 4.0 &  6.0 &   2 \\
 6.0 &  1.0 &   8 \\
 6.0 &  4.0 &   2 \\
\bottomrule
\end{tabular}
\caption{\textbf{x}: (1,-1,0)}
\end{subtable}
\hfill
\begin{subtable}[t]{3cm}\centering
\begin{tabular}[t]{@{}rrr@{}}
\toprule
   i &    j &   n \\
\midrule
 1.0 &  1.0 &  20 \\
 1.0 &  3.0 &   1 \\
 1.0 &  4.0 &  10 \\
 1.0 &  6.0 &   6 \\
 3.0 &  1.0 &   1 \\
 3.0 &  4.0 &   1 \\
 4.0 &  1.0 &  10 \\
 4.0 &  3.0 &   1 \\
 4.0 &  4.0 &   2 \\
 6.0 &  1.0 &   6 \\
\bottomrule
\end{tabular}
\caption{\textbf{x}: (1,-1,1)}
\end{subtable}
\hfill
\begin{subtable}[t]{3cm}\centering
\begin{tabular}[t]{@{}rrr@{}}
\toprule
   i &    j &   n \\
\midrule
 1.0 &  1.0 &  38 \\
 1.0 &  3.0 &   1 \\
 1.0 &  4.0 &   7 \\
 1.0 &  6.0 &   8 \\
 3.0 &  1.0 &   1 \\
 3.0 &  4.0 &   1 \\
 4.0 &  1.0 &   7 \\
 4.0 &  3.0 &   1 \\
 4.0 &  4.0 &   8 \\
 4.0 &  6.0 &   2 \\
 6.0 &  1.0 &   8 \\
 6.0 &  4.0 &   2 \\
\bottomrule
\end{tabular}
\caption{\textbf{x}: (1,0,-1)}
\end{subtable}
\end{table}
\begin{table}[ht]\ContinuedFloat
\centering
\small
\begin{subtable}[t]{3cm}\centering
\begin{tabular}[t]{@{}rrr@{}}
\toprule
   i &    j &   n \\
\midrule
 1.0 &  1.0 &  50 \\
 1.0 &  3.0 &   2 \\
 1.0 &  4.0 &  10 \\
 1.0 &  6.0 &   9 \\
 3.0 &  1.0 &   2 \\
 4.0 &  1.0 &  10 \\
 4.0 &  4.0 &  12 \\
 4.0 &  6.0 &   5 \\
 6.0 &  1.0 &   9 \\
 6.0 &  4.0 &   5 \\
\bottomrule
\end{tabular}
\caption{\textbf{x}: (1,0,0)}
\end{subtable}
\hfill
\begin{subtable}[t]{3cm}\centering
\begin{tabular}[t]{@{}rrr@{}}
\toprule
   i &    j &   n \\
\midrule
 1.0 &  1.0 &  34 \\
 1.0 &  3.0 &   2 \\
 1.0 &  4.0 &   8 \\
 1.0 &  6.0 &   7 \\
 3.0 &  1.0 &   2 \\
 4.0 &  1.0 &   8 \\
 4.0 &  4.0 &   8 \\
 4.0 &  6.0 &   3 \\
 6.0 &  1.0 &   7 \\
 6.0 &  4.0 &   3 \\
\bottomrule
\end{tabular}
\caption{\textbf{x}: (1,0,1)}
\end{subtable}
\hfill
\begin{subtable}[t]{3cm}\centering
\begin{tabular}[t]{@{}rrr@{}}
\toprule
   i &    j &   n \\
\midrule
 1.0 &  1.0 &  32 \\
 1.0 &  3.0 &   1 \\
 1.0 &  4.0 &   6 \\
 1.0 &  6.0 &   4 \\
 3.0 &  1.0 &   1 \\
 3.0 &  4.0 &   1 \\
 4.0 &  1.0 &   6 \\
 4.0 &  3.0 &   1 \\
 4.0 &  6.0 &   3 \\
 6.0 &  1.0 &   4 \\
 6.0 &  4.0 &   3 \\
\bottomrule
\end{tabular}
\caption{\textbf{x}: (1,1,-1)}
\end{subtable}
\hfill
\begin{subtable}[t]{3cm}\centering
\begin{tabular}[t]{@{}rrr@{}}
\toprule
   i &    j &   n \\
\midrule
 1.0 &  1.0 &  44 \\
 1.0 &  3.0 &   2 \\
 1.0 &  4.0 &   8 \\
 1.0 &  6.0 &   5 \\
 3.0 &  1.0 &   2 \\
 4.0 &  1.0 &   8 \\
 4.0 &  4.0 &   2 \\
 4.0 &  6.0 &   5 \\
 6.0 &  1.0 &   5 \\
 6.0 &  4.0 &   5 \\
\bottomrule
\end{tabular}
\caption{\textbf{x}: (1,1,0)}
\end{subtable}
\end{table}
\begin{table}[ht]\ContinuedFloat
\centering
\small
\begin{subtable}[t]{3cm}\centering
\begin{tabular}[t]{@{}rrr@{}}
\toprule
   i &    j &   n \\
\midrule
 1.0 &  1.0 &  32 \\
 1.0 &  3.0 &   1 \\
 1.0 &  4.0 &   6 \\
 1.0 &  6.0 &   6 \\
 3.0 &  1.0 &   1 \\
 3.0 &  4.0 &   1 \\
 4.0 &  1.0 &   6 \\
 4.0 &  3.0 &   1 \\
 4.0 &  4.0 &   2 \\
 4.0 &  6.0 &   1 \\
 6.0 &  1.0 &   6 \\
 6.0 &  4.0 &   1 \\
\bottomrule
\end{tabular}
\caption{\textbf{x}: (1,1,1)}
\end{subtable}
\caption{Grey-level co-occurrence matrices extracted volumetrically (3D) from the digital phantom using Chebyshev distance 1. \textbf{x} indicates the direction in $(x,y,z)$ coordinates.}
\end{table}

\FloatBarrier

\section{Grey level co-occurrence matrix (3D, merged)}
\begin{table}[ht]
\centering
\small
\begin{subtable}[t]{3cm}\centering
\begin{tabular}[t]{@{}rrr@{}}
\toprule
   i &    j &    n \\
\midrule
 1.0 &  1.0 &  536 \\
 1.0 &  3.0 &   14 \\
 1.0 &  4.0 &  105 \\
 1.0 &  6.0 &   61 \\
 3.0 &  1.0 &   14 \\
 3.0 &  4.0 &    5 \\
 3.0 &  6.0 &    6 \\
 4.0 &  1.0 &  105 \\
 4.0 &  3.0 &    5 \\
 4.0 &  4.0 &   64 \\
 4.0 &  6.0 &   28 \\
 6.0 &  1.0 &   61 \\
 6.0 &  3.0 &    6 \\
 6.0 &  4.0 &   28 \\
 6.0 &  6.0 &   16 \\
\bottomrule
\end{tabular}
\end{subtable}
\caption{Merged grey-level co-occurrence matrix extracted volumetrically (3D) from the digital phantom using Chebyshev distance 1.}
\end{table}

\FloatBarrier

\section{Grey level run length matrix (2D)}
\begin{table}[ht]
\centering
\small
\begin{subtable}[t]{3cm}\centering
\begin{tabular}[t]{@{}rrr@{}}
\toprule
   i &    r &    n \\
\midrule
 1.0 &  1.0 &  1.0 \\
 1.0 &  2.0 &  2.0 \\
 1.0 &  4.0 &  1.0 \\
 4.0 &  1.0 &  2.0 \\
 4.0 &  2.0 &  3.0 \\
 6.0 &  3.0 &  1.0 \\
\bottomrule
\end{tabular}
\caption{\textbf{x}: (0,1,0)\\slice: 1 of 4}
\end{subtable}
\hfill
\begin{subtable}[t]{3cm}\centering
\begin{tabular}[t]{@{}rrr@{}}
\toprule
   i &    r &    n \\
\midrule
 1.0 &  2.0 &  2.0 \\
 1.0 &  4.0 &  2.0 \\
 3.0 &  1.0 &  1.0 \\
 4.0 &  1.0 &  4.0 \\
 6.0 &  1.0 &  2.0 \\
\bottomrule
\end{tabular}
\caption{\textbf{x}: (0,1,0)\\slice: 2 of 4}
\end{subtable}
\hfill
\begin{subtable}[t]{3cm}\centering
\begin{tabular}[t]{@{}rrr@{}}
\toprule
   i &    r &    n \\
\midrule
 1.0 &  1.0 &  1.0 \\
 1.0 &  3.0 &  3.0 \\
 1.0 &  4.0 &  1.0 \\
 4.0 &  1.0 &  2.0 \\
 6.0 &  1.0 &  1.0 \\
\bottomrule
\end{tabular}
\caption{\textbf{x}: (0,1,0)\\slice: 3 of 4}
\end{subtable}
\hfill
\begin{subtable}[t]{3cm}\centering
\begin{tabular}[t]{@{}rrr@{}}
\toprule
   i &    r &    n \\
\midrule
 1.0 &  2.0 &  1.0 \\
 1.0 &  3.0 &  3.0 \\
 1.0 &  4.0 &  1.0 \\
 4.0 &  1.0 &  2.0 \\
 6.0 &  1.0 &  1.0 \\
\bottomrule
\end{tabular}
\caption{\textbf{x}: (0,1,0)\\slice: 4 of 4}
\end{subtable}
\end{table}
\begin{table}[ht]\ContinuedFloat
\centering
\small
\begin{subtable}[t]{3cm}\centering
\begin{tabular}[t]{@{}rrr@{}}
\toprule
   i &    r &    n \\
\midrule
 1.0 &  1.0 &  7.0 \\
 1.0 &  2.0 &  1.0 \\
 4.0 &  1.0 &  5.0 \\
 4.0 &  3.0 &  1.0 \\
 6.0 &  1.0 &  3.0 \\
\bottomrule
\end{tabular}
\caption{\textbf{x}: (1,-1,0)\\slice: 1 of 4}
\end{subtable}
\hfill
\begin{subtable}[t]{3cm}\centering
\begin{tabular}[t]{@{}rrr@{}}
\toprule
   i &    r &    n \\
\midrule
 1.0 &  1.0 &  6.0 \\
 1.0 &  2.0 &  3.0 \\
 3.0 &  1.0 &  1.0 \\
 4.0 &  1.0 &  4.0 \\
 6.0 &  1.0 &  2.0 \\
\bottomrule
\end{tabular}
\caption{\textbf{x}: (1,-1,0)\\slice: 2 of 4}
\end{subtable}
\hfill
\begin{subtable}[t]{3cm}\centering
\begin{tabular}[t]{@{}rrr@{}}
\toprule
   i &    r &    n \\
\midrule
 1.0 &  1.0 &  5.0 \\
 1.0 &  2.0 &  3.0 \\
 1.0 &  3.0 &  1.0 \\
 4.0 &  1.0 &  2.0 \\
 6.0 &  1.0 &  1.0 \\
\bottomrule
\end{tabular}
\caption{\textbf{x}: (1,-1,0)\\slice: 3 of 4}
\end{subtable}
\hfill
\begin{subtable}[t]{3cm}\centering
\begin{tabular}[t]{@{}rrr@{}}
\toprule
   i &    r &    n \\
\midrule
 1.0 &  1.0 &  3.0 \\
 1.0 &  2.0 &  3.0 \\
 1.0 &  3.0 &  2.0 \\
 4.0 &  1.0 &  2.0 \\
 6.0 &  1.0 &  1.0 \\
\bottomrule
\end{tabular}
\caption{\textbf{x}: (1,-1,0)\\slice: 4 of 4}
\end{subtable}
\end{table}
\begin{table}[ht]\ContinuedFloat
\centering
\small
\begin{subtable}[t]{3cm}\centering
\begin{tabular}[t]{@{}rrr@{}}
\toprule
   i &    r &    n \\
\midrule
 1.0 &  1.0 &  5.0 \\
 1.0 &  2.0 &  2.0 \\
 4.0 &  1.0 &  4.0 \\
 4.0 &  2.0 &  2.0 \\
 6.0 &  1.0 &  3.0 \\
\bottomrule
\end{tabular}
\caption{\textbf{x}: (1,0,0)\\slice: 1 of 4}
\end{subtable}
\hfill
\begin{subtable}[t]{3cm}\centering
\begin{tabular}[t]{@{}rrr@{}}
\toprule
   i &    r &    n \\
\midrule
 1.0 &  1.0 &  2.0 \\
 1.0 &  2.0 &  5.0 \\
 3.0 &  1.0 &  1.0 \\
 4.0 &  2.0 &  2.0 \\
 6.0 &  1.0 &  2.0 \\
\bottomrule
\end{tabular}
\caption{\textbf{x}: (1,0,0)\\slice: 2 of 4}
\end{subtable}
\hfill
\begin{subtable}[t]{3cm}\centering
\begin{tabular}[t]{@{}rrr@{}}
\toprule
   i &    r &    n \\
\midrule
 1.0 &  1.0 &  1.0 \\
 1.0 &  2.0 &  4.0 \\
 1.0 &  5.0 &  1.0 \\
 4.0 &  2.0 &  1.0 \\
 6.0 &  1.0 &  1.0 \\
\bottomrule
\end{tabular}
\caption{\textbf{x}: (1,0,0)\\slice: 3 of 4}
\end{subtable}
\hfill
\begin{subtable}[t]{3cm}\centering
\begin{tabular}[t]{@{}rrr@{}}
\toprule
   i &    r &    n \\
\midrule
 1.0 &  1.0 &  1.0 \\
 1.0 &  2.0 &  2.0 \\
 1.0 &  5.0 &  2.0 \\
 4.0 &  2.0 &  1.0 \\
 6.0 &  1.0 &  1.0 \\
\bottomrule
\end{tabular}
\caption{\textbf{x}: (1,0,0)\\slice: 4 of 4}
\end{subtable}
\end{table}
\begin{table}[ht]\ContinuedFloat
\centering
\small
\begin{subtable}[t]{3cm}\centering
\begin{tabular}[t]{@{}rrr@{}}
\toprule
   i &    r &    n \\
\midrule
 1.0 &  1.0 &  3.0 \\
 1.0 &  2.0 &  3.0 \\
 4.0 &  1.0 &  6.0 \\
 4.0 &  2.0 &  1.0 \\
 6.0 &  1.0 &  3.0 \\
\bottomrule
\end{tabular}
\caption{\textbf{x}: (1,1,0)\\slice: 1 of 4}
\end{subtable}
\hfill
\begin{subtable}[t]{3cm}\centering
\begin{tabular}[t]{@{}rrr@{}}
\toprule
   i &    r &    n \\
\midrule
 1.0 &  1.0 &  2.0 \\
 1.0 &  2.0 &  5.0 \\
 3.0 &  1.0 &  1.0 \\
 4.0 &  1.0 &  4.0 \\
 6.0 &  1.0 &  2.0 \\
\bottomrule
\end{tabular}
\caption{\textbf{x}: (1,1,0)\\slice: 2 of 4}
\end{subtable}
\hfill
\begin{subtable}[t]{3cm}\centering
\begin{tabular}[t]{@{}rrr@{}}
\toprule
   i &    r &    n \\
\midrule
 1.0 &  1.0 &  3.0 \\
 1.0 &  2.0 &  4.0 \\
 1.0 &  3.0 &  1.0 \\
 4.0 &  1.0 &  2.0 \\
 6.0 &  1.0 &  1.0 \\
\bottomrule
\end{tabular}
\caption{\textbf{x}: (1,1,0)\\slice: 3 of 4}
\end{subtable}
\hfill
\begin{subtable}[t]{3cm}\centering
\begin{tabular}[t]{@{}rrr@{}}
\toprule
   i &    r &    n \\
\midrule
 1.0 &  1.0 &  2.0 \\
 1.0 &  2.0 &  3.0 \\
 1.0 &  3.0 &  1.0 \\
 1.0 &  4.0 &  1.0 \\
 4.0 &  1.0 &  2.0 \\
 6.0 &  1.0 &  1.0 \\
\bottomrule
\end{tabular}
\caption{\textbf{x}: (1,1,0)\\slice: 4 of 4}
\end{subtable}
\caption{Grey-level run length matrices extracted from the $xy$ plane (2D) of the digital phantom. \textbf{x} indicates the direction in $(x,y,z)$ coordinates.}
\end{table}

\FloatBarrier

\section{Grey level run length matrix (2D, merged)}
\begin{table}[ht]
\centering
\small
\begin{subtable}[t]{3cm}\centering
\begin{tabular}[t]{@{}rrr@{}}
\toprule
   i &    r &     n \\
\midrule
 1.0 &  1.0 &  16.0 \\
 1.0 &  2.0 &   8.0 \\
 1.0 &  4.0 &   1.0 \\
 4.0 &  1.0 &  17.0 \\
 4.0 &  2.0 &   6.0 \\
 4.0 &  3.0 &   1.0 \\
 6.0 &  1.0 &   9.0 \\
 6.0 &  3.0 &   1.0 \\
\bottomrule
\end{tabular}
\caption{slice: 1 of 4}
\end{subtable}
\hfill
\begin{subtable}[t]{3cm}\centering
\begin{tabular}[t]{@{}rrr@{}}
\toprule
   i &    r &     n \\
\midrule
 1.0 &  1.0 &  10.0 \\
 1.0 &  2.0 &  15.0 \\
 1.0 &  4.0 &   2.0 \\
 3.0 &  1.0 &   4.0 \\
 4.0 &  1.0 &  12.0 \\
 4.0 &  2.0 &   2.0 \\
 6.0 &  1.0 &   8.0 \\
\bottomrule
\end{tabular}
\caption{slice: 2 of 4}
\end{subtable}
\hfill
\begin{subtable}[t]{3cm}\centering
\begin{tabular}[t]{@{}rrr@{}}
\toprule
   i &    r &     n \\
\midrule
 1.0 &  1.0 &  10.0 \\
 1.0 &  2.0 &  11.0 \\
 1.0 &  3.0 &   5.0 \\
 1.0 &  4.0 &   1.0 \\
 1.0 &  5.0 &   1.0 \\
 4.0 &  1.0 &   6.0 \\
 4.0 &  2.0 &   1.0 \\
 6.0 &  1.0 &   4.0 \\
\bottomrule
\end{tabular}
\caption{slice: 3 of 4}
\end{subtable}
\hfill
\begin{subtable}[t]{3cm}\centering
\begin{tabular}[t]{@{}rrr@{}}
\toprule
   i &    r &    n \\
\midrule
 1.0 &  1.0 &  6.0 \\
 1.0 &  2.0 &  9.0 \\
 1.0 &  3.0 &  6.0 \\
 1.0 &  4.0 &  2.0 \\
 1.0 &  5.0 &  2.0 \\
 4.0 &  1.0 &  6.0 \\
 4.0 &  2.0 &  1.0 \\
 6.0 &  1.0 &  4.0 \\
\bottomrule
\end{tabular}
\caption{slice: 4 of 4}
\end{subtable}
\caption{Merged grey-level run length matrices extracted from the $xy$ plane (2D) of the digital phantom.}
\end{table}

\FloatBarrier

\section{Grey level run length matrix (3D)}
\begin{table}[ht]
\centering
\small
\begin{subtable}[t]{3cm}\centering
\begin{tabular}[t]{@{}rrr@{}}
\toprule
   i &    r &    n \\
\midrule
 1.0 &  1.0 &  1.0 \\
 1.0 &  2.0 &  6.0 \\
 1.0 &  3.0 &  3.0 \\
 1.0 &  4.0 &  7.0 \\
 3.0 &  1.0 &  1.0 \\
 4.0 &  1.0 &  4.0 \\
 4.0 &  2.0 &  2.0 \\
 4.0 &  4.0 &  2.0 \\
 6.0 &  1.0 &  1.0 \\
 6.0 &  2.0 &  1.0 \\
 6.0 &  4.0 &  1.0 \\
\bottomrule
\end{tabular}
\caption{\textbf{x}: (0,0,1)}
\end{subtable}
\hfill
\begin{subtable}[t]{3cm}\centering
\begin{tabular}[t]{@{}rrr@{}}
\toprule
   i &    r &     n \\
\midrule
 1.0 &  1.0 &  11.0 \\
 1.0 &  2.0 &  15.0 \\
 1.0 &  3.0 &   3.0 \\
 3.0 &  1.0 &   1.0 \\
 4.0 &  1.0 &  14.0 \\
 4.0 &  2.0 &   1.0 \\
 6.0 &  1.0 &   5.0 \\
 6.0 &  2.0 &   1.0 \\
\bottomrule
\end{tabular}
\caption{\textbf{x}: (0,1,-1)}
\end{subtable}
\hfill
\begin{subtable}[t]{3cm}\centering
\begin{tabular}[t]{@{}rrr@{}}
\toprule
   i &    r &     n \\
\midrule
 1.0 &  1.0 &   2.0 \\
 1.0 &  2.0 &   5.0 \\
 1.0 &  3.0 &   6.0 \\
 1.0 &  4.0 &   5.0 \\
 3.0 &  1.0 &   1.0 \\
 4.0 &  1.0 &  10.0 \\
 4.0 &  2.0 &   3.0 \\
 6.0 &  1.0 &   4.0 \\
 6.0 &  3.0 &   1.0 \\
\bottomrule
\end{tabular}
\caption{\textbf{x}: (0,1,0)}
\end{subtable}
\hfill
\begin{subtable}[t]{3cm}\centering
\begin{tabular}[t]{@{}rrr@{}}
\toprule
   i &    r &     n \\
\midrule
 1.0 &  1.0 &  10.0 \\
 1.0 &  2.0 &   5.0 \\
 1.0 &  3.0 &   6.0 \\
 1.0 &  4.0 &   3.0 \\
 3.0 &  1.0 &   1.0 \\
 4.0 &  1.0 &  14.0 \\
 4.0 &  2.0 &   1.0 \\
 6.0 &  1.0 &   5.0 \\
 6.0 &  2.0 &   1.0 \\
\bottomrule
\end{tabular}
\caption{\textbf{x}: (0,1,1)}
\end{subtable}
\end{table}
\begin{table}[ht]\ContinuedFloat
\centering
\small
\begin{subtable}[t]{3cm}\centering
\begin{tabular}[t]{@{}rrr@{}}
\toprule
   i &    r &     n \\
\midrule
 1.0 &  1.0 &  22.0 \\
 1.0 &  2.0 &  11.0 \\
 1.0 &  3.0 &   2.0 \\
 3.0 &  1.0 &   1.0 \\
 4.0 &  1.0 &  16.0 \\
 6.0 &  1.0 &   7.0 \\
\bottomrule
\end{tabular}
\caption{\textbf{x}: (1,-1,-1)}
\end{subtable}
\hfill
\begin{subtable}[t]{3cm}\centering
\begin{tabular}[t]{@{}rrr@{}}
\toprule
   i &    r &     n \\
\midrule
 1.0 &  1.0 &  21.0 \\
 1.0 &  2.0 &  10.0 \\
 1.0 &  3.0 &   3.0 \\
 3.0 &  1.0 &   1.0 \\
 4.0 &  1.0 &  13.0 \\
 4.0 &  3.0 &   1.0 \\
 6.0 &  1.0 &   7.0 \\
\bottomrule
\end{tabular}
\caption{\textbf{x}: (1,-1,0)}
\end{subtable}
\hfill
\begin{subtable}[t]{3cm}\centering
\begin{tabular}[t]{@{}rrr@{}}
\toprule
   i &    r &     n \\
\midrule
 1.0 &  1.0 &  30.0 \\
 1.0 &  2.0 &  10.0 \\
 3.0 &  1.0 &   1.0 \\
 4.0 &  1.0 &  14.0 \\
 4.0 &  2.0 &   1.0 \\
 6.0 &  1.0 &   7.0 \\
\bottomrule
\end{tabular}
\caption{\textbf{x}: (1,-1,1)}
\end{subtable}
\hfill
\begin{subtable}[t]{3cm}\centering
\begin{tabular}[t]{@{}rrr@{}}
\toprule
   i &    r &     n \\
\midrule
 1.0 &  1.0 &  16.0 \\
 1.0 &  2.0 &  12.0 \\
 1.0 &  3.0 &   2.0 \\
 1.0 &  4.0 &   1.0 \\
 3.0 &  1.0 &   1.0 \\
 4.0 &  1.0 &   8.0 \\
 4.0 &  2.0 &   4.0 \\
 6.0 &  1.0 &   7.0 \\
\bottomrule
\end{tabular}
\caption{\textbf{x}: (1,0,-1)}
\end{subtable}
\end{table}
\begin{table}[ht]\ContinuedFloat
\centering
\small
\begin{subtable}[t]{3cm}\centering
\begin{tabular}[t]{@{}rrr@{}}
\toprule
   i &    r &     n \\
\midrule
 1.0 &  1.0 &   9.0 \\
 1.0 &  2.0 &  13.0 \\
 1.0 &  5.0 &   3.0 \\
 3.0 &  1.0 &   1.0 \\
 4.0 &  1.0 &   4.0 \\
 4.0 &  2.0 &   6.0 \\
 6.0 &  1.0 &   7.0 \\
\bottomrule
\end{tabular}
\caption{\textbf{x}: (1,0,0)}
\end{subtable}
\hfill
\begin{subtable}[t]{3cm}\centering
\begin{tabular}[t]{@{}rrr@{}}
\toprule
   i &    r &     n \\
\midrule
 1.0 &  1.0 &  19.0 \\
 1.0 &  2.0 &  12.0 \\
 1.0 &  3.0 &   1.0 \\
 1.0 &  4.0 &   1.0 \\
 3.0 &  1.0 &   1.0 \\
 4.0 &  1.0 &   8.0 \\
 4.0 &  2.0 &   4.0 \\
 6.0 &  1.0 &   7.0 \\
\bottomrule
\end{tabular}
\caption{\textbf{x}: (1,0,1)}
\end{subtable}
\hfill
\begin{subtable}[t]{3cm}\centering
\begin{tabular}[t]{@{}rrr@{}}
\toprule
   i &    r &     n \\
\midrule
 1.0 &  1.0 &  20.0 \\
 1.0 &  2.0 &  12.0 \\
 1.0 &  3.0 &   2.0 \\
 3.0 &  1.0 &   1.0 \\
 4.0 &  1.0 &  16.0 \\
 6.0 &  1.0 &   7.0 \\
\bottomrule
\end{tabular}
\caption{\textbf{x}: (1,1,-1)}
\end{subtable}
\hfill
\begin{subtable}[t]{3cm}\centering
\begin{tabular}[t]{@{}rrr@{}}
\toprule
   i &    r &     n \\
\midrule
 1.0 &  1.0 &  10.0 \\
 1.0 &  2.0 &  15.0 \\
 1.0 &  3.0 &   2.0 \\
 1.0 &  4.0 &   1.0 \\
 3.0 &  1.0 &   1.0 \\
 4.0 &  1.0 &  14.0 \\
 4.0 &  2.0 &   1.0 \\
 6.0 &  1.0 &   7.0 \\
\bottomrule
\end{tabular}
\caption{\textbf{x}: (1,1,0)}
\end{subtable}
\end{table}
\begin{table}[ht]\ContinuedFloat
\centering
\small
\begin{subtable}[t]{3cm}\centering
\begin{tabular}[t]{@{}rrr@{}}
\toprule
   i &    r &     n \\
\midrule
 1.0 &  1.0 &  19.0 \\
 1.0 &  2.0 &  14.0 \\
 1.0 &  3.0 &   1.0 \\
 3.0 &  1.0 &   1.0 \\
 4.0 &  1.0 &  14.0 \\
 4.0 &  2.0 &   1.0 \\
 6.0 &  1.0 &   7.0 \\
\bottomrule
\end{tabular}
\caption{\textbf{x}: (1,1,1)}
\end{subtable}
\caption{Grey-level run length matrices extracted volumetrically (3D) from the digital phantom. \textbf{x} indicates the direction in $(x,y,z)$ coordinates.}
\end{table}

\FloatBarrier

\section{Grey level run length matrix (3D, merged)}
\begin{table}[ht]
\centering
\small
\begin{subtable}[t]{3cm}\centering
\begin{tabular}[t]{@{}rrr@{}}
\toprule
   i &    r &      n \\
\midrule
 1.0 &  1.0 &  190.0 \\
 1.0 &  2.0 &  140.0 \\
 1.0 &  3.0 &   31.0 \\
 1.0 &  4.0 &   18.0 \\
 1.0 &  5.0 &    3.0 \\
 3.0 &  1.0 &   13.0 \\
 4.0 &  1.0 &  149.0 \\
 4.0 &  2.0 &   24.0 \\
 4.0 &  3.0 &    1.0 \\
 4.0 &  4.0 &    2.0 \\
 6.0 &  1.0 &   78.0 \\
 6.0 &  2.0 &    3.0 \\
 6.0 &  3.0 &    1.0 \\
 6.0 &  4.0 &    1.0 \\
\bottomrule
\end{tabular}
\end{subtable}
\caption{Merged grey-level run length matrix extracted volumetrically (3D) from the digital phantom.}
\end{table}

\FloatBarrier

\section{Grey level size zone matrix (2D)}
\begin{table}[ht]
\centering
\small
\begin{subtable}[t]{3cm}\centering
\begin{tabular}[t]{@{}rrr@{}}
\toprule
   i &  s &  n \\
\midrule
 1.0 &  3 &  1 \\
 1.0 &  6 &  1 \\
 4.0 &  2 &  1 \\
 4.0 &  6 &  1 \\
 6.0 &  3 &  1 \\
\bottomrule
\end{tabular}
\caption{slice: 1 of 4}
\end{subtable}
\hfill
\begin{subtable}[t]{3cm}\centering
\begin{tabular}[t]{@{}rrr@{}}
\toprule
   i &  s &  n \\
\midrule
 1.0 &  4 &  1 \\
 1.0 &  8 &  1 \\
 3.0 &  1 &  1 \\
 4.0 &  2 &  2 \\
 6.0 &  1 &  2 \\
\bottomrule
\end{tabular}
\caption{slice: 2 of 4}
\end{subtable}
\hfill
\begin{subtable}[t]{3cm}\centering
\begin{tabular}[t]{@{}rrr@{}}
\toprule
   i &   s &  n \\
\midrule
 1.0 &  14 &  1 \\
 4.0 &   2 &  1 \\
 6.0 &   1 &  1 \\
\bottomrule
\end{tabular}
\caption{slice: 3 of 4}
\end{subtable}
\hfill
\begin{subtable}[t]{3cm}\centering
\begin{tabular}[t]{@{}rrr@{}}
\toprule
   i &   s &  n \\
\midrule
 1.0 &  15 &  1 \\
 4.0 &   2 &  1 \\
 6.0 &   1 &  1 \\
\bottomrule
\end{tabular}
\caption{slice: 4 of 4}
\end{subtable}
\caption{Grey level size zone matrices extracted from the $xy$ plane (2D) of the digital phantom.}
\end{table}

\FloatBarrier

\section{Grey level size zone matrix (3D)}
\begin{table}[ht]
\centering
\small
\begin{subtable}[t]{3cm}\centering
\begin{tabular}[t]{@{}rrr@{}}
\toprule
   i &   s &  n \\
\midrule
 1.0 &  50 &  1 \\
 3.0 &   1 &  1 \\
 4.0 &   2 &  1 \\
 4.0 &  14 &  1 \\
 6.0 &   7 &  1 \\
\bottomrule
\end{tabular}
\end{subtable}
\caption{Grey level size zone matrix extracted volumetrically (3D) from the digital phantom.}
\end{table}

\FloatBarrier

\section{Grey level distance zone matrix (2D)}
\begin{table}[ht]
\centering
\small
\begin{subtable}[t]{3cm}\centering
\begin{tabular}[t]{@{}rrr@{}}
\toprule
   i &    d &  n \\
\midrule
 1.0 &  1.0 &  2 \\
 4.0 &  1.0 &  2 \\
 6.0 &  1.0 &  1 \\
\bottomrule
\end{tabular}
\caption{slice: 1 of 4}
\end{subtable}
\hfill
\begin{subtable}[t]{3cm}\centering
\begin{tabular}[t]{@{}rrr@{}}
\toprule
   i &    d &  n \\
\midrule
 1.0 &  1.0 &  2 \\
 3.0 &  2.0 &  1 \\
 4.0 &  1.0 &  2 \\
 6.0 &  1.0 &  1 \\
 6.0 &  2.0 &  1 \\
\bottomrule
\end{tabular}
\caption{slice: 2 of 4}
\end{subtable}
\hfill
\begin{subtable}[t]{3cm}\centering
\begin{tabular}[t]{@{}rrr@{}}
\toprule
   i &    d &  n \\
\midrule
 1.0 &  1.0 &  1 \\
 4.0 &  1.0 &  1 \\
 6.0 &  1.0 &  1 \\
\bottomrule
\end{tabular}
\caption{slice: 3 of 4}
\end{subtable}
\hfill
\begin{subtable}[t]{3cm}\centering
\begin{tabular}[t]{@{}rrr@{}}
\toprule
   i &    d &  n \\
\midrule
 1.0 &  1.0 &  1 \\
 4.0 &  1.0 &  1 \\
 6.0 &  1.0 &  1 \\
\bottomrule
\end{tabular}
\caption{slice: 4 of 4}
\end{subtable}
\caption{Grey level distance zone matrices extracted from the $xy$ plane (2D) of the digital phantom.}
\end{table}

\FloatBarrier

\section{Grey level distance zone matrix (3D)}
\begin{table}[ht]
\centering
\small
\begin{subtable}[t]{3cm}\centering
\begin{tabular}[t]{@{}rrr@{}}
\toprule
   i &    d &  n \\
\midrule
 1.0 &  1.0 &  1 \\
 3.0 &  1.0 &  1 \\
 4.0 &  1.0 &  2 \\
 6.0 &  1.0 &  1 \\
\bottomrule
\end{tabular}
\end{subtable}
\caption{Grey level distance zone matrix extracted volumetrically (3D) from the digital phantom.}
\end{table}

\FloatBarrier

\section{Neighbourhood grey tone difference matrix (2D)}
\begin{table}[ht]
\centering
\small
\begin{subtable}[t]{3cm}\centering
\begin{tabular}[t]{@{}rrr@{}}
\toprule
   i &       s &  n \\
\midrule
 1.0 &  14.575 &  9 \\
 4.0 &   5.775 &  8 \\
 6.0 &   7.325 &  3 \\
\bottomrule
\end{tabular}
\caption{slice: 1 of 4}
\end{subtable}
\hfill
\begin{subtable}[t]{3cm}\centering
\begin{tabular}[t]{@{}rrr@{}}
\toprule
   i &          s &   n \\
\midrule
 1.0 &  11.928571 &  12 \\
 3.0 &   0.375000 &   1 \\
 4.0 &   4.800000 &   4 \\
 6.0 &   8.000000 &   2 \\
\bottomrule
\end{tabular}
\caption{slice: 2 of 4}
\end{subtable}
\hfill
\begin{subtable}[t]{3cm}\centering
\begin{tabular}[t]{@{}rrr@{}}
\toprule
   i &         s &   n \\
\midrule
 1.0 &  7.985714 &  14 \\
 4.0 &  4.650000 &   2 \\
 6.0 &  5.000000 &   1 \\
\bottomrule
\end{tabular}
\caption{slice: 3 of 4}
\end{subtable}
\hfill
\begin{subtable}[t]{3cm}\centering
\begin{tabular}[t]{@{}rrr@{}}
\toprule
   i &         s &   n \\
\midrule
 1.0 &  7.582143 &  15 \\
 4.0 &  4.650000 &   2 \\
 6.0 &  5.000000 &   1 \\
\bottomrule
\end{tabular}
\caption{slice: 4 of 4}
\end{subtable}
\caption{Neighbourhood grey tone difference matrices extracted from the $xy$ plane (2D) of the digital phantom using Chebyshev distance 1.}
\end{table}

\FloatBarrier

\section{Neighbourhood grey tone difference matrix (3D)}
\begin{table}[ht]
\centering
\small
\begin{subtable}[t]{3cm}\centering
\begin{tabular}[t]{@{}rrr@{}}
\toprule
   i &          s &   n \\
\midrule
 1.0 &  39.946954 &  50 \\
 3.0 &   0.200000 &   1 \\
 4.0 &  20.825401 &  16 \\
 6.0 &  24.127005 &   7 \\
\bottomrule
\end{tabular}
\end{subtable}
\caption{Neighbourhood grey tone difference matrix extracted volumetrically (3D) from the digital phantom using Chebyshev distance 1.}
\end{table}

\FloatBarrier

\section{Neighbouring grey level dependence matrix (2D)}
\begin{table}[ht]
\centering
\small
\begin{subtable}[t]{3cm}\centering
\begin{tabular}[t]{@{}rrr@{}}
\toprule
   i &    j &  s \\
\midrule
 1.0 &  2.0 &  3 \\
 1.0 &  3.0 &  1 \\
 1.0 &  4.0 &  3 \\
 1.0 &  5.0 &  2 \\
 4.0 &  2.0 &  2 \\
 4.0 &  3.0 &  4 \\
 4.0 &  4.0 &  2 \\
 6.0 &  2.0 &  2 \\
 6.0 &  3.0 &  1 \\
\bottomrule
\end{tabular}
\caption{slice: 1 of 4}
\end{subtable}
\hfill
\begin{subtable}[t]{3cm}\centering
\begin{tabular}[t]{@{}rrr@{}}
\toprule
   i &    j &  s \\
\midrule
 1.0 &  3.0 &  2 \\
 1.0 &  4.0 &  6 \\
 1.0 &  6.0 &  4 \\
 3.0 &  1.0 &  1 \\
 4.0 &  2.0 &  4 \\
 6.0 &  1.0 &  2 \\
\bottomrule
\end{tabular}
\caption{slice: 2 of 4}
\end{subtable}
\hfill
\begin{subtable}[t]{3cm}\centering
\begin{tabular}[t]{@{}rrr@{}}
\toprule
   i &    j &  s \\
\midrule
 1.0 &  3.0 &  1 \\
 1.0 &  4.0 &  5 \\
 1.0 &  5.0 &  3 \\
 1.0 &  6.0 &  3 \\
 1.0 &  7.0 &  2 \\
 4.0 &  2.0 &  2 \\
 6.0 &  1.0 &  1 \\
\bottomrule
\end{tabular}
\caption{slice: 3 of 4}
\end{subtable}
\hfill
\begin{subtable}[t]{3cm}\centering
\begin{tabular}[t]{@{}rrr@{}}
\toprule
   i &    j &  s \\
\midrule
 1.0 &  3.0 &  1 \\
 1.0 &  4.0 &  3 \\
 1.0 &  5.0 &  3 \\
 1.0 &  6.0 &  4 \\
 1.0 &  7.0 &  1 \\
 1.0 &  8.0 &  3 \\
 4.0 &  2.0 &  2 \\
 6.0 &  1.0 &  1 \\
\bottomrule
\end{tabular}
\caption{slice: 4 of 4}
\end{subtable}
\caption{Neighbouring grey level dependence matrices extracted from the $xy$ plane (2D) of the digital phantom using Chebyshev distance 1 and coarseness 0.}
\end{table}

\FloatBarrier

\section{Neighbouring grey level dependence matrix (3D)}
\begin{table}[ht]
\centering
\small
\begin{subtable}[t]{3cm}\centering
\begin{tabular}[t]{@{}rrr@{}}
\toprule
   i &     j &  s \\
\midrule
 1.0 &   5.0 &  2 \\
 1.0 &   6.0 &  2 \\
 1.0 &   7.0 &  1 \\
 1.0 &   8.0 &  6 \\
 1.0 &   9.0 &  4 \\
 1.0 &  10.0 &  6 \\
 1.0 &  11.0 &  5 \\
 1.0 &  12.0 &  5 \\
 1.0 &  13.0 &  3 \\
 1.0 &  14.0 &  2 \\
 1.0 &  15.0 &  5 \\
 1.0 &  16.0 &  3 \\
 1.0 &  17.0 &  3 \\
 1.0 &  18.0 &  2 \\
 1.0 &  21.0 &  1 \\
 3.0 &   1.0 &  1 \\
 4.0 &   2.0 &  2 \\
 4.0 &   4.0 &  2 \\
 4.0 &   5.0 &  6 \\
 4.0 &   6.0 &  4 \\
 4.0 &   7.0 &  2 \\
 6.0 &   2.0 &  1 \\
 6.0 &   3.0 &  4 \\
 6.0 &   4.0 &  1 \\
 6.0 &   5.0 &  1 \\
\bottomrule
\end{tabular}
\end{subtable}
\caption{Neighbouring grey level dependence matrix extracted volumetrically (3D) from the digital phantom using Chebyshev distance 1 and coarseness 0.}
\end{table}

\FloatBarrier

\newpage

\bibliography{Bibliography}

\end{document}